\documentclass[sigconf,prologue,table]{acmart}

\usepackage{booktabs} 

\usepackage[ruled]{algorithm2e} 

\SetAlFnt{\small}
\SetAlCapFnt{\small}
\SetAlCapNameFnt{\small}
\SetAlCapHSkip{0pt}

\usepackage{amsmath}
\usepackage{amsfonts}
\usepackage{enumitem}
\usepackage[ruled]{algorithm2e}
\usepackage{soul}
\usepackage{microtype}
\usepackage{units}
\usepackage{gensymb}
\usepackage{array}
\usepackage{subcaption}
\usepackage{xcolor}
\usepackage{pifont}
\usepackage{multirow}
\usepackage{float}

\newcommand{\cmark}{\ding{51}}%
\newcommand{\xmark}{\ding{55}}%

\usepackage{dsfont}
\usepackage{wrapfig}
\usepackage{makecell}

\definecolor{azure}{rgb}{0.0, 0.5, 1.0}

\definecolor{deeplilac}{rgb}{0.6, 0.33, 0.73}

\newlength{\tmpintextsep}
\newlength{\tmpcolumnsep}
\tmpintextsep=\intextsep
\tmpcolumnsep=\columnsep

\newcommand{\figref}[1]{Fig.\ \ref{#1}}
\newcommand{\secref}[1]{Sec.\ \ref{#1}}
\newcommand{\tabref}[1]{Table\ \ref{#1}}

\newcommand{\Ghat}{\hat{G}}

\newcommand{\What}{\hat{W}}

\newsavebox{\largestimage}
\copyrightyear{2023} 
\acmYear{2023} 
\setcopyright{rightsretained} 
\acmConference[SA Conference Papers '23]{SIGGRAPH Asia 2023 Conference Papers}{December 12--15, 2023}{Sydney, NSW, Australia}
\acmBooktitle{SIGGRAPH Asia 2023 Conference Papers (SA Conference Papers '23), December 12--15, 2023, Sydney, NSW, Australia}
\acmDOI{10.1145/3610548.3618174}
\acmISBN{979-8-4007-0315-7/23/12}

\begin{document}

\title{UVDoc: Neural Grid-based Document Unwarping}

\author{Floor Verhoeven}
\orcid{0000-0003-3768-0460}
\affiliation{%
  \institution{ETH Zurich}
  \country{Switzerland}
}
\email{floor.verhoeven@inf.ethz.ch}

\author{Tanguy Magne}
\orcid{0009-0001-0231-026X}
\affiliation{%
  \institution{ETH Zurich}
  \country{Switzerland}
}
\email{tanguy.magne@inf.ethz.ch}

\author{Olga Sorkine-Hornung}
\orcid{0000-0002-8089-3974}
\affiliation{%
  \institution{ETH Zurich}
  \country{Switzerland}
}
\email{sorkine@inf.ethz.ch}

\begin{teaserfigure}
\newlength{\teaserImgWidth}
\setlength{\teaserImgWidth}{0.156\textwidth}
\newlength{\teaserImgHeight}
\setlength{\teaserImgHeight}{0.22\linewidth}
\newcommand{\horSpace}{\hspace{1.2pt}}
\newcommand{\lastSpace}{\hspace{1.5pt}}
\newcommand{\verSpace}{2.5pt}
\captionsetup[subfigure]{labelformat=empty}
\begin{center}
\begin{subfigure}[t]{0.65\linewidth}
\centering
    \includegraphics[height=\teaserImgHeight, width=\teaserImgWidth]{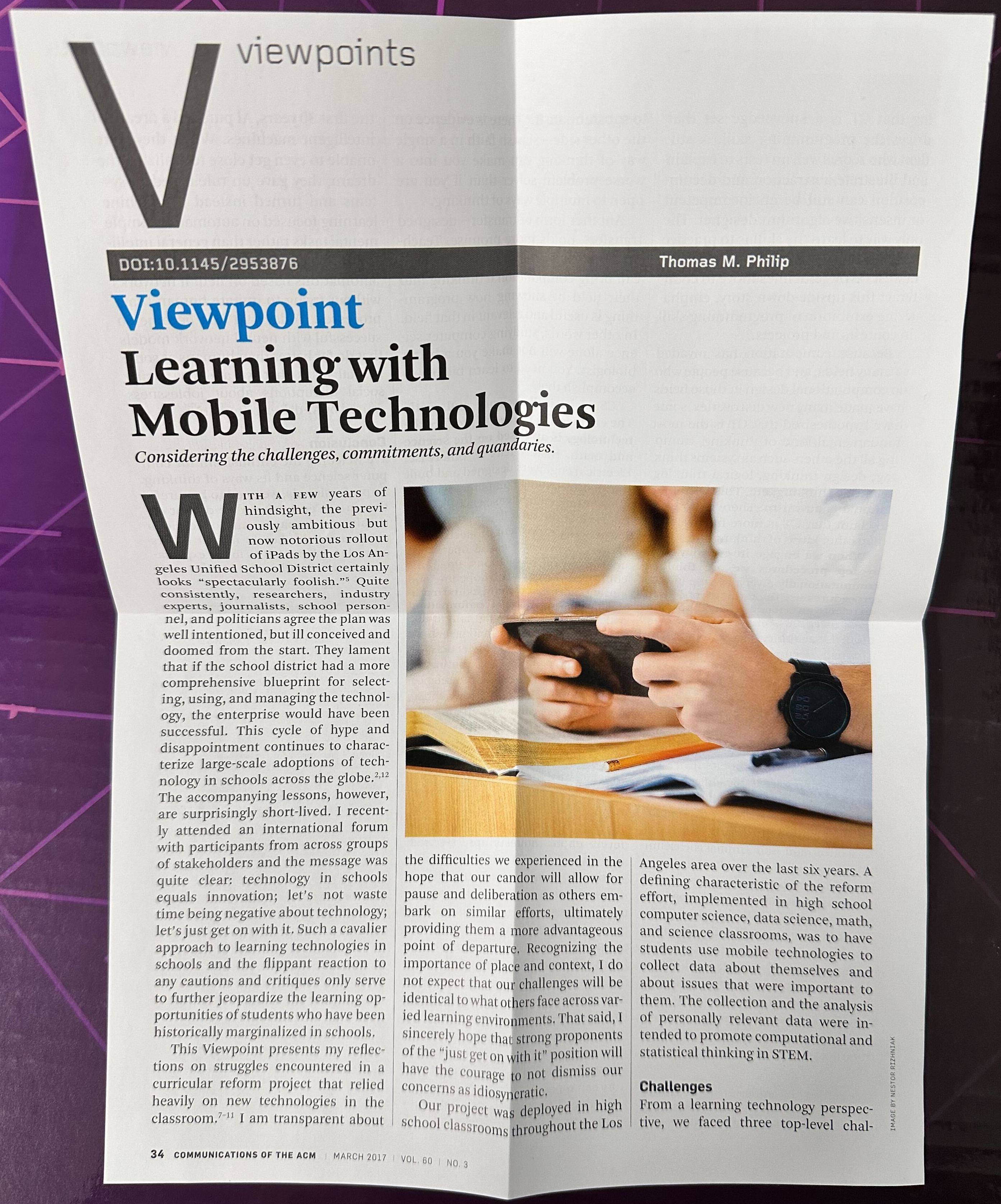}
    \horSpace
    \includegraphics[height=\teaserImgHeight, width=\teaserImgWidth]{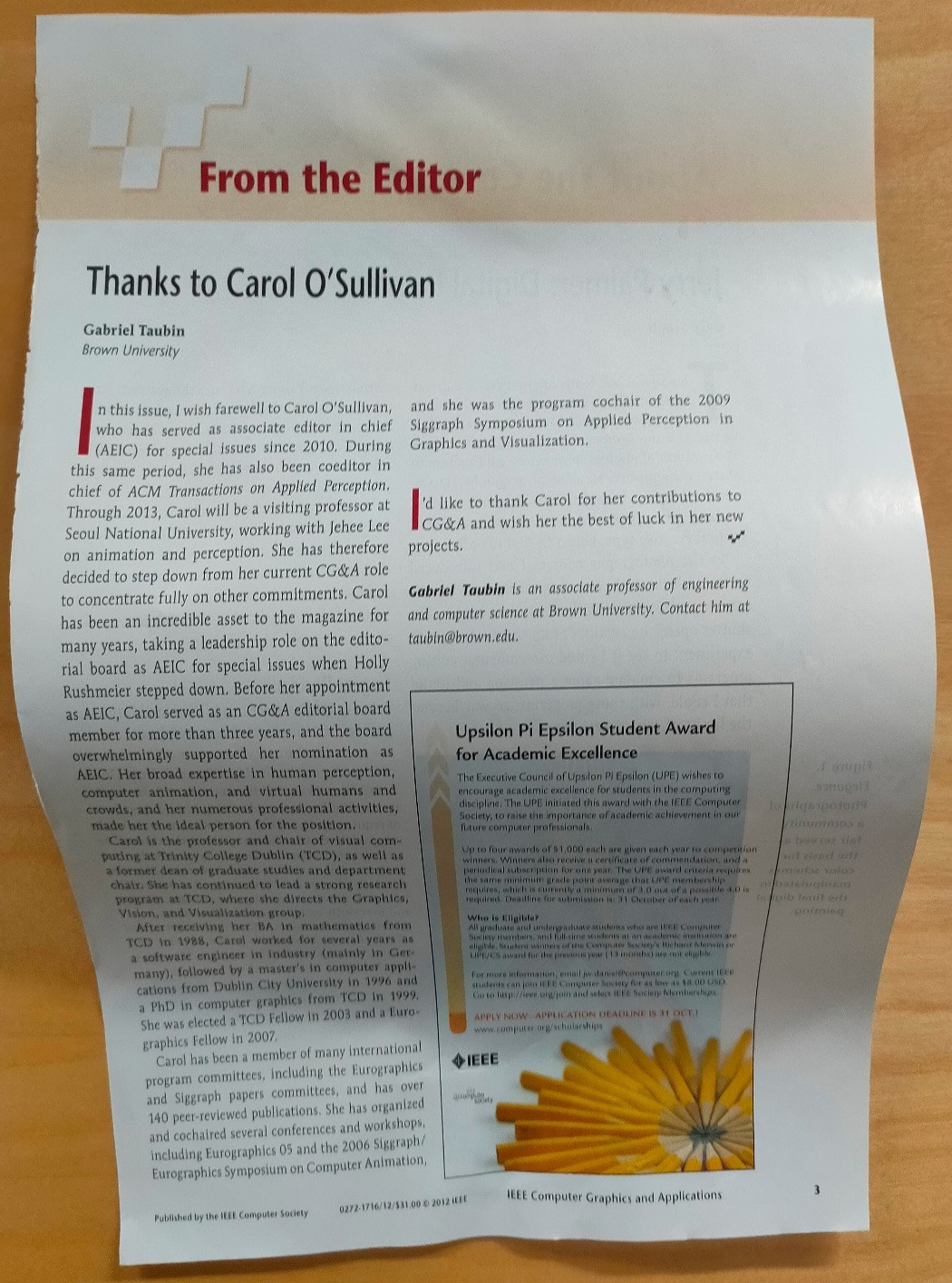}
    \horSpace
    \includegraphics[height=\teaserImgHeight, width=\teaserImgWidth]{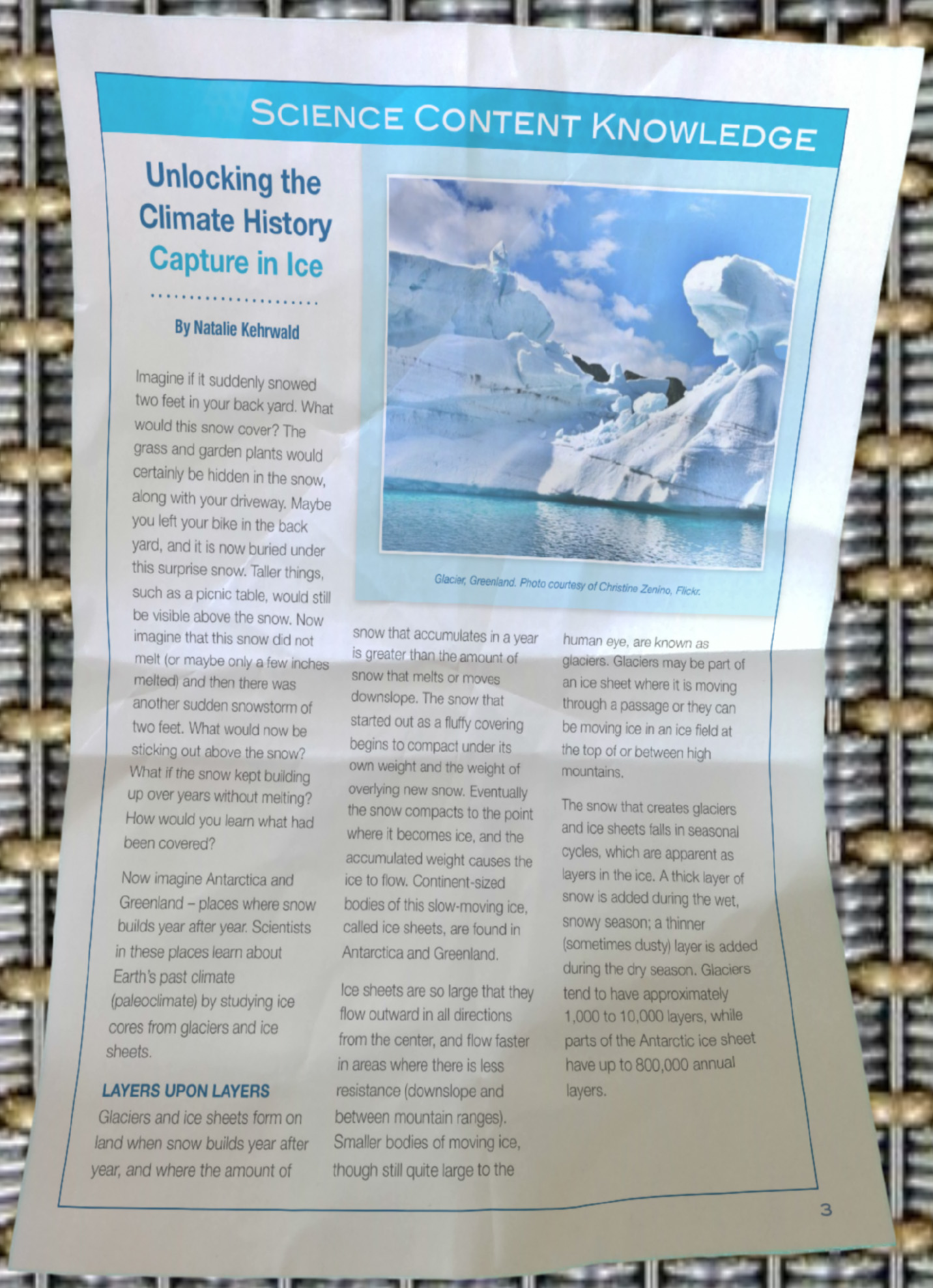}
    \horSpace
    \includegraphics[height=\teaserImgHeight, width=\teaserImgWidth]{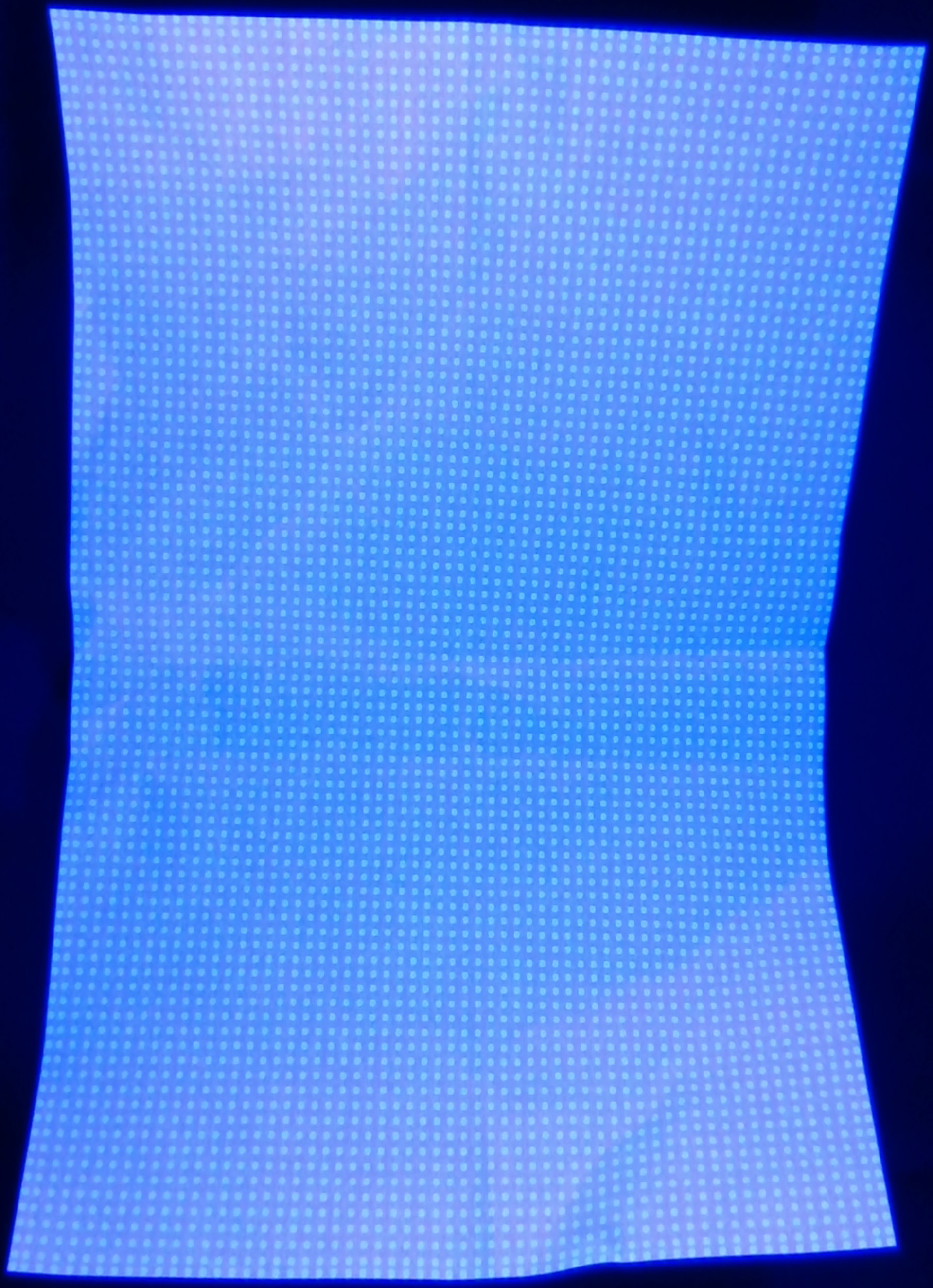}
\\\vspace{\verSpace}
    \includegraphics[height=\teaserImgHeight, width=\teaserImgWidth]{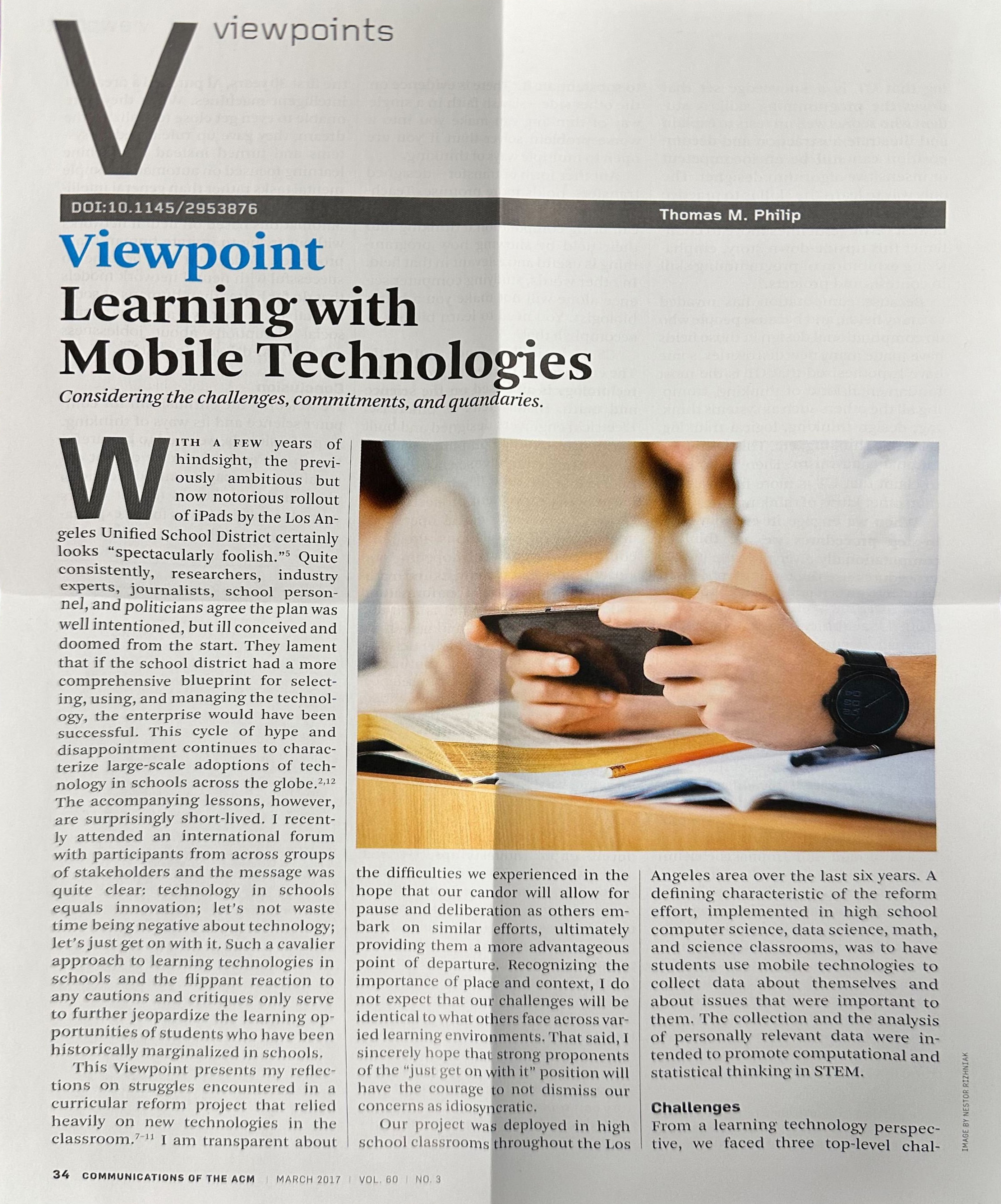}
    \horSpace
    \includegraphics[height=\teaserImgHeight, width=\teaserImgWidth]{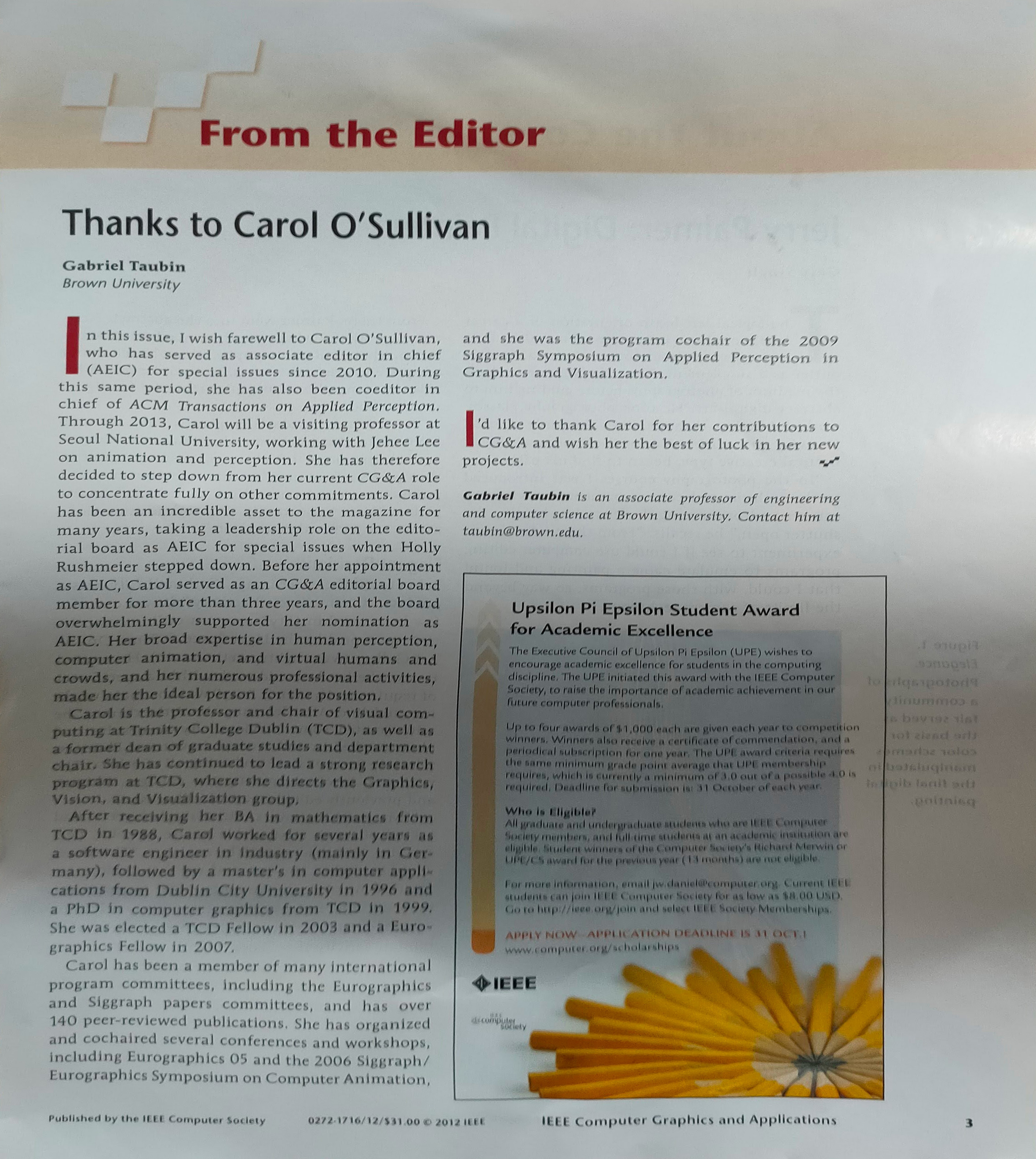}
    \horSpace
    \includegraphics[height=\teaserImgHeight, width=\teaserImgWidth]{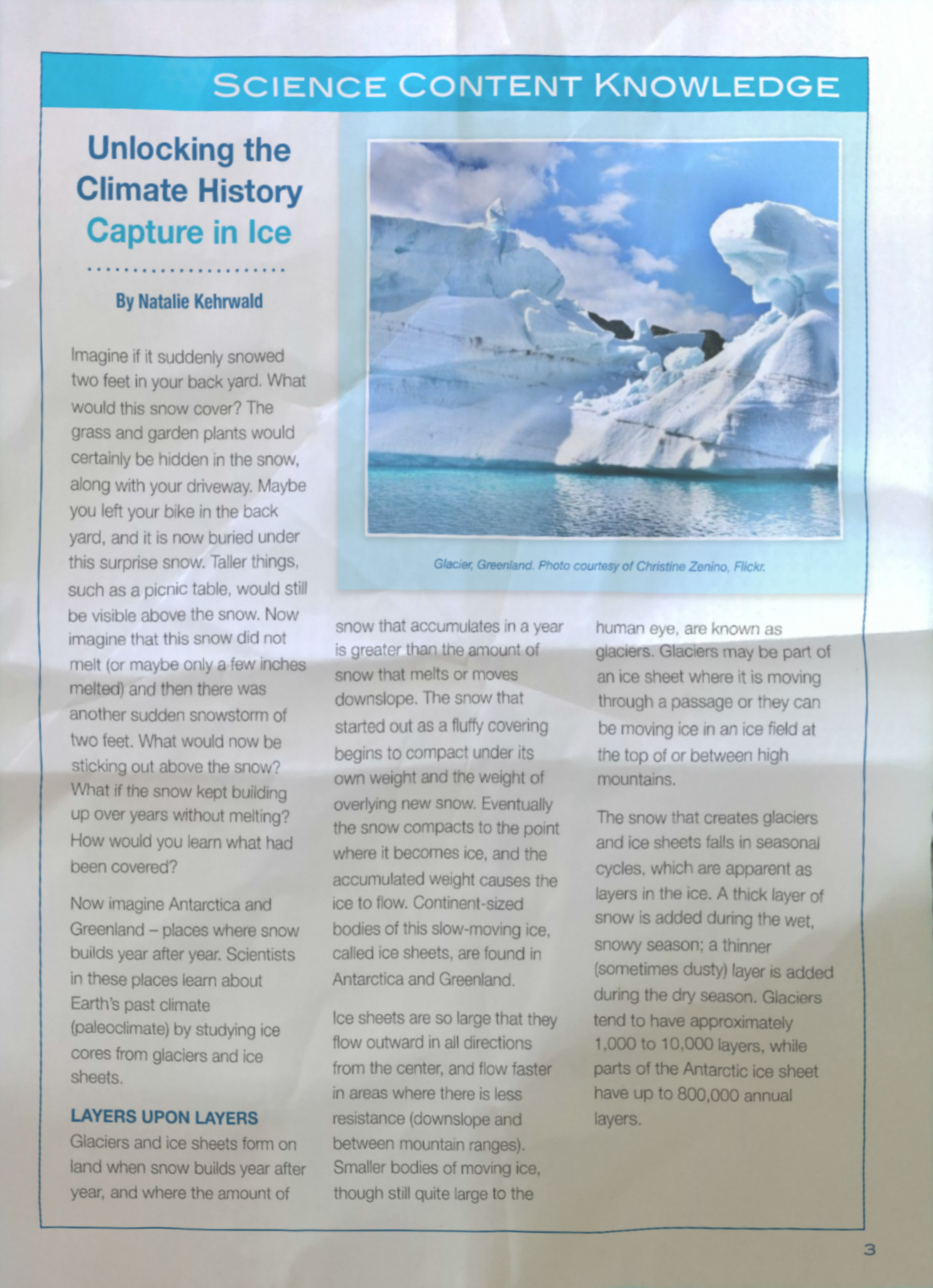}
    \horSpace
    \includegraphics[height=\teaserImgHeight, width=\teaserImgWidth]{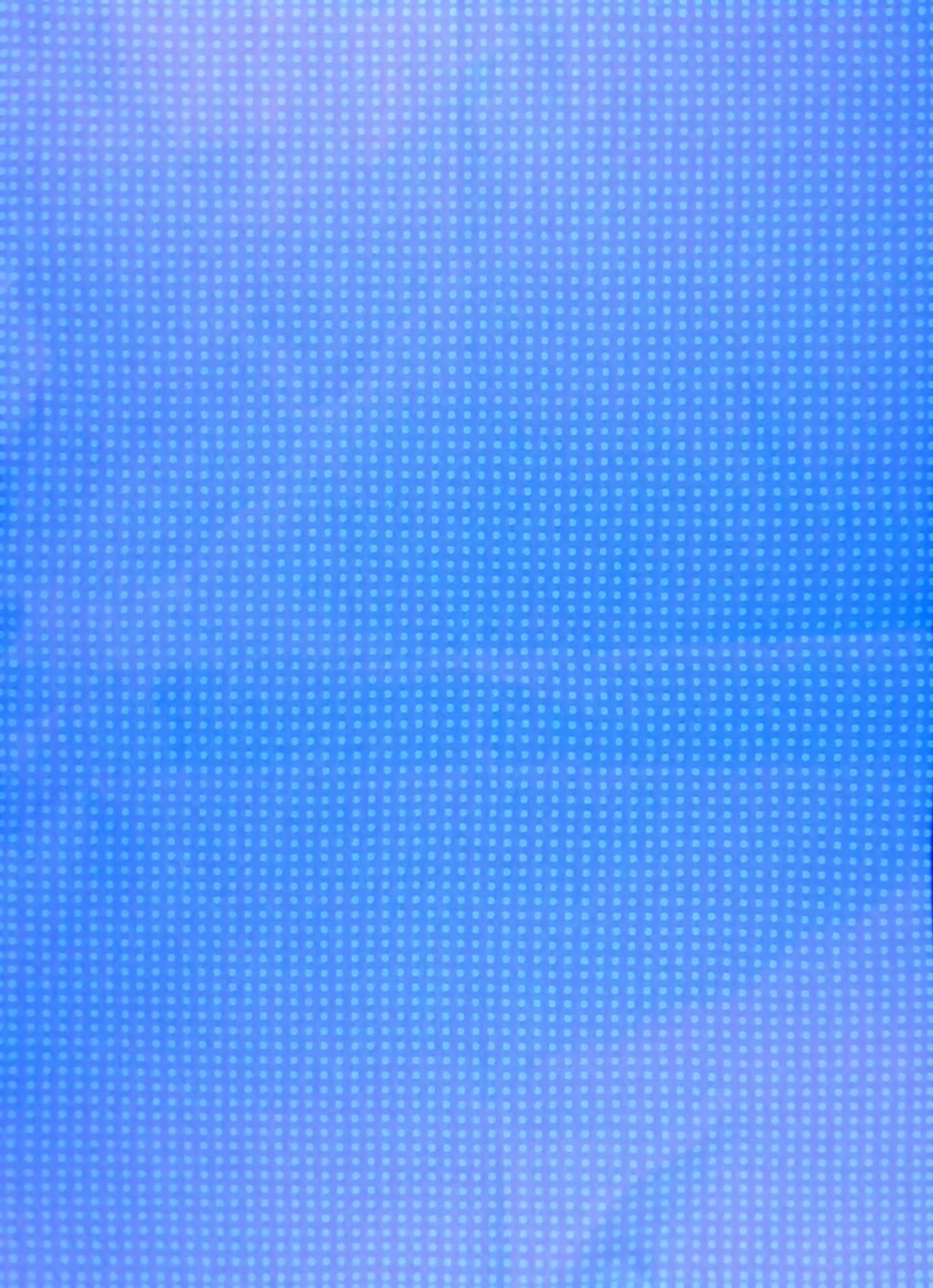}
    \caption{Input images (top) and our unwarping (bottom)}
\end{subfigure}
    \hspace{1.1pt}
\begin{subfigure}[t]{0.33\linewidth}
\centering
    \includegraphics[height=\teaserImgHeight, width=\teaserImgWidth]{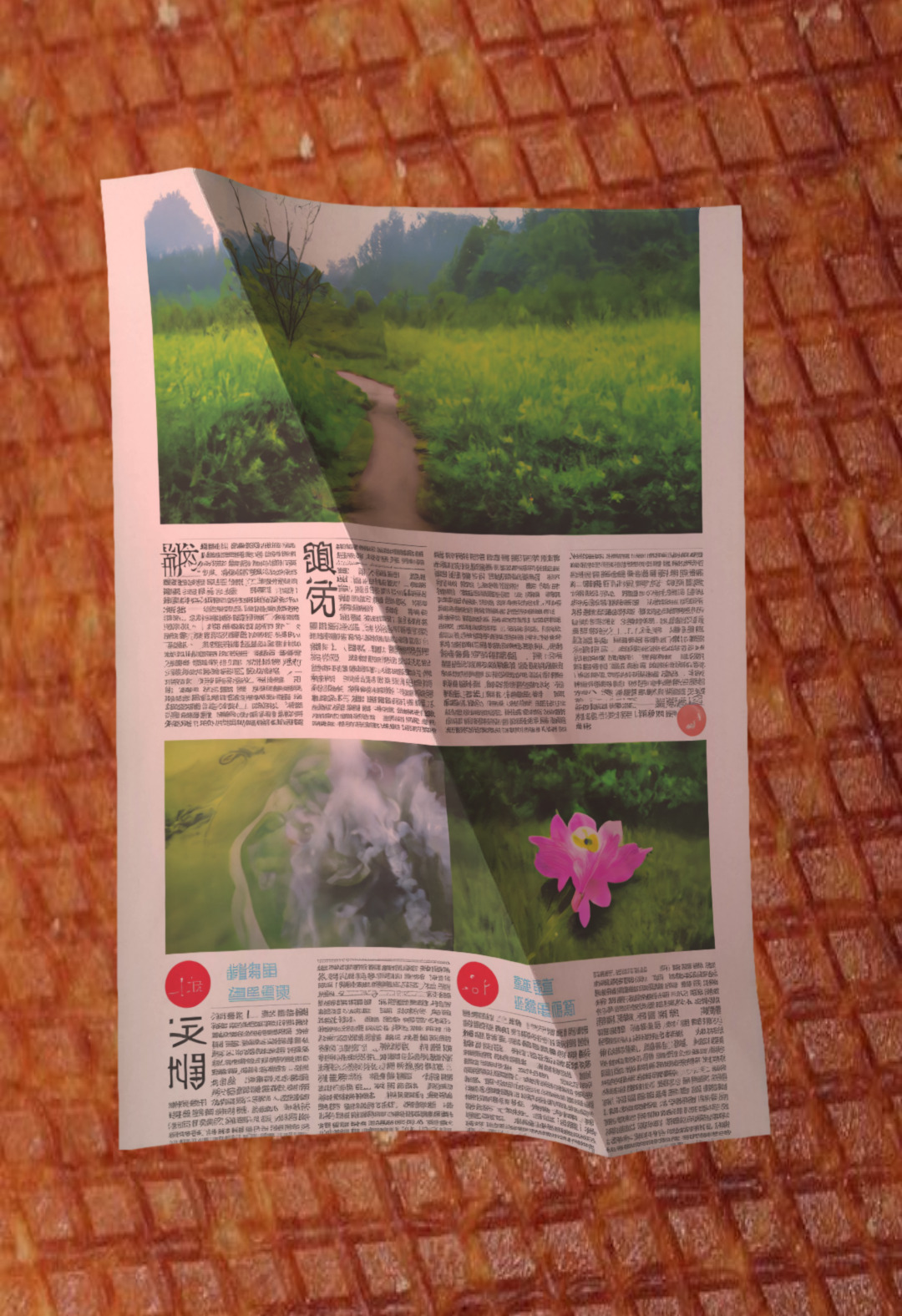}
    \lastSpace
    \includegraphics[height=\teaserImgHeight, width=\teaserImgWidth]{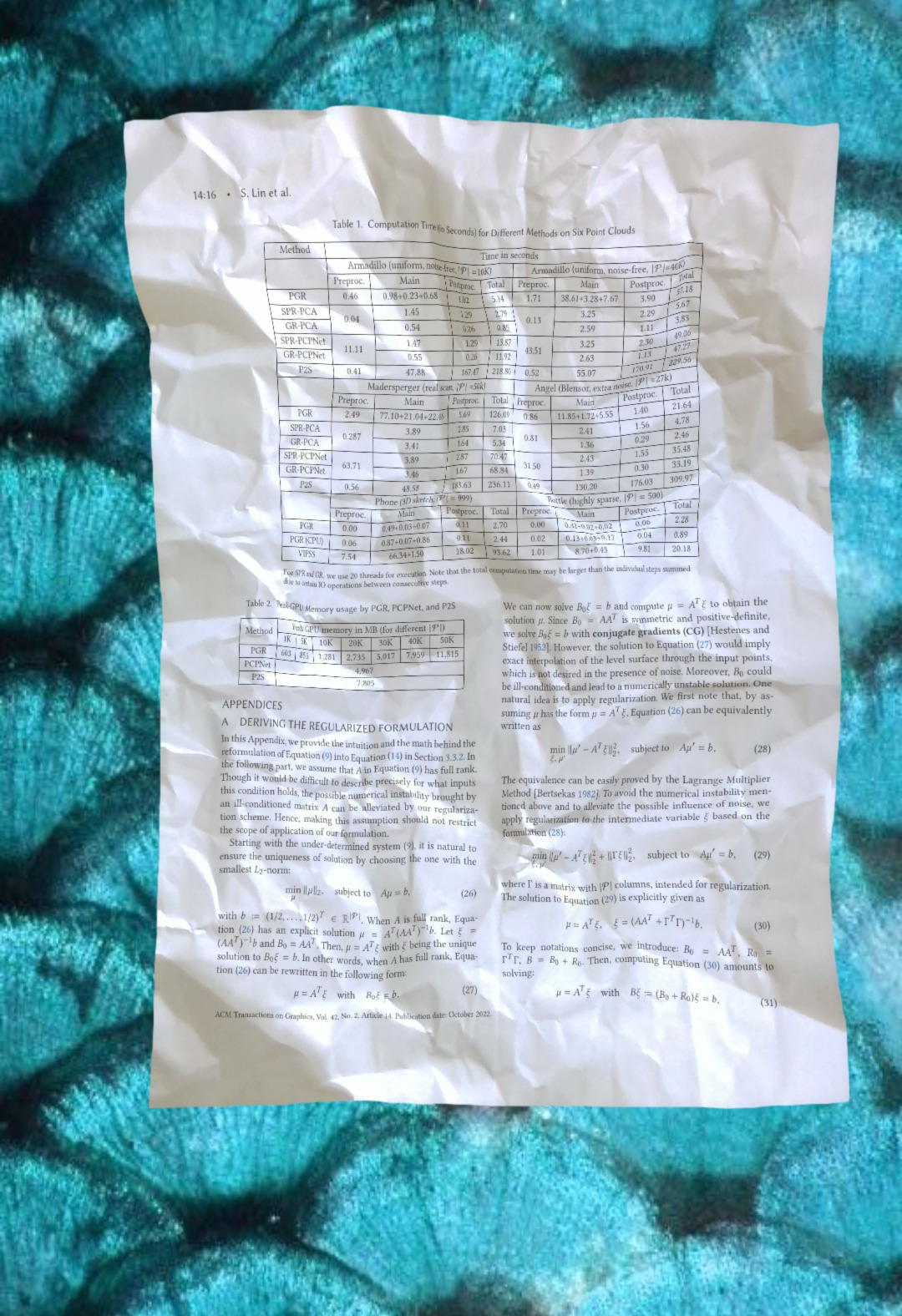}
\\\vspace{\verSpace}
    \includegraphics[height=\teaserImgHeight, width=\teaserImgWidth]{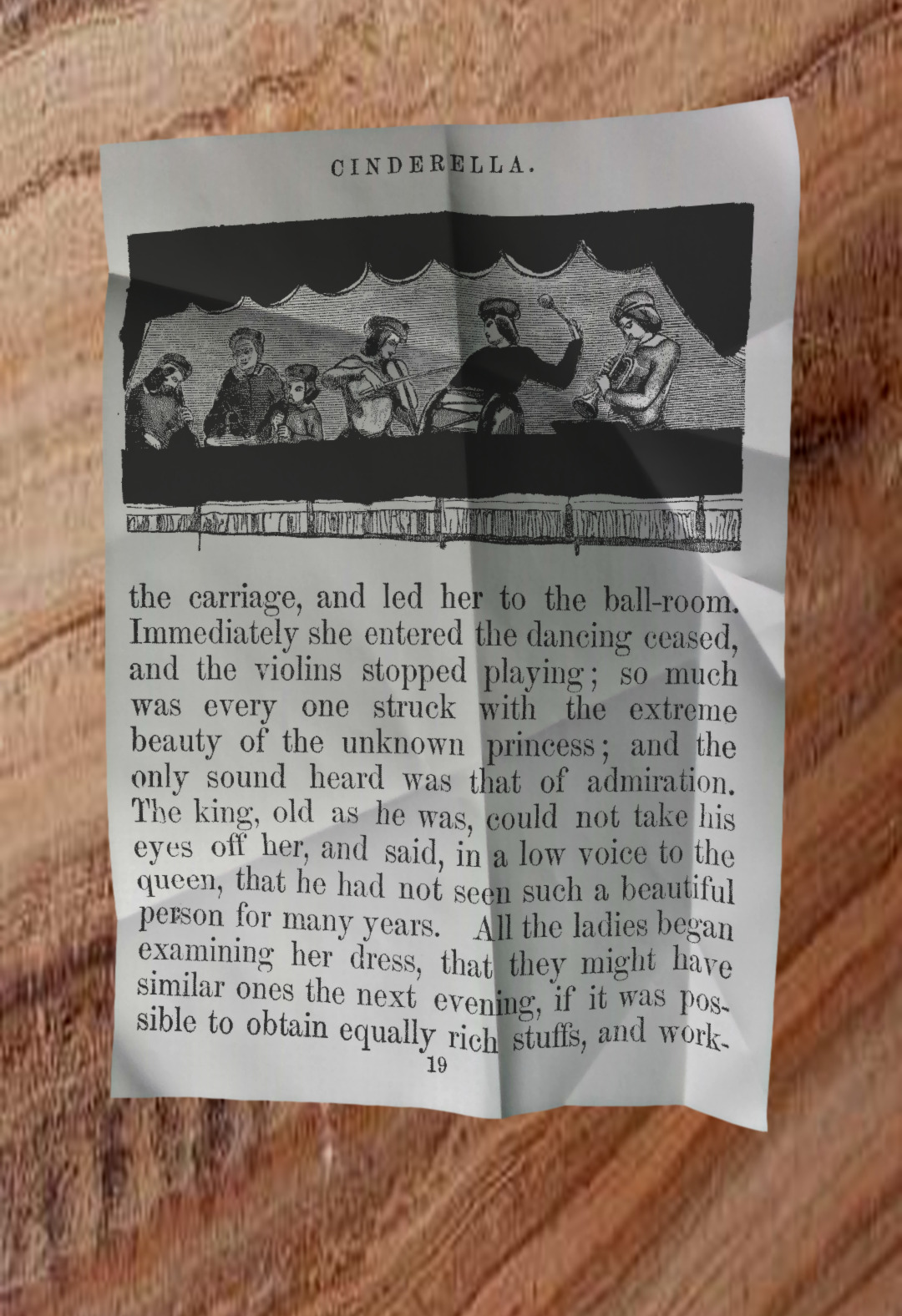}
   \lastSpace
    \includegraphics[height=\teaserImgHeight, width=\teaserImgWidth]{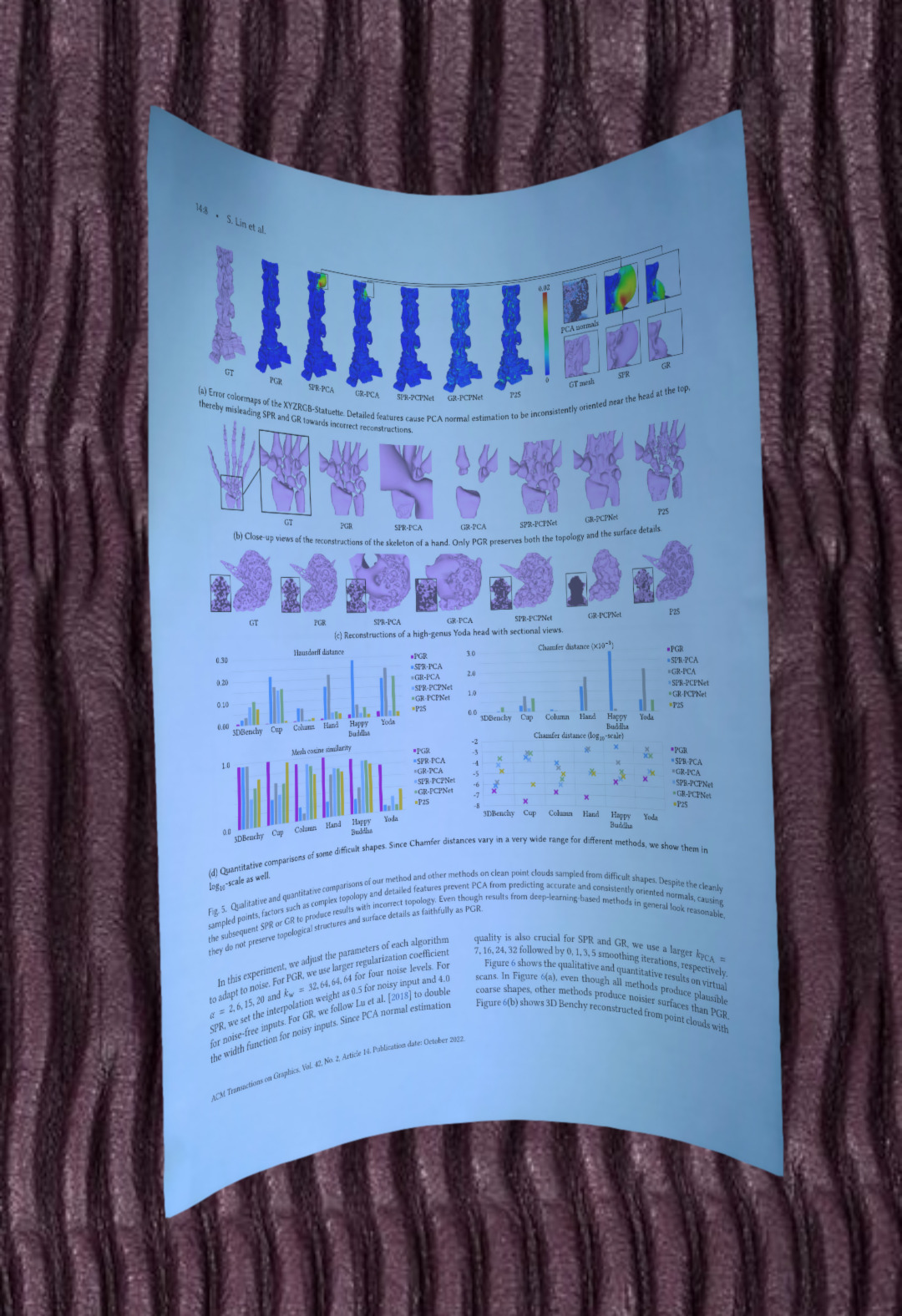}
    \caption{Samples from our UVDoc dataset}
\end{subfigure}    
\end{center}
\vspace{-9pt}
\centering
\caption{Unwarping results and data samples from our UVDoc dataset. 
The first two columns are examples where the input images were captured on a mobile phone. The third column is a sample from our UVDoc dataset and the fourth shows the corresponding UV-lit image along with its unwarping (obtained based on the unwarping grid predicted for the image in the third column). The last two columns show a few more examples from our UVDoc dataset.
}
\label{fig:teaser}
\end{teaserfigure}

\begin{abstract}
Restoring the original, flat appearance of a printed document from casual photographs of bent and wrinkled pages is a common everyday problem.
In this paper we propose a novel method for grid-based single-image document unwarping. Our method performs geometric distortion correction via a fully convolutional deep neural network that learns to predict the 3D grid mesh of the document and the corresponding 2D unwarping grid in a dual-task fashion, implicitly encoding the coupling between the shape of a 3D piece of paper and its 2D image.
In order to allow unwarping models to train on data that is more realistic in appearance than the commonly used synthetic Doc3D dataset, we create and publish our own dataset, called UVDoc, which combines pseudo-photorealistic document images with physically accurate 3D shape and unwarping function annotations.
Our dataset is labeled with all the information necessary to train our unwarping network, without having to engineer separate loss functions that can deal with the lack of ground-truth typically found in \emph{document in the wild} datasets.
We perform an in-depth evaluation that demonstrates that with the inclusion of our novel pseudo-photorealistic dataset, our relatively small network architecture achieves state-of-the-art results on the DocUNet benchmark. We show that the pseudo-photorealistic nature of our UVDoc dataset allows for new and better evaluation methods, such as lighting-corrected MS-SSIM. We provide a novel benchmark dataset that facilitates such evaluations, and propose a metric that quantifies line straightness after unwarping.
Our code, results and UVDoc dataset are available at \url{https://github.com/tanguymagne/UVDoc}.
\end{abstract}



\maketitle


\section{Introduction}
The task of physical document digitization, e.g.\ for financial administration, is increasingly being done in a casual setting with the use of smartphones rather than the more traditional in-office flatbed scanners. However, the appearance of these casually captured images usually differs greatly from flatbed-scans due to varying camera angles, unconstrained illumination conditions and physical deformations of the paper, such as folding, crumpling and curving. These appearance variations pose a problem for post-processing steps, such as optical character recognition (OCR). Document image rectification is therefore an important step in the modernized document digitization pipeline, making layout extraction and OCR performance comparable to that of the traditional pipeline.

Several research efforts have been undertaken to tackle the problem of document unwarping using either model- or data-driven approaches. The model-driven approaches typically try to fit a constrained, piecewise-smooth surface to the imaged document. This geometric optimization is generally slow and has limited approximation capabilities, making it unsuitable for practical applications.
Data-driven approaches instead rely on training a neural network to perform the unwarping. These methods are fast at runtime but {typically} require a large amount of high-quality training data, which {can be} difficult to obtain.
{The available training data can roughly be categorized as either synthetic or \emph{in the wild} document images. The former group is generated by rendering images using 3D scans of real document geometries, whilst the latter simply consists of photographs of actual deformed documents. The synthetic category has the problem that dense 3D capture is often noisy, and rendering photorealistic paper can be challenging, making the appearance of the generated data samples non-realistic as a result. The challenge with the latter category is that ground truth data, most notably the ground truth unwarping function, is difficult to obtain.}

Our main contribution is UVDoc, a dataset that aims to decrease the domain gap between the synthetic Doc3D dataset \cite{DocUNet} usually used to train models for the task of document unwarping, and real document photographs. Our dataset contains 20,000 pseudo-photorealistic images of documents, and is equipped with all the required information to train a coarse grid-based document unwarping neural net. We achieve photorealistic appearance by using image compositing instead of rendering, thereby retaining the shading and material appearance from the original image capture. As our dataset is tailored to a coarse grid-based approach, it is easy to produce even though it includes numerous ground-truth annotations.
We offer a new benchmark dataset whose rich ground-truth annotations allow for evaluation of the unwarping performance without the entanglement of shading artifacts, as well as a new metric that measures the straightness of lines in the unwarped image.

We train a small deep convolutional neural network that performs document image unwarping from a single RGB image. Its design is chosen specifically to make use of our UVDoc dataset. It uses a dual-head approach to predict both a 3D grid mesh representing the 3D shape of the document, as well as a 2D unwarping grid representing a coarse backward map. The backward mapping acts as an inverse parameterization; it indicates at each output pixel, which pixel coordinates should be sampled from the input image to produce the unwarped image.
This dual-task approach encodes an implicit coupling between the 2D and the 3D grid, just like there is a physical coupling between the 3D document shape and its 2D image. Since we learn a coarse 2D unwarping grid instead of a dense unwarping flow, our network size is greatly reduced compared to state-of-the-art methods. 

Using our own relatively small model, and training on a combination of the large Doc3D synthetic dataset and our own custom UVDoc data, we obtain state-of-the-art performance on the DocUNet benchmark for most evaluation criteria. Moreover, we show that the addition of our UVDoc dataset improves the performance of existing document unwarping methods.
\section{Related Work}
\label{sec:related_work}
Document image unwarping is a widely studied topic
We divide previous work into two categories: model-based and data-driven approaches.

\subsection{Model-based document unwarping}
\label{sec:geomrect}
Early works take a geometric modeling approach and try to unwarp document images by first creating a 3D reconstruction of the document surface, which is then flattened onto the plane by solving an optimization problem. 
These works commonly obtain an estimate of the 3D document surface with the help of auxiliary equipment, such as structured light \cite{Brown2001, Brown2004}, two structured laser beams \cite{Meng2014} or laser range scanners \cite{Zhang2008}. 
Other model-based methods use multi-view images instead of hardware to estimate the 3D shape of the document surface \cite{Yamashita2004, Ulges2004, Tsoi2007, Koo2009, You2018, Luo2022}.
Finally, \citet{tian2011rectification} exploit the structure of the document, such as lines, to reconstruct the 3D geometry through a shape-from-texture method.

Once the 3D reconstruction of the document surface is in place, different methods are used to flatten it to the plane. Brown and Seales ~\shortcite{Brown2001, Brown2004} and Zhang et al.~\shortcite{Zhang2008} flatten the document surface using a simulation of a stiff mass-spring system falling down to a plane under gravity. Another common technique is to fit a (piecewise-)smooth parametric surface to the estimated 3D document surface and flatten it according to a  parameterization. This approach can involve generalized cylinders \cite{Koo2009, Zhang2004, Kim2015, Kil2017, Meng2018VF, Nachappa2022}, generalized ruled surfaces \cite{Tsoi2007, Meng2014}, smooth developable surfaces \cite{Liang2005, Liang2008} and NURBS \cite{Yamashita2004, Zhang2005}. 

Parametric approaches often heavily rely on the texture flow of the text lines in the document to estimate parametric line directions, making them less suitable for documents that only contain sparse text. 
Additionally, their optimization-based nature makes these methods slow and unsuitable for real-time applications, and their dependence on auxiliary equipment makes their use in real-world scenarios inconvenient and costly.

\subsection{Data-driven document unwarping}
\label{secref:datarect}

Data-driven document unwarping methods work directly on a single RGB image of a document, employing deep learning to infer a 2D displacement fields or a coarse grid that can be used to unwarp the distorted input image. Ma et al.~\shortcite{DocUNet} are one of the first to propose such a network, using two chained U-Nets to predict the forward mapping (the first estimates an initial guess, and the second refines it). DewarpNet \cite{DewarpNet} also employs two chained networks inferring first the 3D coordinates and then the backward mapping from the 3D coordinates. They create the large synthetic Doc3D dataset with rich annotations to make this possible. Xu et al.~\shortcite{siamese} build on this approach, but use siamese losses to additionally encourage deformation and texture consistency.

Several other ideas have been implemented. Patch-based methods \cite{DRIC, PieceWise_unwarping} predict the displacement field independently on different parts of the image, thus better handling local distortions at the cost of having to properly stitch the different patches together. Iterative methods \cite{Marior, DocScanner} progressively refine the predicted warping flow field and predict a foreground segmentation mask before starting the iterative rectification process, removing the burden of localizing the document boundaries from the unwarping network. Several methods based on textlines also use foreground segmentation. Jiang et al.~\shortcite{RDGR} use these pieces of information as explicit constraints of an optimization problem. Feng et al.~\shortcite{DocGeoNet} feed a concatenation of textlines and 3D shape features into a network that predicts the displacement map. Recent work by Das et al.~\shortcite{Das2022} learns a texture parameterization for neural representations through differentiable rendering, using multi-view input in a data-driven approach.

A variety of different network architectures have been proposed, such as fully convolutional neural networks \cite{FullyCNN}, pyramid encoder-decoder networks \cite{AGUN}, or transformers \cite{DocTr}. Other works use transformer architectures to tackle specific use cases, such as partially visible documents \cite{DocTr++} or invoices \cite{Hertlein2023}. 

Recently works by Xie et al.~\shortcite{DDControlPoints}, Xue et al.~\shortcite{FDRNet} and Ma et al.~\shortcite{PaperEdge} follow the approach of predicting a coarse backward mapping.
Some of these are capable of learning from images captured \emph{in the wild}, either by direct comparison of the Fourier-filtered unwarped images \cite{FDRNet}, or by designing a specific loss on pairs of slightly perturbed images \cite{PaperEdge}.

Our dual-task-based network architecture enforces the model to predict physically plausible shapes and unwarping grids. It processes input images in a single stage without any segmentation pre-processing and predicts a coarse backward mapping rather than a dense displacement field, making it very efficient.

\subsubsection{Datasets}
The datasets used for training the methods mentioned above can be split into two categories; real and synthetic, with the latter being most commonly used. Earlier works \cite{DocUNet, FullyCNN, DDControlPoints} use synthetic datasets generated based on non-physically plausible 2D deformations. More accurate are datasets based on 3D deformations \cite{DRIC, DewarpNet}, such as Doc3D. Most recent works are trained using Doc3D or its variations with richer annotations \cite{DocGeoNet}, or cropped images \cite{DocTr++}. Even though the 3D shapes in Doc3D are more realistic than those in \cite{DocUNet}, since they are based on depth captures of actual deformed papers, they are heavily smoothed compared to the original document shapes. The rendered appearance is also not very photorealistic, which causes performance degradation when using the network on actual photographs. In contrast, our UVDoc dataset, which is made from real, captured sheets of paper, is more realistic both visually and geometrically.

Datasets of real photographs of deformed documents, paired with their flatbed scans ground truth, have recently gathered increased interest. Some of these datasets provide segmentation information \citep{PaperEdge} but most do not come with further annotations \cite{FDRNet}. These datasets are closer in appearance to real document images, but since they are equipped with very few annotations, they require the design of custom loss functions to train models. In comparison, our dataset is equipped with a lot of annotations while being visually similar to \textit{in-the-wild} data. See comparative summary in \tabref{tab:dataset_comparison}.


\section{The UVDoc dataset}
\label{sec:dataset}
\emergencystretch 3em
We create our own dataset, UVDoc, containing 20k pseudo-photorealistic images of warped documents. Our motivation is to obtain a dataset of photorealistic document images that has more ground truth information available than \emph{document in the wild} images, and more realistic appearance than synthetically generated renderings. This allows for a stronger supervision signal than what is available for general \emph{document in the wild} data and benefits from more realistic appearance. We compare the main characteristics of our dataset against other available datasets in \tabref{tab:dataset_comparison}.

\begin{table}
    \centering
    \caption{Comparison between the different document unwarping datasets. The last column indicates whether the ground-truth Backward Mapping (BM) between the distorted and the unwarped document is available.}
    \vspace*{-2mm}
    \begin{tabular}{l r l c}
    \toprule
      Dataset & \# Samples & Type & BM \\ \midrule
      Doc3D \cite{DewarpNet} & 100,000 & Synthetic & \cmark \\
      DIW \cite{PaperEdge} & 5,000 & Real & \xmark \\
      WarpDoc \cite{FDRNet} & 1,020 & Real & \xmark\ \\
      Ours & 20,000 & Pseudo-real & \cmark \\\bottomrule
    \end{tabular}
    \label{tab:dataset_comparison}
\end{table}

 \definecolor{one}{HTML}{2b9eb3}
\definecolor{two}{HTML}{f8333c}
\definecolor{three}{HTML}{fcab10}
\definecolor{four}{HTML}{44af69}

\begin{figure}
    \centering
    \begin{subfigure}[t]{\linewidth}
        \centering
        \includegraphics[width=\textwidth]{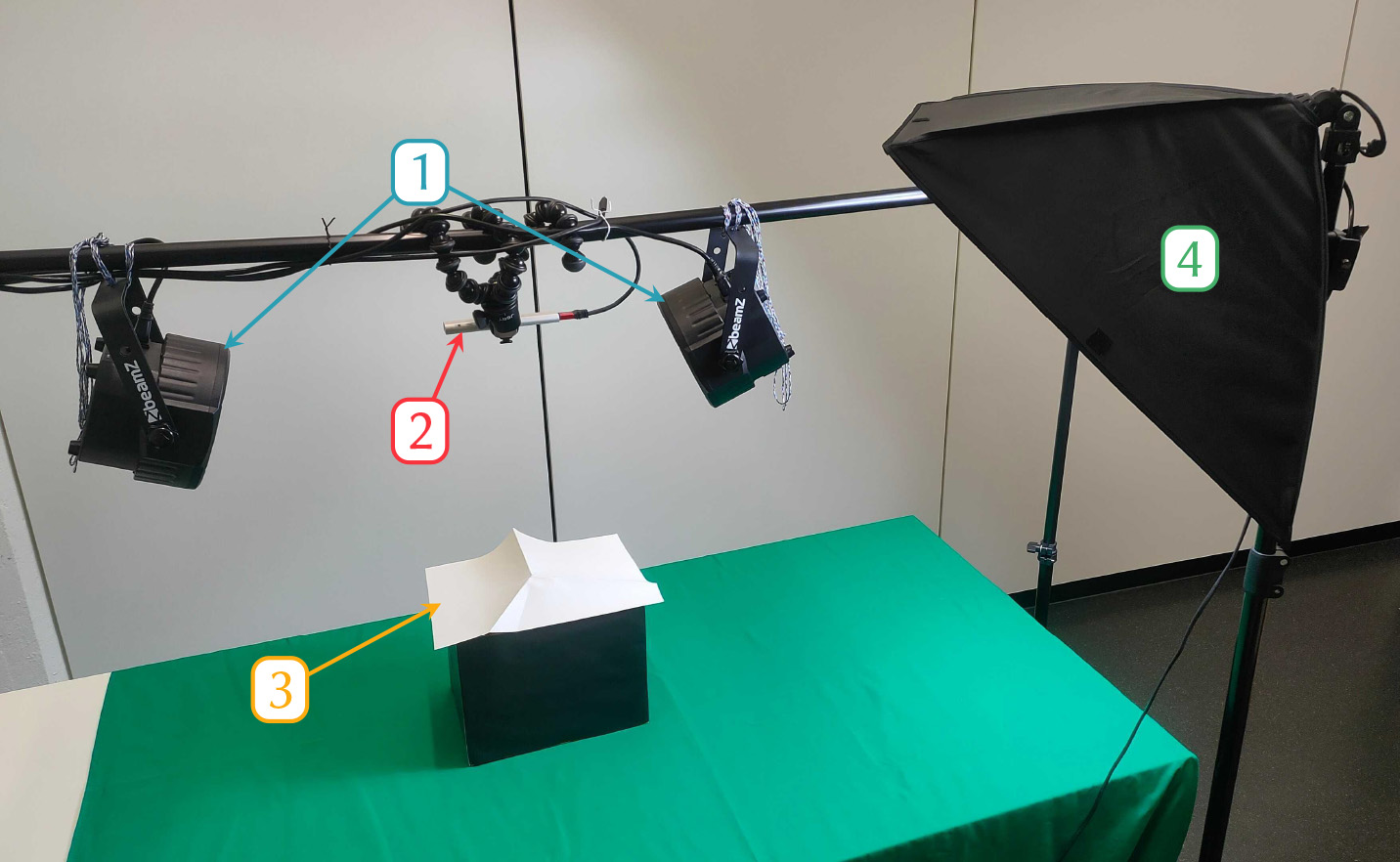}
    \end{subfigure}
    \par\medskip
    \begin{subfigure}[t]{\linewidth}
        \centering
        \begin{subfigure}[t]{0.32\textwidth}
            \centering
            \includegraphics[width=\textwidth]{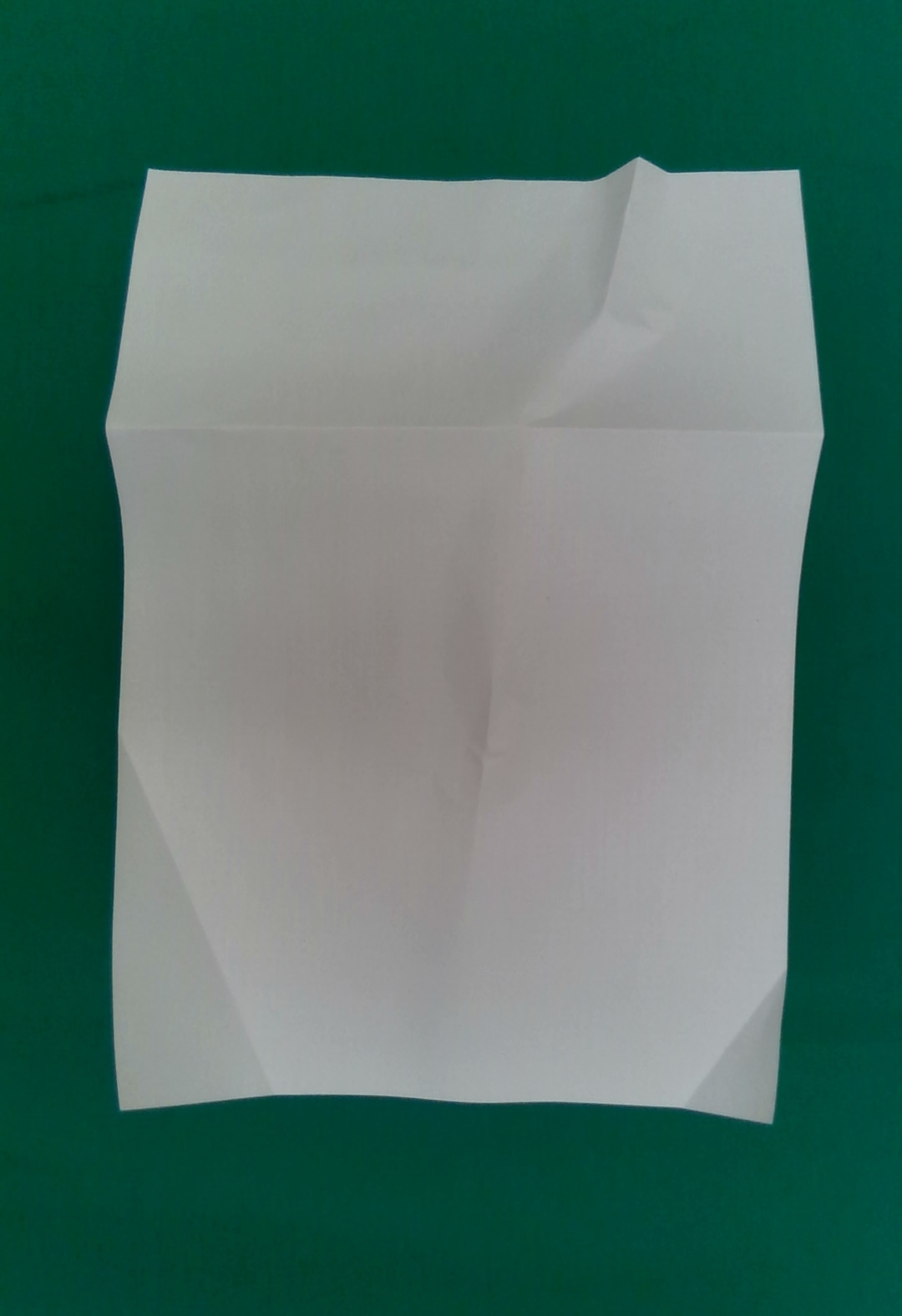}\\
            \vspace*{-2mm}
            \caption*{\small normal light}
        \end{subfigure}
        \hfill
        \begin{subfigure}[t]{0.32\textwidth}
            \centering
             \includegraphics[width=\textwidth]{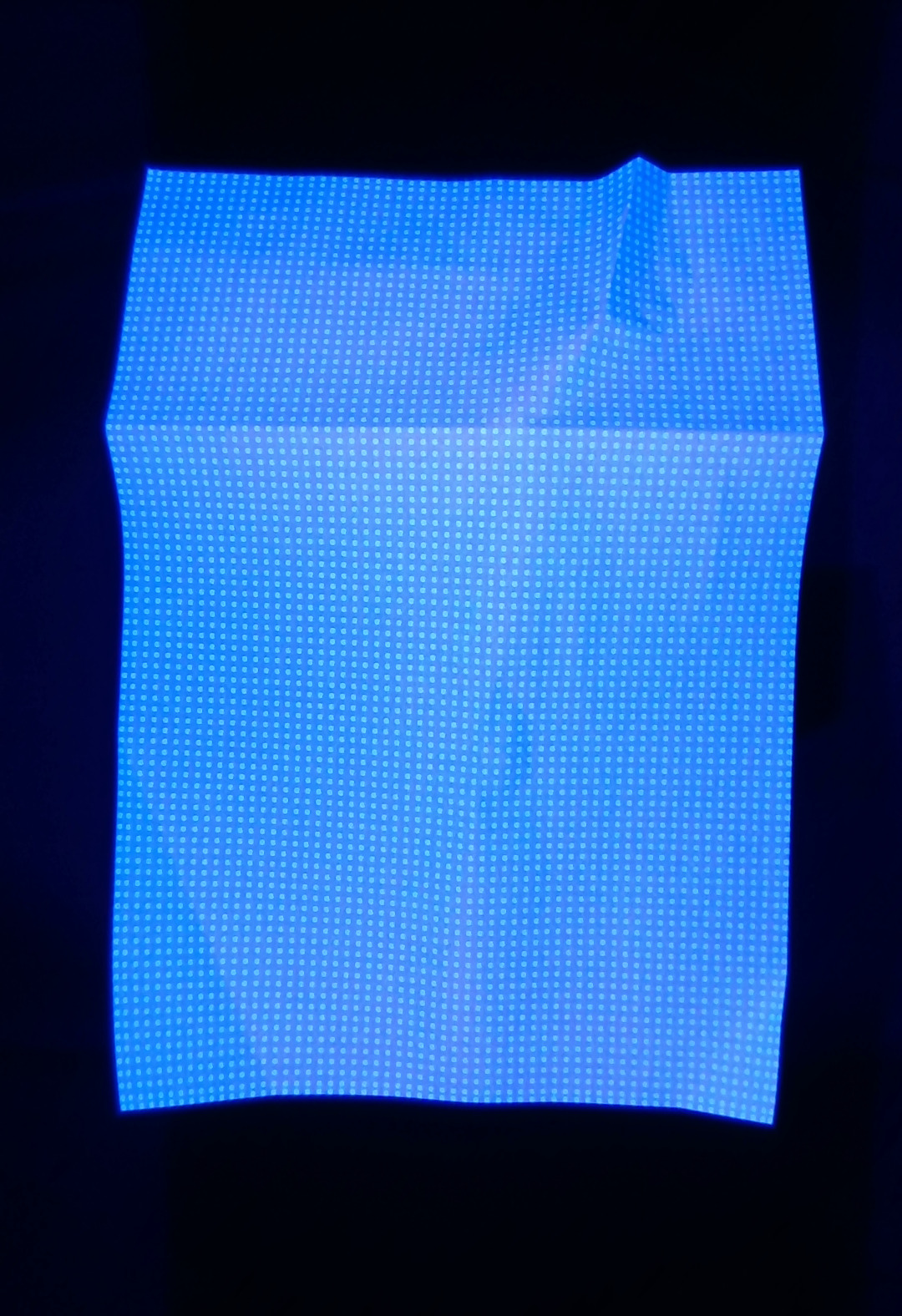}\\
            \vspace*{-2mm}
            \caption*{\small UV light}
        \end{subfigure}
        \hfill
        \begin{subfigure}[t]{0.32\textwidth}
            \centering
            \includegraphics[width=\textwidth]{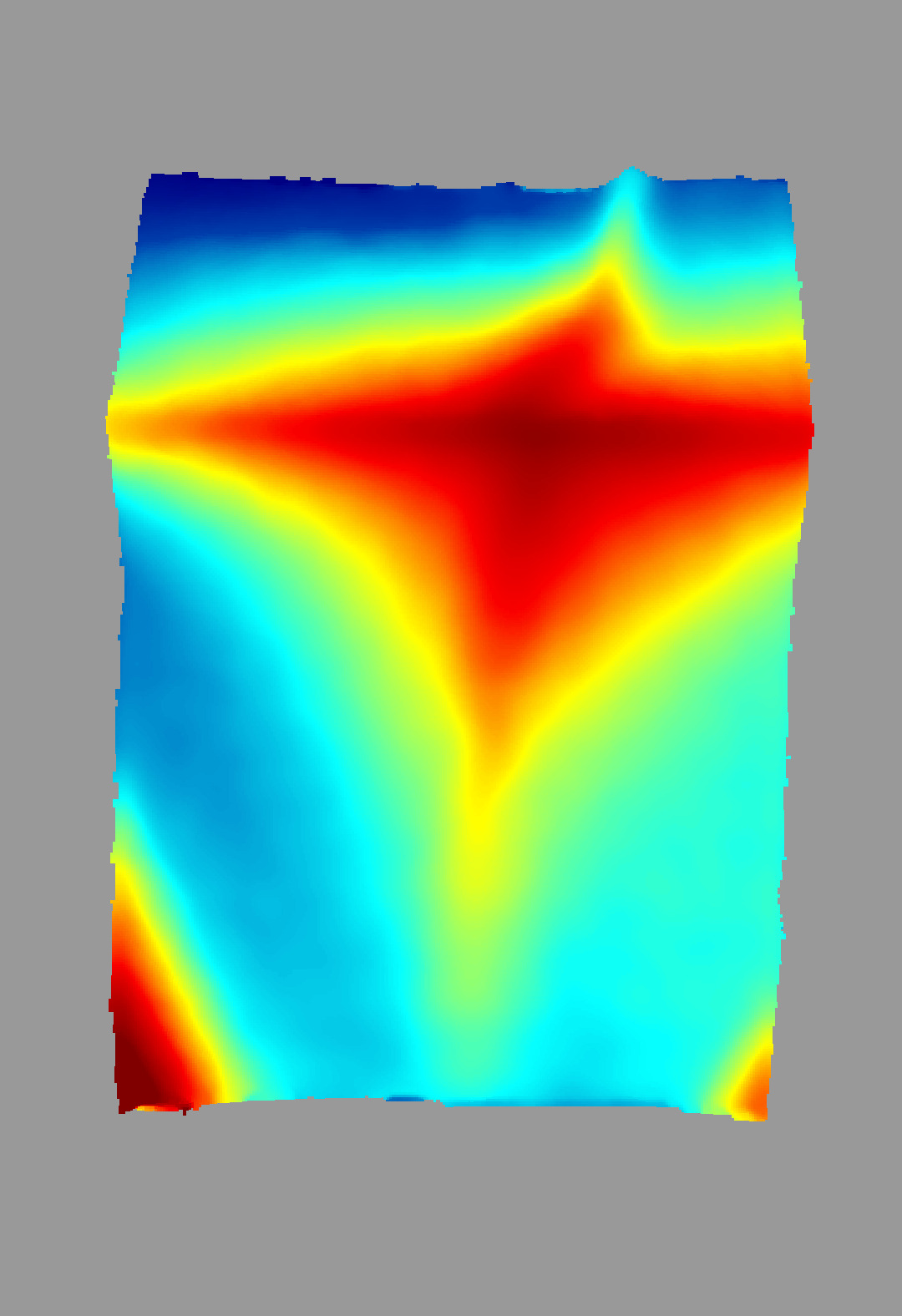}\\
            \vspace*{-2mm}
            \caption*{\small depth image}
        \end{subfigure}
    \end{subfigure}
    \vspace*{-3mm}
    \caption{An overview of our data capture setup and sample data acquired in the process. The top shows our capture setup: \textcolor{one}{[1]} UV lights, \textcolor{two}{[2]} SR305 depth camera, \textcolor{three}{[3]} deformed sheet of paper, \textcolor{four}{[4]} regular light. The bottom shows a capture sample including RGB images of the normally lit and UV-lit paper, and its depth image.}\label{fig:ourdatasetsamples}
\end{figure}

\paragraph{Capture}
We print regular grids of dots, with grid size of $89\times61$, on A4-sized pieces of paper using an inkjet printer with UV ink that is invisible to an RGB camera in regular light, but becomes visible in UV light in an otherwise dark room. We opt for this grid aspect ratio to obtain an equally spaced grid in both horizontal and vertical direction, and to approximate the aspect ratio of A4 paper in portrait mode, the most common paper type that documents are printed on. {Note that on the paper boundary, we deviate slightly from a perfectly regular grid by offsetting the border dots a little, so that they fully fit on the paper and can be detected more easily.} We fold and bend the pieces of paper in various ways to emulate common deformations.
We then capture pairs of RGB-D images of deformed papers using the Intel RealSense SR305 depth camera: one image in regular lighting and one in UV lighting (\figref{fig:ourdatasetsamples}). We use two commercially available \unit[30]{W}, \unit[395]{nm} UV lamps and one \unit[72]{W}, \unit[395]{nm} UV lamp to reduce the amount of shadows in the UV-lit image. We also use a regular light with adjustable color temperature and brightness to create varying lighting conditions. We control the camera and the lights using a laptop and remote switches, so that there is no movement between the two captured frames, and the depth and pixel information is aligned. We capture various types of deformed paper, such as curved, folded, and crumpled, and we also vary the lighting conditions. The dataset contains a total of 1008 distinct geometries, which we augment to 4032 geometries by applying horizontal and/or vertical flips to each sample.

\paragraph{Recovering the grid} 
Using the UV-lit image, where the printed grid is visible, we obtain the pixel coordinates of the grid points on the deformed piece of paper. To detect them we use OpenCV's \texttt{SimpleBlobDetector}, coupled with manual annotation for extreme cases where the automatic detection fails (less than 0.5\% of the points need to be manually annotated). 
Once all points have been detected, we compute their correspondences to the vertices of a regular grid, which is equivalent to ordering them as an $89\times61$ grid. The technical details of solving this ordering problem are described in the supplementary material. 

We call the ordered grid the 2D unwarping grid. Combining the coordinates of the 2D unwarping grid with the depth values at these same pixel coordinates and the intrinsics of the camera, we construct a 3D grid mesh corresponding to the 3D shape of the piece of paper. 

\begin{figure}[t]
    \centering
         \includegraphics[width=\linewidth]{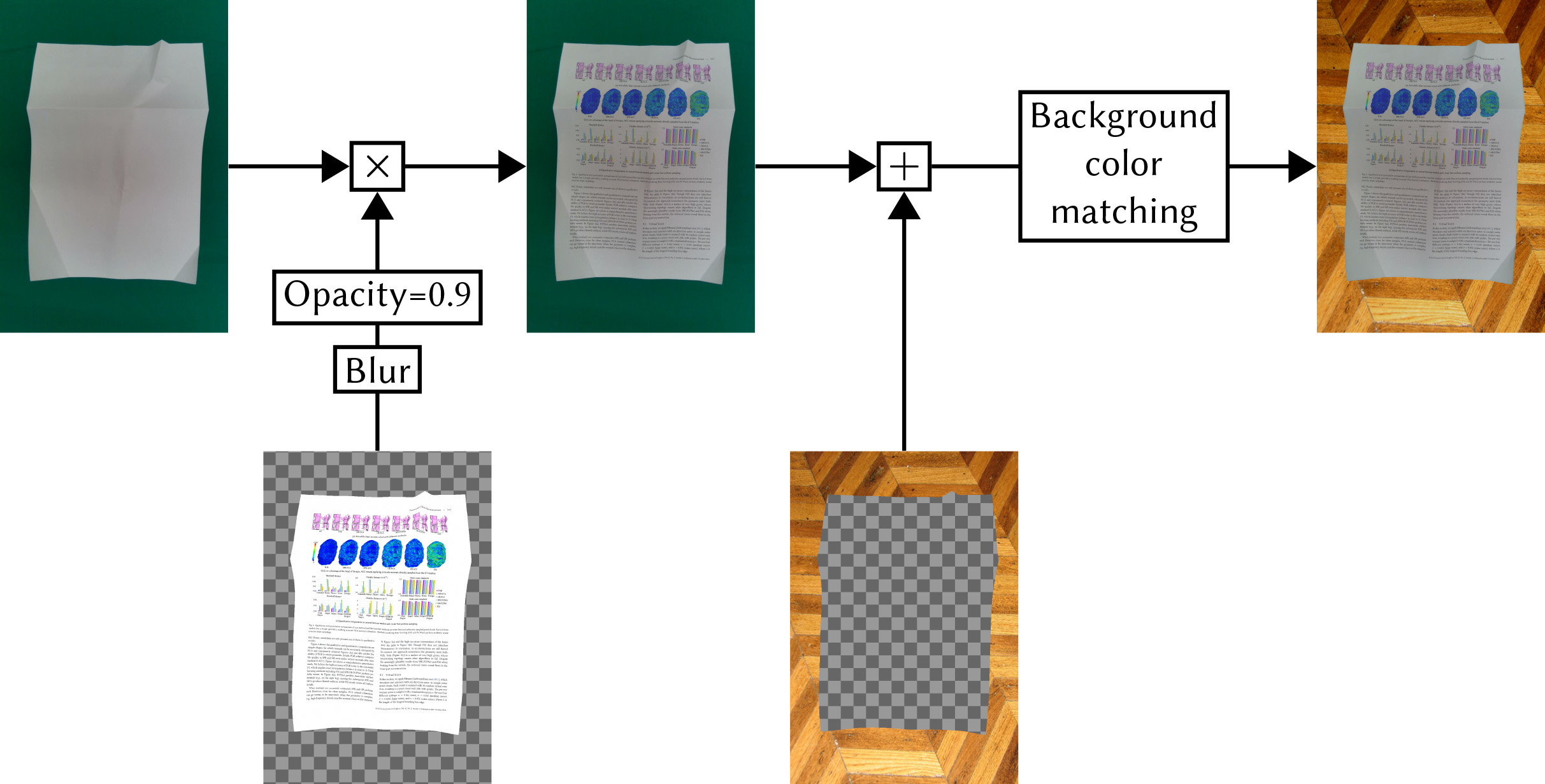}
        \vspace*{-3mm}
    \caption{The pipeline used to create a sample of our UVDoc dataset. It combines the captured image of a blank paper, the texture and the background.}\label{fig:samplepipeline}
\end{figure}

\paragraph{Pseudo-photorealistic image generation} 
 Since we have a known mapping between the 2D unwarping grid and the original regular 2D grid, we can construct a coarse $uv$-parameterization of the 3D grid mesh. We use bilinear interpolation for the $uv$-parameterization when applying a texture to the geometry and for the 2D unwarping grid when performing the unwarping, to obtain a full-resolution dense backward mapping. 
 
The $uv$-parameterization is used to apply a document texture on top of the image of the blank warped paper. The document textures include books and scientific articles sampled from the web, as well as other types of documents such as magazines, invoices, and music sheets, generated using a text-to-image model \cite{DeepFloydIF}. As illustrated in \figref{fig:samplepipeline}, we blend the document texture with the lighting-baked blank document image by multiplying the two images. This gives a pseudo-photorealistic combination between the lighting and the texture. We also replace the background in the image with a background sampled from the \emph{Describable Textures} dataset \cite{DTD}. Finally, we apply color correction to match the hue of the background to the hue of the document and we also equalize the brightness of the background to the foreground. Using this approach, we create a dataset of 20,000 images in total.
We provide the original lighting-baked blank document images along with the $uv$-parameterization, so users of the dataset can easily replace the document and the background textures if desired. 

At the end of our data capture pipeline, we are equipped with a ground-truth 2D unwarping grid, a $uv$-parameterization and a 3D grid mesh for each sample in our dataset. Since we use the \emph{physical} $uv$-parameterization recorded via the 2D grid, rather than a parameterization designed by a rendering engine, our texture gets deformed and applied with greater physical accuracy. Additionally, by circumventing a rendering pipeline, our images \emph{look like real paper}, which is hard to simulate when rendering.
The full UVDoc dataset is available at \url{https://github.com/tanguymagne/UVDoc}.


\section{Method}
\label{sec:method}

To completely unwarp a document, we assume that the input photograph is taken from a camera position in which the document's 3D shape can be represented as a height field, i.e., the entire document is visible and there are no occlusions and foldovers.

\begin{figure}[t]
	\centering
	\begin{tabular}{c}
		\includegraphics[width=0.9\linewidth]{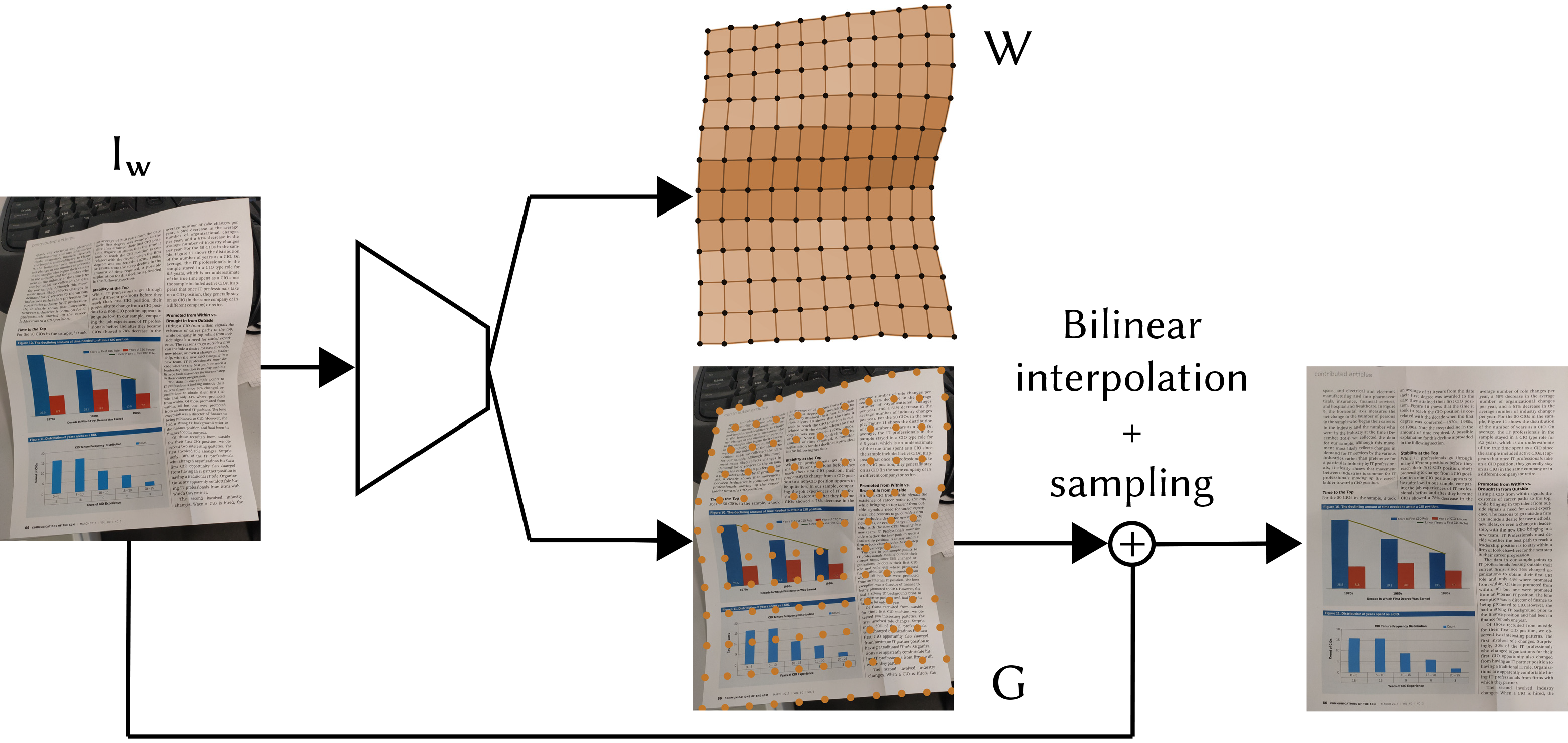}
    \end{tabular}
	\caption{Our unwarping pipeline. We start with an RGB image of a warped document and feed it into our encoder-style network. The network predicts both a 3D grid mesh (top branch), as well as a 2D unwarping grid (bottom branch) in parallel. The 2D unwarping grid is then bilinearly interpolated to the desired output image resolution and is used to sample pixels from the input image to obtain the final unwarped document image.
 \label{fig:pipeline}}
\end{figure}

We use a dual-head network to predict a $45 \times 31$ 2D unwarping grid $G$ containing pixel coordinates, and a $45 \times 31$ grid mesh of 3D shape coordinates $W$ from a warped 488-by-712 input image $I_w$. We do not predict $G$ at the full ground-truth resolution in an attempt to keep our network as compact as possible.
As illustrated in \figref{fig:pipeline}, the 2D unwarping grid $G$ encodes the deformation that leads to the unwarped document: grid-point $G_{i,j}$ holds the pixel coordinates (relative to the image, in the range $[-1,1]$) of the pixel that will be placed at position $(i,j)$ in the unwarped image (up to constant scaling). The grid $G$ can also be seen as a coarse backward mapping. Finally, $G$ is  bilinearly interpolated to the original image size. This upsampled backward mapping is used to generate the full-resolution unwarped image. The 3D grid mesh $W$ is not used for the unwarping, but we incorporate it in training with an $L_1$ loss as a regularization term. This helps the network understand the underlying geometry of the document and improve the unwarping performance (see ablations studies in \secref{subsec:ablation} and \tabref{tab:ablation}).

\subsection{Network architecture}
We use a relatively straightforward dual-head, fully convolutional encoder architecture inspired by the encoder part of the architecture used in \cite{FullyCNN}. The input image goes through two convolutional downsampling layers that each use a $5\times5$ kernel and reduce the image size by a factor of two. This is followed by three dilated residual blocks, which lead to a spatial pyramid with stacked dilated convolutions. Finally, two heads with two convolutional layers predict $G$ and $W$, respectively. We give a detailed graphical overview of our architecture in the supplemental.

\subsection{Training loss}
We denote the ground-truth variables as their regular symbols (e.g., $G$) and their predicted counterparts with a hat (e.g., $\Ghat$). Our training loss is a combination of $L_1$ losses on both the 2D unwarping grid $G$ and the 3D grid mesh $W$, as well as an image reconstruction loss $\mathcal{L}_r$:
\begin{equation}\label{eq:training-loss}
   \mathcal{L}   =  \alpha \|G-\Ghat\|_1  + \beta\|W-\What\|_1 + \gamma\mathcal{L}_r,
\end{equation}
where $\alpha, \beta, \gamma$ are weights used to balance the influence of the individual loss terms.

$\mathcal{L}_r$ is an $L_1$ loss between the ground truth unwarped image and the image unwarped using the predicted unwarping grid $G$. For Doc3D samples, the reconstruction loss is computed directly on the unwarping of the input image $I_w$, which includes shading. For our UVDoc samples, to allow the network to focus on the content of the document rather than on shading artifacts, we compute the reconstruction loss using the unwarping of the unshaded document and the ground truth document texture.
We provide further training details for our method in the supplementary material.
\definecolor{best}{HTML}{D6AF36}
\definecolor{second}{HTML}{A7A7AD}
\definecolor{third}{HTML}{824A02}
\newcommand{\best}[1]{\textbf{#1}}
\newcommand{\second}[1]{\underline{#1}}
\newcommand{\third}[1]{\textit{#1}}

\section{Evaluation metrics and the UVDoc benchmark}

We discuss common existing evaluation metrics for document unwarping and propose a new benchmark (UVDoc), along with a new metric that provides faithful evaluation even in the presence of varied shading.

\begin{figure}
    \centering
    \begin{subfigure}[t]{\linewidth}
        \centering
        \begin{subfigure}[t]{0.49\textwidth}
            \centering
            \includegraphics[width=0.49\textwidth]{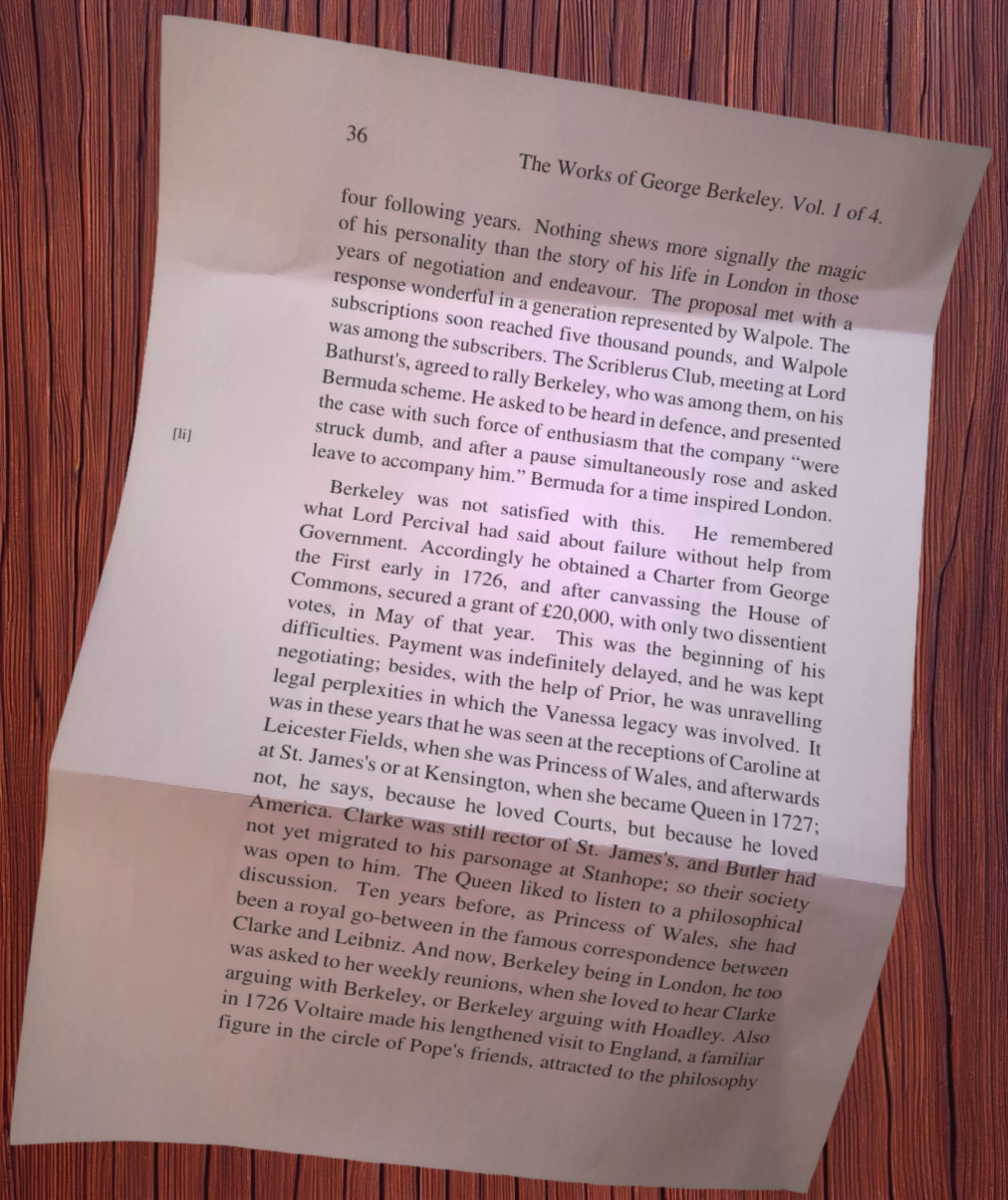}
            \includegraphics[width=0.49\textwidth]{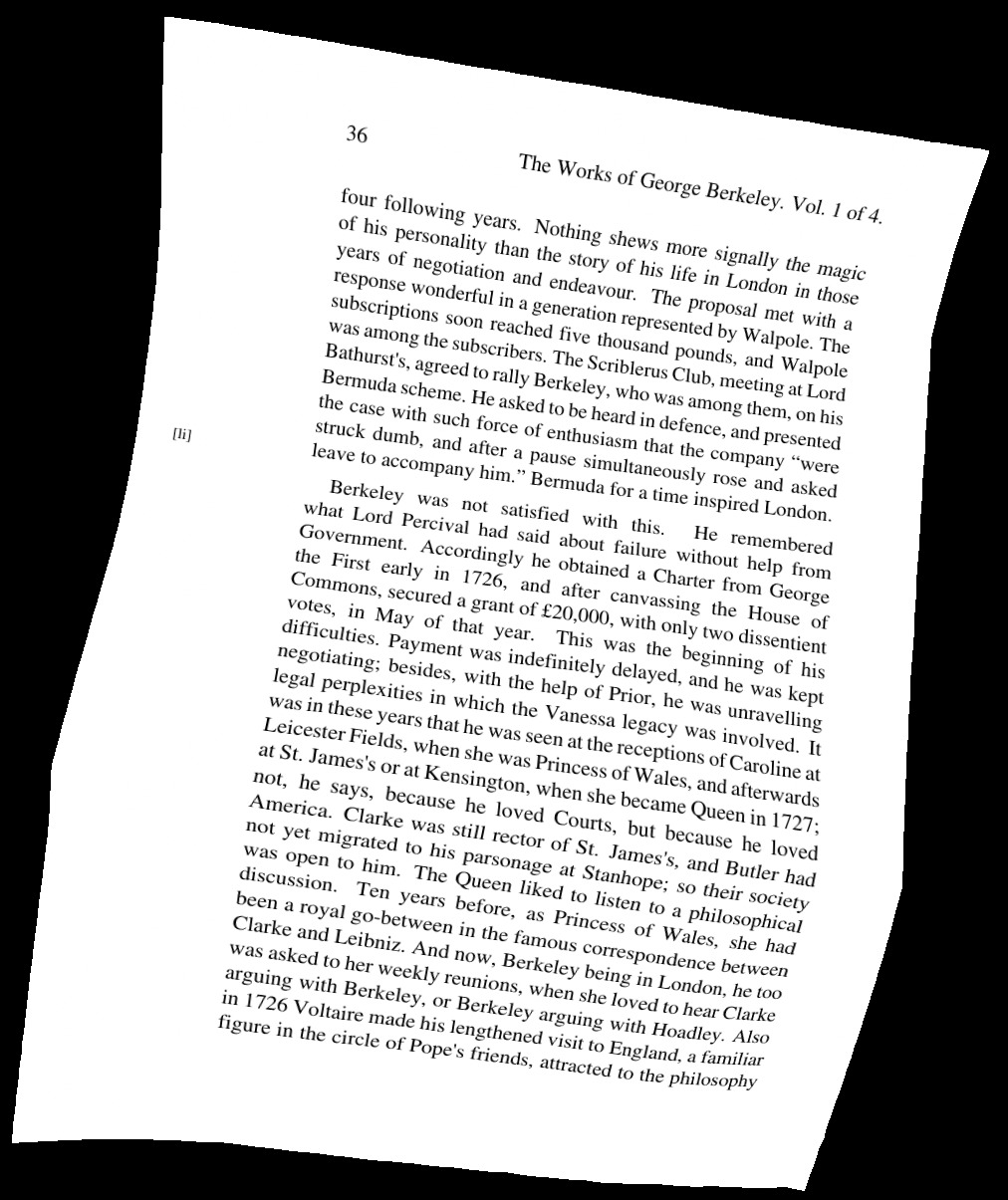}
            \caption*{Warped document \\ and texture}
        \end{subfigure}
        \begin{subfigure}[t]{0.49\textwidth}
            \centering
            \frame{\includegraphics[width=0.49\textwidth]{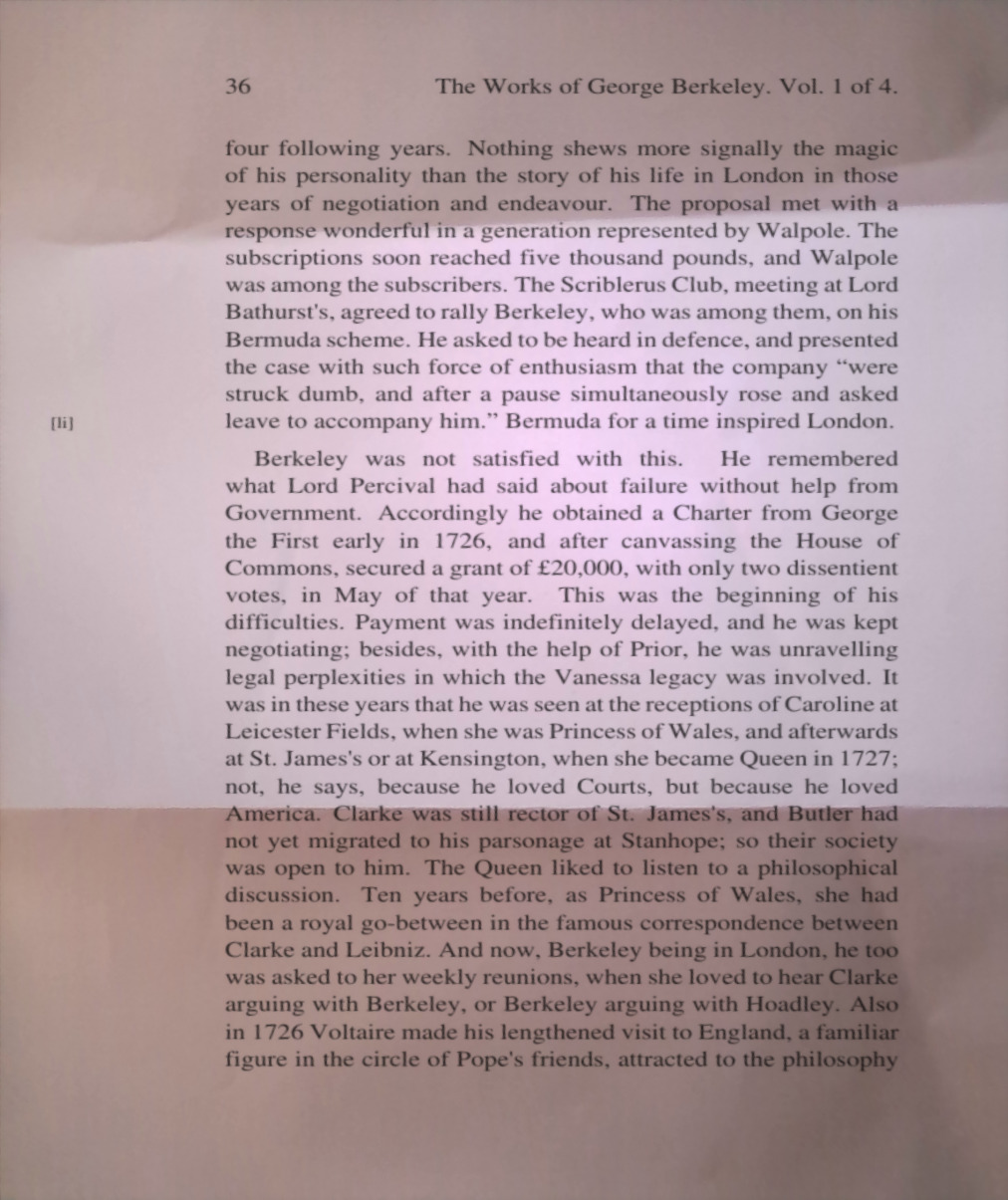}}
            \frame{\includegraphics[width=0.49\textwidth]{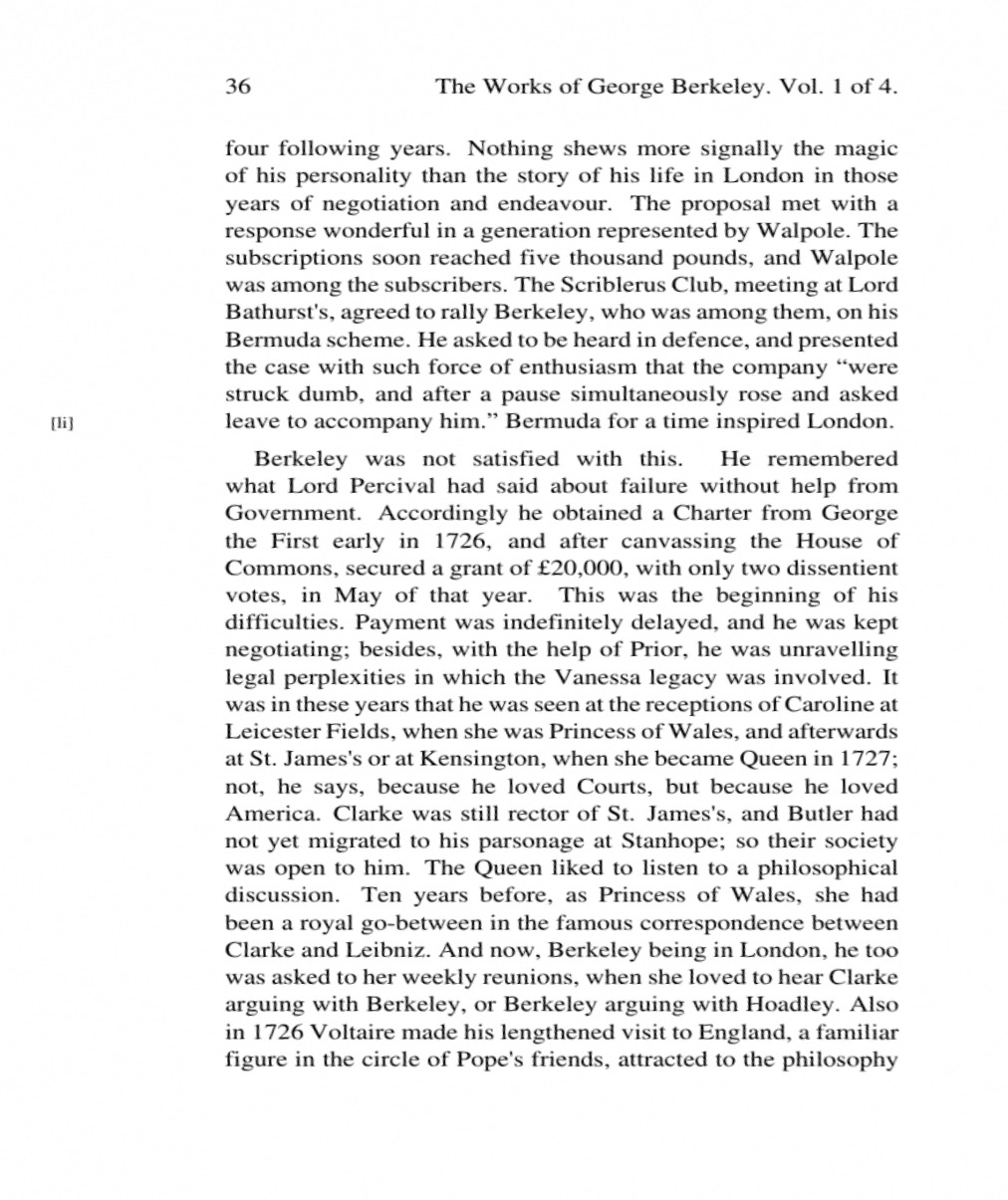}}
            \caption*{Unwarped document \\ and texture}
        \end{subfigure}
    \end{subfigure}
    \vspace{-7pt}
    \caption{The shaded and unshaded version of a sample from the UVDoc benchmark, identical up to shading. The shaded version and unshaded version have a CER of 0.439 and 0.004 respectively and ED of 959 and 14. Note that the unshaded version has non-zero CER and ED as it is compared to the original texture, while it has been warped and unwarped using our coarse bilinearly interpolated 2D grid, and thus includes some artifacts.}\label{fig:shadingOCR}
\end{figure}

\subsection{UVDoc benchmark}
\label{sec:UVDocbenchmark}
To foster more detailed evaluation of document unwarping methods in the future, we create the UVDoc benchmark dataset. This benchmark is generated in a similar fashion to the UVDoc dataset but contains other geometries, document textures and backgrounds, not included in the main dataset. The benchmark consists of 50 images. Thanks to our pseudo-photorealistic data generation pipeline, we have access to pairs of warped images with and without lighting (and thus shadows) baked in, see \figref{fig:shadingOCR}. This setup provides new opportunities for meaningful metrics. For each sample in the benchmark, an unwarping pipeline can predict the unwarping function for the shaded image, and then apply it to the unshaded image. The unwarped unshaded image can then be compared to the ground-truth original image texture. This way the effect of illumination can be removed and the unwarping deformation can be better evaluated. 

\subsection{Evaluation metrics}
In our objective evaluations, we employ image similarity metrics as well as optical character recognition (OCR) performance. Following \cite{DocUNet} and \cite{DewarpNet}, we use multi-scale structural similarity (MS-SSIM) and local distortion (LD) as metrics for the image similarity evaluation. We also employ the aligned distortion (AD) metric (introduced by \cite{PaperEdge}), which corrects some of the flaws of the previous metrics. We evaluate OCR performance using the character error rate (CER) and edit distance (ED).

As OCR engines are typically targeted towards use with images originating from flatbed scanners, they are ill-suited for text recognition on images with lighting variations and shadows \cite{tesseract}. The two images in \figref{fig:shadingOCR} are identical up to shading but result in vastly different OCR performances. We therefore want to point out that OCR performance on the DocUNet dataset should be interpreted with care, since its baked-in shading plays such a large role.
More details regarding our evaluation metrics can be found in the supplementary material.

\begin{figure}
	\centering
	\begin{subfigure}[t]{0.32\linewidth}
		\centering
		\includegraphics[width=\textwidth, height=1.4\textwidth]{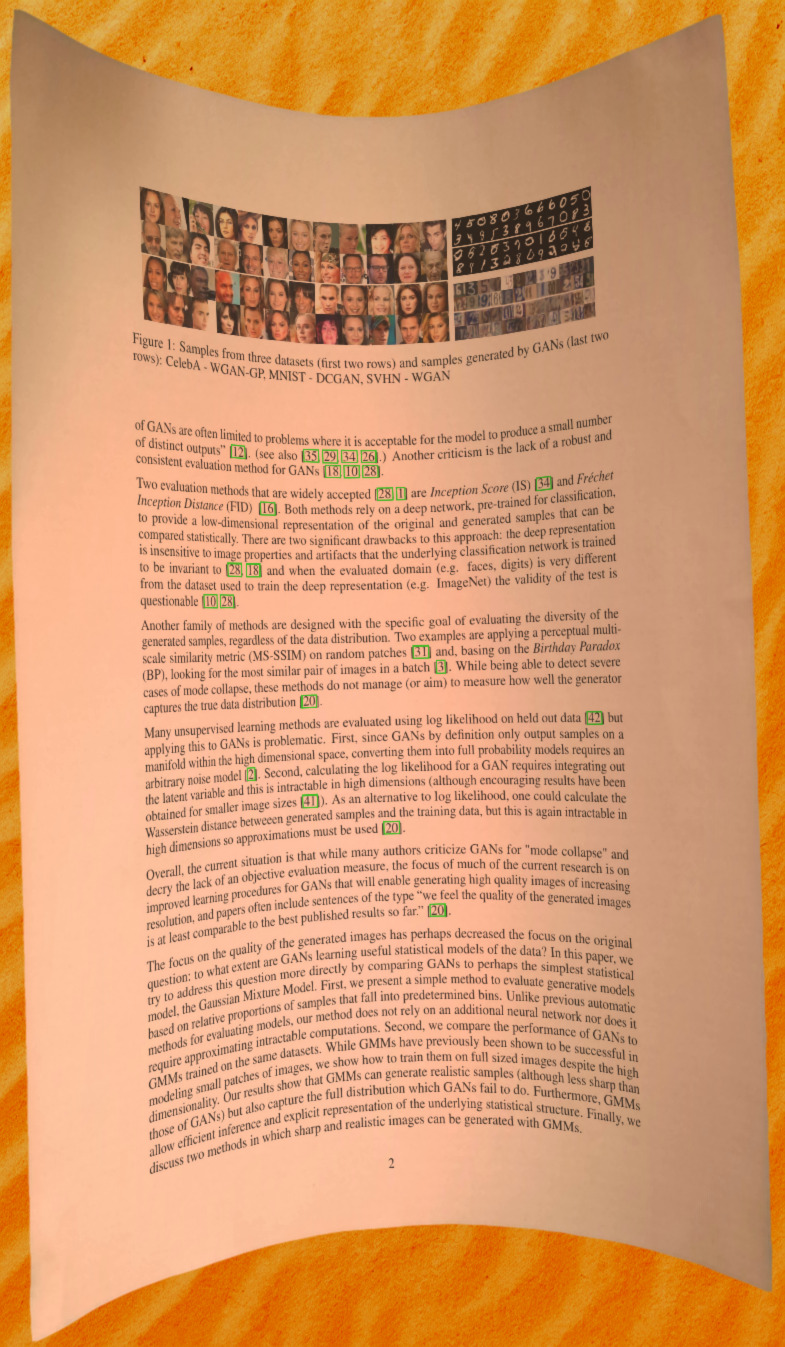}
		\vspace*{-5mm}
		\caption*{\footnotesize \centering Warped document image}
	\end{subfigure}
	\begin{subfigure}[t]{0.32\linewidth}
		\centering
		\includegraphics[width=\textwidth, height=1.4\textwidth]{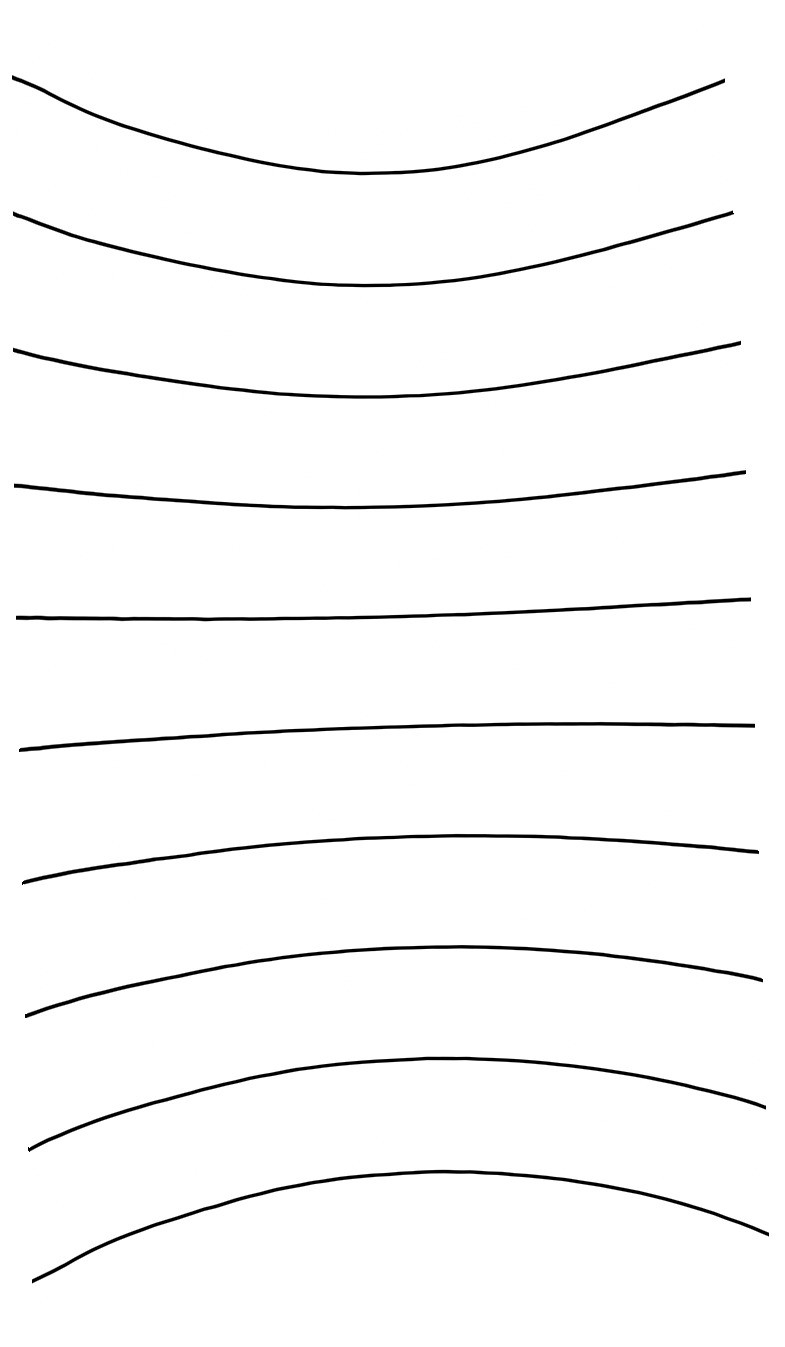}
		\vspace*{-5mm}
		\caption*{\footnotesize \centering Warped horizontal \newline lines}
	\end{subfigure}
	\begin{subfigure}[t]{0.32\linewidth}
		\centering
		\includegraphics[width=\textwidth, height=1.4\textwidth]{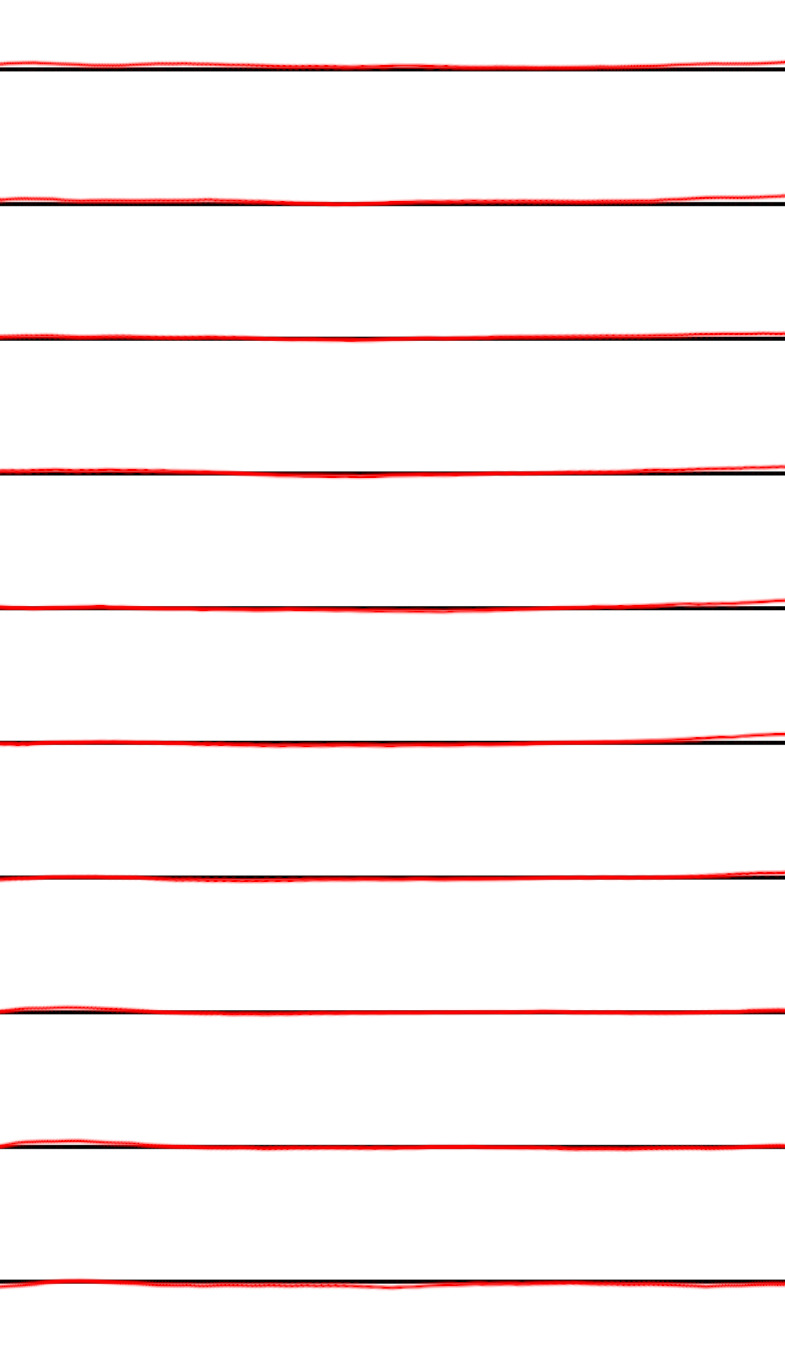}
		\vspace*{-5mm}
		\caption*{\footnotesize \centering Unwarped horizontal lines}
	\end{subfigure}
	\vspace*{-2mm}
	\caption{Our new horizontal line metric is the standard deviation of the $y$ coordinate of warped horizontal lines (middle) unwarped using the predicted backward mapping (right in red, ground-truth in black). }\label{fig:new_metric}
\end{figure}

\subsubsection{Line straightness metric}
\label{sec:linemetric}

Our UVDoc benchmark is annotated with not just a ground-truth unwarped image but also the ground-truth warping function, which allows us to design a new metric that evaluates the straightness of lines in the unwarped image. We generate triplets of images consisting of a warped document image and two images containing warped horizontal and vertical lines. These three images are generated using the same geometry and thus correspond to the same ground-truth unwarping function. We can now predict the unwarping function from the warped document image, and then apply it to the warped line images, giving us the unwarped horizontal and vertical lines. A perfectly predicted unwarping function should map the lines (which are 1 pixel thick) back to exactly horizontal and vertical lines. By measuring the average standard deviation of the lines we obtain a measure of how well the unwarping function maps horizontal (resp.\  vertical) lines to horizontal (resp.\ vertical) lines; see \figref{fig:new_metric} for a visual explanation of the process. Note that unlike other metrics that are commonly used in the document unwarping field to measure distortion (such as LD and AD), this metric does not rely on the usage of dense SIFT flow, which is slow to compute and can give unstable results due to shading artifacts.

\section{Experiments}
\label{sec:experiments}

\subsection{Evaluation}

\begin{table}
\centering
    \caption{Quantitative unwarping performance comparisons on the \hbox{DocUNet} benchmark dataset. \textbf{Bold} font indicates best, \underline{underline} indicates second-best and \textit{italic} indicates third-best score. The last column compares the network sizes, expressed in number of parameters (millions). We compare our results against DewarpNet \cite{DewarpNet}, DispFlow \cite{FullyCNN}, DocTr \cite{DocTr}, PW Unwarping \cite{PieceWise_unwarping}, DDCP \cite{DDControlPoints}, FDRNet \cite{FDRNet}, RDGR \cite{RDGR}, Marior \cite{Marior}, PaperEdge \cite{PaperEdge} and DocGeoNet \cite{DocGeoNet}.}
    \vspace*{-2mm}
    \resizebox{\linewidth}{!}{%
    \begin{tabular}{l c c c c c c c c c}
    \toprule
    Method & MS-SSIM $\uparrow$ & LD $\downarrow$ & AD $\downarrow$ &  CER $\downarrow$ & ED $\downarrow$ & Para. \\
    \midrule
    DewarpNet & 0.472 & 8.38 & 0.396 & 0.217 & 834 & 86.9M \\ 
    DispFlow & 0.432 & \third{7.62} & 0.396 & 0.292 &  1216 & 23.6M \\ 
    DocTr & \third{0.509} & 7.78 &  \second{0.366} & \third{0.181} & \second{712} & 26.9M \\ 
    PW Unwarping & 0.490 & 8.65 & 0.431 & 0.252  & 987 & -\\
    DDCP & 0.473 & 8.93 & 0.423  & 0.278 & 1118 & 13.3M \\ 
    FDRNet & \second{0.543} & 8.08 & 0.396  & 0.214 & 878 & -\\
    RDGR & 0.495 & 8.50 & 0.432 & \best{0.171} & \third{732} & -\\
    Marior & 0.476 & \second{7.37} & 0.404 &0.200 & 797 & -\\
    PaperEdge & 0.472 & 7.98 & \third{0.367} & 0.193 & 763 & 36.6M \\ 
    DocGeoNet & 0.504 & 7.70 & 0.378 & 0.190 & 736 & 24.8M \\ \bottomrule 
    Ours & \best{0.544} & \best{6.83} & \best{0.315} & \second{0.172} & \best{707} & 8M 
    \end{tabular}}
\label{tab:comparisontable}
\end{table}

\begin{table}
\centering
    \caption{Quantitative unwarping performance comparisons on our \hbox{UVDoc} benchmark dataset. \textbf{Bold} font indicates best, \underline{underline} indicates second-best and \textit{italic} indicates third-best score. Refer to \tabref{tab:comparisontable} for the list of referenced methods.}
    \vspace*{-2mm}
    \resizebox{\linewidth}{!}{%
    \begin{tabular}{l c c c c c c c c c c}
    \toprule
    Method & MS-SSIM $\uparrow$ &  AD $\downarrow$ & CER $\downarrow$ & ED $\downarrow$ & H-line $\downarrow$ & V-line $\downarrow$ \\
    \midrule
    DewarpNet & 0.600 & 0.189 & 0.115 & 338 & 3.22 & 4.32 \\ 
    DocTr & \third{0.684} & \third{0.176} & \best{0.065} & \second{192} & \third{2.42} & \second{3.32} \\
    DDCP & 0.591 & 0.334 &  0.117 & 362 & 4.20 & 4.88  \\
    RDGR & 0.603 & 0.314 & \best{0.065} &  \best{190} & 4.03 & 5.87 \\
    DocGeoNet & \second{0.714} & \second{0.167} & \second{0.066} & \third{196} & \second{2.24} & \third{3.91}  \\ \bottomrule
    Ours & \best{0.784} & \best{0.122} & \third{0.072} & 202 & \best{1.82} & \best{2.48} 
    \end{tabular}}
\label{tab:comparisontableUVDoc}
\end{table}

We evaluate our network on the DocUNet benchmark dataset \cite{DocUNet} as well as on our own UVDoc benchmark, described in \secref{sec:UVDocbenchmark}. The DocUNet benchmark is composed of 65 documents. For each of them, 2 deformed images in a real-world scenario are provided. The ground truth flatbed-scans are also provided for comparison. Note that similarly to Feng et al.~\shortcite{DocGeoNet} we exclude the two images of document 64, as the real world images are rotated by 180 degrees. We also exclude this document when computing the quantitative results for previous works.

\paragraph{Quantitative evaluation}
We compare our method with several state-of-the-art deep learning methods. For each of them, we use the DocUNet result images published by the authors. We also evaluate the methods that additionally published their pre-trained models on our UVDoc benchmark. All metric scores are evaluated using Tesseract v4.0.0, pytesseract v0.3.10, MATLAB R2022a, Levenshtein v0.21.0 and jiwer v3.0.1. The results are presented in Tables \ref{tab:comparisontable} and \ref{tab:comparisontableUVDoc}.

Compared to previous works, our method achieves state-of-the-art MS-SSIM, LD, AD and ED performance and a second-best CER score on the DocUNet benchmark. On the UVDoc benchmark our method achieves state-of-the-art performance on visual metrics (MS-SSIM and LD). On OCR metrics, our network performs close to state-of-the-art. Small differences should be interpreted with care, as OCR scores suffer from high standard deviation (see \tabref{tab:ablation}). This is due to the Tesseract OCR engine being sensitive to small changes in input, as explained in its documentation \cite{tesseract}. Our method also ranks best on the horizontal and vertical line straightness metrics (\secref{sec:linemetric}), indicating better unwarping of the geometric features.

Our approach builds on a grid-based unwarping method, thanks to which our network is significantly smaller in size than current state-of-the-art methods, whilst still achieving state-of-the-art performance. We compare our network size to previous works in the last column of \tabref{tab:comparisontable}.

\begin{table*}
\small
\centering
    \caption{Quantitative comparison on the \hbox{DocUNet} and UVDoc benchmark datasets for DewarpNet \shortcite{DewarpNet} finetuned with and without the UVDoc data. We finetuned the pre-trained models for 10 epochs with Doc3D only, and for 5 epochs with Doc3D+UVDoc to equalize the number of optimization steps.}
    \vspace{-4pt}
    \begin{tabular}{l c c c c c >{\hspace{0pc}} c c c c c c}
    \toprule
    & \multicolumn{5}{c}{DocUNet} & \multicolumn{6}{c}{UVDoc}\\
    \cmidrule[0.1pt](lr){2-6}\cmidrule[0.1pt](lr){7-12}
     & MS-SSIM $\uparrow$ & LD $\downarrow$ & AD $\downarrow$ & CER $\downarrow$ & ED $\downarrow$ & MS-SSIM $\uparrow$ & AD $\downarrow$ & CER $\downarrow$ & ED $\downarrow$ & H-line $\downarrow$ & V-line $\downarrow$\\
    \midrule
    Doc3D only & {0.475} & 8.40 & 0.411 & \best{0.222} & \best{856} & 0.659 &  0.211 & 0.085 & 265 & 3.48 & 4.75\\ 
    Doc3D + UVDoc &  \best{0.504} & \best{7.68} & \best{0.400} & 0.228 & 878 & \best{0.725} & \best{0.151} & \best{0.075} & \best{232} & \best{2.88} & \best{3.56}\\\bottomrule
    \end{tabular}
\label{tab:ablation_DewarpNet_finetune}
\end{table*}

\begin{table*}
\small
\centering
    \caption{Ablations about losses and data used. The reported values are averages and standard deviations over 10 repetitions of training with otherwise constant parameters on the \hbox{DocUNet} and UVDoc benchmark datasets. Settings used in our final model are underlined.}
    \vspace{-4pt}
    \resizebox{\textwidth}{!}{
    \begin{tabular}{l c c c c c c c c c c c c}
    \toprule
    & & \multicolumn{5}{c}{DocUNet} & \multicolumn{6}{c}{UVDoc}\\
    \cmidrule[0.1pt](lr){3-7}\cmidrule[0.1pt](lr){8-13}
     & & MS-SSIM $\uparrow$ & LD $\downarrow$ & AD $\downarrow$ & CER $\downarrow$ & ED $\downarrow$ & MS-SSIM $\uparrow$ & AD $\downarrow$ & CER $\downarrow$ & ED $\downarrow$ & H-line $\downarrow$ & V-line $\downarrow$ \\
    \midrule
    \multicolumn{2}{l}{\underline{Doc3D + UVDoc}}   & \best{0.536$\pm$0.006} & \best{6.96$\pm$0.17} & \best{0.325$\pm$0.006} & \best{0.195$\pm$0.012} & \best{745$\pm$34} & \best{0.762$\pm$0.014} & \best{0.129$\pm$0.008} & \best{0.070$\pm$0.010} & \best{205$\pm$23} & \best{1.85$\pm$0.06} & \best{2.53$\pm$0.06} \\     
\multicolumn{2}{l}{Doc3D only} & 0.492$\pm$0.004 & 7.99$\pm$0.13 & 0.360$\pm$0.007 & 0.197$\pm$0.018 & 757$\pm$57 & 0.669$\pm$0.015 & 0.178$\pm$0.013 & 0.078$\pm$0.013 &  220$\pm$30 & 2.42$\pm$0.03 & 3.85$\pm$0.16 \\
    \midrule
    \multirow{2}{*}{3D grid} & \underline{w/} & \textbf{0.536$\pm$0.006} & \textbf{6.96$\pm$0.17} & \textbf{0.325$\pm$0.006} & 0.195$\pm$0.012 & 745$\pm$34 & \textbf{0.762$\pm$0.014} & \textbf{0.129$\pm$0.008} & 0.070$\pm$0.010 & 205$\pm$23 & \textbf{1.85$\pm$0.06} & \textbf{2.53$\pm$0.06} \\ 
    & w/o & 0.531$\pm$0.005 & 7.04$\pm$0.16 & 0.331$\pm$0.004 & \textbf{0.189$\pm$0.017} & \textbf{743$\pm$54} & 0.747$\pm$0.010 & 0.148$\pm$0.011 & \textbf{0.068$\pm$0.010} & \textbf{201$\pm$22} &  1.87$\pm$0.08 & {2.59$\pm$0.08}  \\ 
    \midrule
    \multirow{2}{*}{$\mathcal{L}_r$} & \underline{w/} & \textbf{0.536$\pm$0.006} & {6.96$\pm$0.17} & \textbf{0.325$\pm$0.006} & \textbf{0.195$\pm$0.012} & \textbf{745$\pm$34} & \textbf{0.762$\pm$0.014} & \textbf{0.129$\pm$0.008} & 0.070$\pm$0.010 & 205$\pm$23 & \textbf{1.85$\pm$0.06} & \textbf{2.53$\pm$0.06} \\     
& w/o & {0.533$\pm$0.005} & \textbf{6.87$\pm$0.13} & {0.327$\pm$0.005} & 0.199$\pm$0.015 & 764$\pm$67 & 0.746$\pm$0.010 & 0.136$\pm$0.012 & \textbf{0.065$\pm$0.006} & \textbf{196$\pm$12} & 1.89$\pm$0.09 & 2.56$\pm$0.13 \\ 
    \bottomrule
    \end{tabular}}
\label{tab:ablation}
\end{table*}

In addition to the performance of our own method, we also evaluate the effect of adding our UVDoc data to the DewarpNet \cite{DewarpNet} architecture. We compare the performance of the pre-trained DewarpNet models fine-tuned for 10 epochs on the Doc3D data with the performance of the pre-trained models fine-tuned for 5 epochs on a combination of Doc3D and UVDoc data. As shown in \tabref{tab:ablation_DewarpNet_finetune}, adding the UVDoc data into the fine-tuning process greatly improves all metrics except for the OCR performance on the \emph{shaded} DocUNet images.

\begin{figure*}
	\vspace{20pt}
	\centering
	\setlength{\tabcolsep}{2pt}
	\renewcommand{\arraystretch}{1.5}
	\begin{tabular}{ccccccccc}
		\includegraphics[width=0.135\linewidth,height=0.173\linewidth]{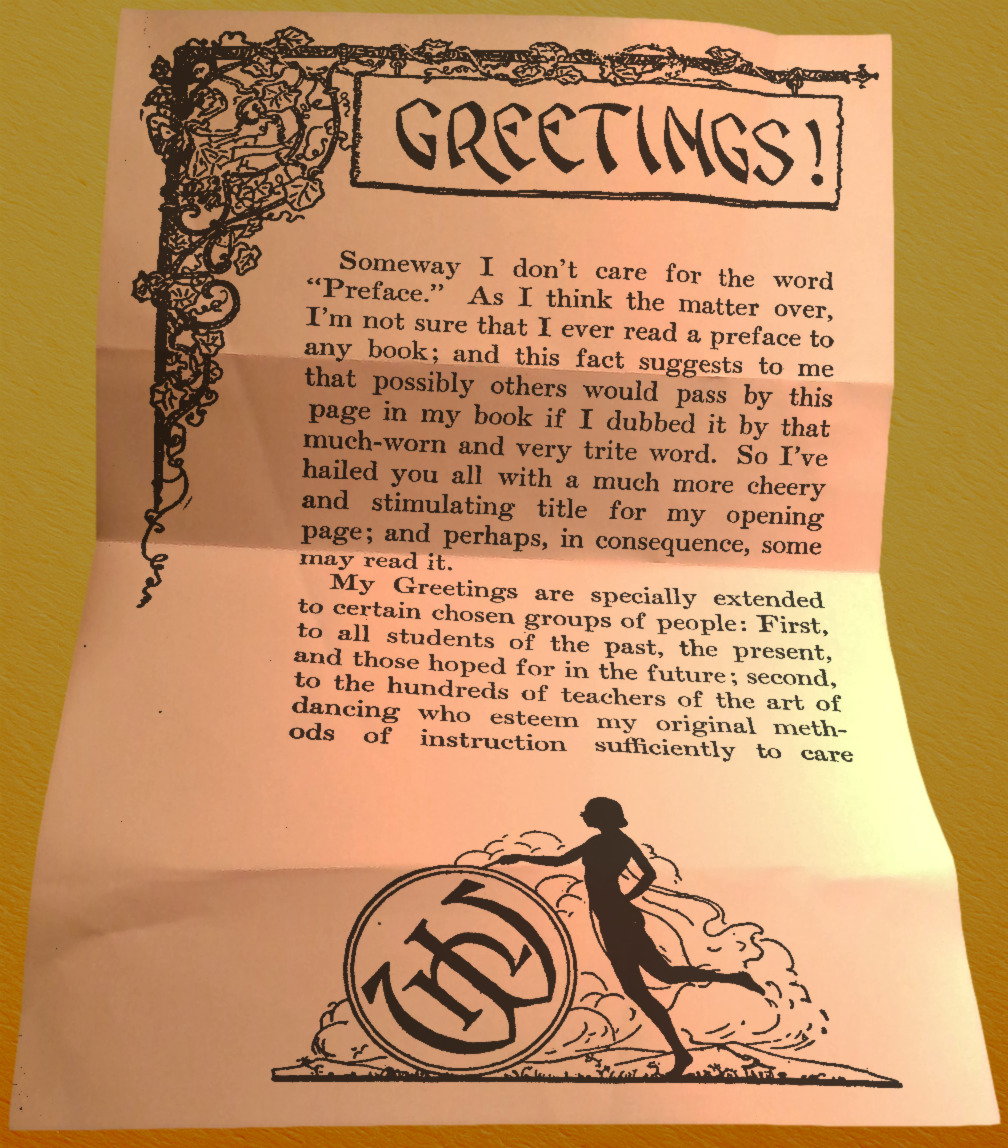} & 
		\frame{\includegraphics[width=0.135\linewidth,height=0.173\linewidth]{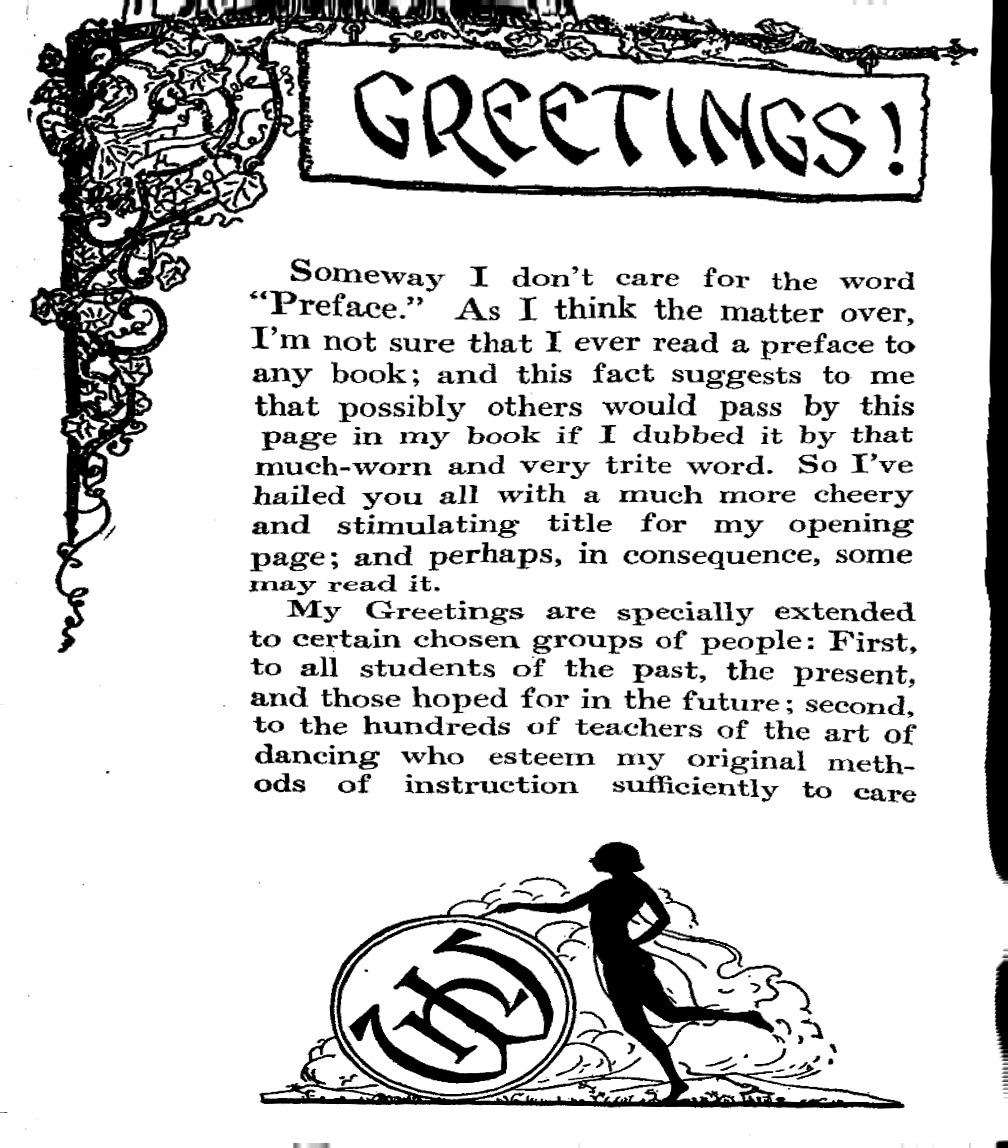}} & 
		\frame{\includegraphics[width=0.135\linewidth,height=0.173\linewidth]{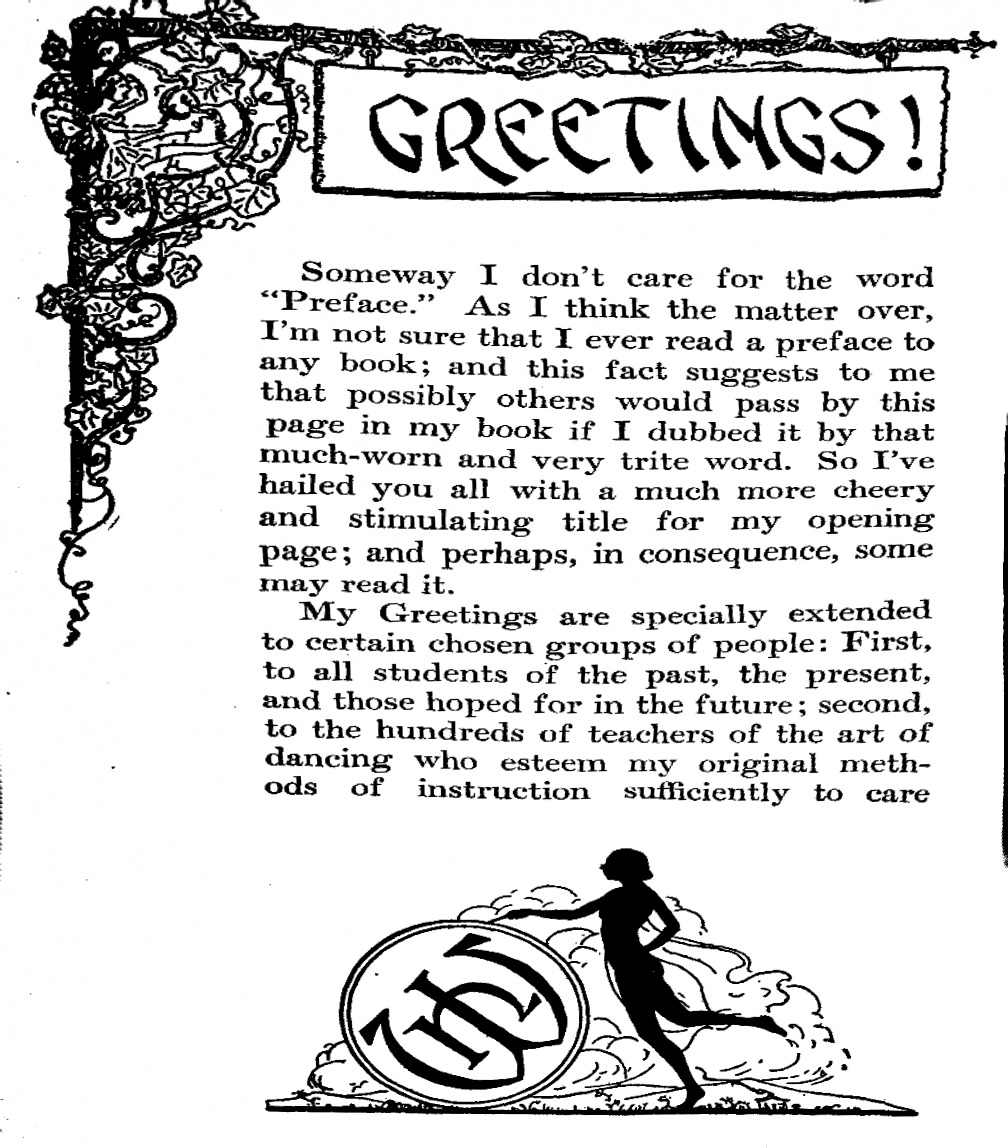}} &
		\frame{\includegraphics[width=0.135\linewidth,height=0.173\linewidth]{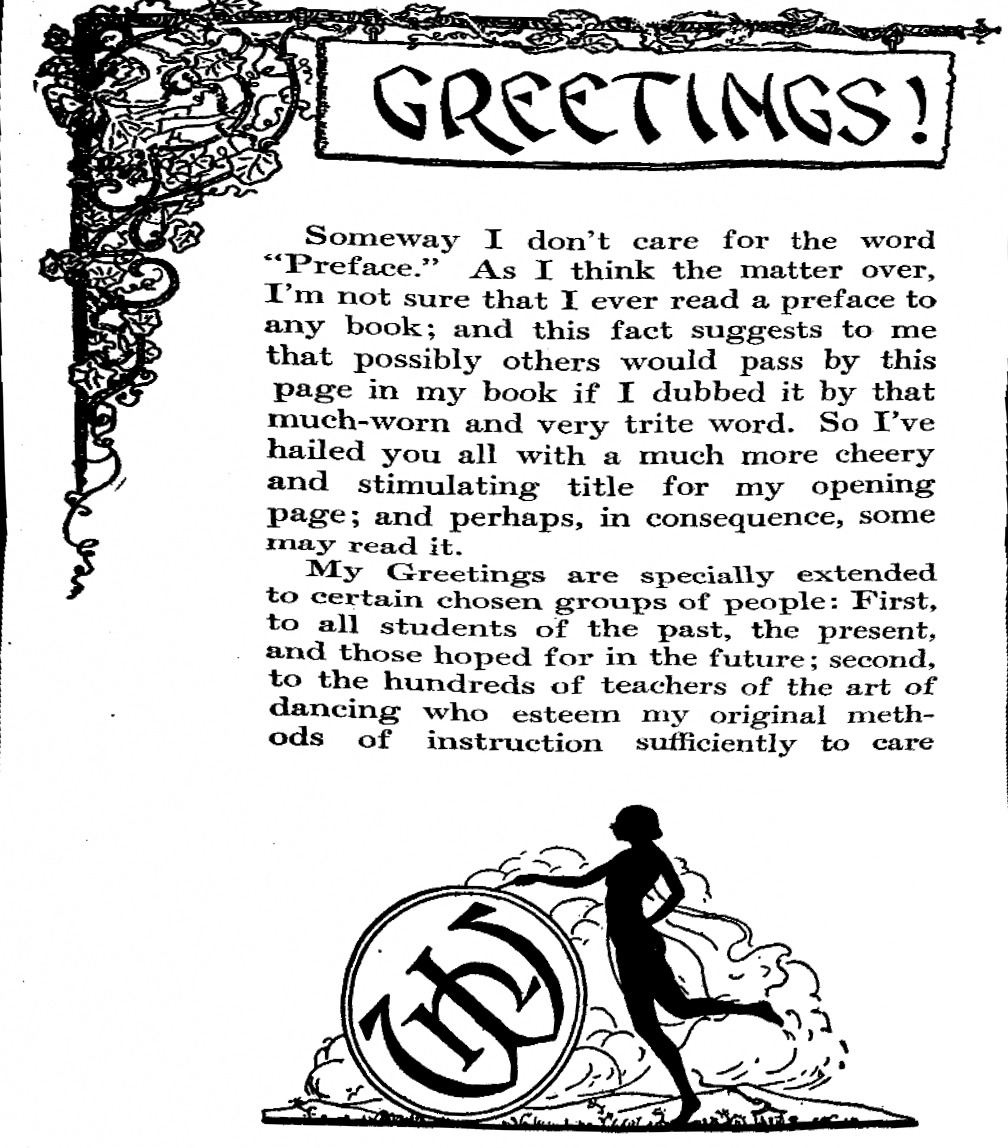}} & 
		\frame{\includegraphics[width=0.135\linewidth,height=0.173\linewidth]{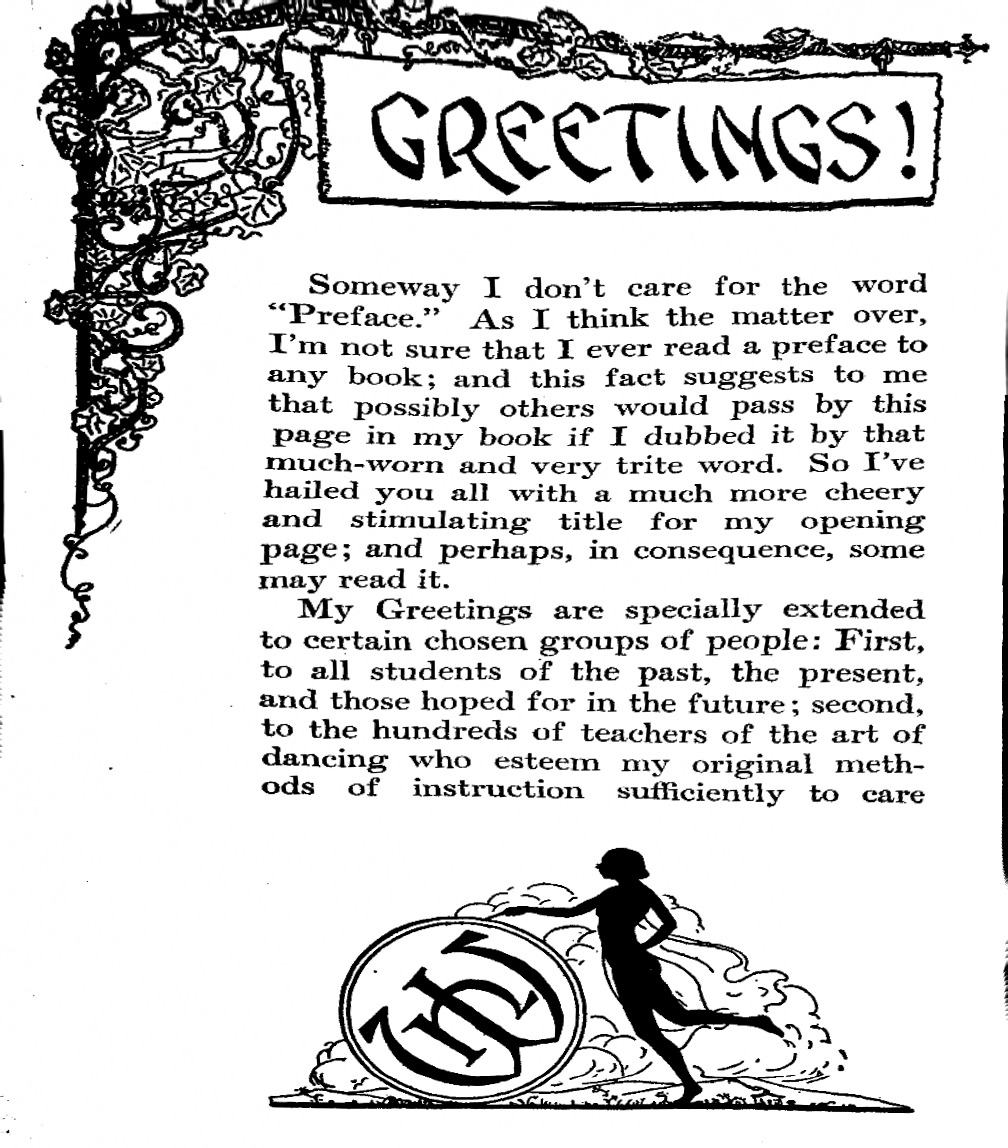}} &
		\frame{\includegraphics[width=0.135\linewidth,height=0.173\linewidth]{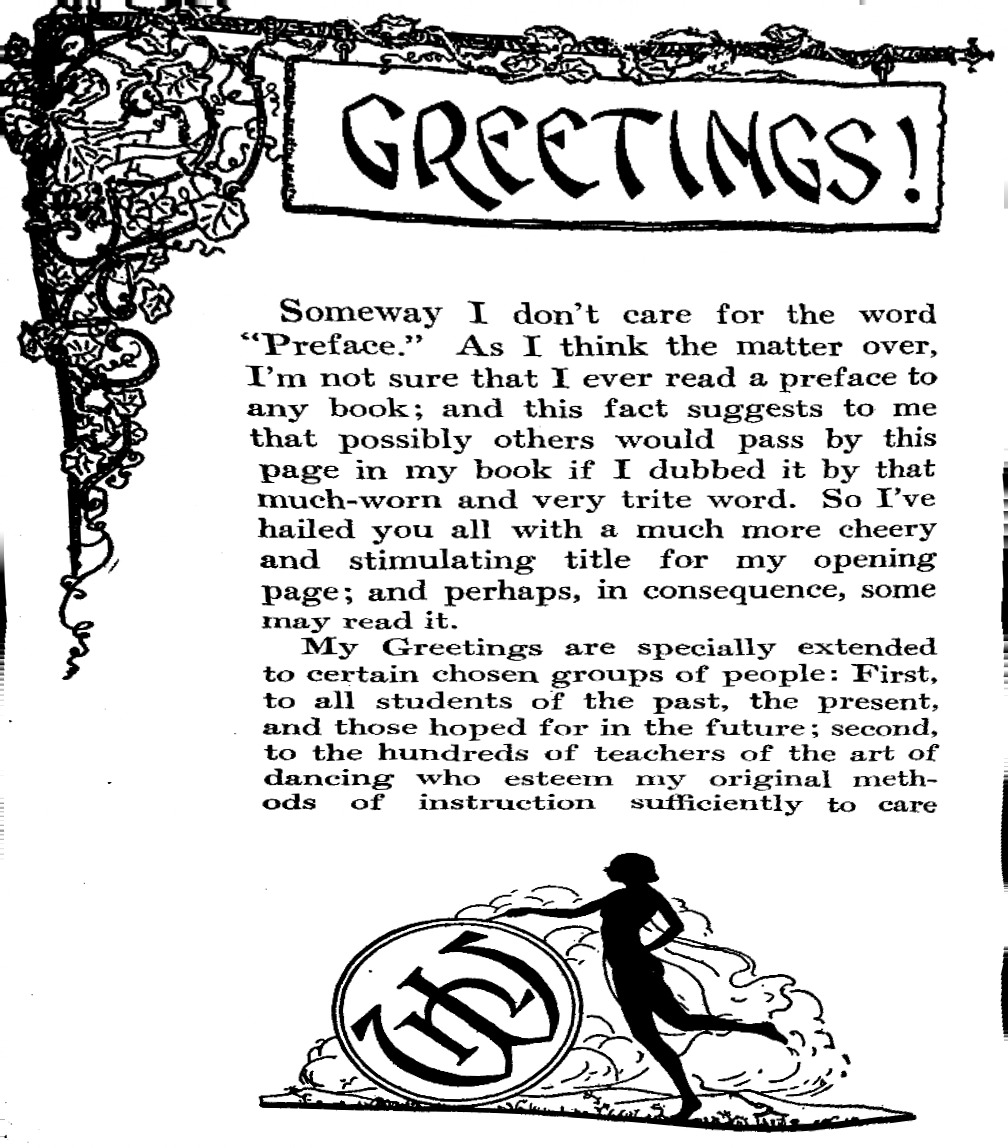}} & 
		\frame{\includegraphics[width=0.135\linewidth,height=0.173\linewidth]{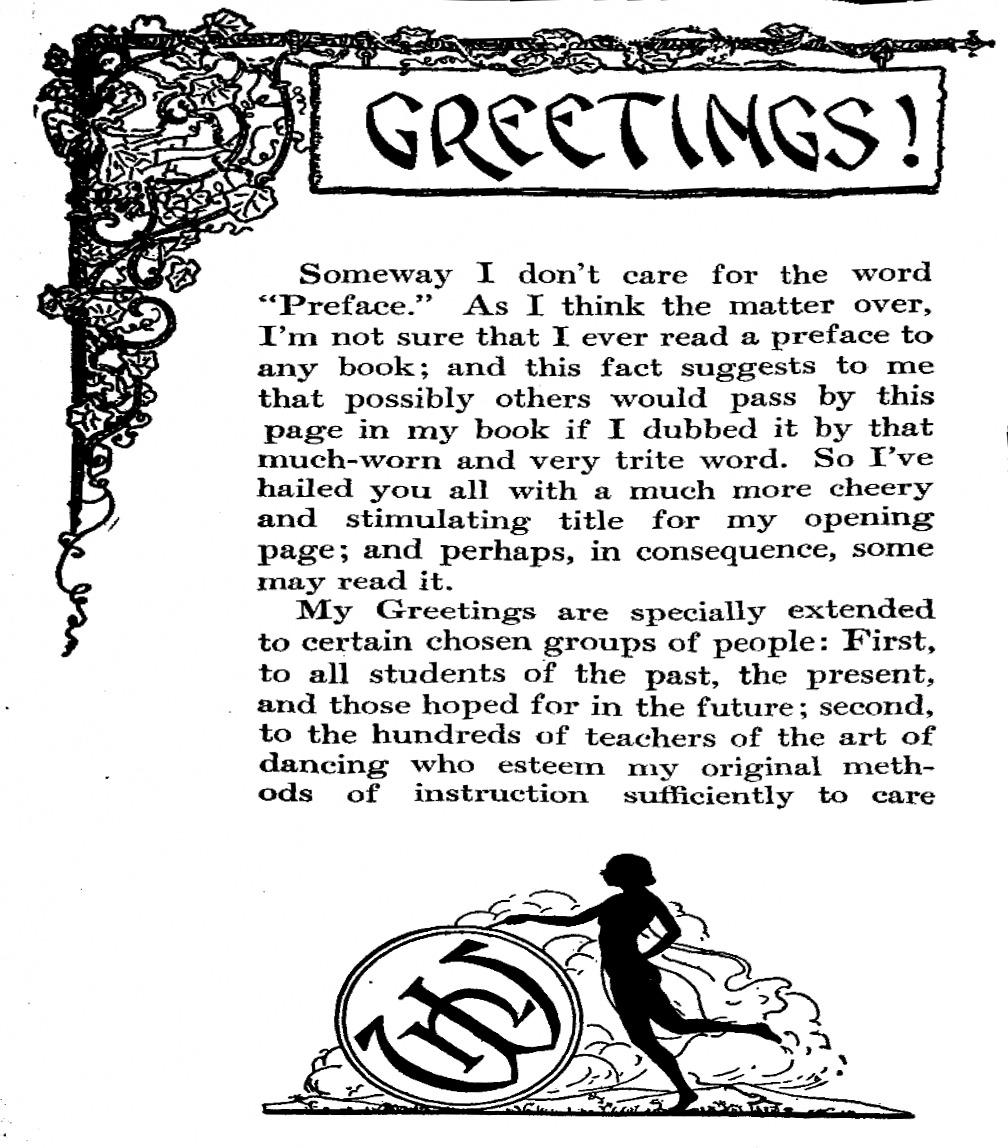}} \\
		\includegraphics[width=0.135\linewidth,height=0.163\linewidth]{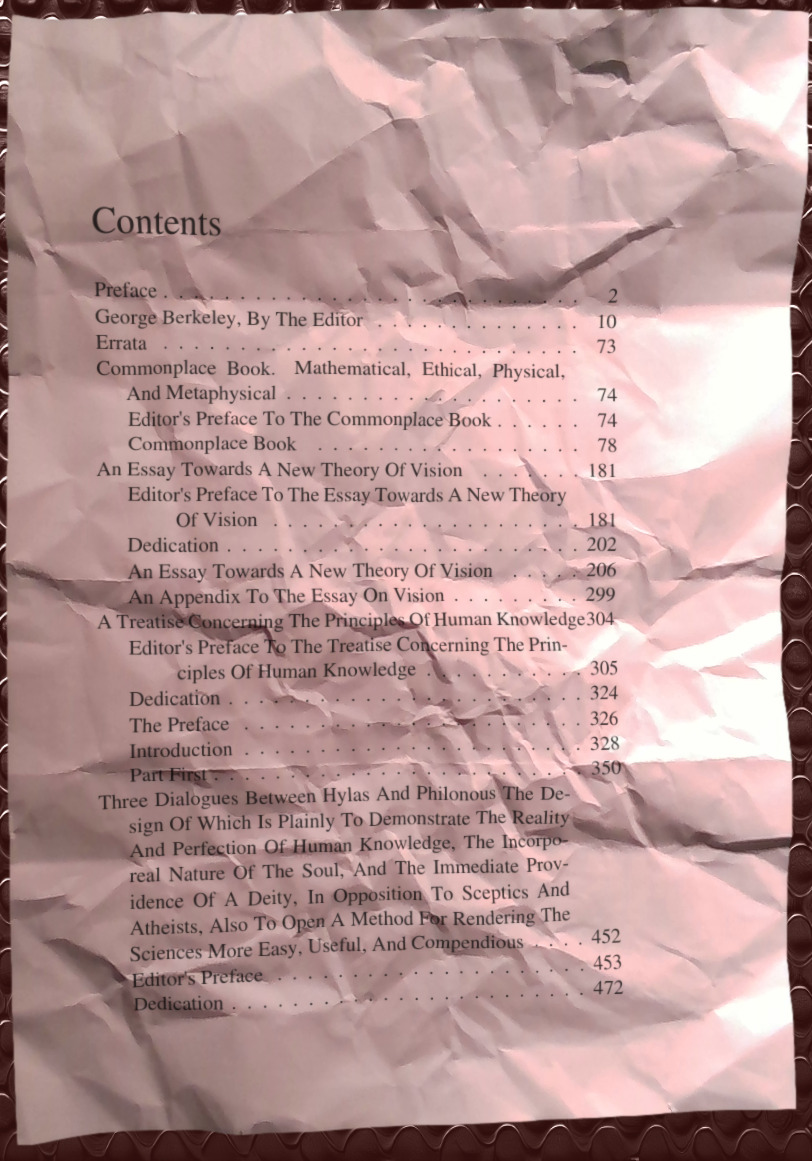} & 
		\frame{\includegraphics[width=0.135\linewidth,height=0.163\linewidth]{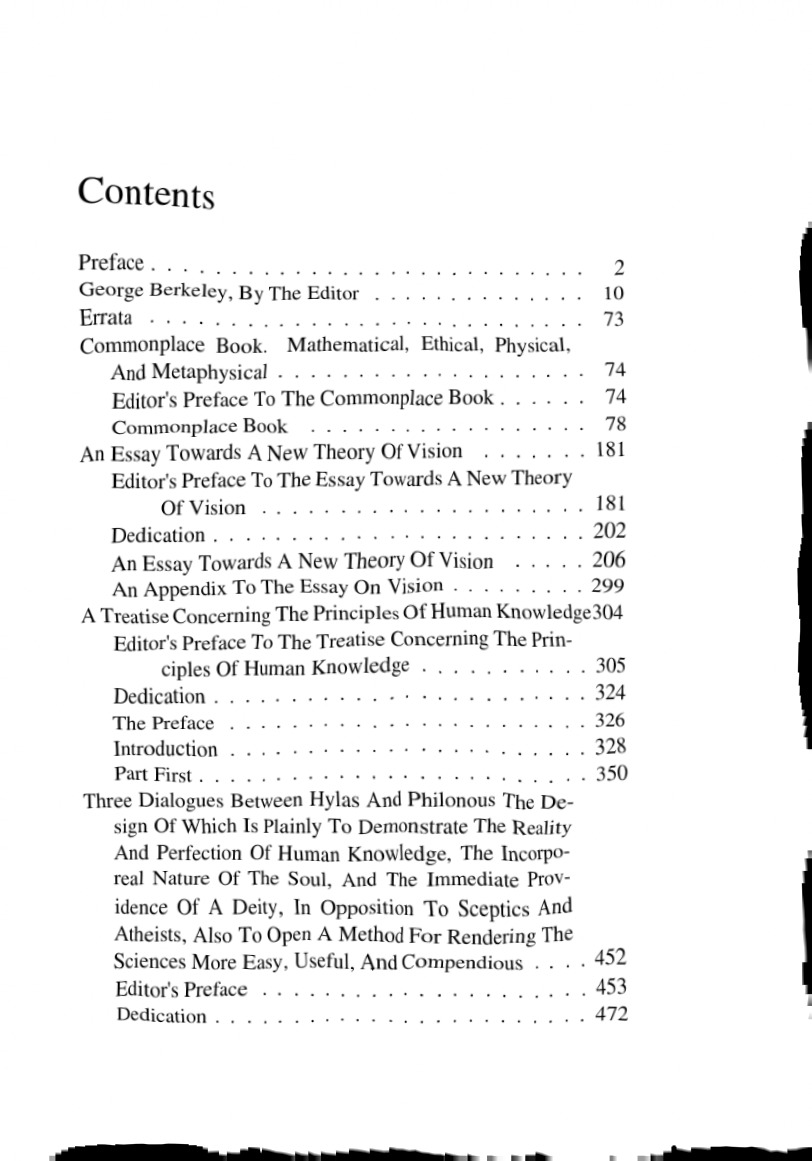}} & 
		\frame{\includegraphics[width=0.135\linewidth,height=0.163\linewidth]{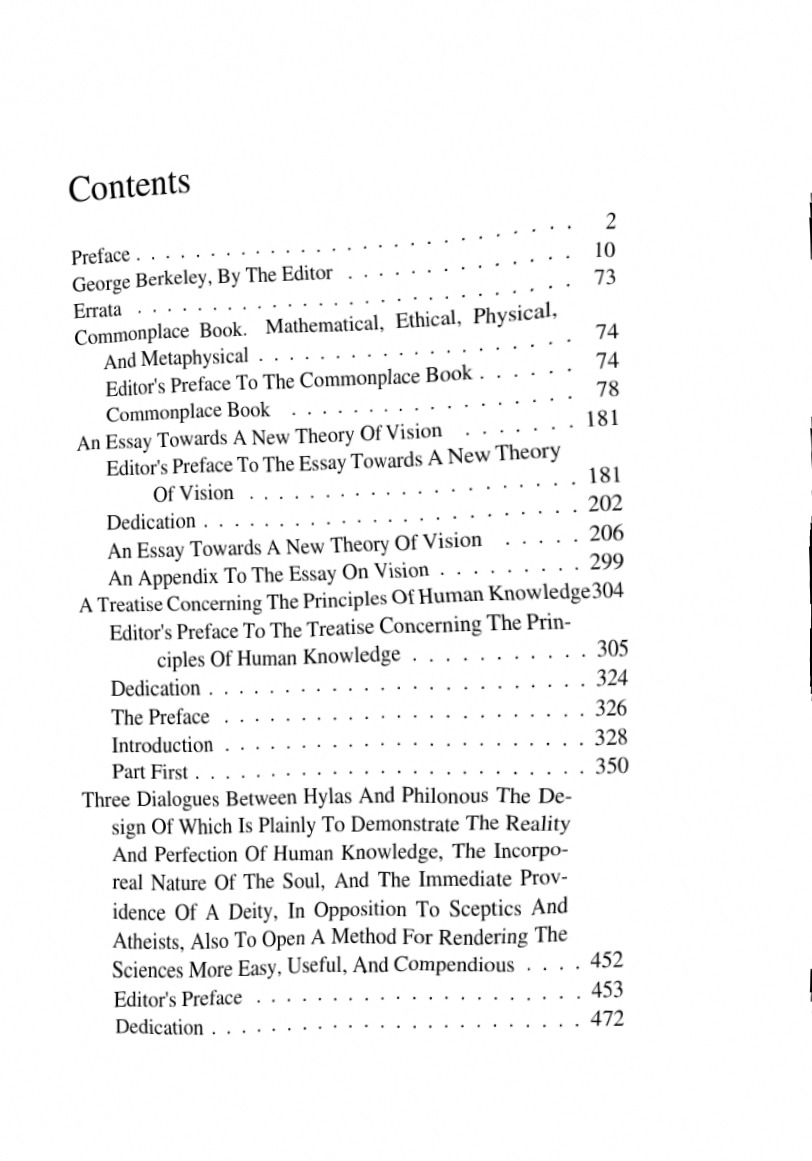}} &
		\frame{\includegraphics[width=0.135\linewidth,height=0.163\linewidth]{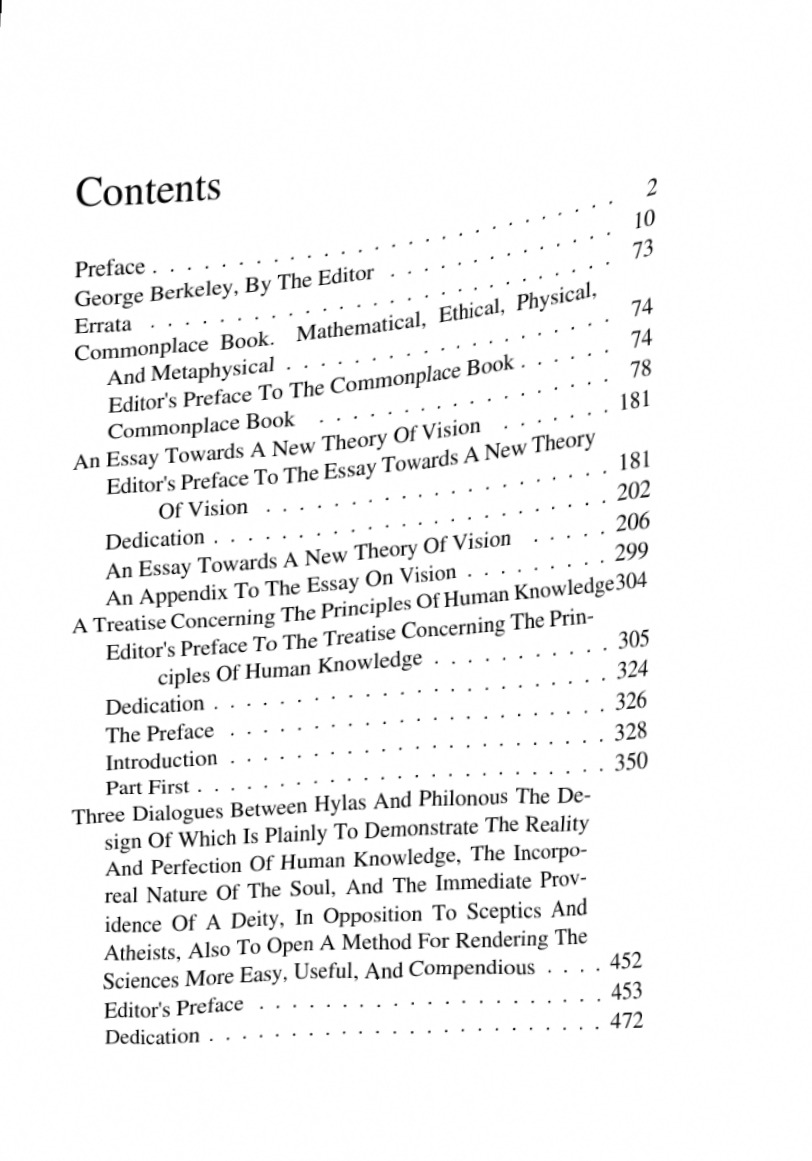}} & 
		\frame{\includegraphics[width=0.135\linewidth,height=0.163\linewidth]{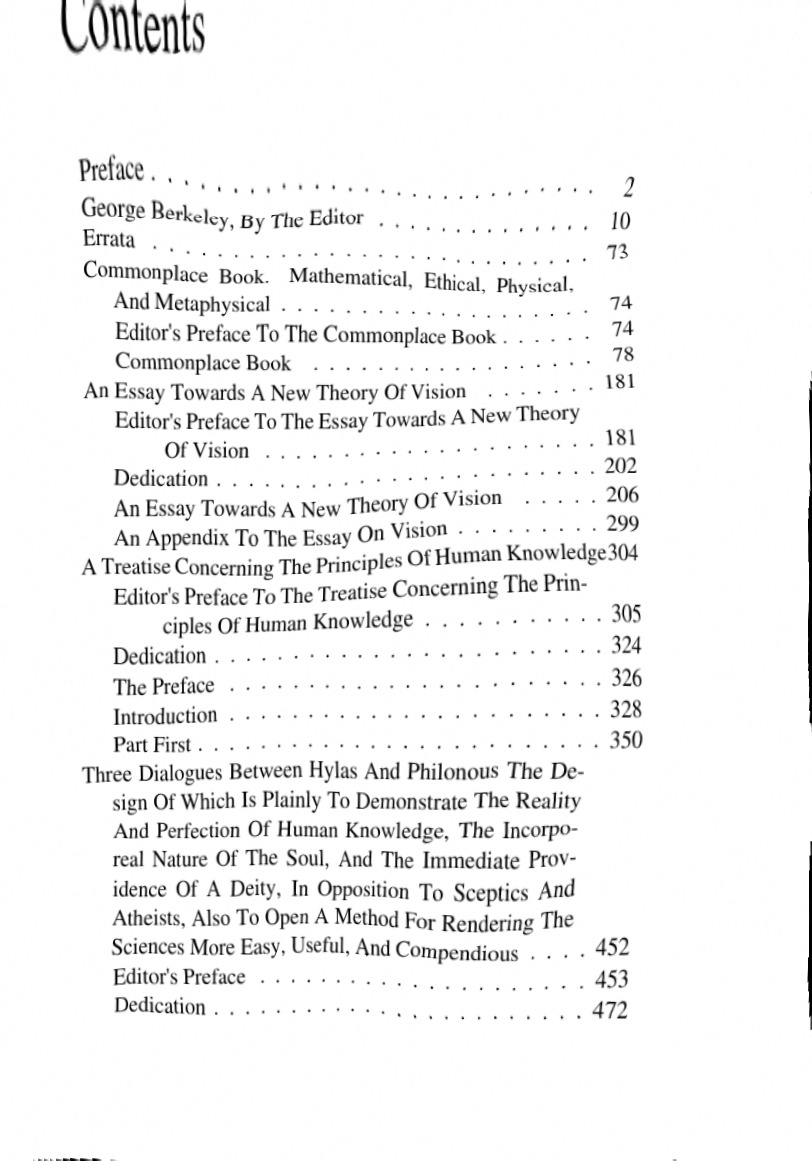}} &
		\frame{\includegraphics[width=0.135\linewidth,height=0.163\linewidth]{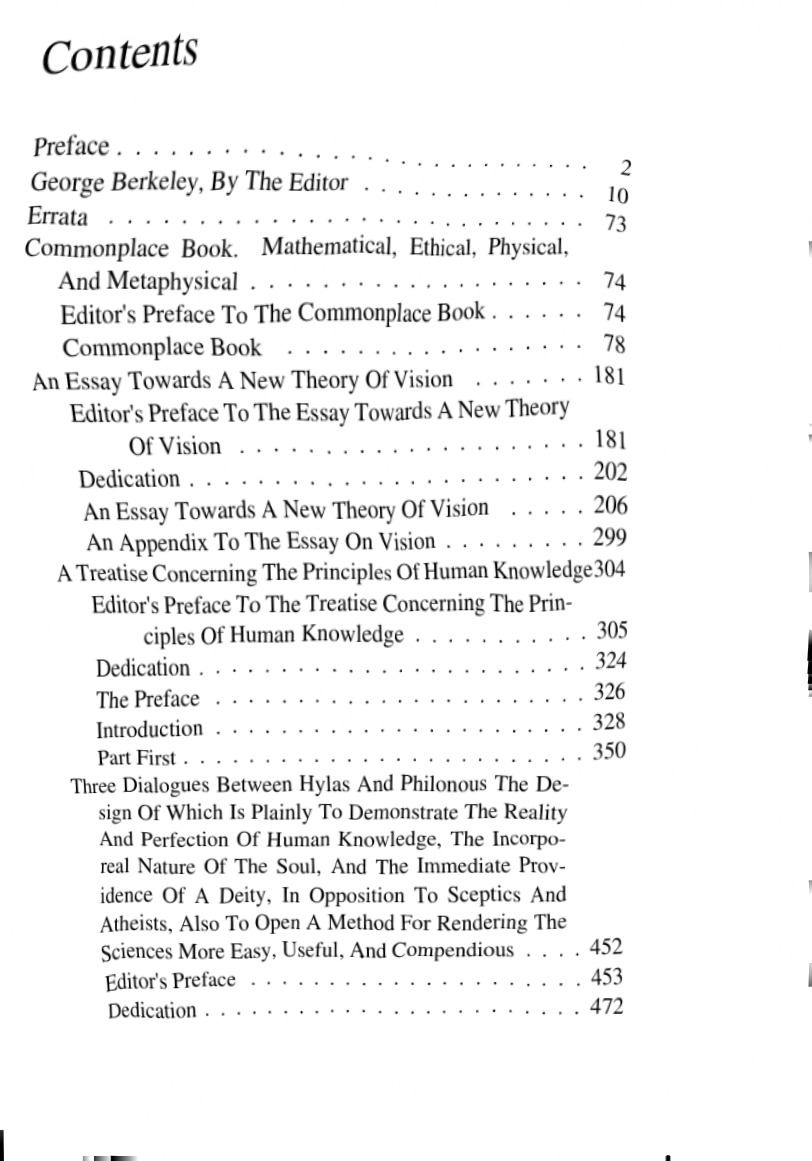}} & 
		\frame{\includegraphics[width=0.135\linewidth,height=0.163\linewidth]{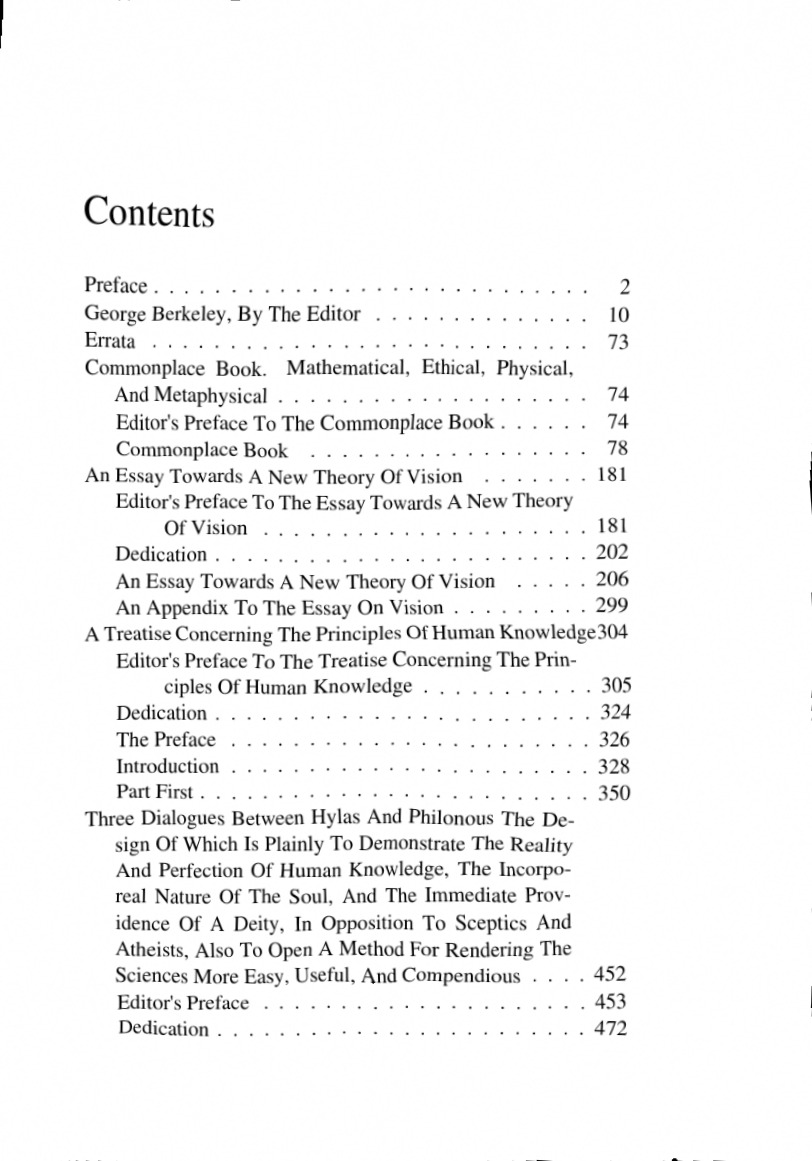}}\\
		\includegraphics[width=0.135\linewidth,height=0.161\linewidth]{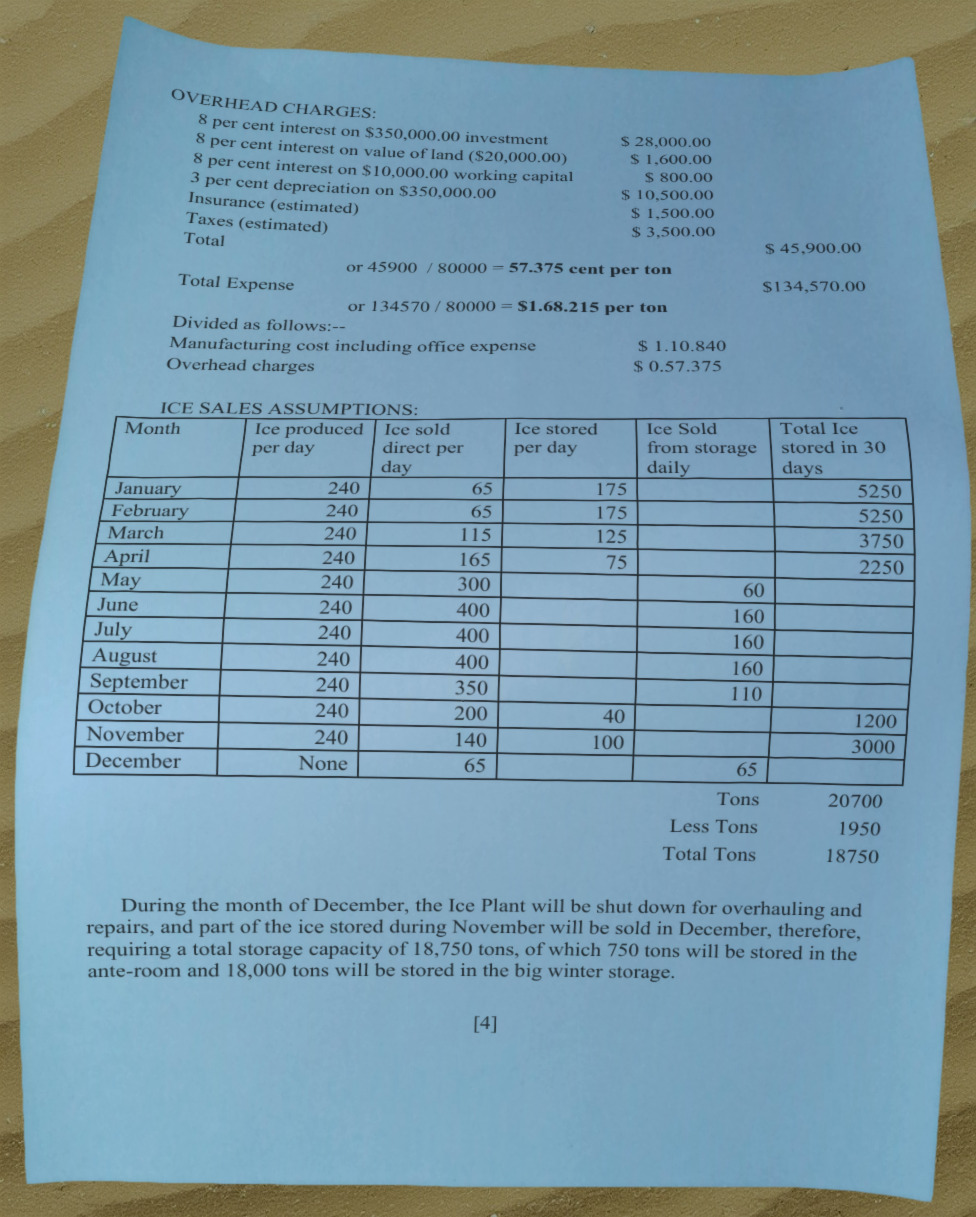} & 
		\frame{\includegraphics[width=0.135\linewidth,height=0.161\linewidth]{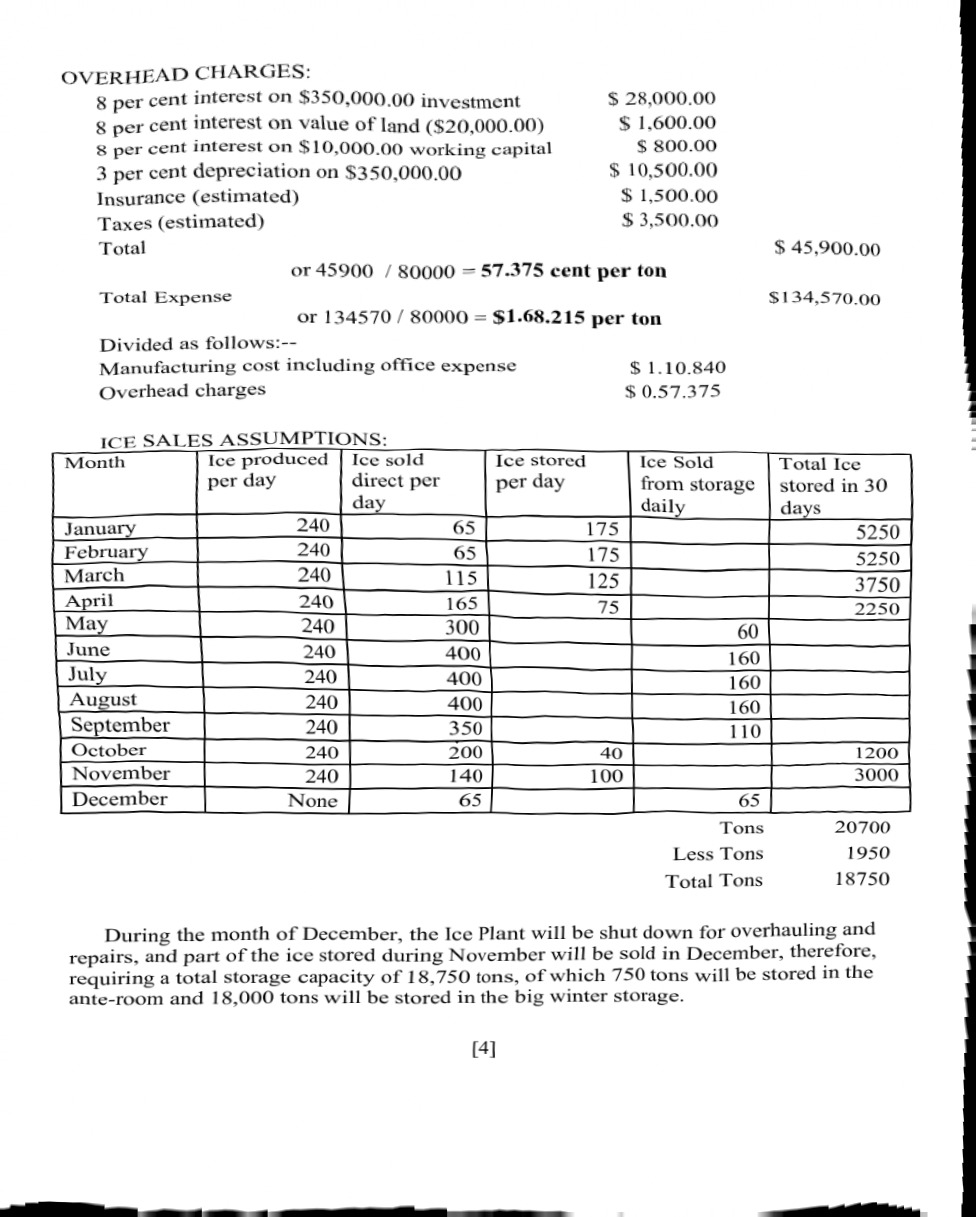}} & 
		\frame{\includegraphics[width=0.135\linewidth,height=0.161\linewidth]{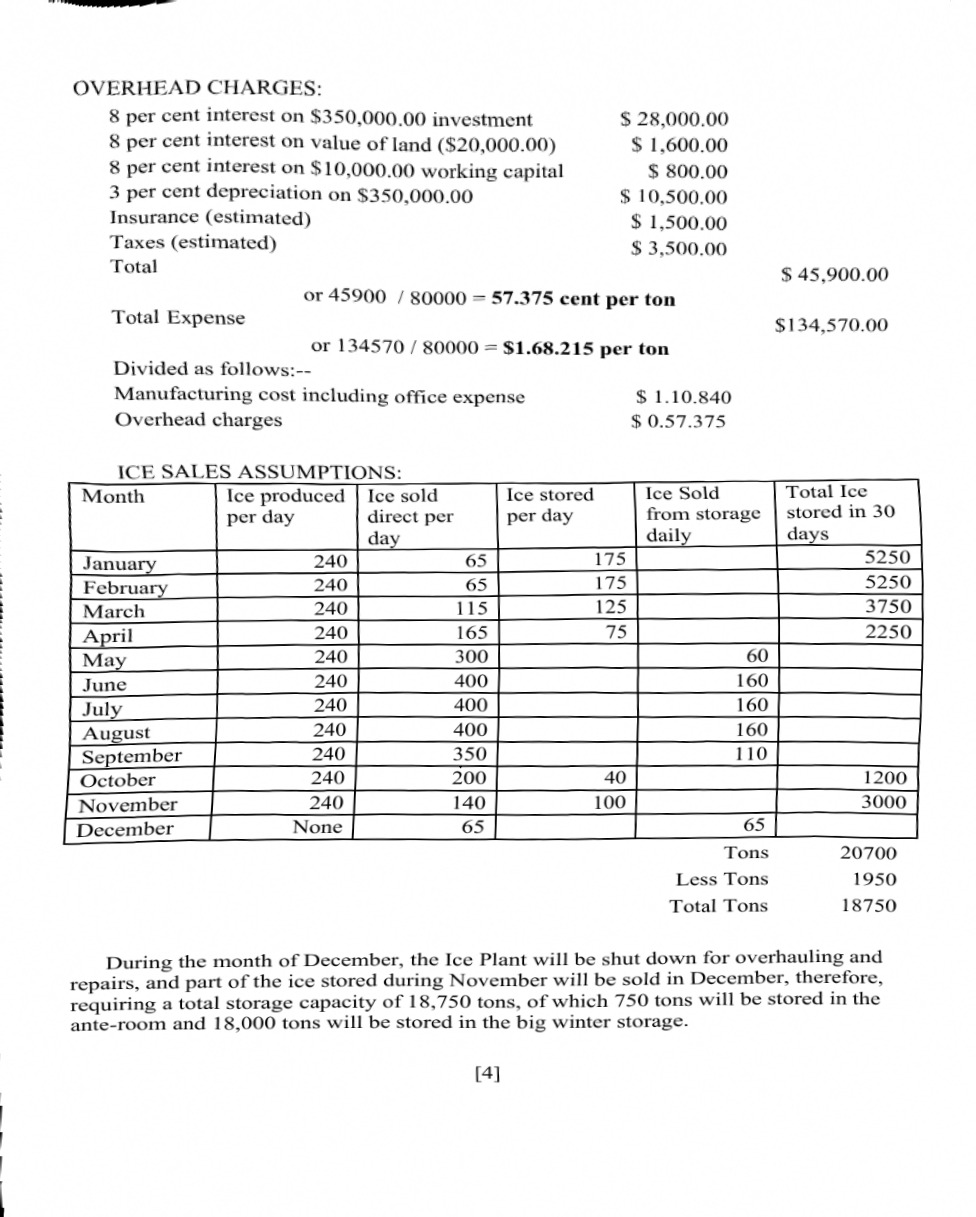}} &
		\frame{\includegraphics[width=0.135\linewidth,height=0.161\linewidth]{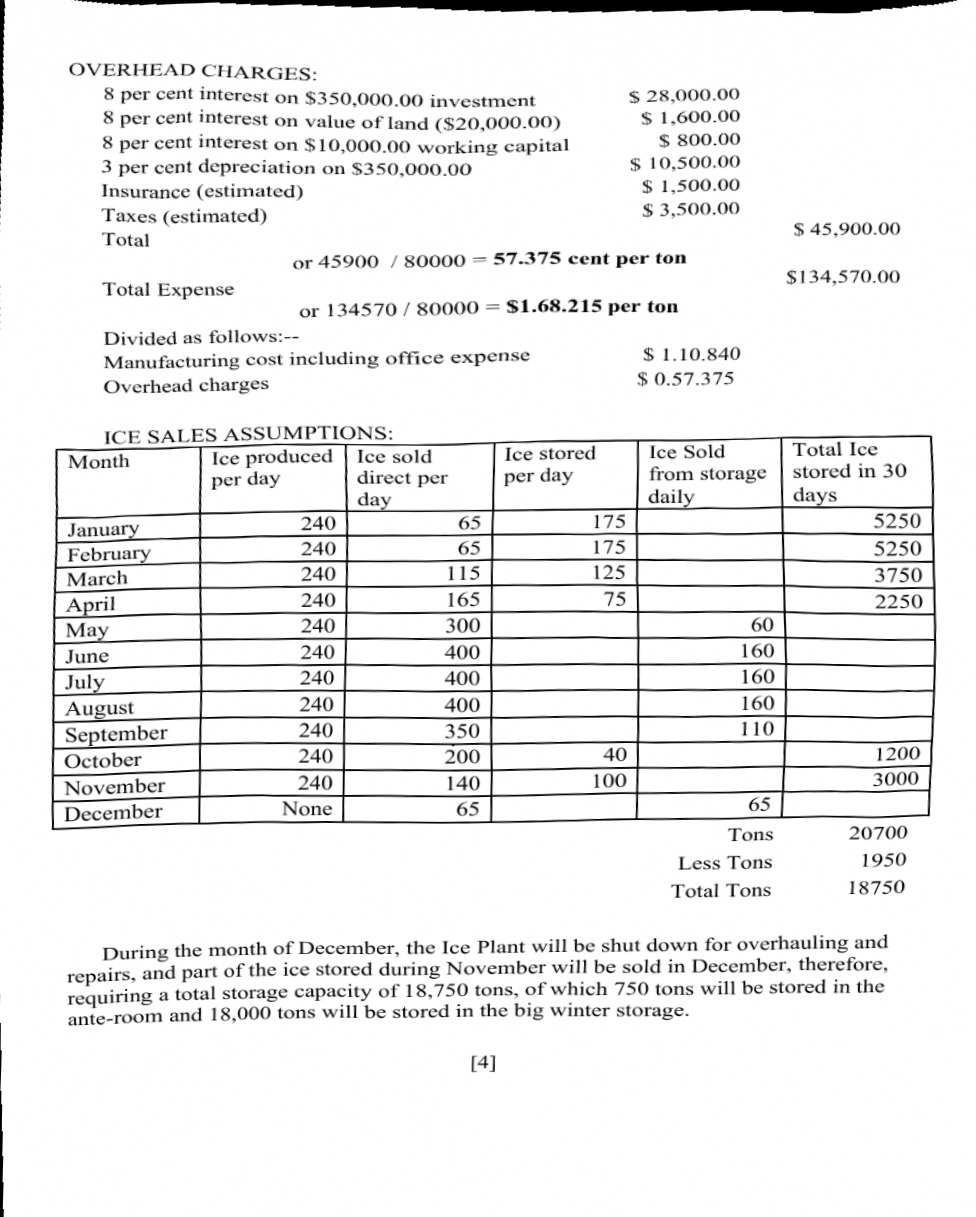}} & 
		\frame{\includegraphics[width=0.135\linewidth,height=0.161\linewidth]{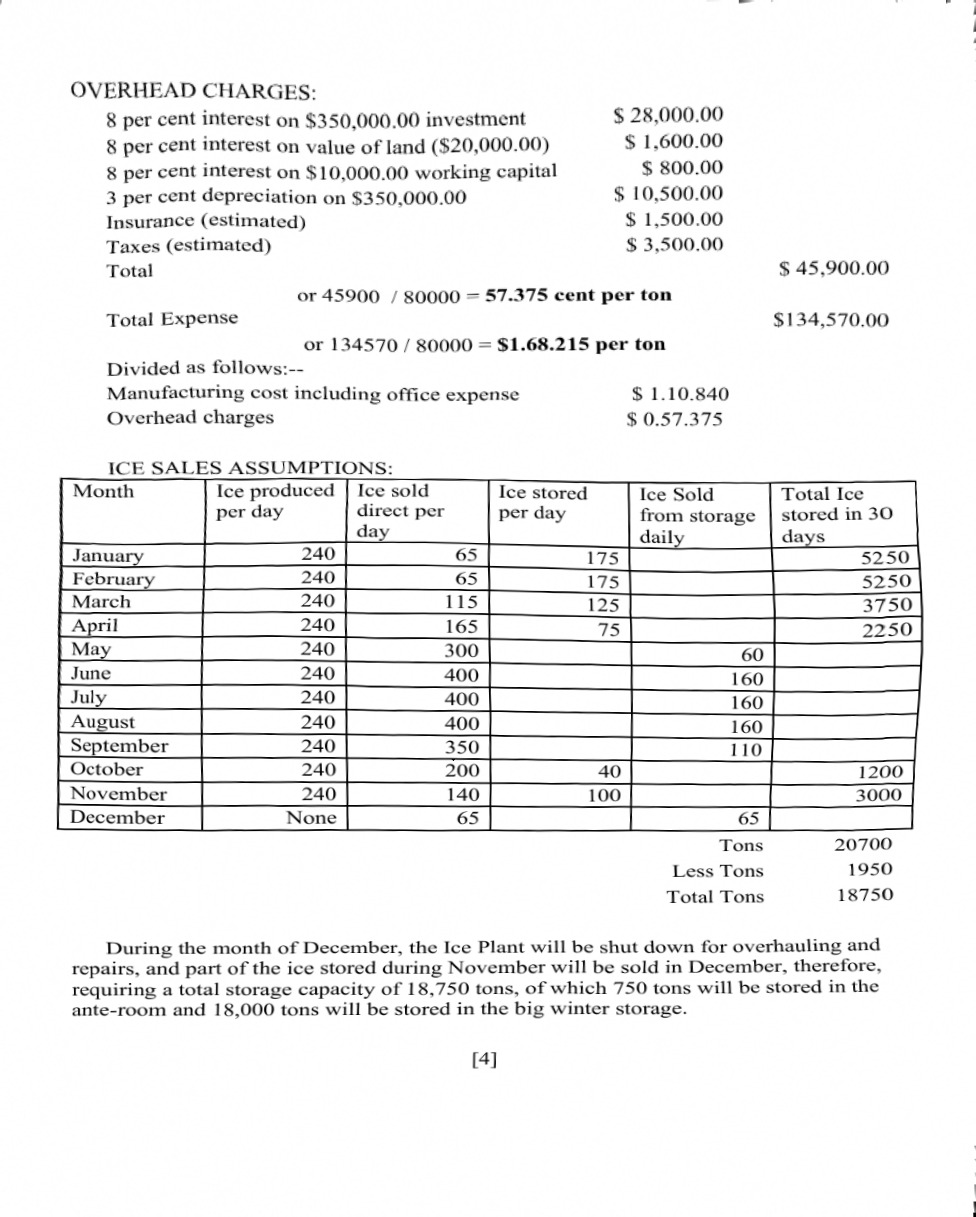}} &
		\frame{\includegraphics[width=0.135\linewidth,height=0.161\linewidth]{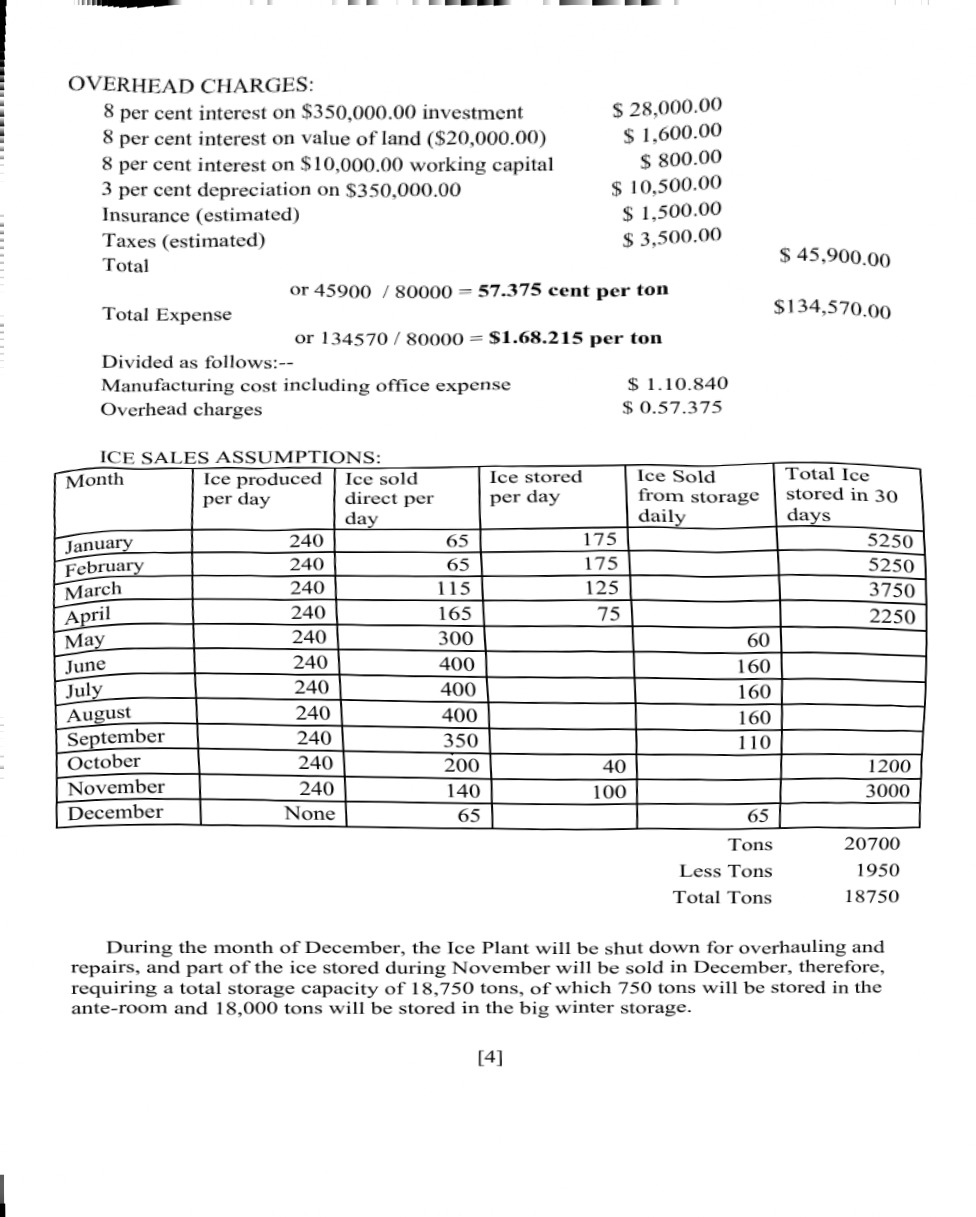}} & 
		\frame{\includegraphics[width=0.135\linewidth,height=0.161\linewidth]{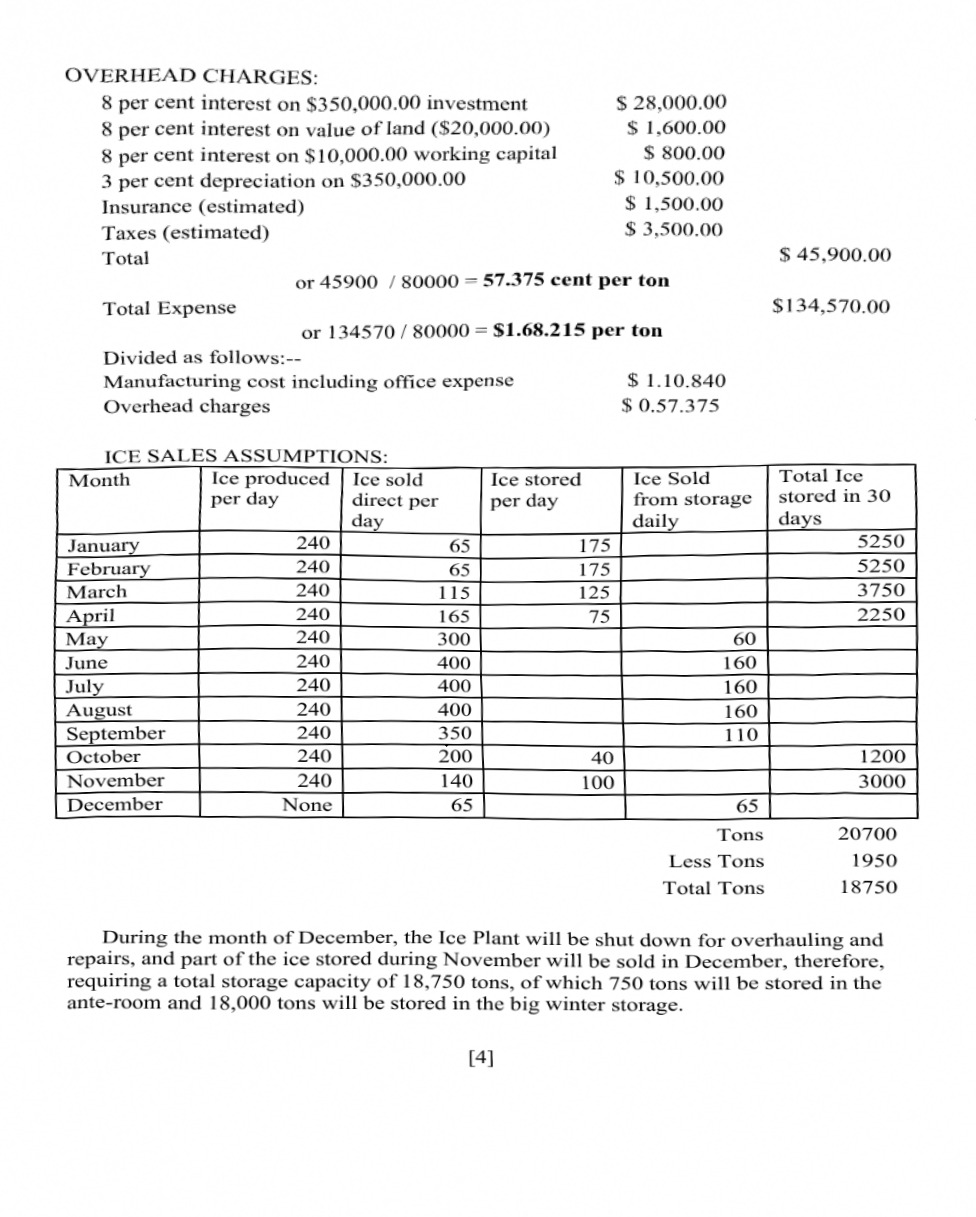}}
		\\[-8pt]
		\small input &
		\small \cite{DewarpNet} &
		\small \cite{DocTr} &
		\small \cite{DDControlPoints} &
		\small \cite{DocGeoNet} &
		\small \cite{RDGR} &
		\small ours \\
	\end{tabular}
	\caption{Qualitative comparisons on the UVDoc benchmark dataset. From left to right: input, DewarpNet \cite{DewarpNet}, DocTr \cite{DocTr}, DDCP \cite{DDControlPoints}, DocGeoNet \cite{DocGeoNet}, RDGR \cite{RDGR}, ours. All input images come from the \emph{shaded} subset and we show the \emph{unshaded} version of the unwarped images to emphasize their structure. \label{fig:unwarpedUVDoc}}
\end{figure*}

\begin{figure*}
	\centering
     \setlength{\tabcolsep}{1pt}
     \renewcommand{\arraystretch}{1.5}
	\begin{tabular}{ccccccccc}
		\includegraphics[width=0.124\linewidth,height=0.1674\linewidth]{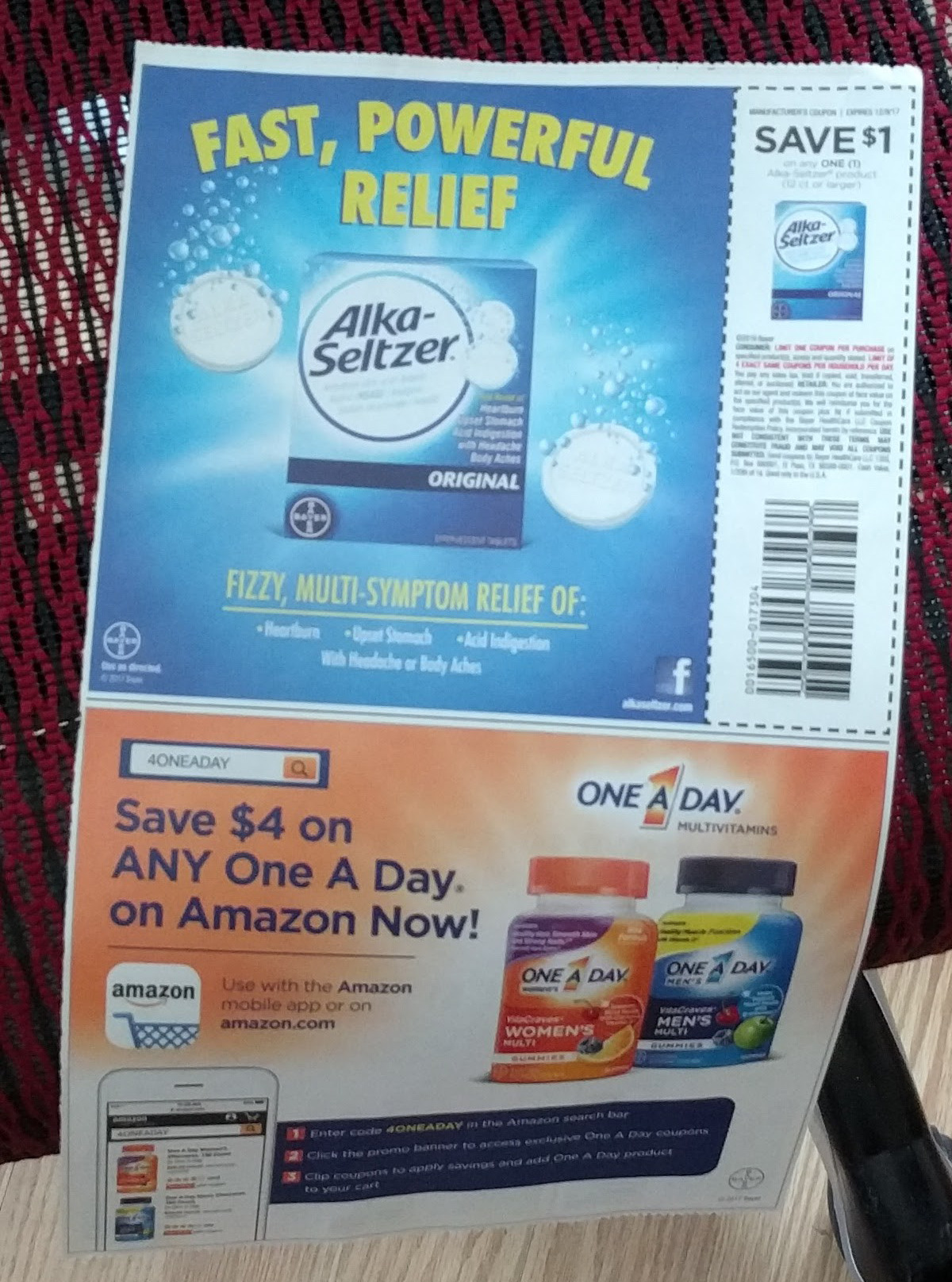} & 
		\includegraphics[width=0.124\linewidth,height=0.1674\linewidth]{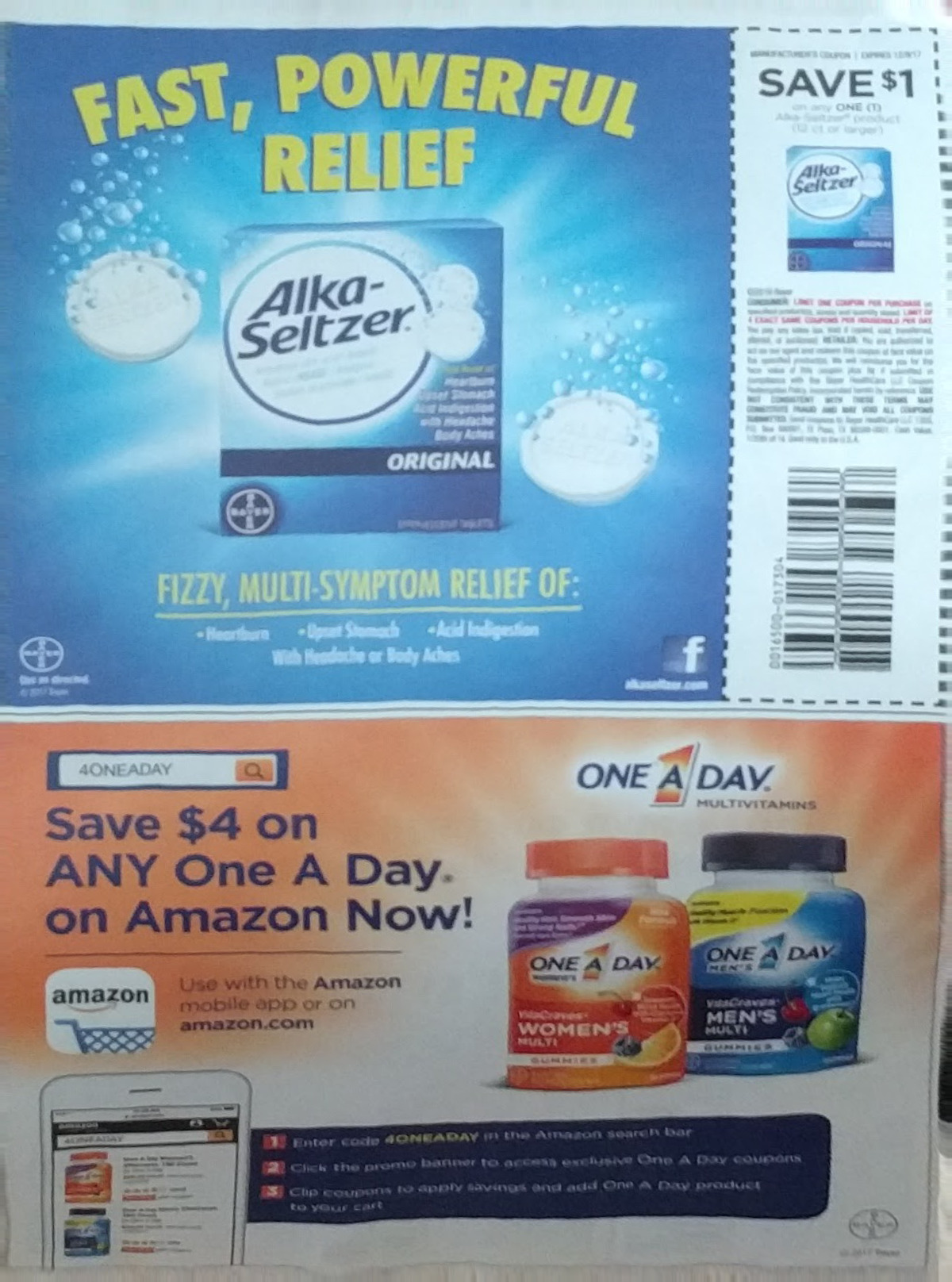} & 
        \includegraphics[width=0.124\linewidth,height=0.1674\linewidth]{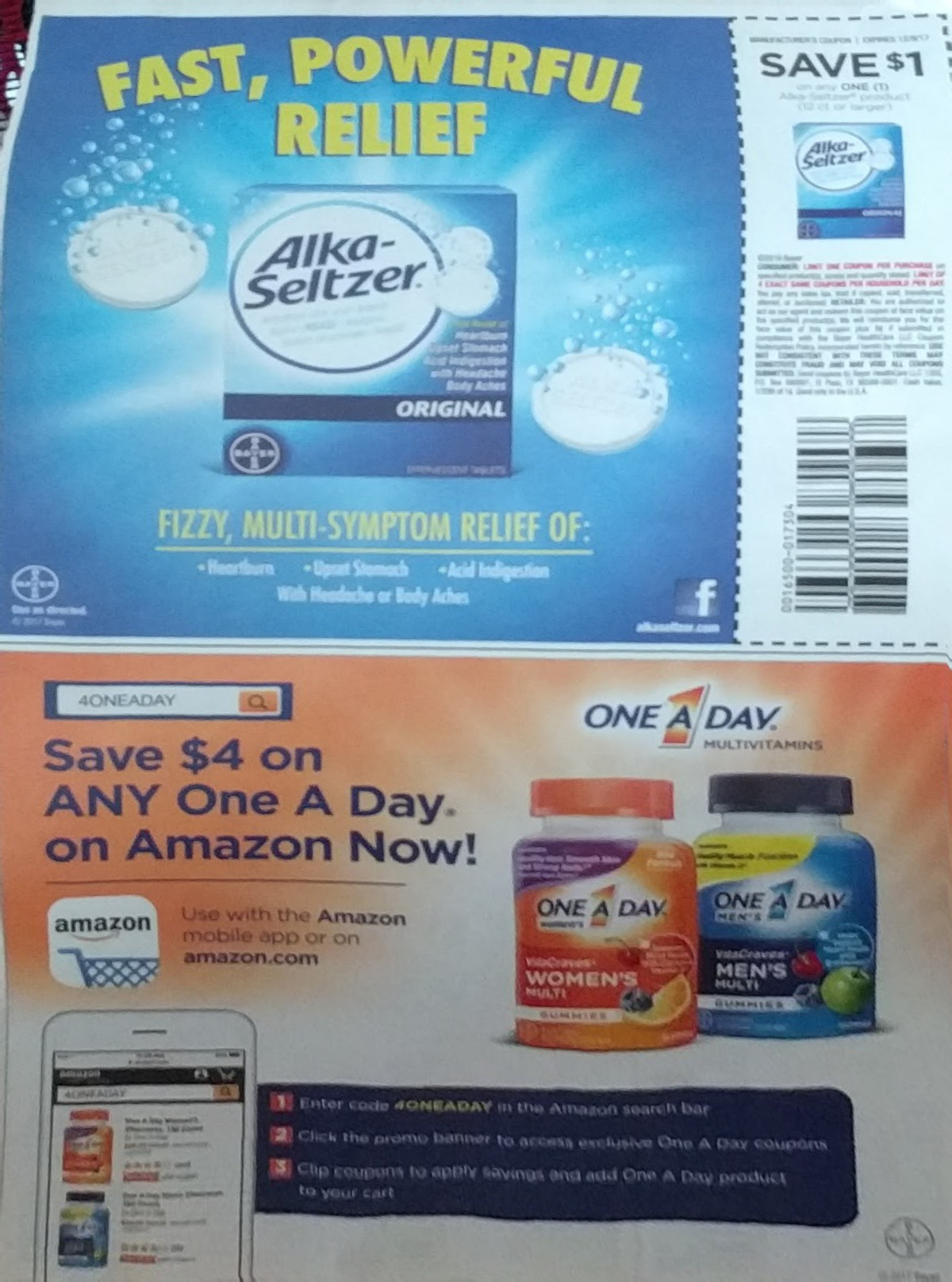} &
        \includegraphics[width=0.124\linewidth,height=0.1674\linewidth]{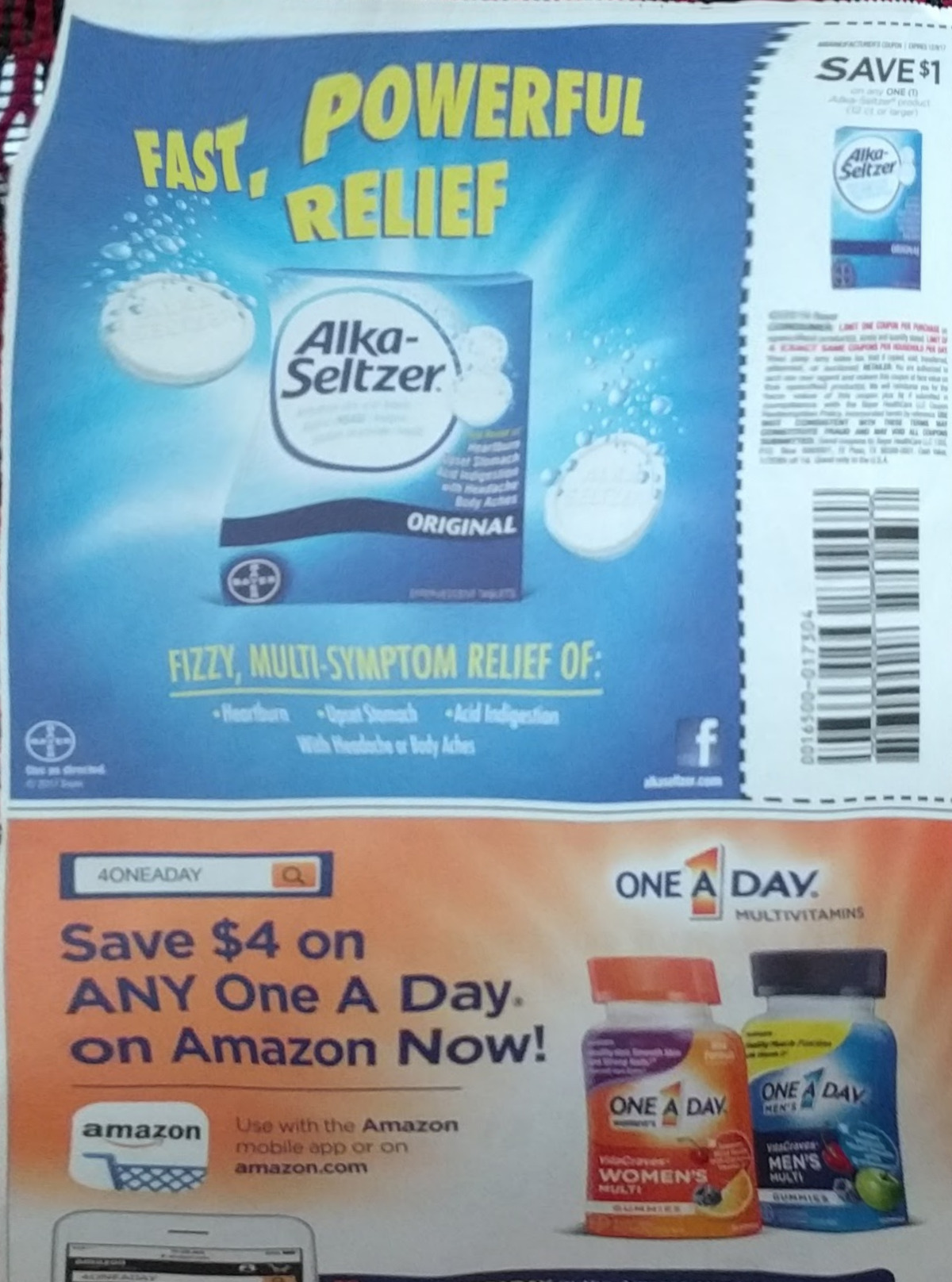} &
        \includegraphics[width=0.124\linewidth,height=0.1674\linewidth]{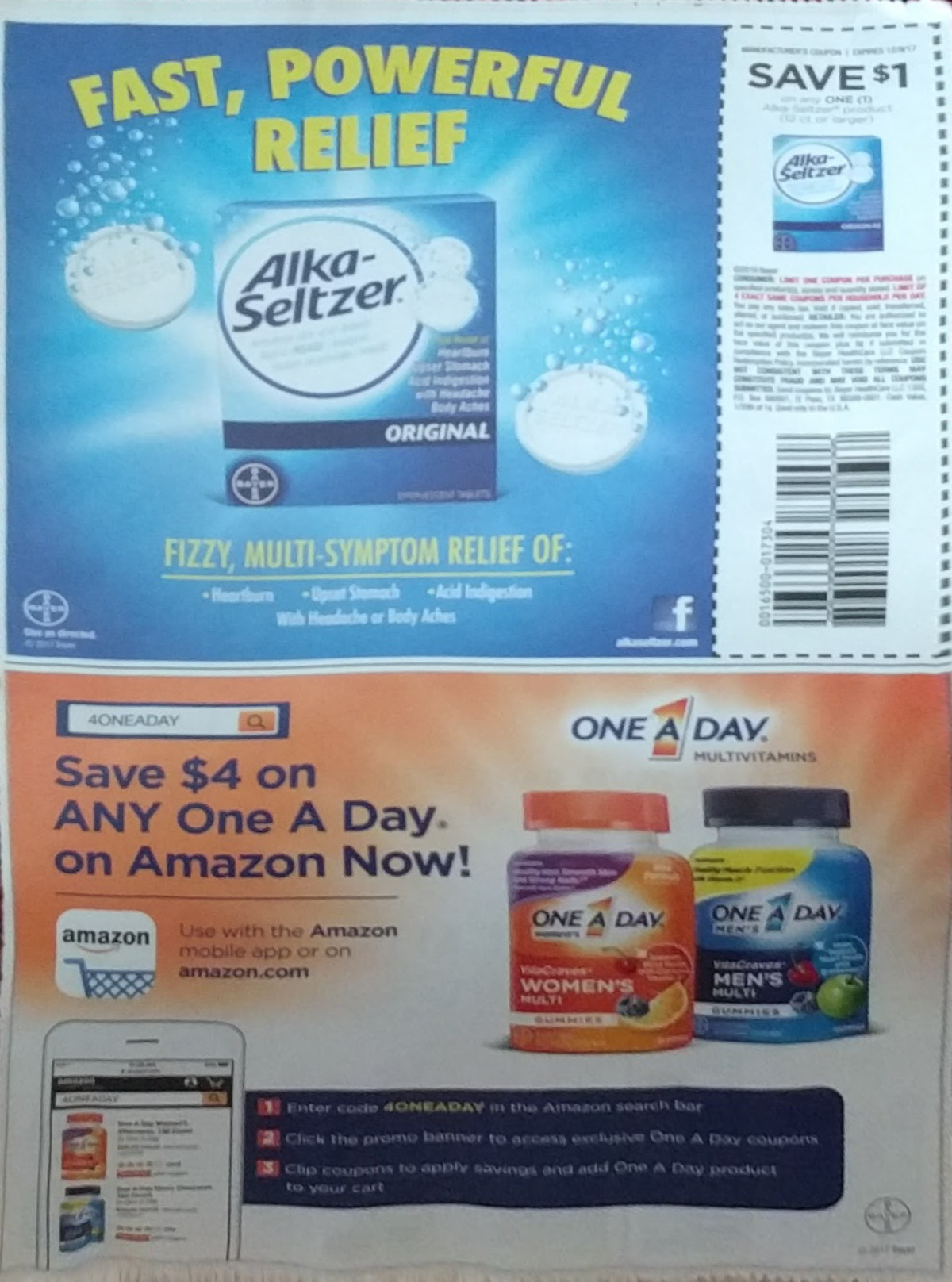} & 
        \includegraphics[width=0.124\linewidth,height=0.1674\linewidth]{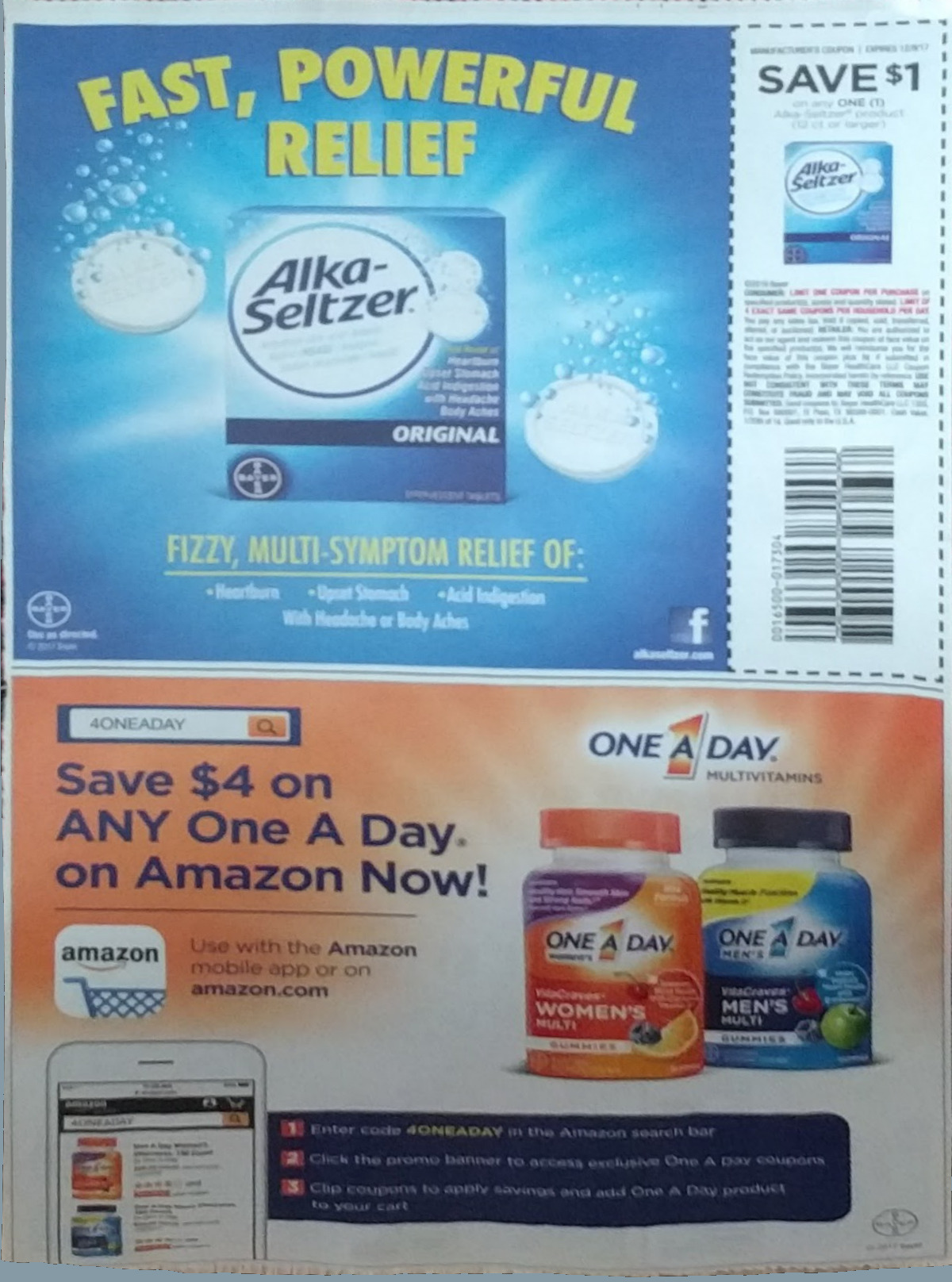} & 
        \includegraphics[width=0.124\linewidth,height=0.1674\linewidth]{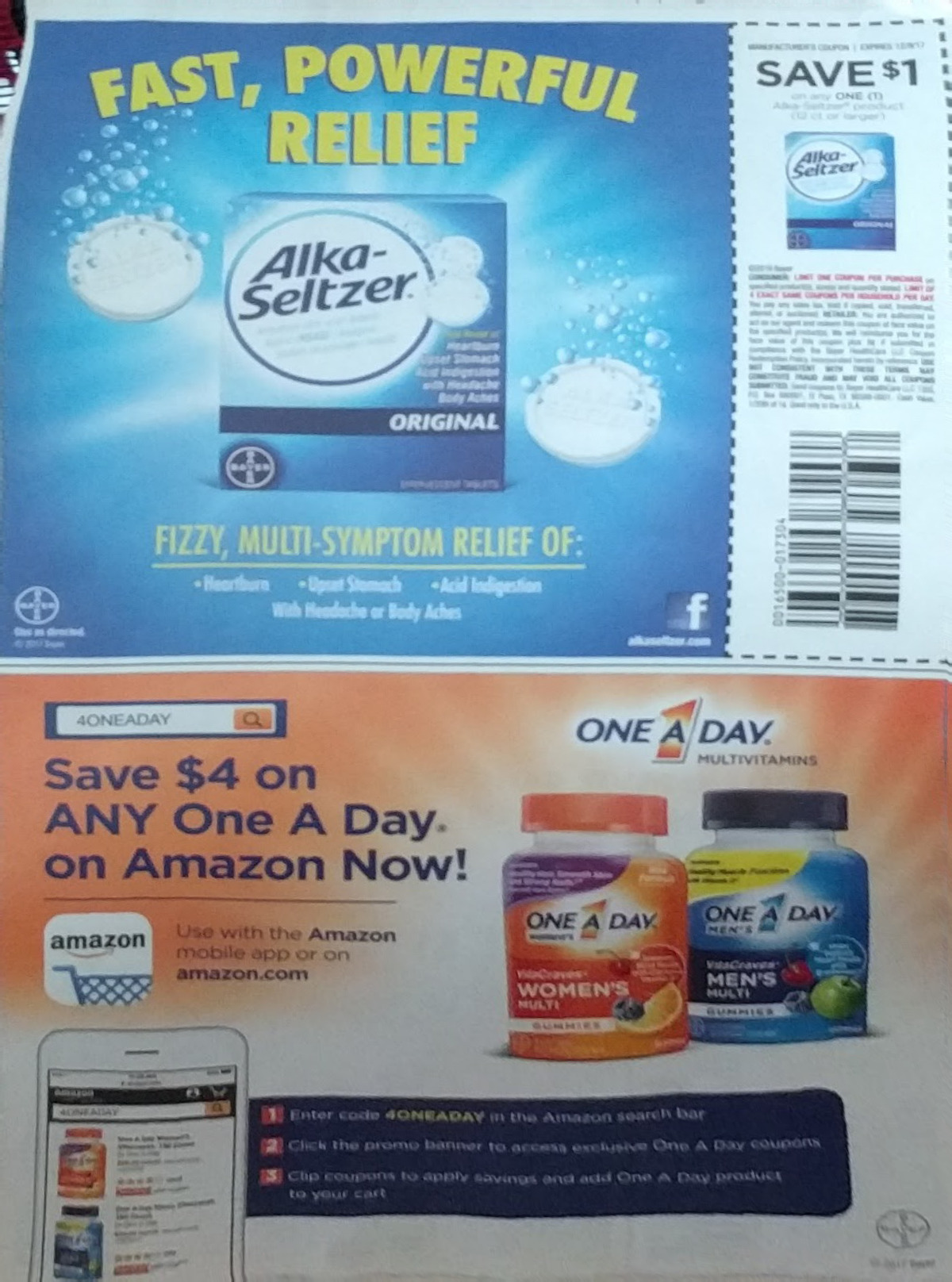} & 
        \includegraphics[width=0.124\linewidth,height=0.1674\linewidth]{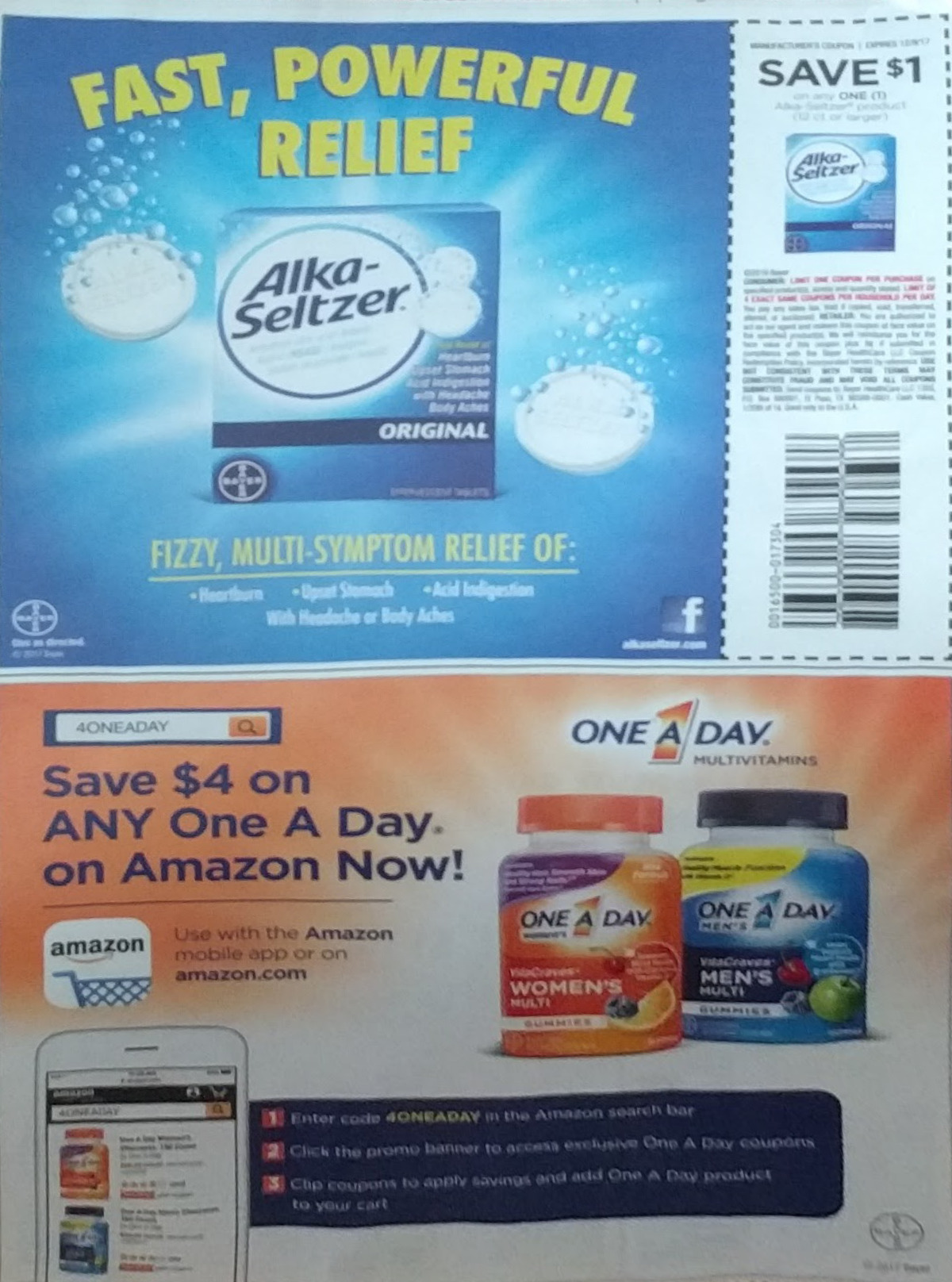} \\
        \includegraphics[width=0.124\linewidth,height=0.15\linewidth]{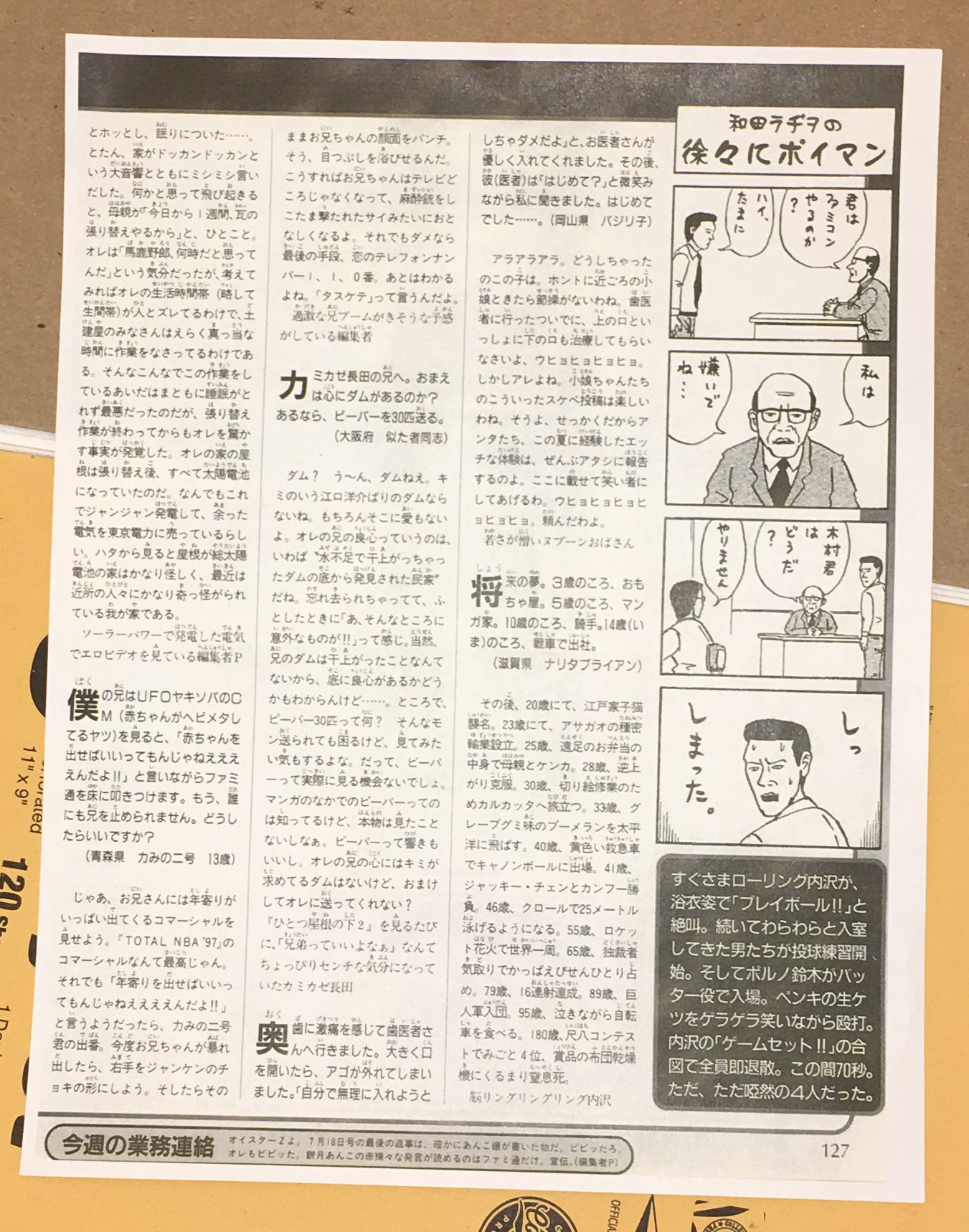} & 
        \includegraphics[width=0.124\linewidth,height=0.15\linewidth]{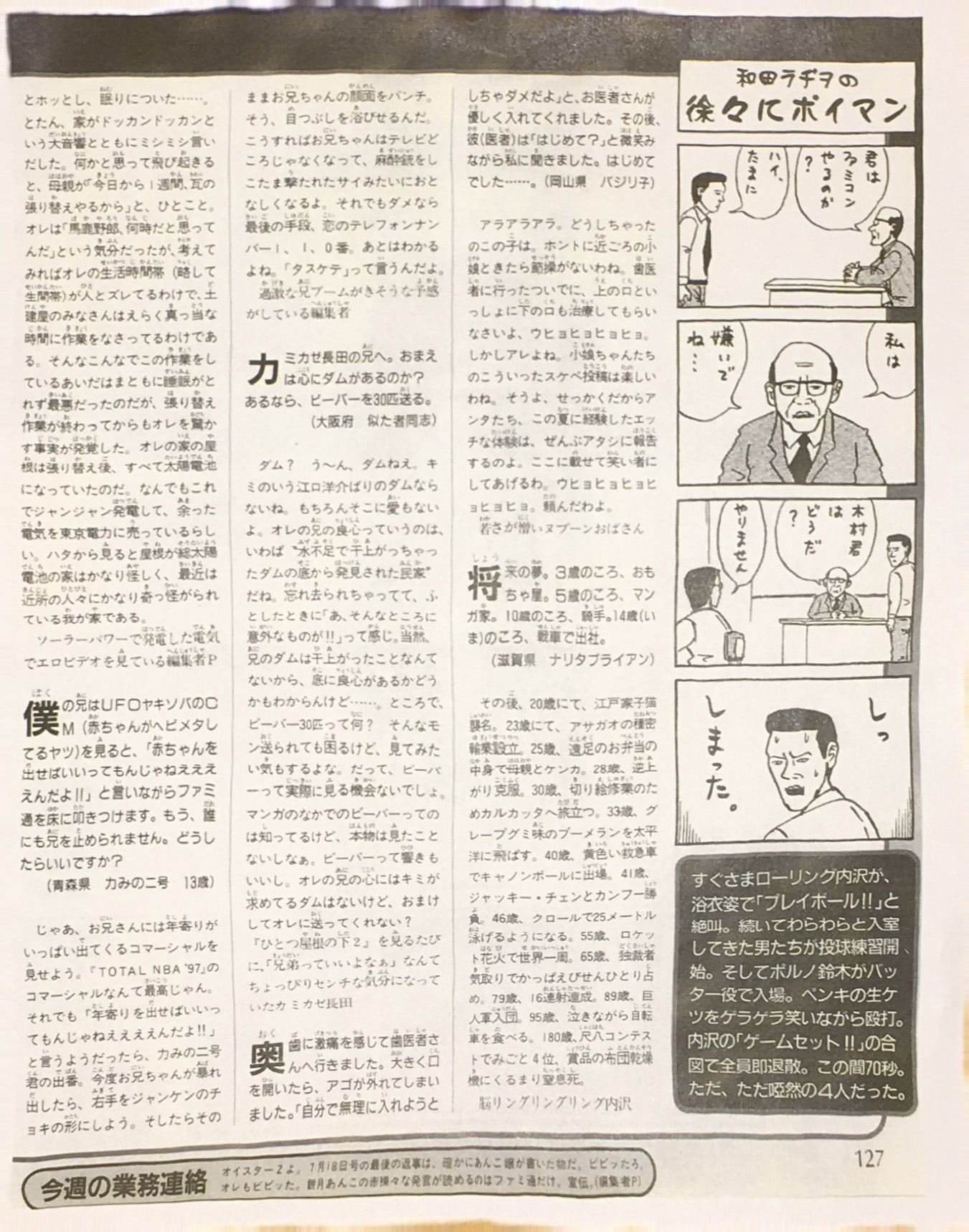} & 
        \includegraphics[width=0.124\linewidth,height=0.15\linewidth]{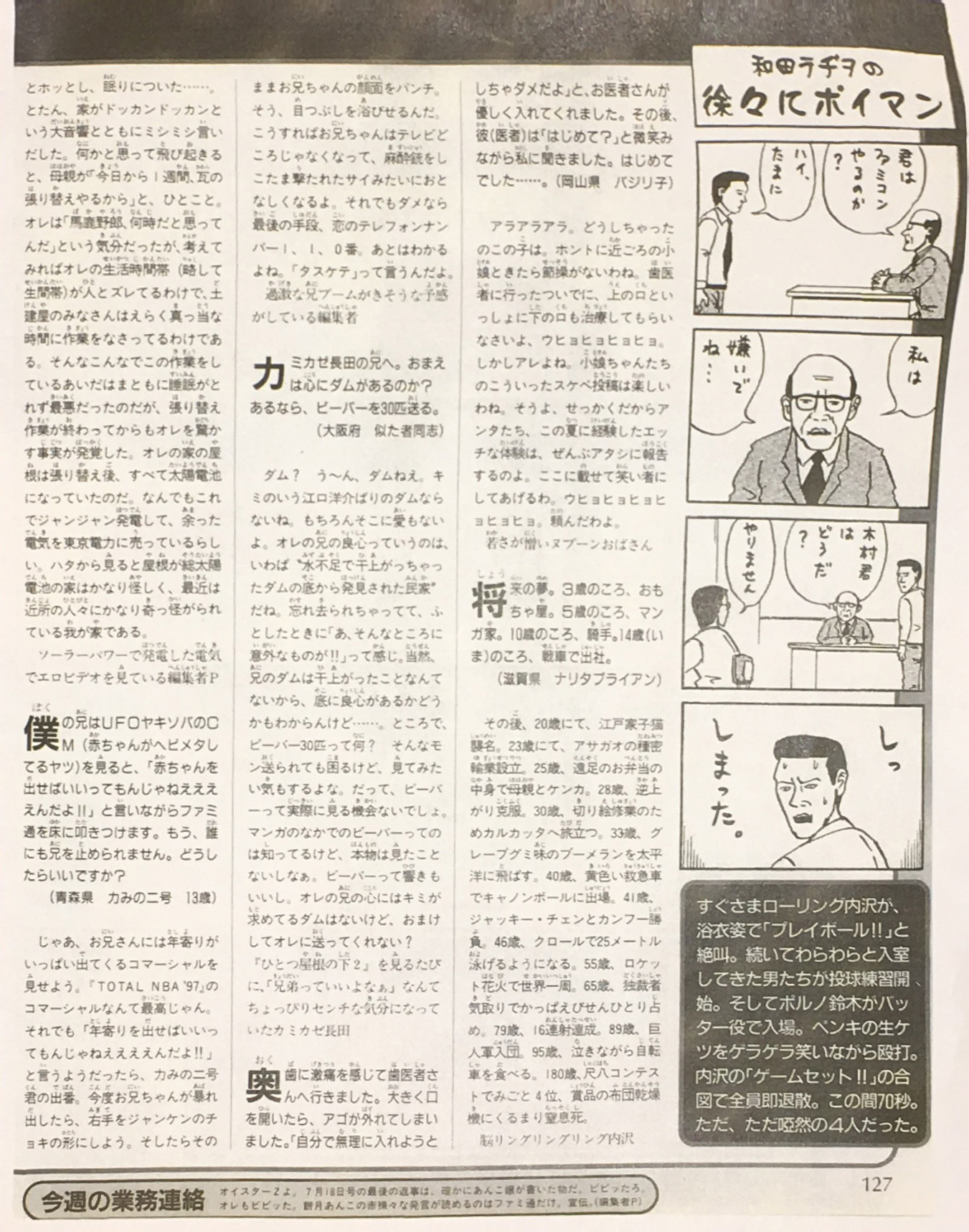} &
        \includegraphics[width=0.124\linewidth,height=0.15\linewidth]{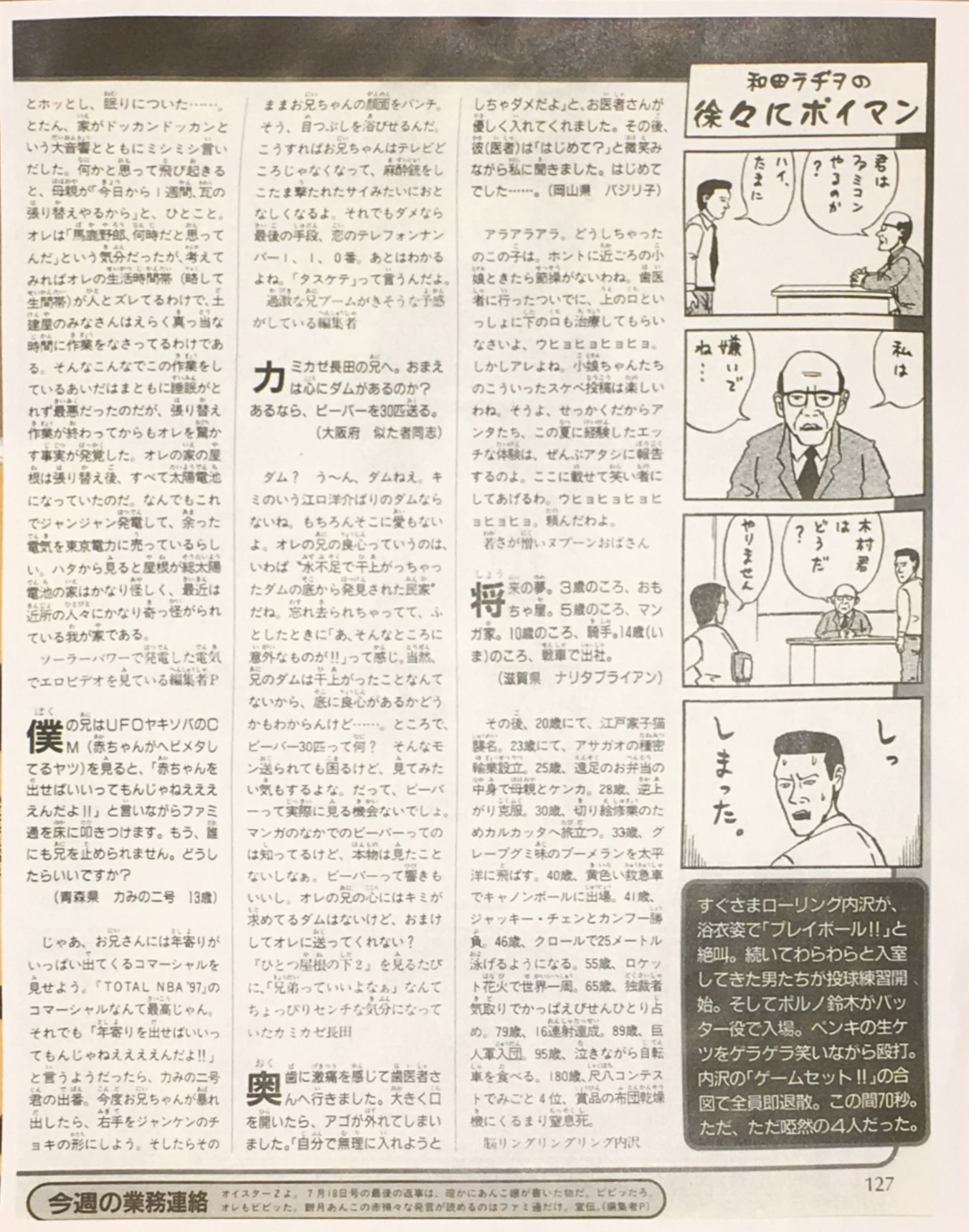} &
        \includegraphics[width=0.124\linewidth,height=0.15\linewidth]{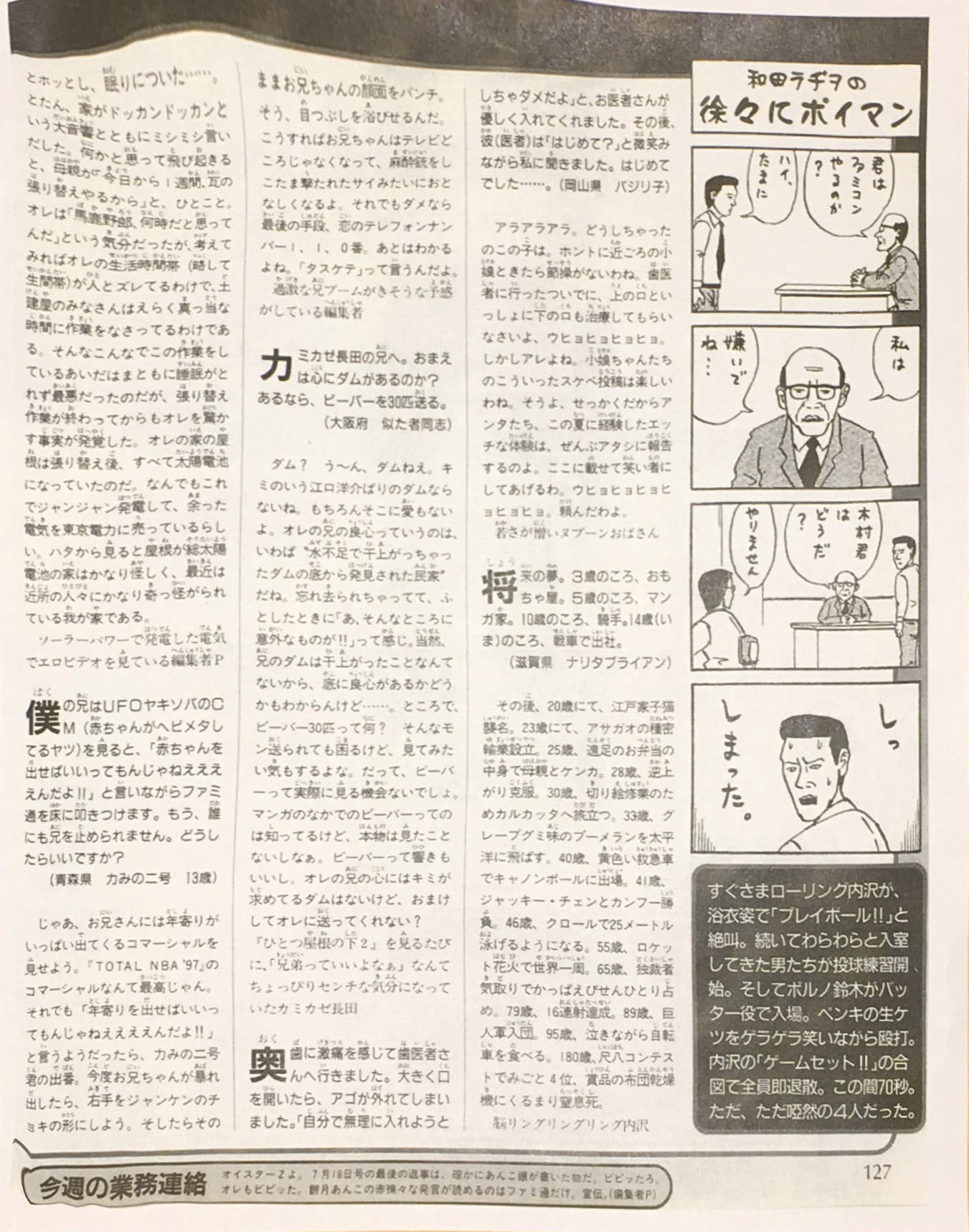} & 
        \includegraphics[width=0.124\linewidth,height=0.15\linewidth]{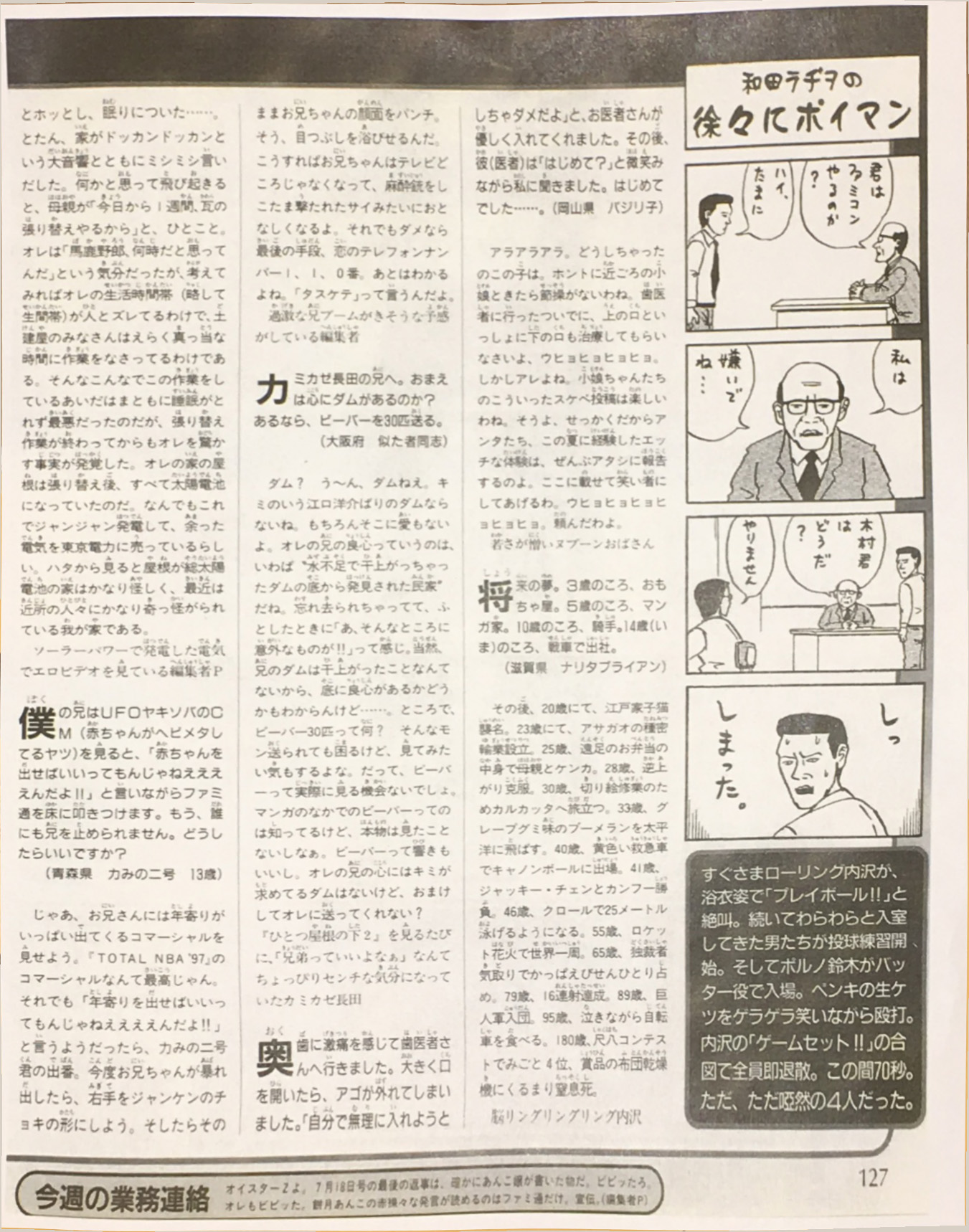} & 
        \includegraphics[width=0.124\linewidth,height=0.15\linewidth]{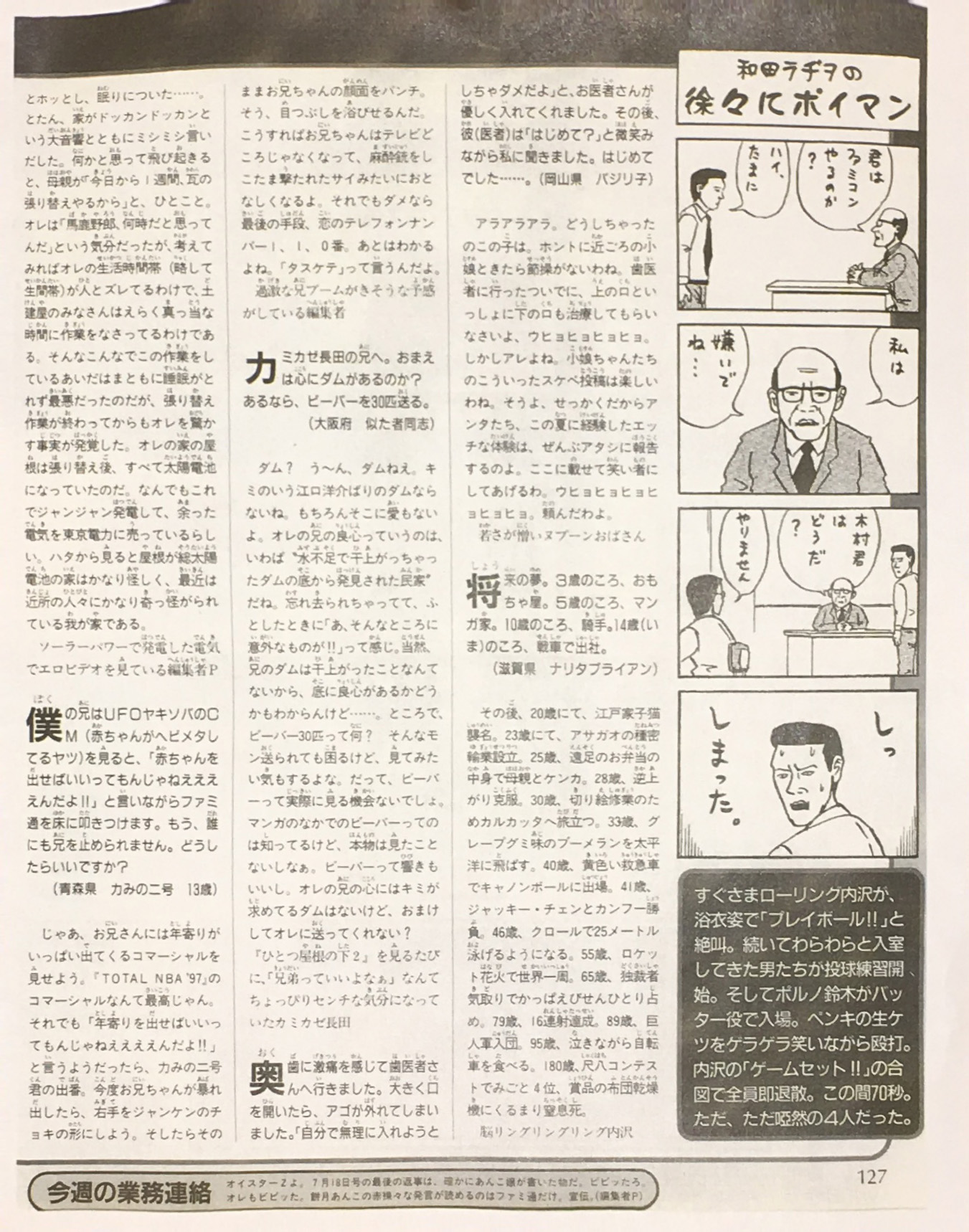} & 
        \includegraphics[width=0.124\linewidth,height=0.15\linewidth]{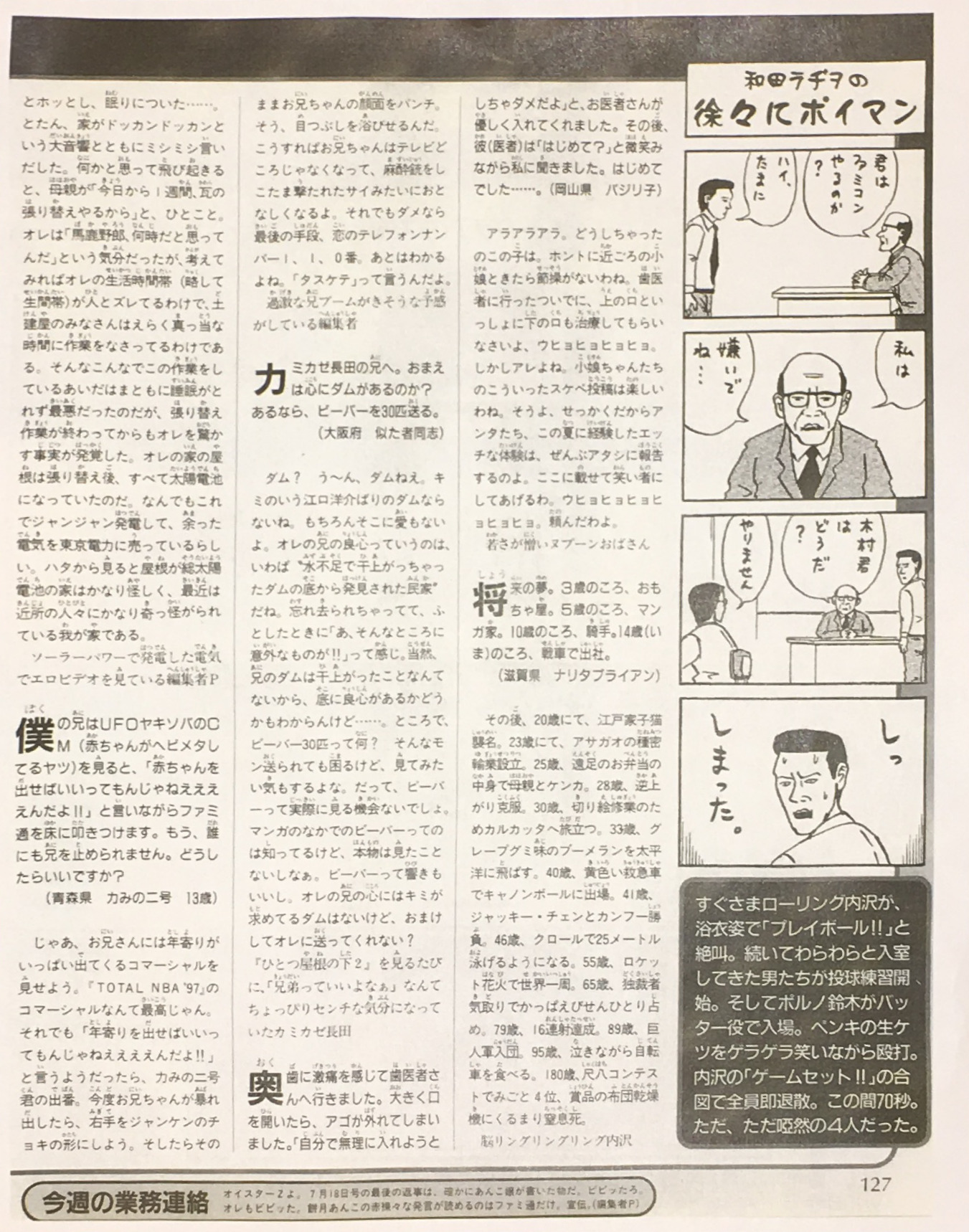} \\
         \includegraphics[width=0.124\linewidth,height=0.148\linewidth]{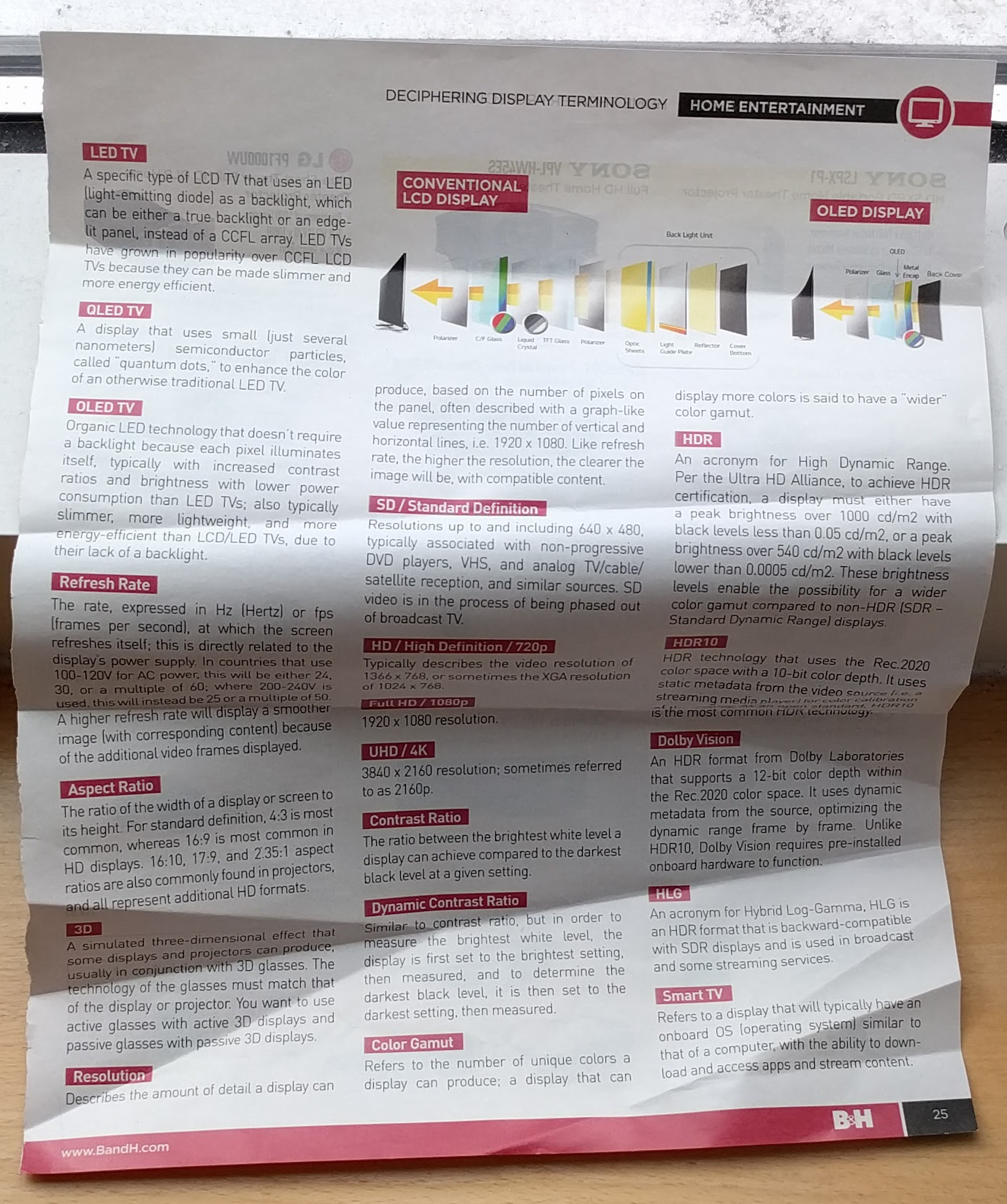} & 
        \includegraphics[width=0.124\linewidth,height=0.148\linewidth]{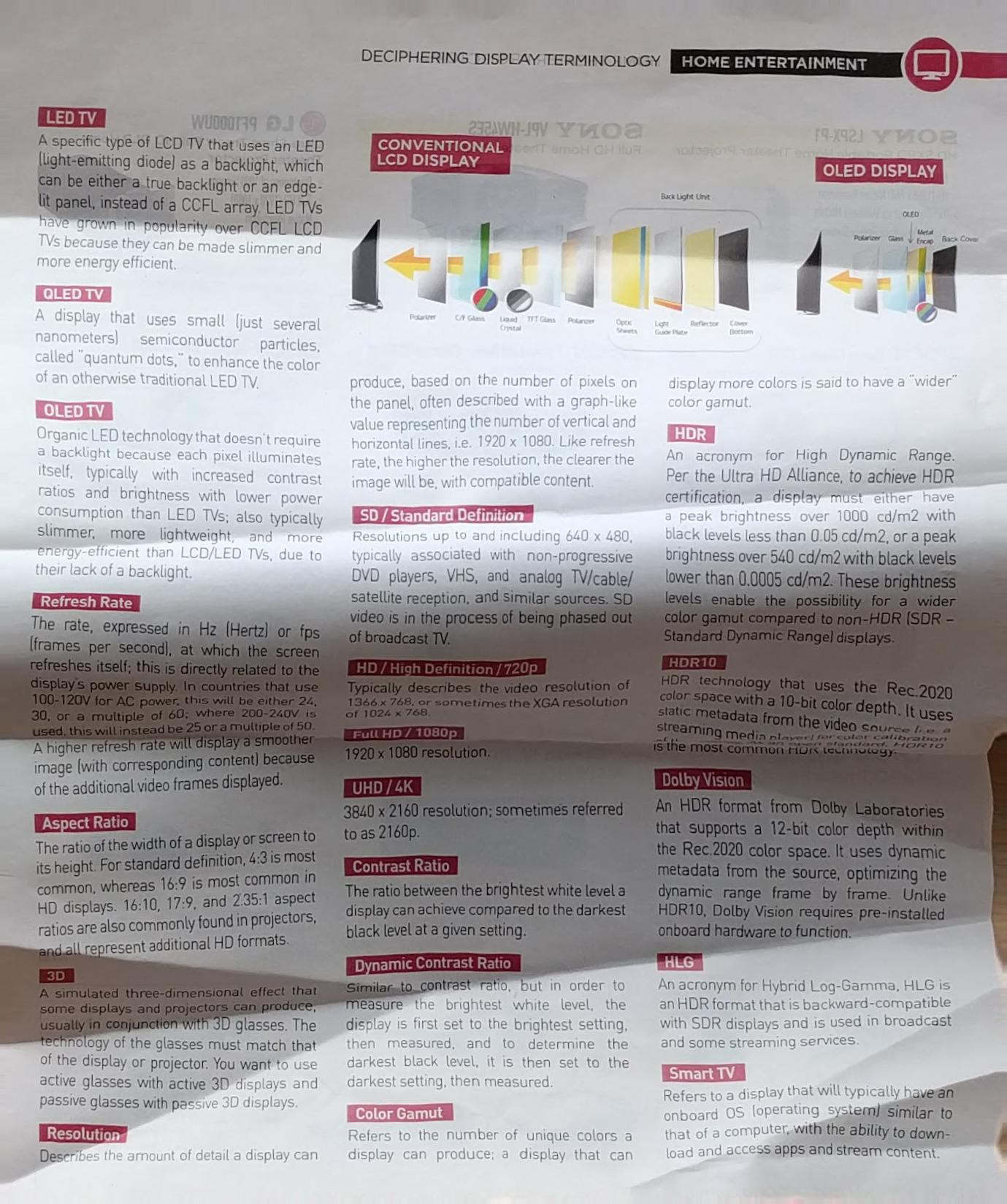} & 
        \includegraphics[width=0.124\linewidth,height=0.148\linewidth]{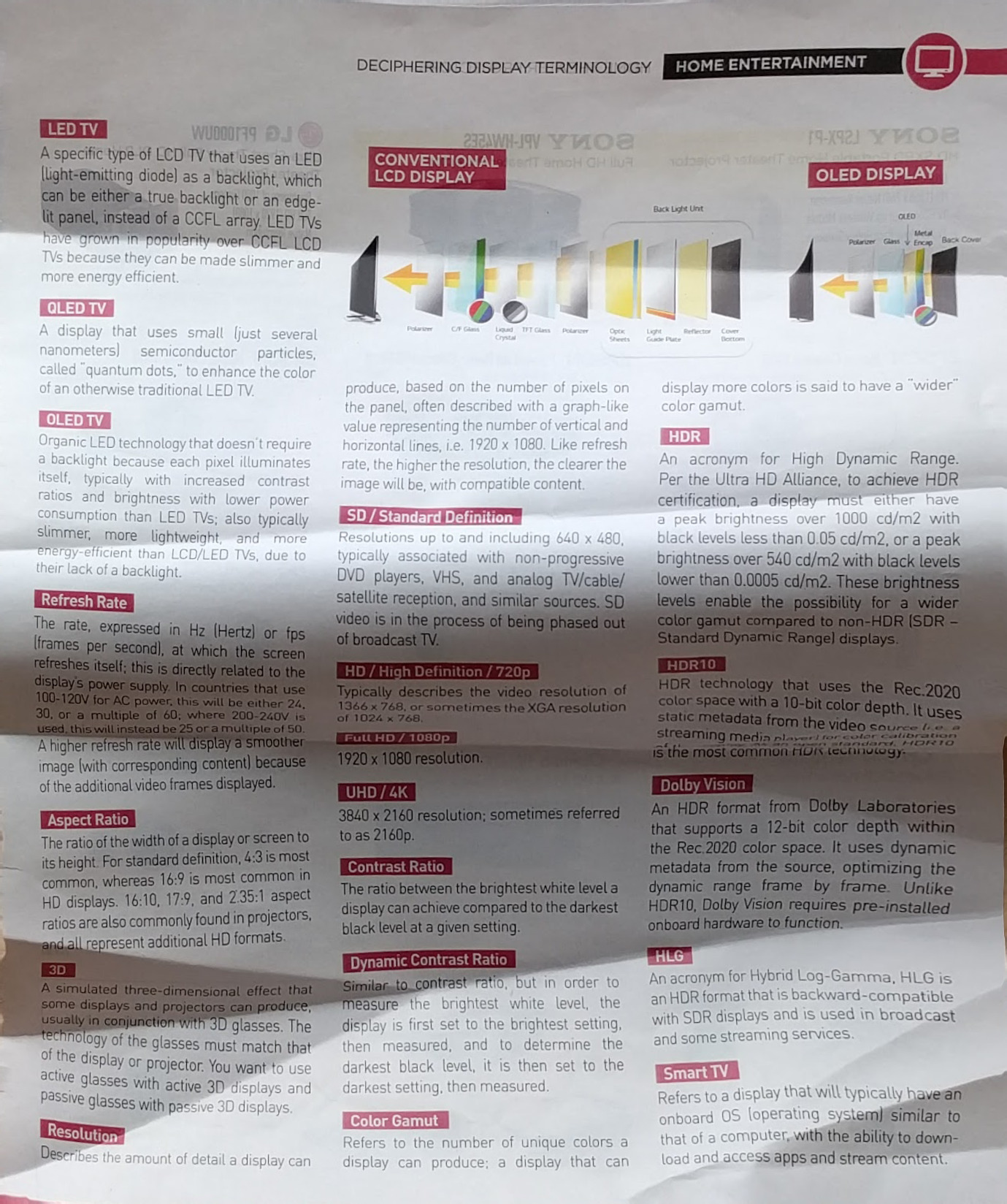} &
        \includegraphics[width=0.124\linewidth,height=0.148\linewidth]{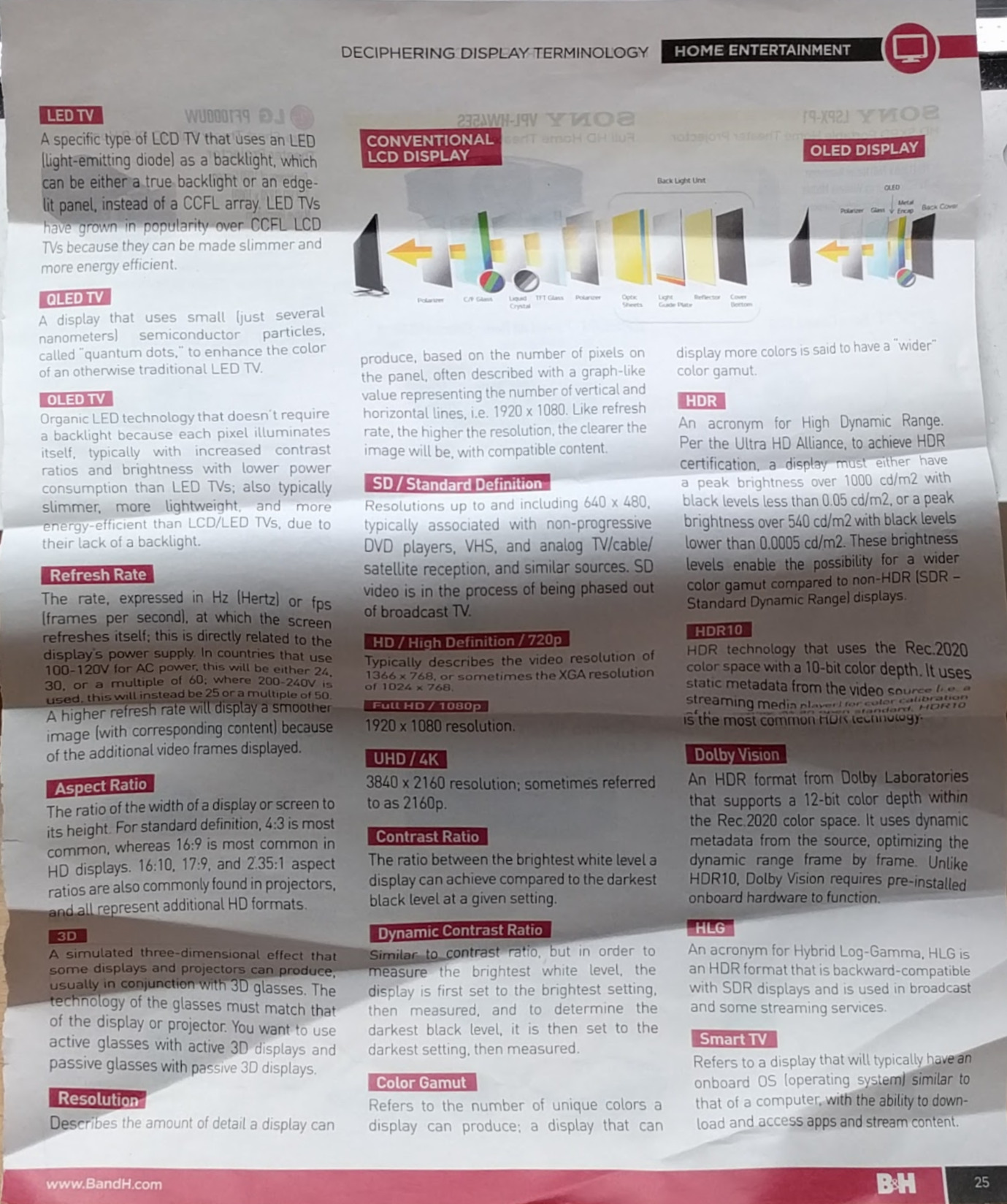} &
        \includegraphics[width=0.124\linewidth,height=0.148\linewidth]{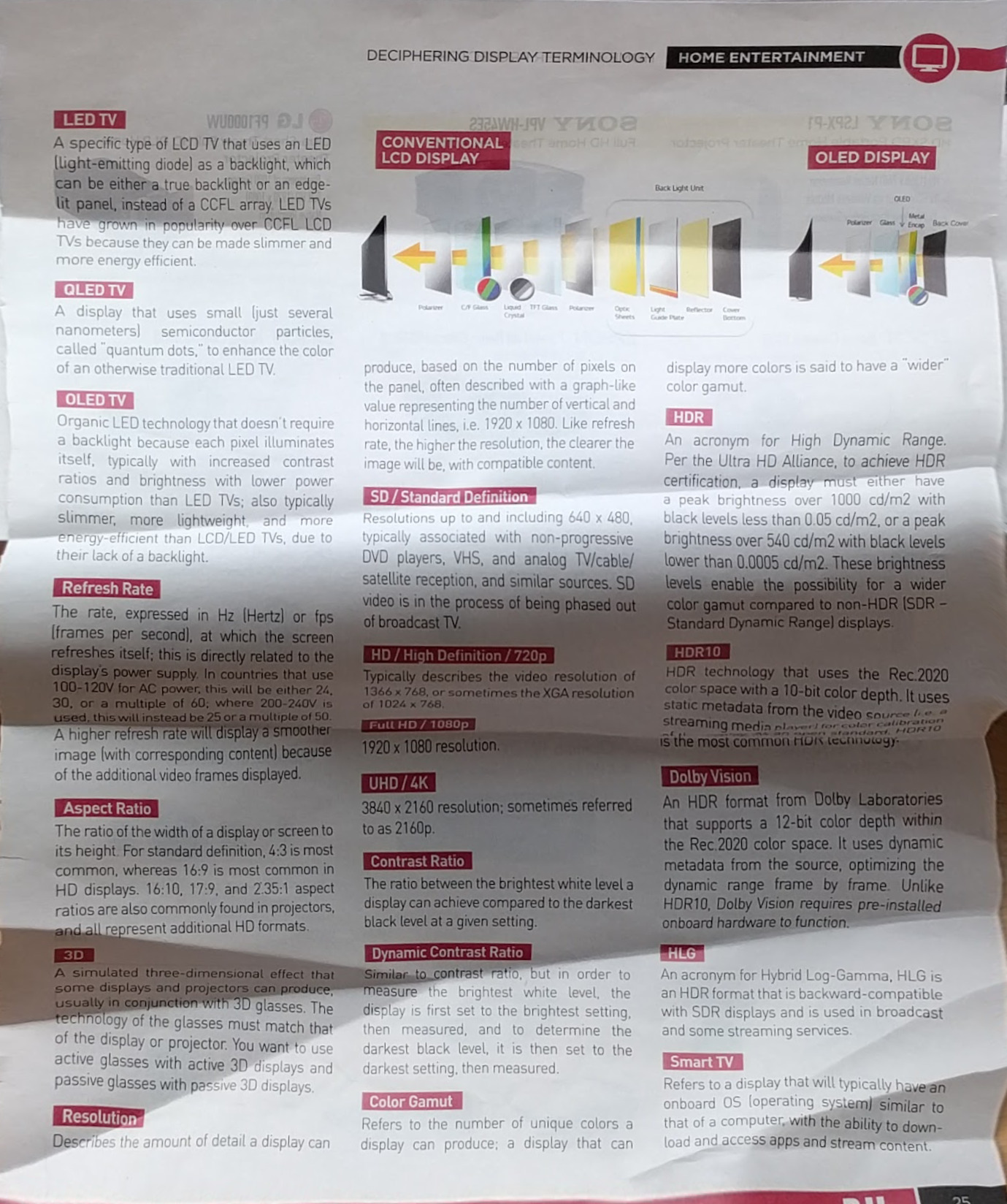} & 
        \includegraphics[width=0.124\linewidth,height=0.148\linewidth]{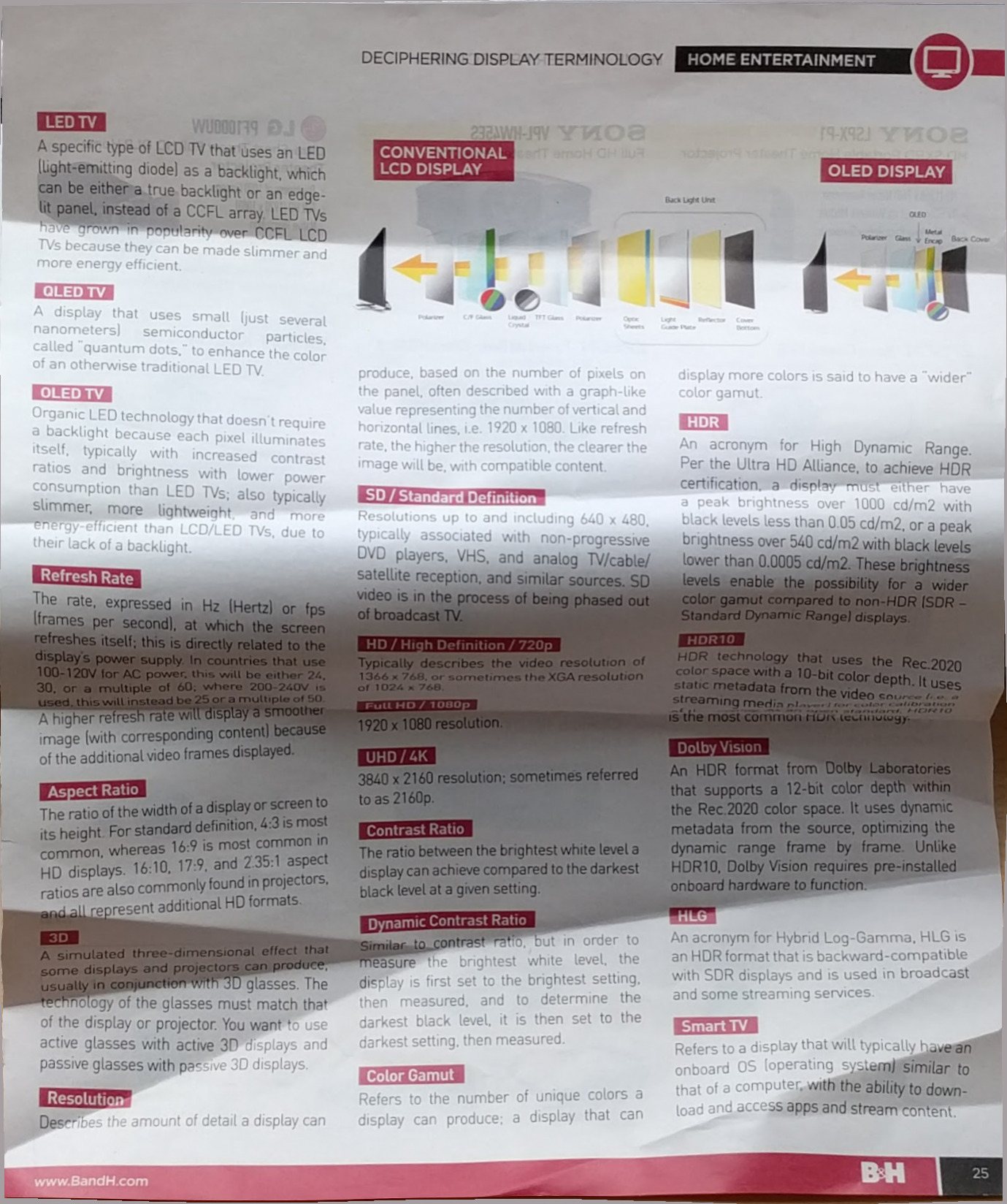} & 
        \includegraphics[width=0.124\linewidth,height=0.148\linewidth]{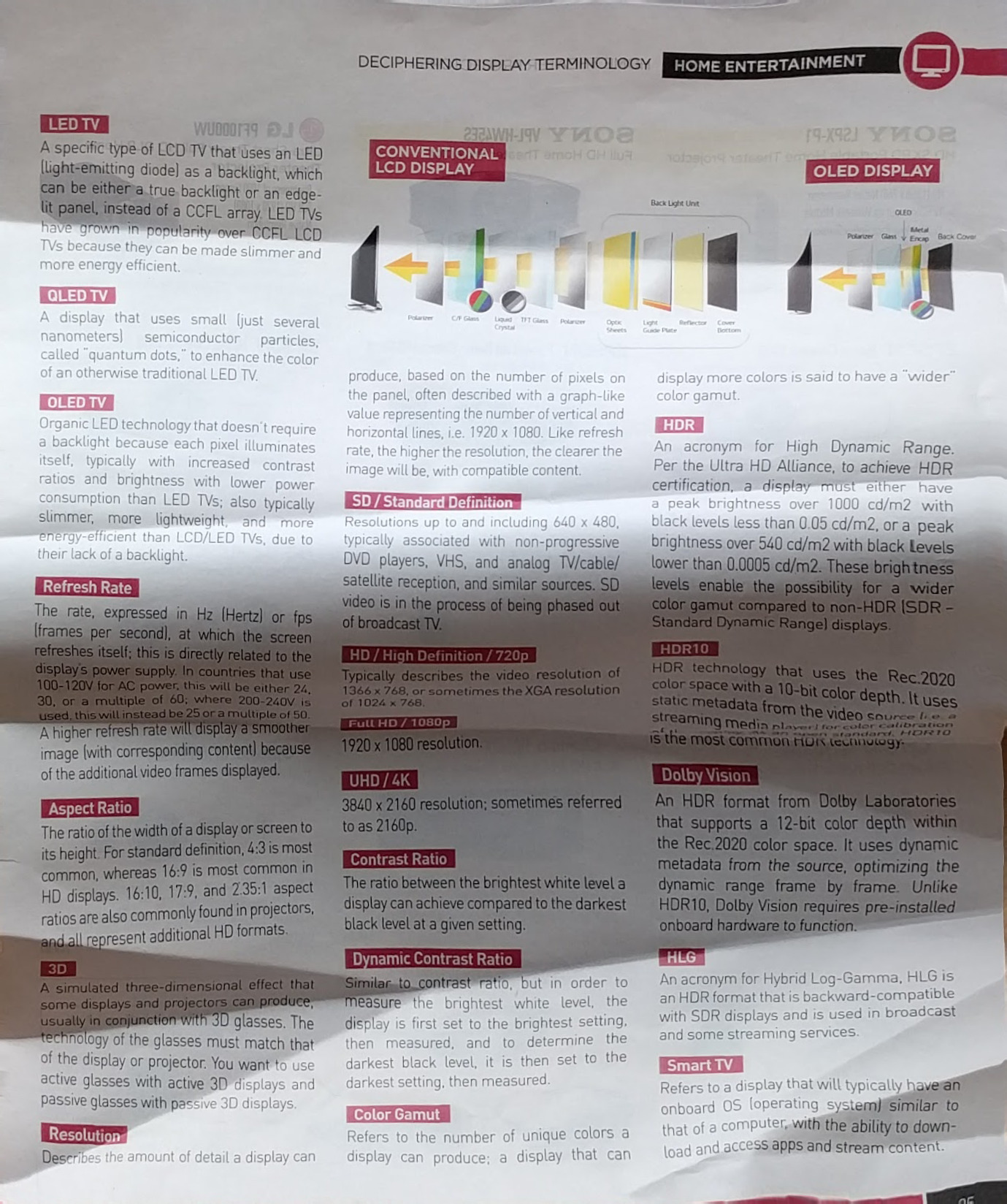} & 
        \includegraphics[width=0.124\linewidth,height=0.148\linewidth]{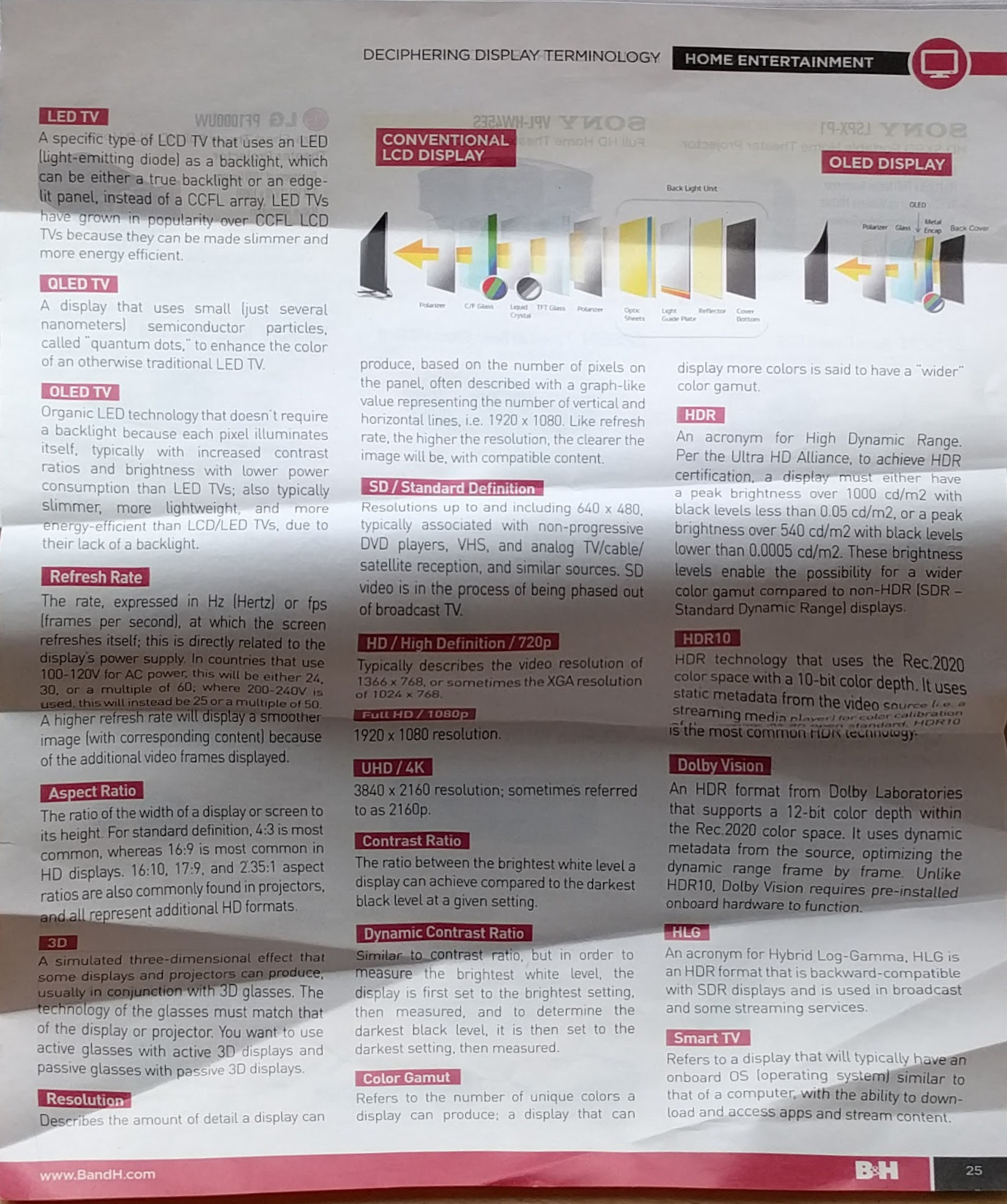} \\
        \includegraphics[width=0.124\linewidth,height=0.146\linewidth]{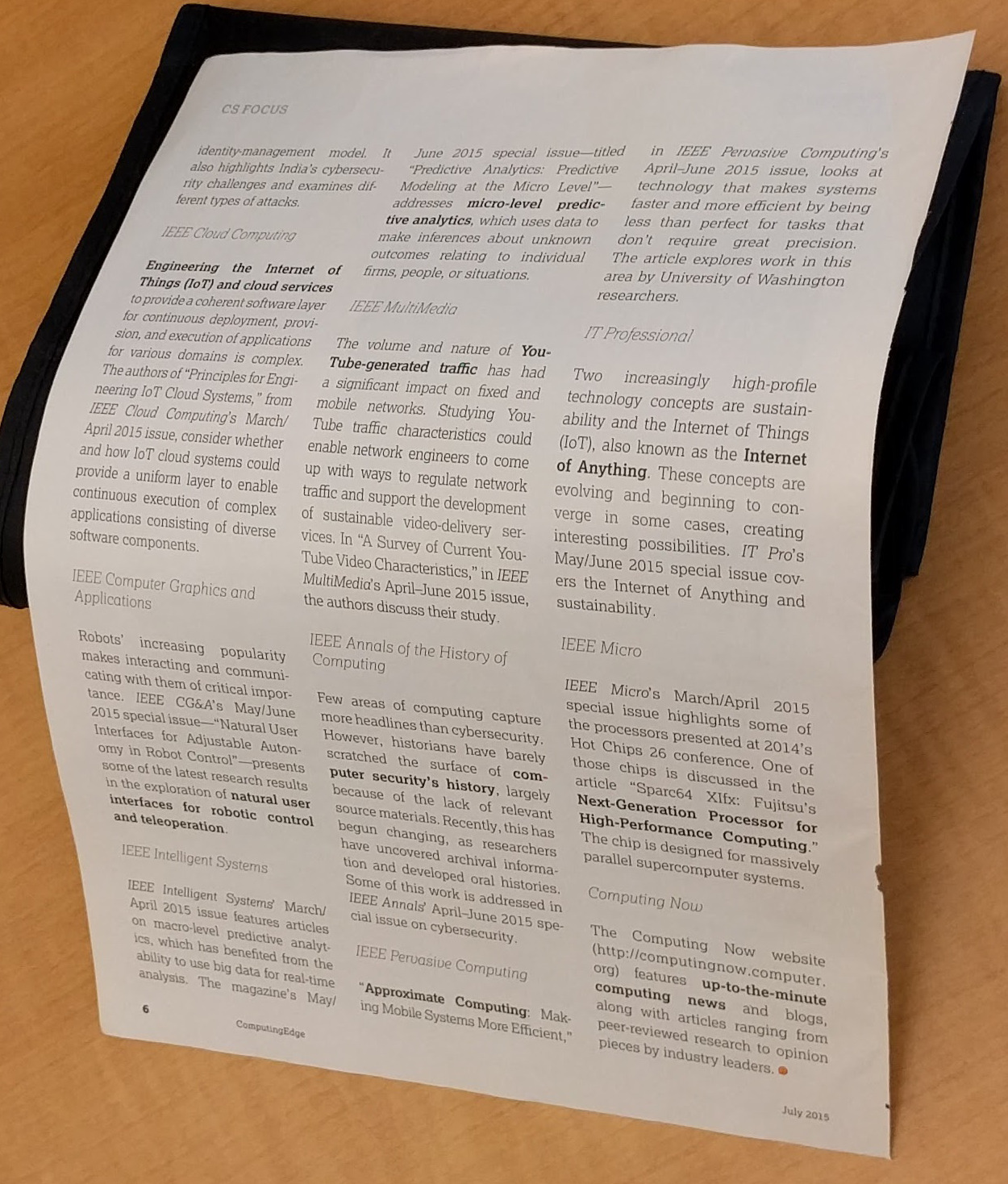} & 
        \includegraphics[width=0.124\linewidth,height=0.146\linewidth]{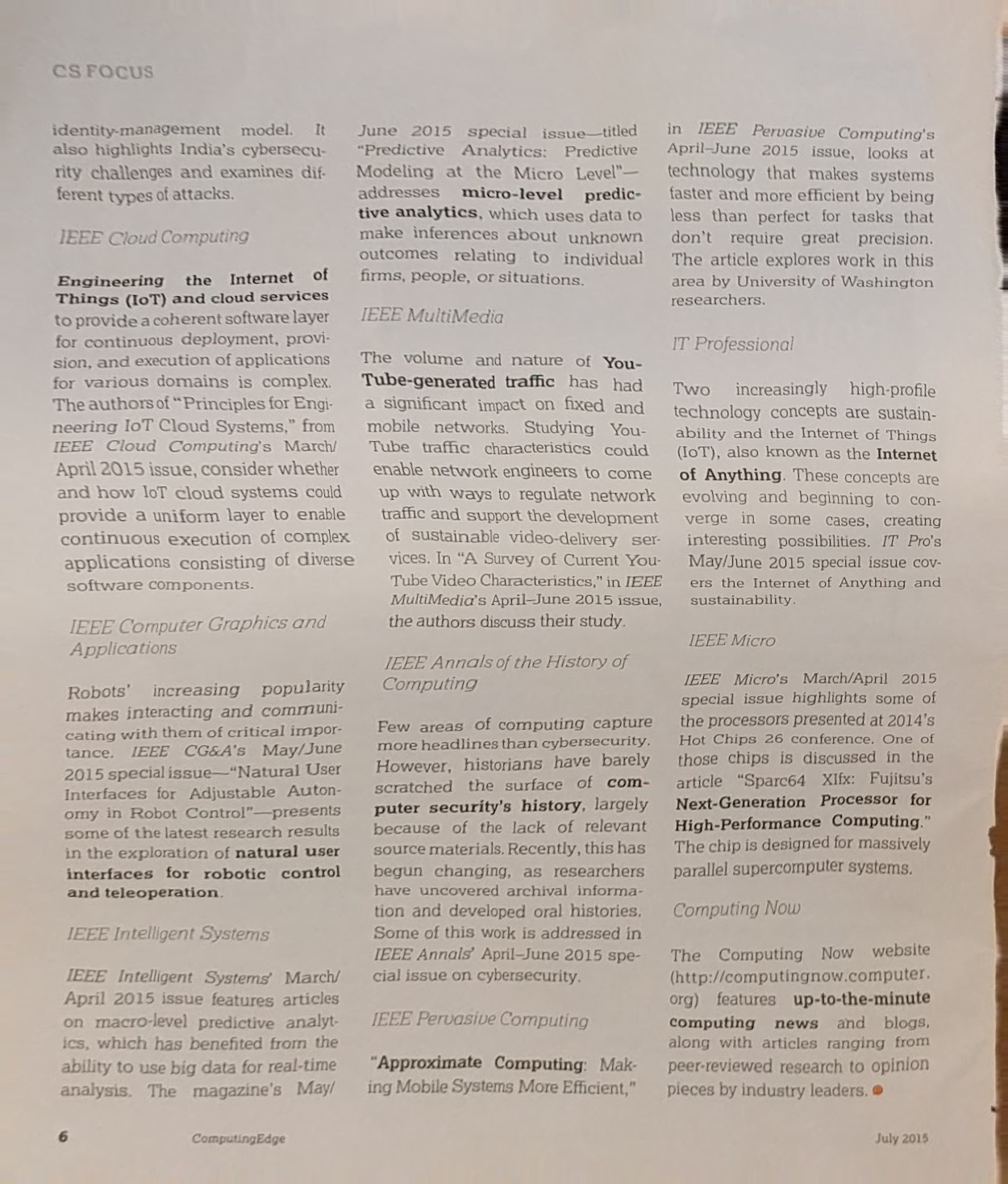} & 
        \includegraphics[width=0.124\linewidth,height=0.146\linewidth]{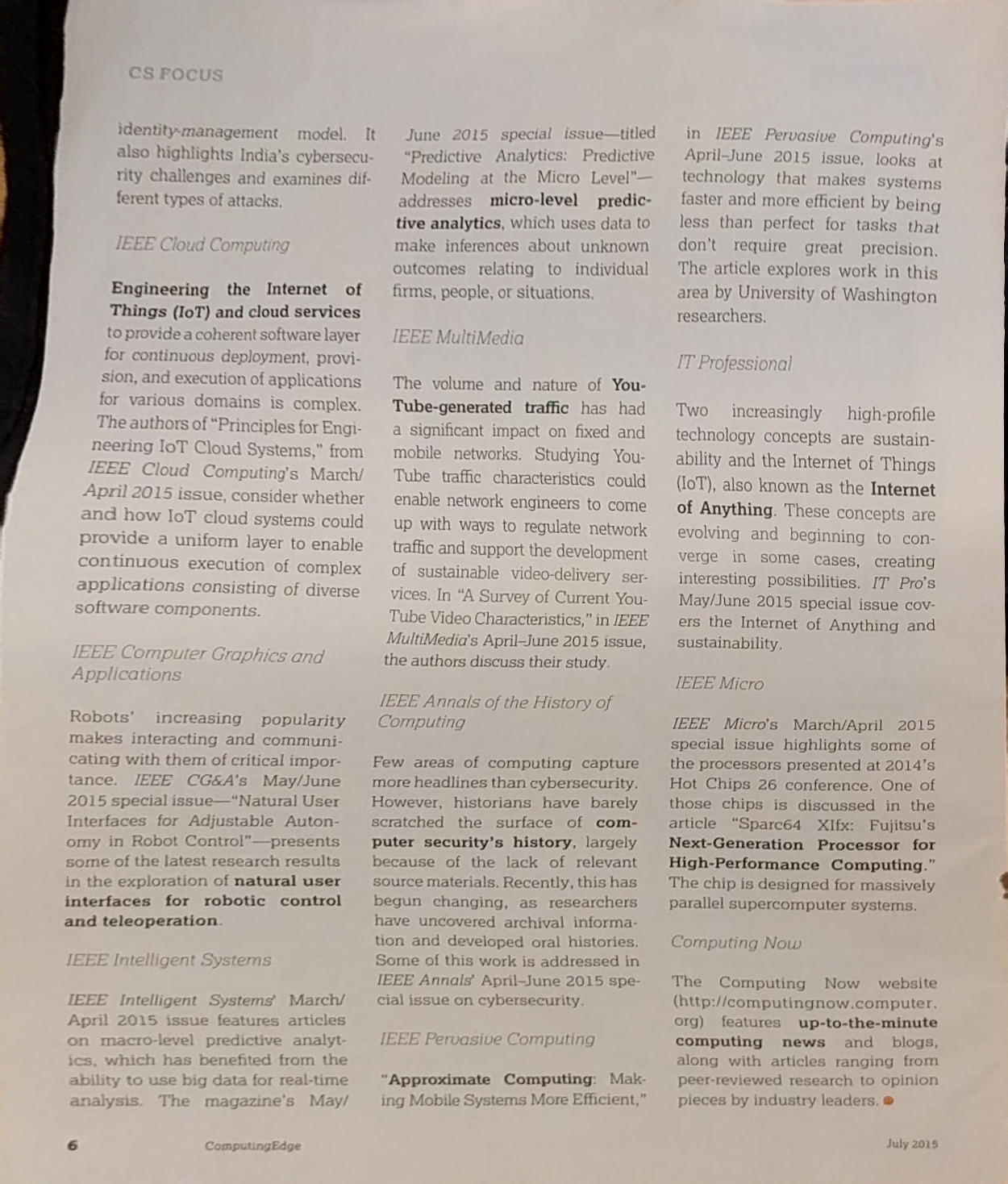} &
        \includegraphics[width=0.124\linewidth,height=0.146\linewidth]{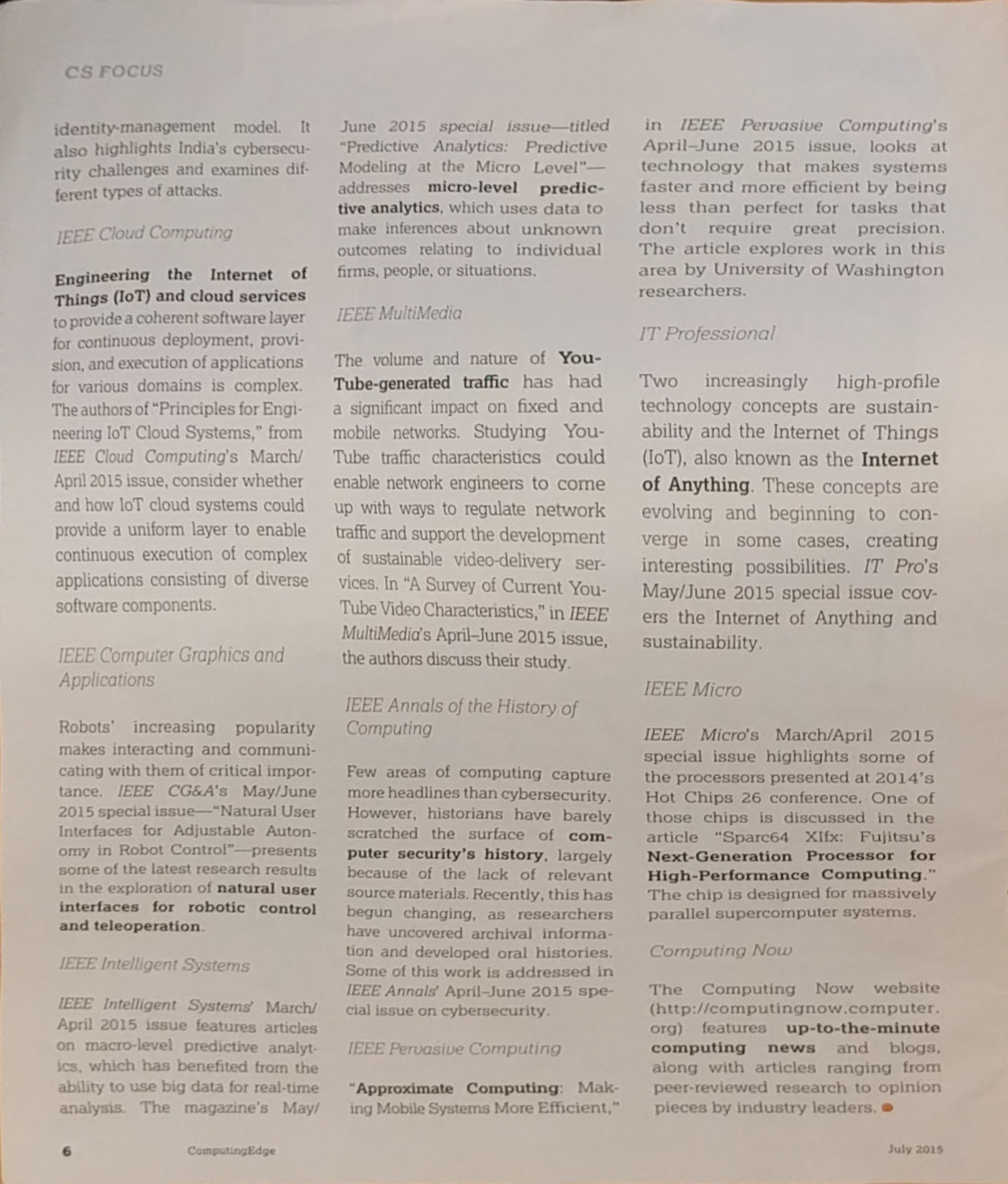} &
        \includegraphics[width=0.124\linewidth,height=0.146\linewidth]{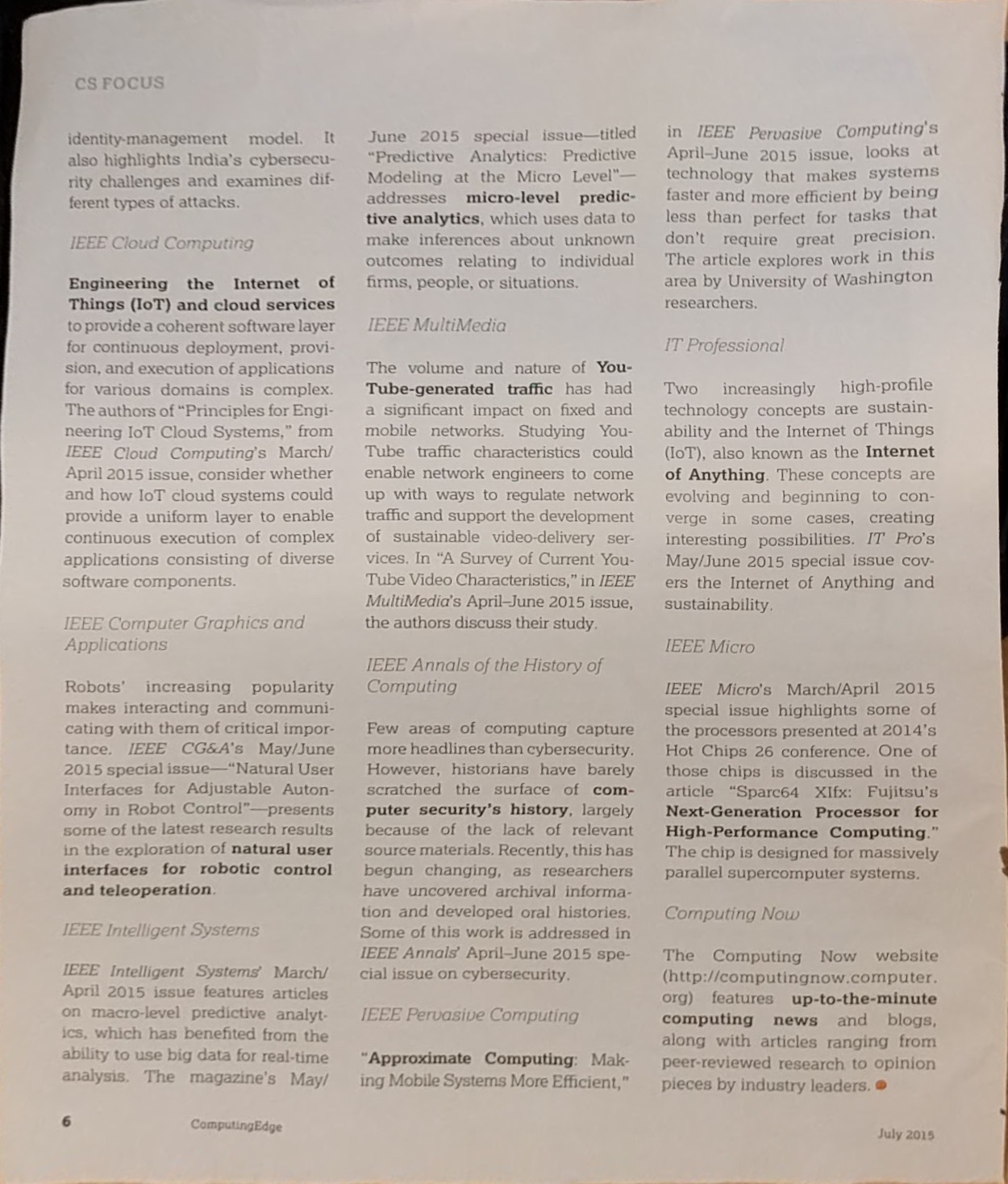} & 
        \includegraphics[width=0.124\linewidth,height=0.146\linewidth]{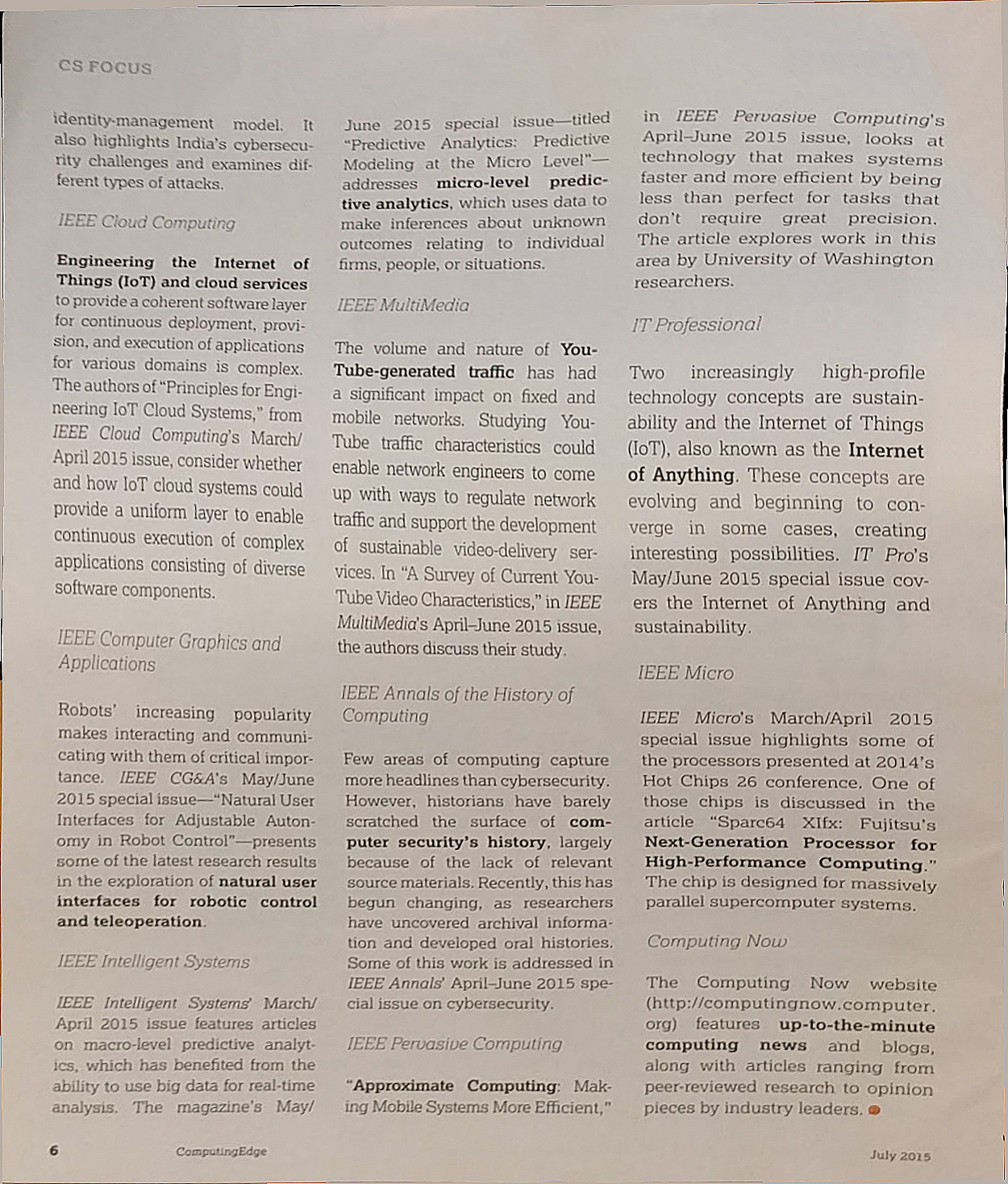} &
        \includegraphics[width=0.124\linewidth,height=0.146\linewidth]{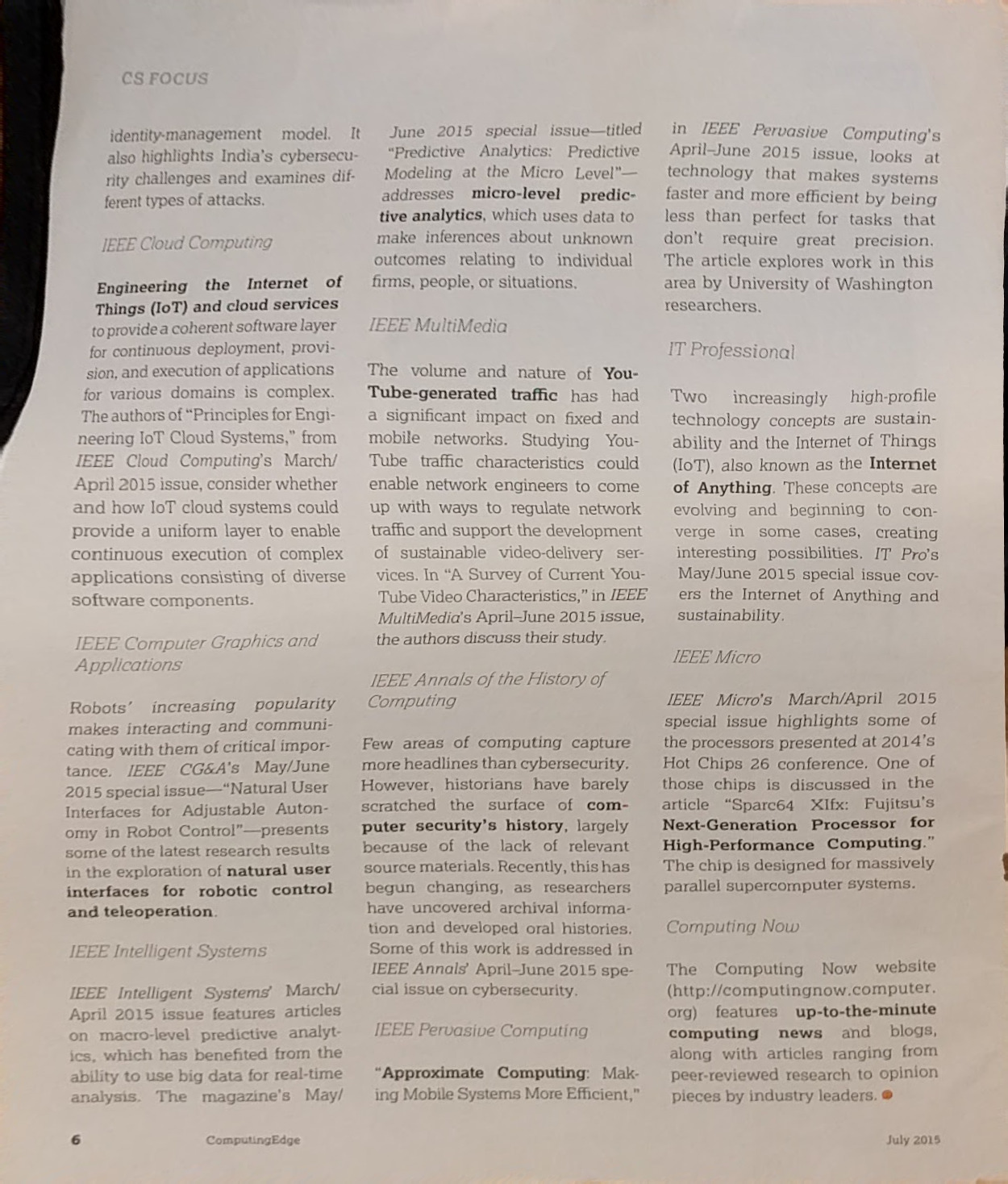} & 
        \includegraphics[width=0.124\linewidth,height=0.146\linewidth]{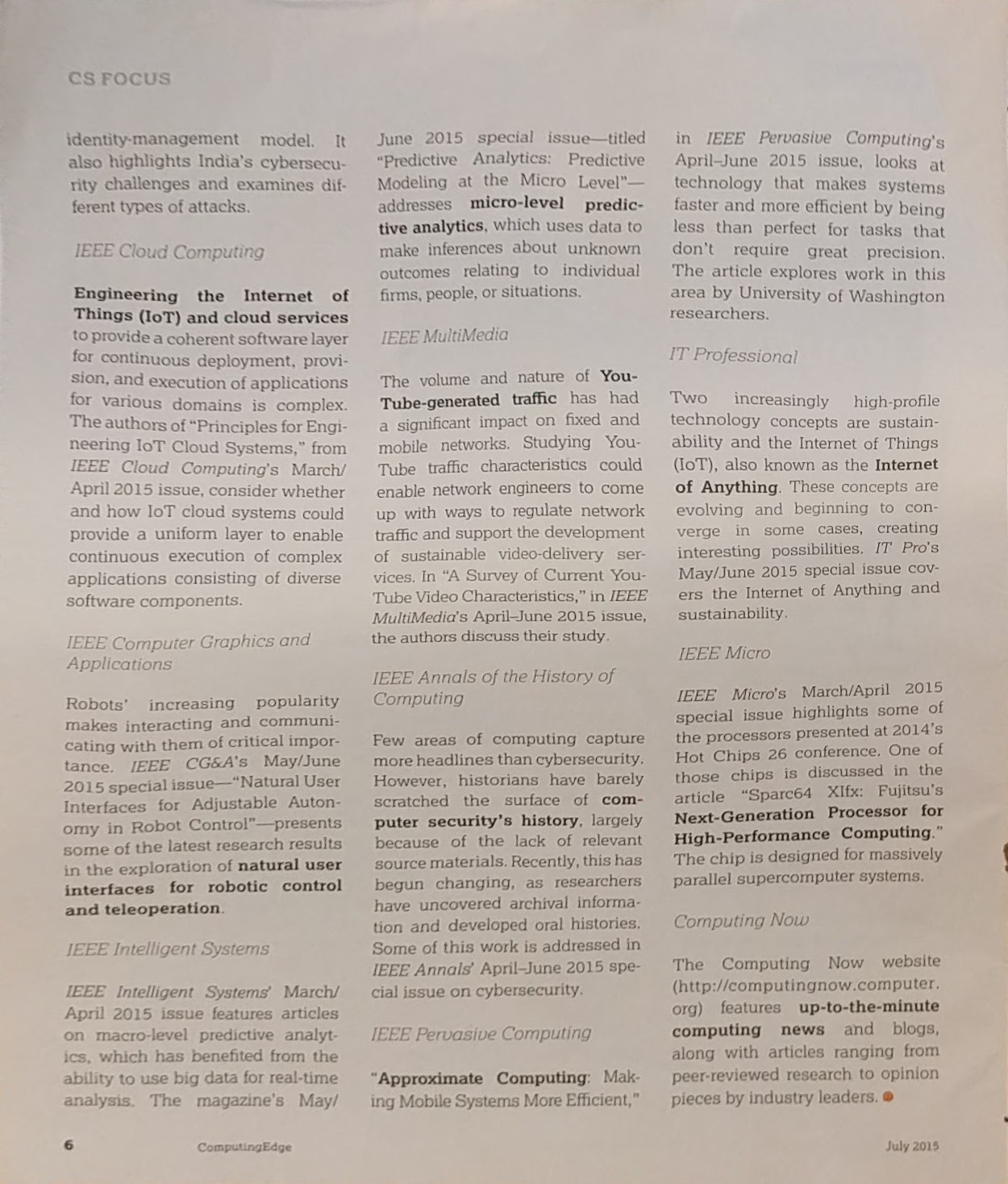} \\
        \includegraphics[width=0.124\linewidth,height=0.15\linewidth]{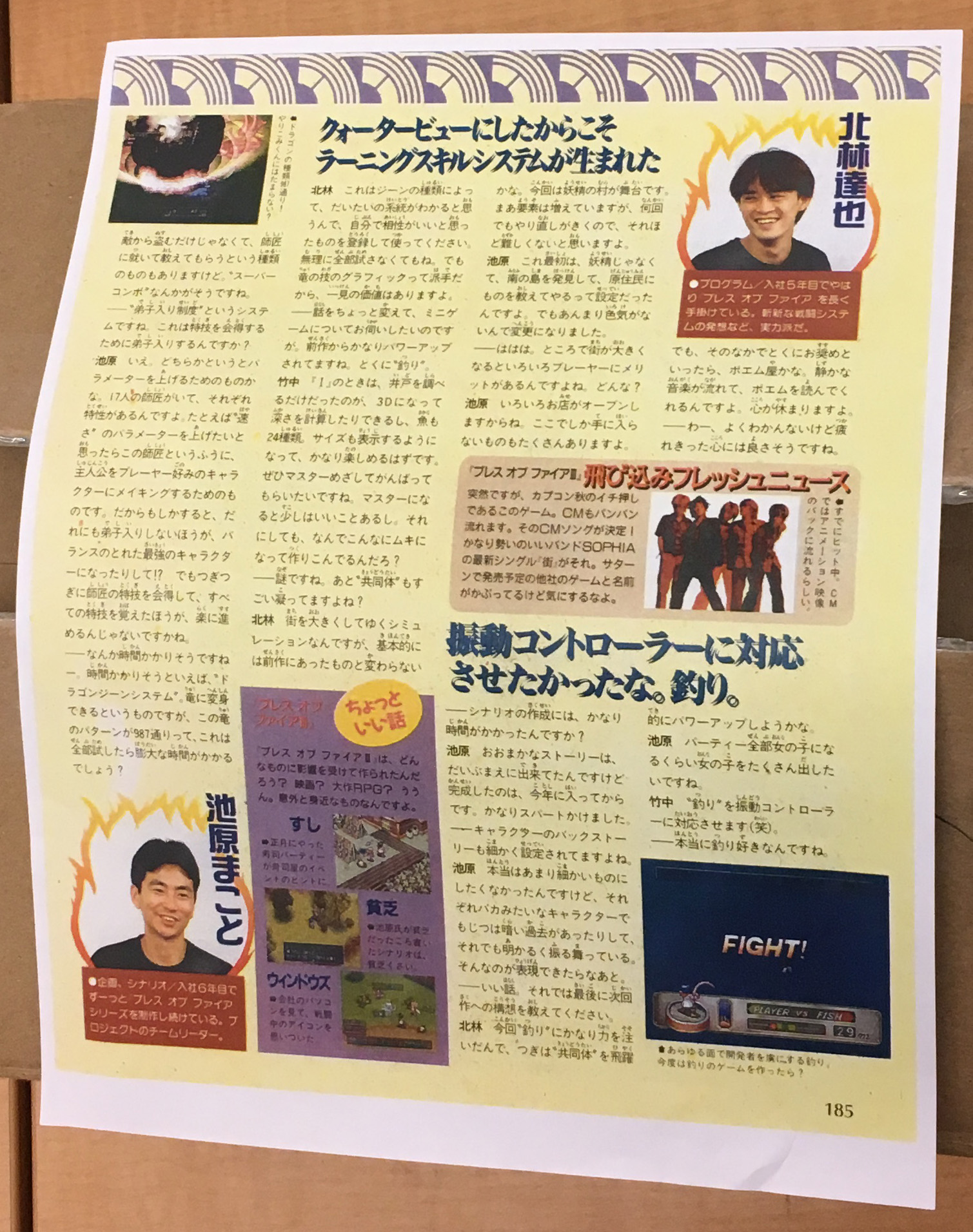} & 
        \includegraphics[width=0.124\linewidth,height=0.15\linewidth]{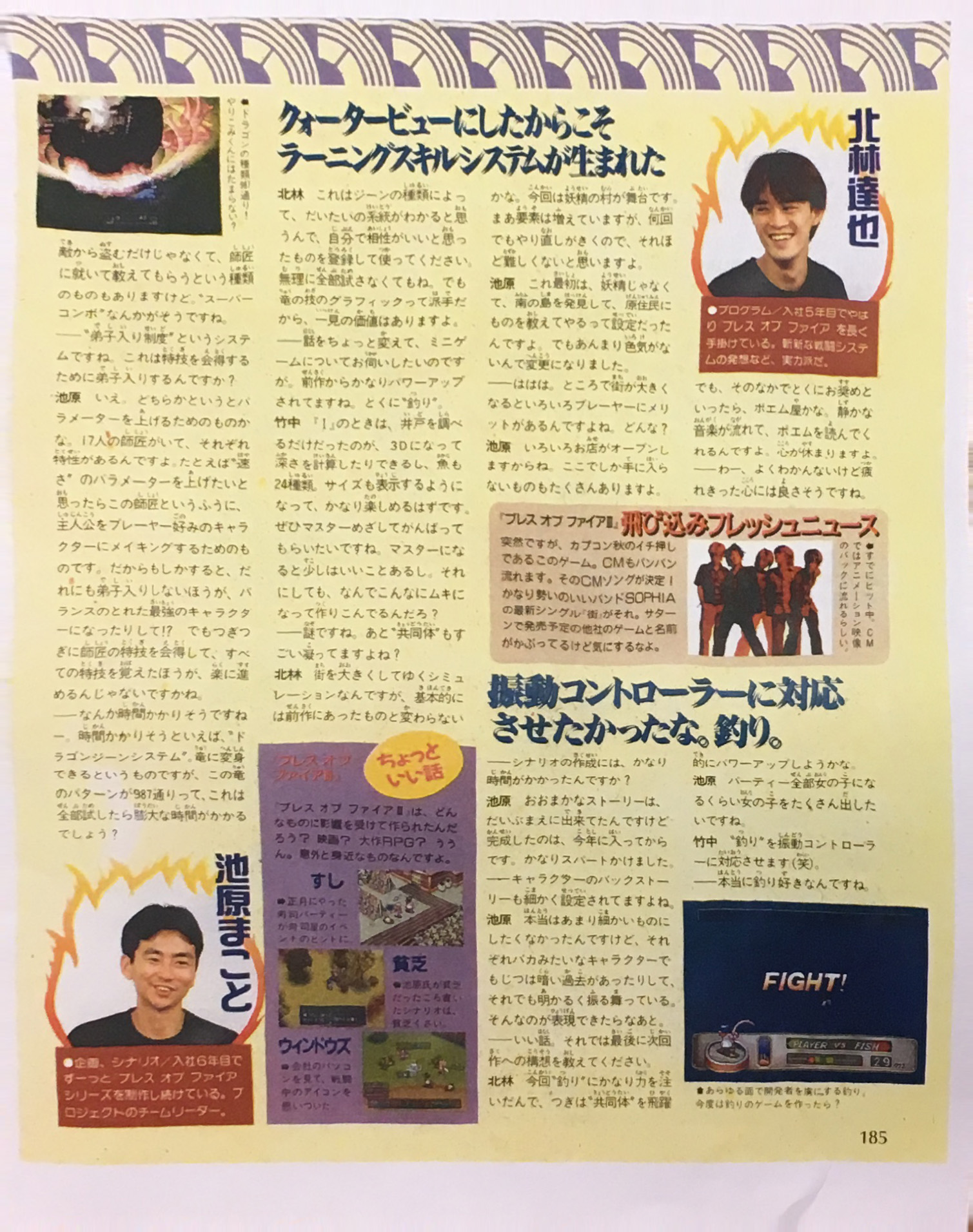} & 
        \includegraphics[width=0.124\linewidth,height=0.15\linewidth]{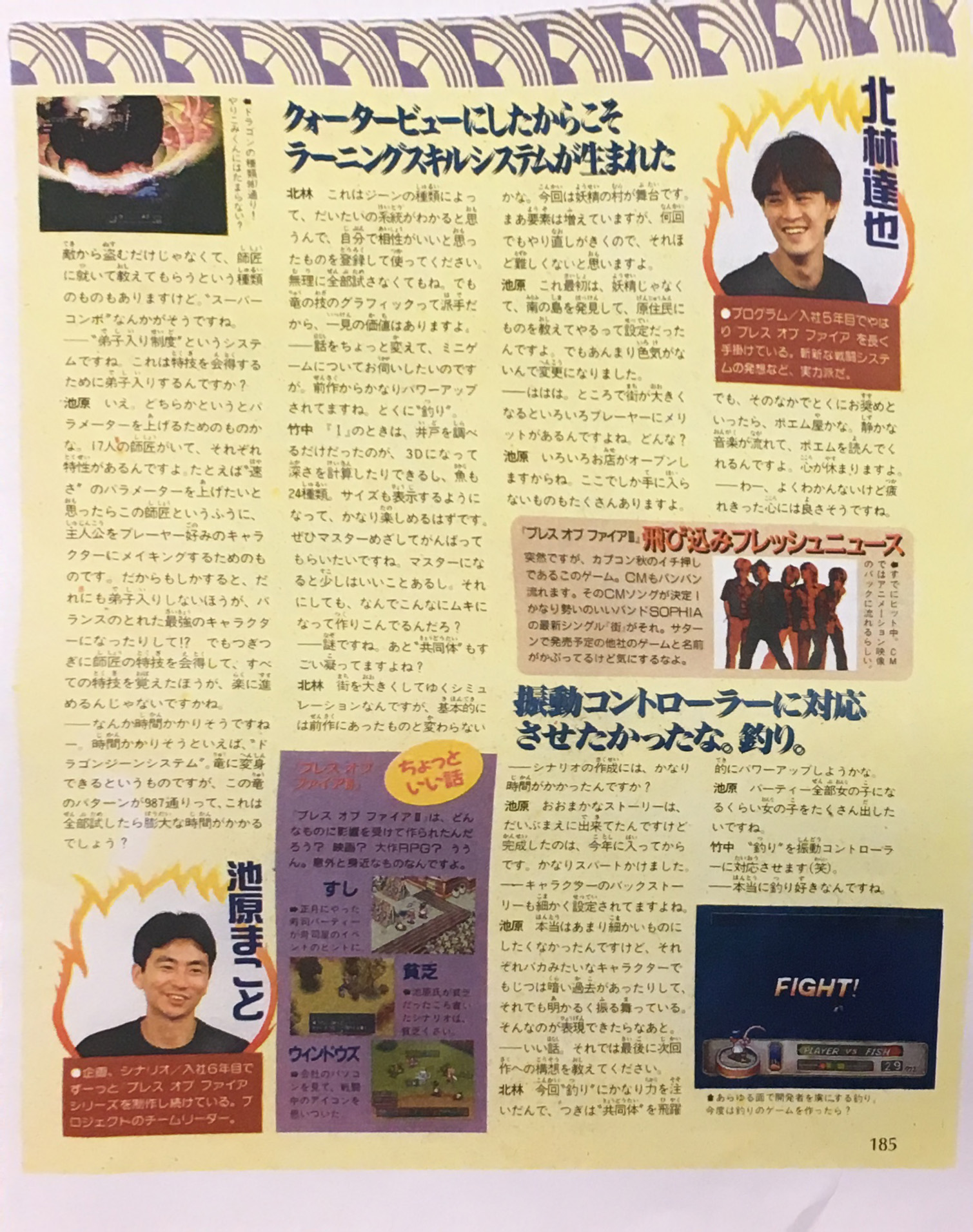} &
        \includegraphics[width=0.124\linewidth,height=0.15\linewidth]{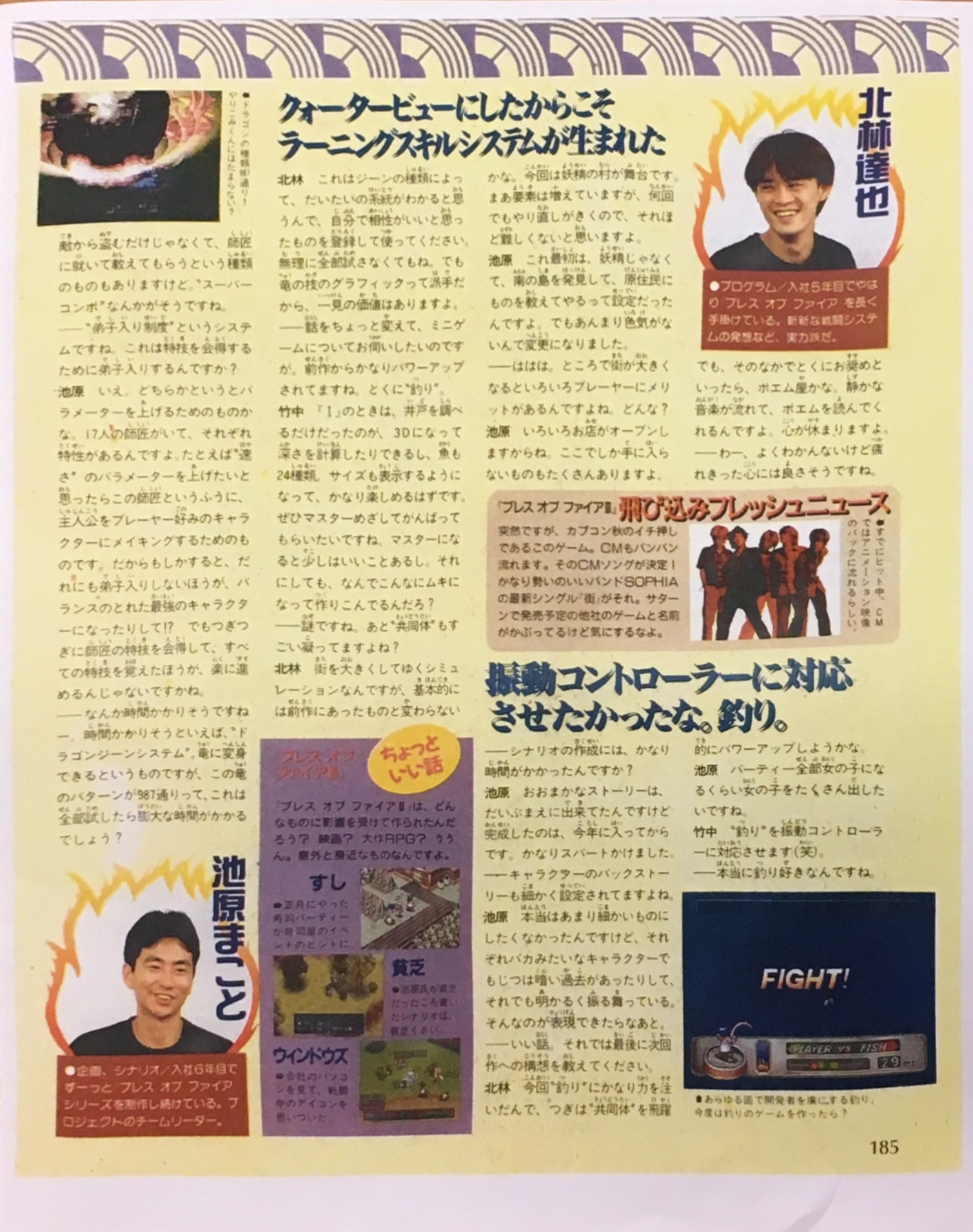} &
        \includegraphics[width=0.124\linewidth,height=0.15\linewidth]{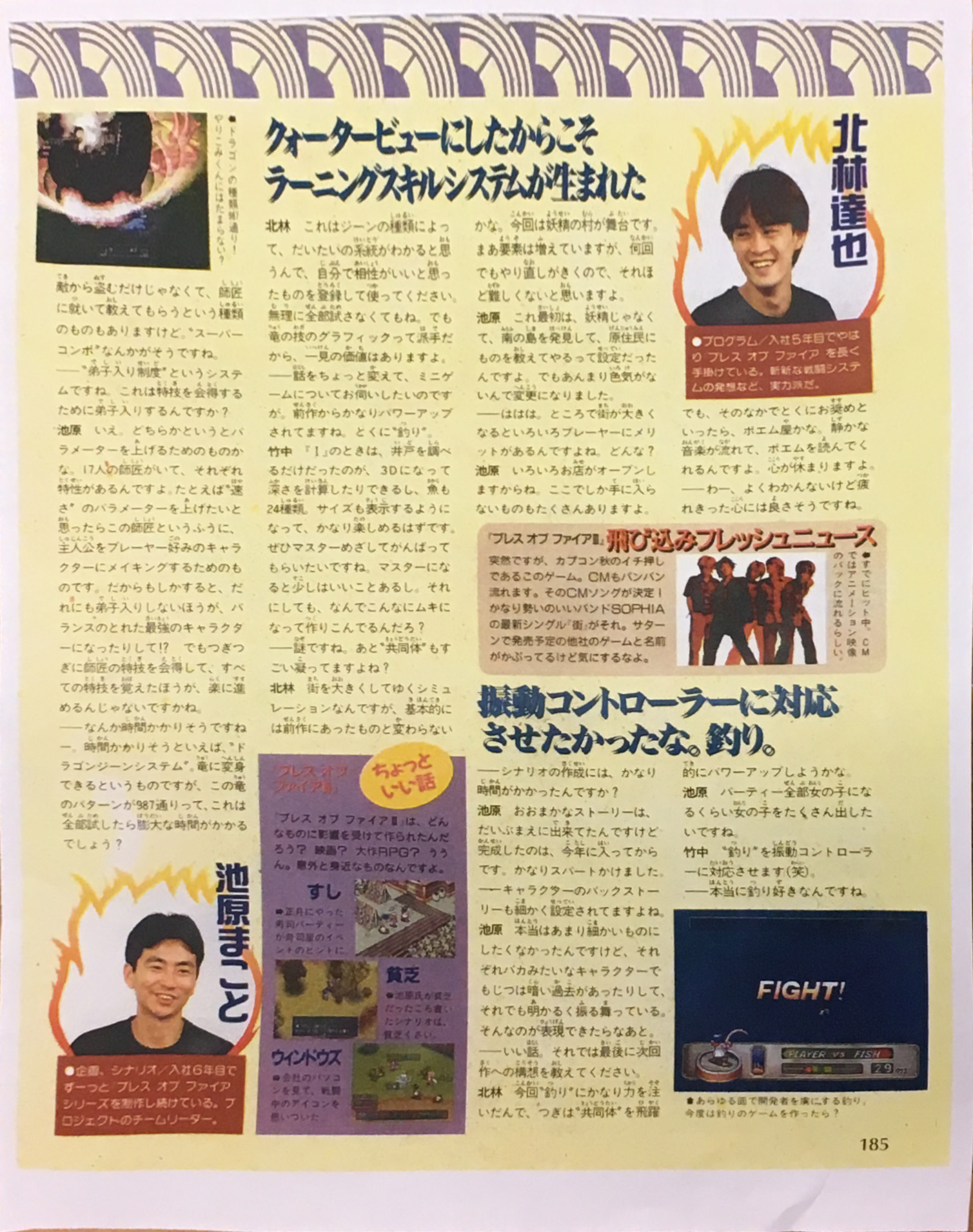} & 
        \includegraphics[width=0.124\linewidth,height=0.15\linewidth]{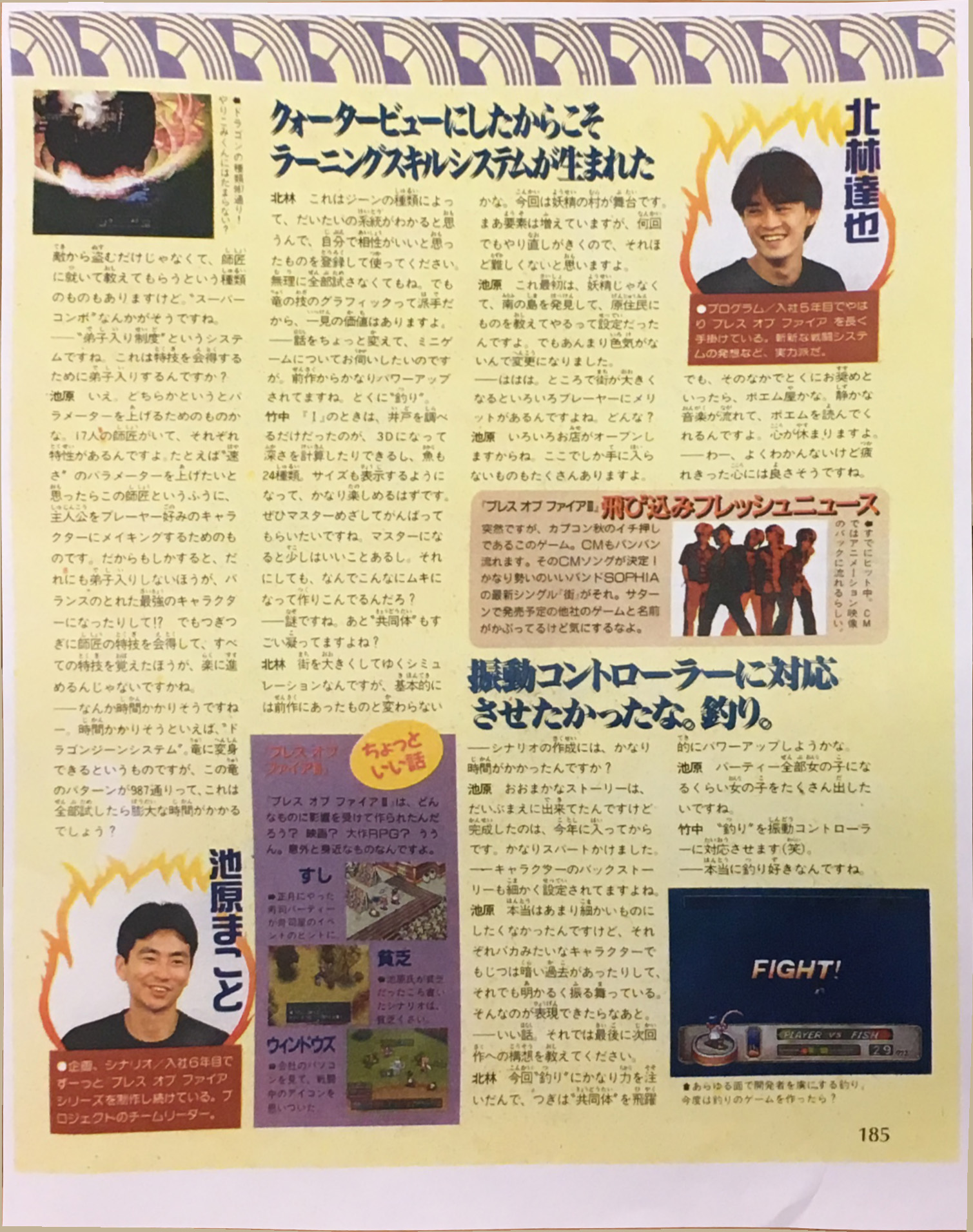} & 
        \includegraphics[width=0.124\linewidth,height=0.15\linewidth]{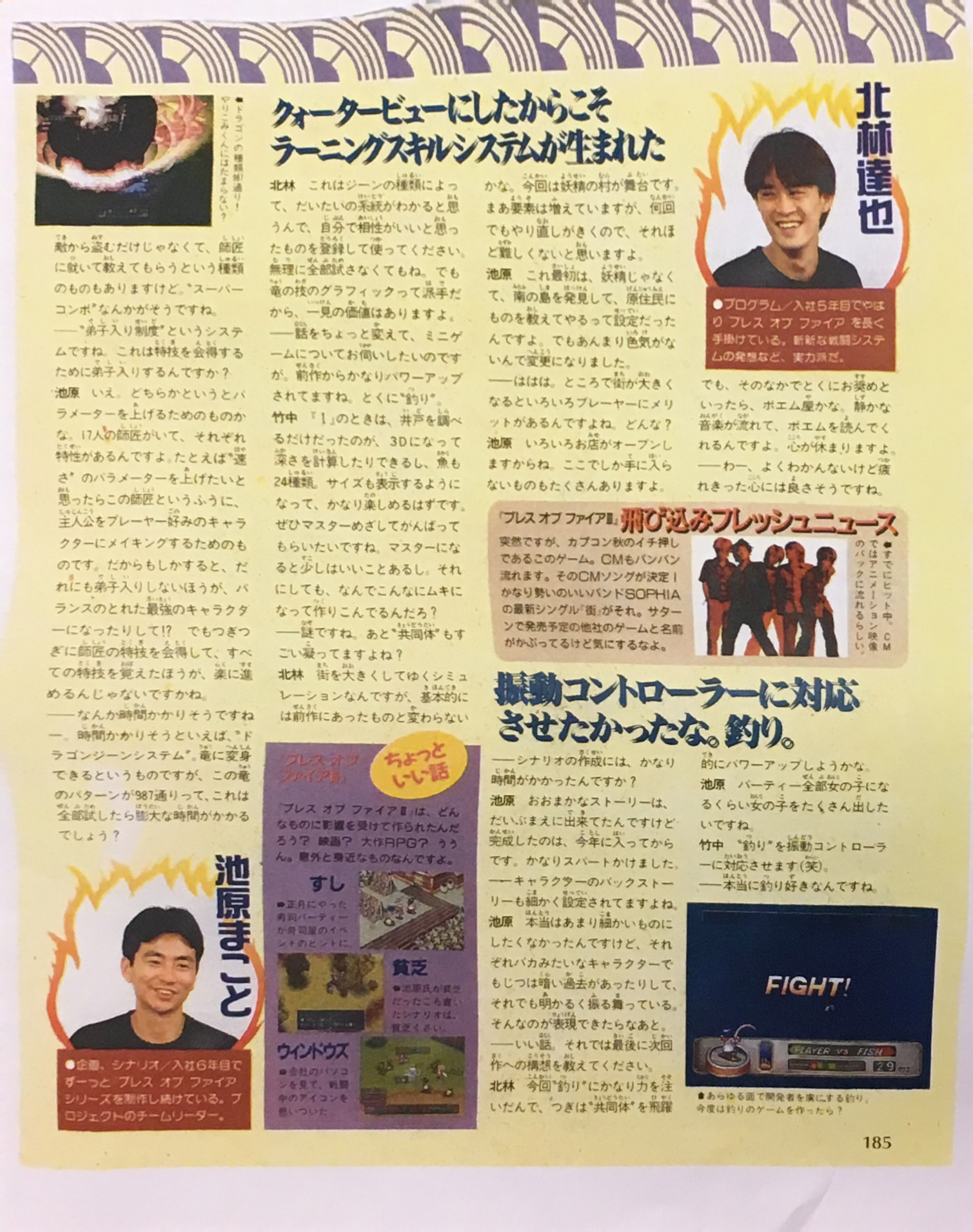} & 
        \includegraphics[width=0.124\linewidth,height=0.15\linewidth]{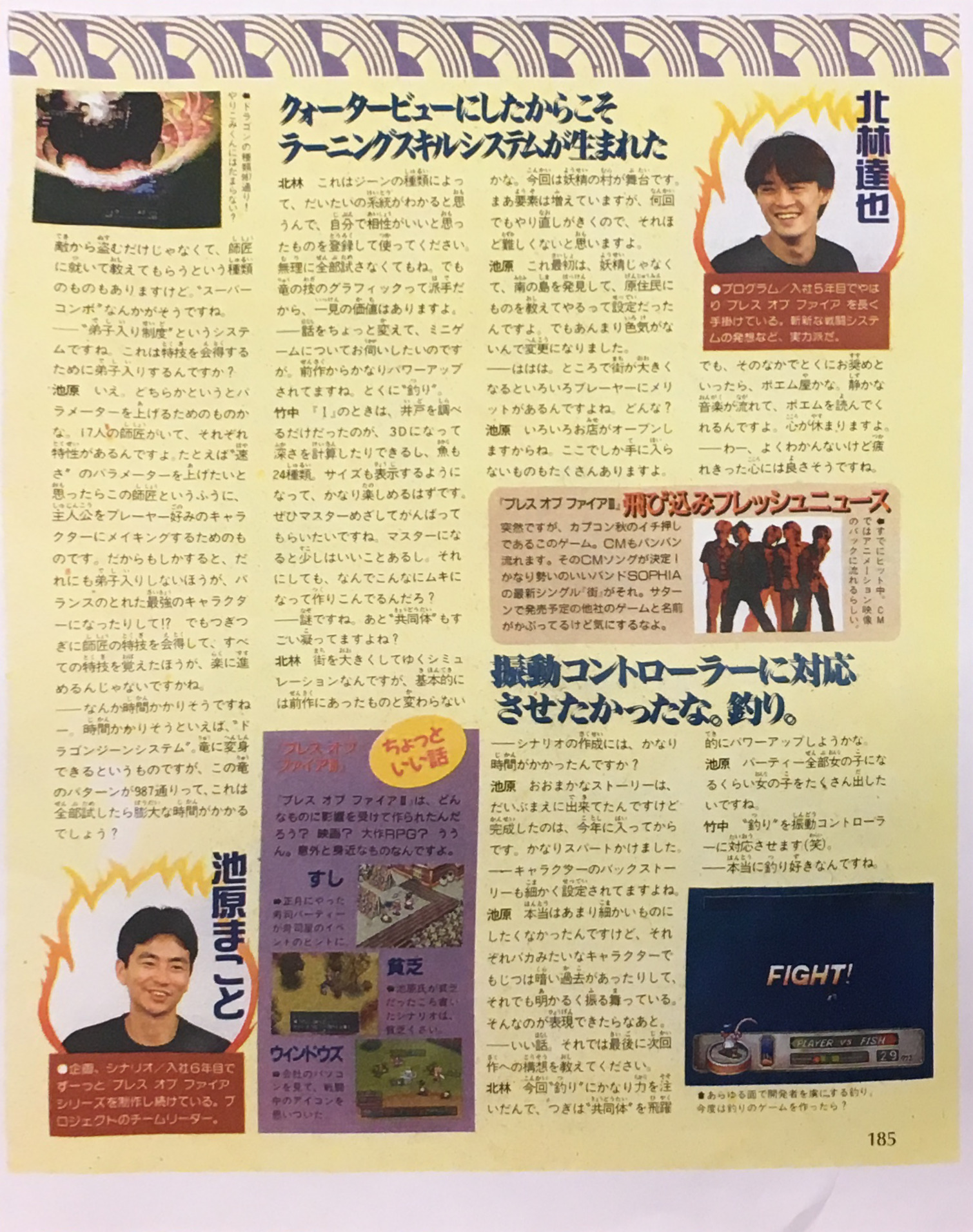} 
        \\[-8pt]
        \small input &
        \small \cite{DewarpNet} &
        \small \cite{DocTr} &
        \small \cite{FDRNet} &
        \small \cite{PaperEdge} &
        \small \cite{Marior} &
        \small \cite{DocGeoNet} &
        \small ours \\
	\end{tabular}
	\caption{Qualitative comparisons on the DocUNet benchmark dataset. From left to right: input, DewarpNet \cite{DewarpNet}, DocTr \cite{DocTr}, FDRNet \cite{FDRNet}, PaperEdge \cite{PaperEdge}, Marior \cite{Marior}, DocGeoNet \cite{DocGeoNet}, ours. All input images come from the \emph{``crop''} subset. \label{fig:unwarpedimages}}
\end{figure*}

\paragraph{Qualitative evaluation}
In addition to the quantitative comparisons made in the previous section, we provide a qualitative comparison to previous works. 
We show a side-by-side comparison of unwarped images by several methods in \figref{fig:unwarpedimages} and \figref{fig:unwarpedUVDoc}. The left-most column shows the input images. The images unwarped by our method are perceptually of high quality and have good unwarping at the borders of the document as well, even though we do not include explicit handling of borders or segmentation, in contrast to \cite{DocTr, PaperEdge, Marior, DocGeoNet}. 
We present more qualitative results on real-world images in \figref{fig:teaser}.
We include the unwarped images for all items in the DocUNet and UVDoc benchmarks in the supplemental material.

\subsection{Ablation study}\label{subsec:ablation}
We show the effectiveness of the dual-task learning, i.e., the combination of  predicting the 3D and the 2D grid meshes in the training process, the effectiveness of the reconstruction loss $\mathcal{L}_r$, as well as the benefit of combined training on both Doc3D and our UVDoc dataset, via ablation experiments. As we notice large variance in the OCR performance, we use averages of 10 repeated experiments with constant settings to perform the ablation study. 

We first show in \tabref{tab:ablation} that training on a combination of the Doc3D and UVDoc datasets considerably improves the performance on all metrics, compared to training only on the Doc3D data. To ensure a fair comparison between the two, we double the number of epochs for the Doc3D-only training (both the number of epochs at a constant learning rate, as well as the number of epochs with linearly decaying learning rate), such that they process the same number of samples and have an equal amount of optimizer steps. In particular, adding the UVDoc data to the training process leads to improvements of 8.9\% for MS-SSIM, 12.9\% for LD and 9.7\% for AD on the DocUNet benchmark. We attribute the improvement in performance to the fact that our data (UVDoc) is closer in appearance to the real document photographs in the DocUNet dataset, as well as the fact that the 3D ground-truth data in our dataset is more physically accurate (albeit coarser), since we do not apply any smoothing to it. 

We also show that the dual-task of learning the 3D grid $W$ along with the 2D grid $G$ improves the performance by comparing it against our full model but with the loss on $W$ removed. \tabref{tab:ablation} shows that including the $L_1$ loss on $W$ greatly improves all non-OCR metrics on both DocUNet and UVDoc benchmarks whilst the OCR metrics remain comparable.

\tabref{tab:ablation} additionally shows the benefits of the reconstruction loss. Adding it improves performances on almost all metrics, and those that worsen only do so slightly. This loss helps the model to target content in the document, making the unwarping better in the areas that matter the most.
\vspace{-0.09cm}
\section{Conclusion}
\label{sec:conclusion}
We presented UVDoc, a new document unwarping dataset that consists of pseudo-photorealistic images of warped documents along with annotated ground-truth 3D shapes and unwarping functions. Our proposed acquisition methodology is simple to implement and uses relatively inexpensive, commonly attainable equipment, enabling easy replication and further expansion of the dataset by others. Since our dataset includes both a shaded and unshaded version of each document image, it allows its users to evaluate unwarping performance without the influence of shading artifacts. 

We show that with the addition of the UVDoc dataset, our dual-task deep learning approach that implicitly encodes the coupling between the document's 3D shape and its appearance in a 2D photograph achieves state-of-the-art performance on the commonly used DocUNet benchmark. Additionally, we introduce the new UVDoc benchmark and a new line straightness metric, on which we also achieve state-of-the-art results.

The shape-from-shading effect can remain quite strong in the unwarped documents and makes some of them appear more distorted to the human eye than they geometrically are. Research into the illumination correction process is therefore of great importance.
Since the pseudo-photorealistic nature of our dataset allows us to decouple the deformation and shading of a warped document image, it could benefit research progress in this field.

\vspace{-0.04cm}
\begin{acks}
We thank Ruben Schenk for his help with data capture. This work was supported in part by the \grantsponsor{ERC}{European Research Council (ERC)}{https://erc.europa.eu/apply-grant/consolidator-grant} under the European Union’s Horizon 2020 Research and Innovation Programme (ERC Consolidator Grant, agreement No.~\grantnum{ERC}{101003104}, MYCLOTH).
\end{acks}


\clearpage
\balance
\bibliographystyle{ACM-Reference-Format}
\bibliography{references}

\appendix	

\section{UVDoc dataset: Ordering the grid}
\label{sec:recoveringrid}

\begin{figure}[H]
	\centering
	\begin{subfigure}[t]{0.49\linewidth}
		\centering
		\includegraphics[width=\textwidth]{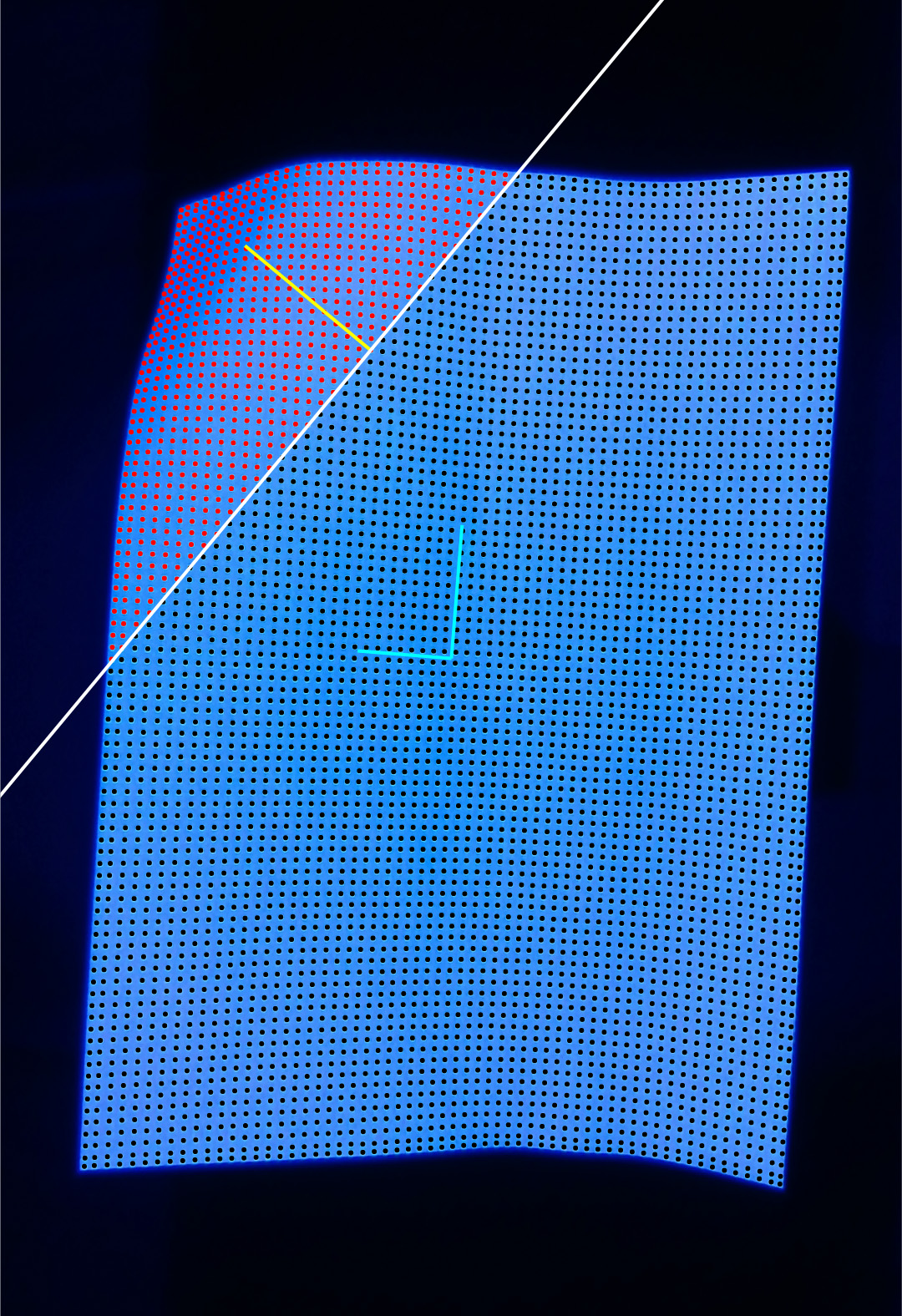}
	\end{subfigure}
	\hfill
	\begin{subfigure}[t]{0.49\linewidth}
		\centering
		\includegraphics[width=\textwidth]{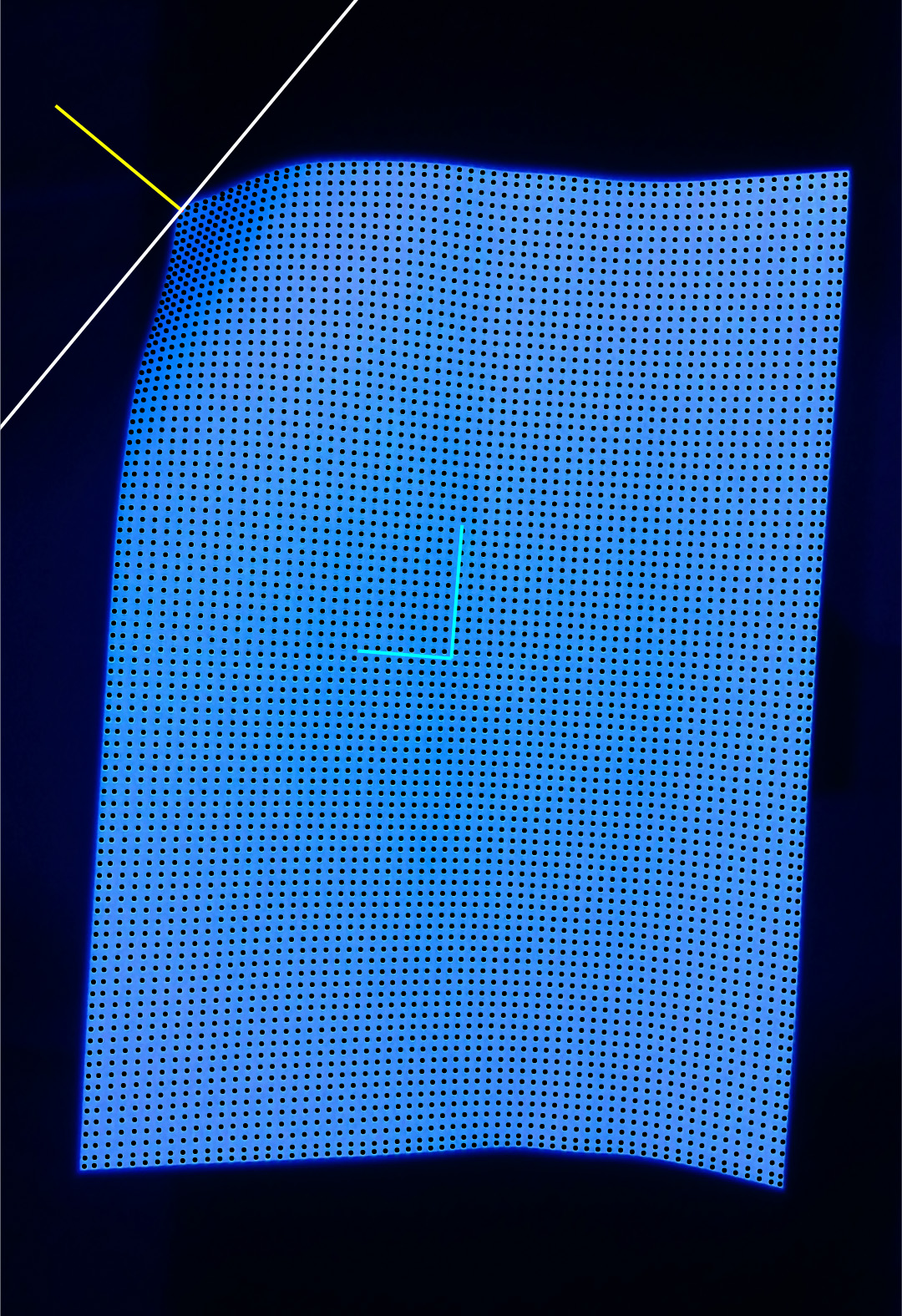}\\
	\end{subfigure}
	\caption{Illustration of the top-left corner identification step. Cyan lines represent the principal components of the grid points, the yellow line is the diagonal direction, and the white line is the orthogonal line defining the dividing half-space. Red points are towards the left of the line and black points towards its right. (Left) There are several red points, this is not the top-left corner. (Right) There are no red points, the top-left corner is the point on top of which the white is located.}\label{fig:findtopleft}
\end{figure}

Using the UV-lit image, where the printed grid is visible, we obtain the pixel coordinates of the grid points on the deformed piece of paper. We then need to compute their correspondences to the vertices of a regular grid, which is equivalent to ordering them as an $89\times61$ grid. We solve the ordering problem in 3 steps:

\begin{enumerate}
    \item \textit{Finding the top-left corner.}
We first find the top-left corner of the grid. 
We compute the two principal components of the detected grid points and define the diagonal direction of the grid as the sum of these two vectors. For each point, we draw a line orthogonal to this diagonal direction and we count the number of points on each side of the line. The top-left corner is then the point that has exactly zero points to its left. The process is illustrated in \figref{fig:findtopleft}.

\item \textit{Ordering border points.}
Next we detect all border points. To this end, we use a segmentation of the paper that we obtain by thresholding the UV-lit image.
Based on this segmentation, we use OpenCV's \textsf{findContours} function to extract an \emph{ordered} contour polyline. For each contour vertex, we find the nearest neighbor point in the set of grid points. We then define our grid border points as the 296 grid points --- the number of points on the border of the grid --- that are most frequently found as nearest neighbor.
Finally, since the contour extracted using OpenCV is ordered, we can also order the detected grid border points.

\item \textit{Ordering interior points.}
The final step is to order the points that lie in the interior of the grid. 
We iteratively identify all points $(i,j)\in [2,88]\times[2,60]$ in row-major ordering, starting from point $(2,2)$ (the top-left interior grid point). We do this (for point $(i,j)$) by finding the three nearest yet-unordered grid points for each of the previously-ordered points $(i-1, j-1)$, $(i, j-1)$, and $(i-1, j)$ (the points to the top-left, top and left of the point we are currently trying to identify). The point that is in the intersection of these three nearest-neighbor sets is chosen as point $(i,j)$. We use the average distance to the three reference points as a tiebreaker in case the intersection contains multiple points.
This point is then considered ordered, and we move on to the next point.

\end{enumerate}

\section{Training details}
\label{sec:trainingdetails}

We obtain the ground-truth $G$ and $W$ for the Doc3D dataset by sampling the ground truth backward maps at a regular grid of $45\times31$ points covering the entire backward map. For our UVDoc dataset (see Sec. 3 of the main paper) we slice the available high-resolution ground truths by a factor of 2.

We use the ADAM optimizer \cite{ADAM} with a batch size of 8. The initial learning rate is set to $2\times10^{-4}$ for 10 epochs and linearly decays to 0 over 10 further epochs. We alternate optimization steps based on a batch of Doc3D data with a batch of our UVDoc data, using the same loss function on both of them.

We visually augment both the Doc3D and our data with noise, color changes and other appearance transformations. Additionally, we augment our data with rotations, since our images are captured from a more uniform angle than the Doc3D data. All images are tightly cropped before being fed to the network.

\begin{table*}
\centering
     \renewcommand{\arraystretch}{1.2}
    \caption{Ablations on training data. The reported values are averages and standard deviations over 10 repetitions of training with otherwise constant parameters. Settings used in our final model are underlined. We show performance on the DocUNet and UVDoc benchmarks. \emph{Doc3D reduced} is a version of the Doc3D dataset with 20,000 samples removed to offset for the additional UVDoc samples. The underlined setting is the one we use.}
    \resizebox{\textwidth}{!}{
    \begin{tabular}{l c c c c c c c c c c c}
    \toprule
    & \multicolumn{5}{c}{DocUNet} & \multicolumn{6}{c}{UVDoc}\\
    \cmidrule[0.1pt](lr){2-6}\cmidrule[0.1pt](lr){7-12}
    Data & MS-SSIM $\uparrow$ & LD $\downarrow$ & AD $\downarrow$ & CER $\downarrow$ & ED $\downarrow$ & MS-SSIM $\uparrow$ & AD $\downarrow$ & CER $\downarrow$ & ED $\downarrow$ & H-line $\downarrow$ & V-line $\downarrow$ \\
    \midrule
    Doc3D & 0.492$\pm$0.004 &  7.99$\pm$0.13 & 0.360$\pm$0.007 & 0.197$\pm$0.018 &  757$\pm$57  & 0.669$\pm$0.015 & 0.178$\pm$0.013 & 0.078$\pm$0.013 &  220$\pm$30 & 2.42$\pm$0.03 &  3.85$\pm$0.16 \\
    Doc3D reduced + UVDoc & 0.535$\pm$0.004 & 7.01$\pm$0.20 & 0.331$\pm$0.008 & 0.206$\pm$0.019 & 797$\pm$69 & \textbf{0.765$\pm$0.009} & 0.138$\pm$0.011 & 0.073$\pm$0.010 & 217$\pm$25 & \textbf{1.84$\pm$0.11} & 2.65$\pm$0.13 \\
    \ul{Doc3D + UVDoc} & \textbf{0.536$\pm$0.006} & \textbf{6.96$\pm$0.17} & \textbf{0.325$\pm$0.006} & \textbf{0.195$\pm$0.012} & \textbf{745$\pm$34} & 0.762$\pm$0.014 & \textbf{0.129$\pm$0.008} & \textbf{0.070$\pm$0.010} & \textbf{205$\pm$23} & 1.85$\pm$0.06 & \textbf{2.53$\pm$0.06} \\
    \bottomrule
    \end{tabular}}
\label{tab:ablation_mixed}
\end{table*}

\begin{figure*}[t]
	\centering
	\begin{tabular}{c}
		\includegraphics[width=1.0\linewidth]{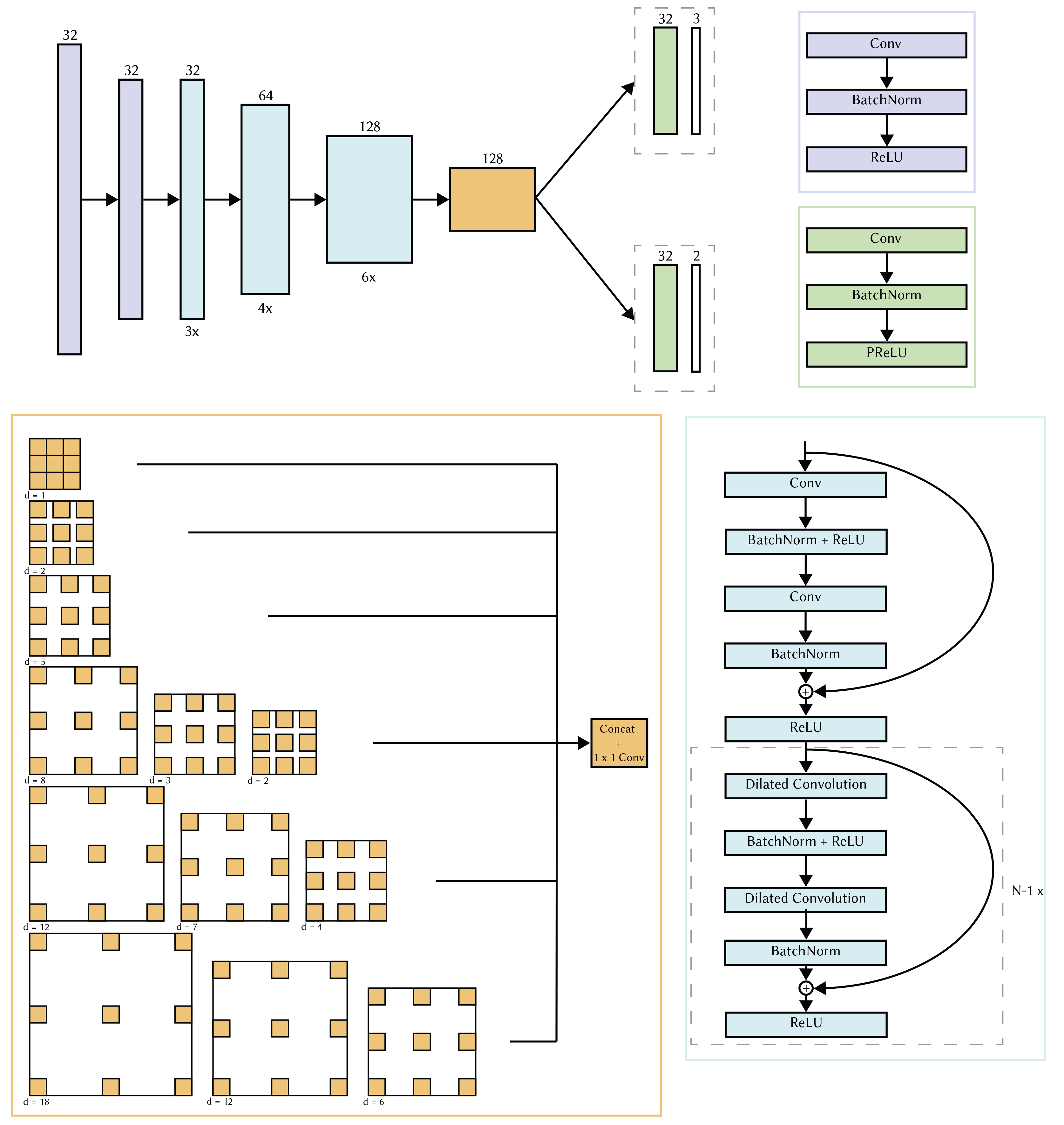}
	\end{tabular}
	\caption{An overview of the architecture of our network.\label{fig:arch}}
\end{figure*}

Empirically, we find that the best set of weights to balance the influence of the individual loss terms as defined in Eq.\ 1 in the main paper are $\alpha=5$ and $\beta=5$. During training $\gamma$ is set to $0.0$ for the first 10 epochs (first half) and then to $1.0$ for the remaining 10 epochs.
We give a detailed graphical overview of our model architecture in \figref{fig:arch}.

\section{Evaluation metrics}

As explained in the main paper, we used image similarity metrics such as MS-SSIM, LD and AD as well as optical character recognition (OCR) performance measured with CER and ED. Details about these metrics are provided below.
The structural similarity measure (SSIM) \cite{SSIM} quantifies the visual similarity between two images by measuring the similarity of mean pixel values and variance within image patches between the two images. The multi-scale variant (MS-SSIM) repeats this process at multiple scales using a Gaussian pyramid and computes a weighted average over the different scales as its final measure. We use the same weights as described in the original implementation \cite{MS_SSIM}. 

LD is computed using a dense SIFT flow mapping \cite{SiftFlow} from the ground truth image to the rectified image. Using this registration, LD is computed as the mean $L_2$ distance between mapped pixels \cite{You2018}, essentially measuring the average local deformation of the unwarped image. 

Aligned distortion (AD) is a more robust variant of the LD metric, introduced in \cite{PaperEdge}. In contrast to LD, AD eliminates the error caused by a global translation and scaling of the image by factoring out the optimal affine transformation out of the SIFT flow distortion. Such a global affine transformation can cause large LD values but does not greatly impact human readability of the image. Additionally, AD weighs the error according to the magnitude of the gradient in the image, emphasizing interesting areas, such as text or image edges, rather than the background.
Prior to computing these similarity metrics, we resize all images, both rectified and ground-truth, to a 598,400-pixel area, as suggested in \cite{DocUNet}.

In addition to the image similarity metrics, we evaluate OCR performance based on character error rate (CER) and editing distance (ED) \cite{ED}.
The CER is defined as the ratio between the ED (the edit distance between the recognized and reference text) and the number of characters in the reference text.
We obtain the reference text by extraction from the flatbed scans of the documents. The full definition for the CER then becomes: $\text{CER} = (s+i+d)/N$, where $s,i,d$ are the number of substitutions, insertions and deletions, respectively, and $N$ is the number of characters in the reference text. 

\section{Additional experiments}

\paragraph{Mixed training} As shown in the main paper, we find that training models on a combination of the Doc3D and UVDoc datasets yields improved performance compared to training on Doc3D alone. However, models trained on a combination of both datasets see more samples and thus more variety than the ones trained on Doc3D only. To verify that the increased number of unique samples is not the cause of the performance gain, we train on a combination of Doc3D and UVDoc datasets, removing 20,000 samples from the Doc3D dataset. This way, the models trained on a combination of the two datasets see equally many samples as the ones trained on Doc3D only. The results of these experiments, along with the results of models trained on Doc3D only and on a combination of the full Doc3D and UVDoc datasets are presented in \tabref{tab:ablation_mixed}.

The models trained on a combination of the reduced Doc3D dataset and UVDoc have slightly worse performance than the models trained on the full datasets. This is expected, as the models are trained with fewer samples. However, the difference between the two is very small. More importantly, the models trained on the full Doc3D dataset alone give very poor results in comparison. Replacing samples from the Doc3D dataset with higher-quality ones from our UVDoc dataset improves its overall performance.


\section{Line unwarping visualization}
\label{sec:newmetric}

Our new UVDoc benchmark, equipped with the ground-truth unwarping function, allows one to warp and unwarp not only the document image but any texture. We can warp the texture based on the ground truth deformation and unwarp it using the predicted deformation. This idea, which we apply to create our new line straightness metric, can also be used to better visualize the structural behavior of an unwarping function. By unwarping the unshaded document texture, we can remove the visual effect of shape-from-shading, giving a better visualization of the remaining geometric distortions. We apply this to visually compare our method with related works in \figref{fig:newmetricline}.

\newcommand{\horSpace}{\hfill\hspace{-2pt}}
\newcommand{\verSpace}{2.5pt}

\begin{figure*}
\vspace{0.4cm}
\captionsetup[subfigure]{labelformat=empty}

\newlength{\teaserSubFigWidth}
\setlength{\teaserSubFigWidth}{0.14\linewidth}
\setlength{\teaserImgWidth}{0.14\linewidth}
\setlength{\teaserImgHeight}{0.20\textwidth}

\begin{center}
\begin{subfigure}[t]{\teaserSubFigWidth}
    \centering
    \includegraphics[height=\teaserImgHeight, width=\teaserImgWidth]{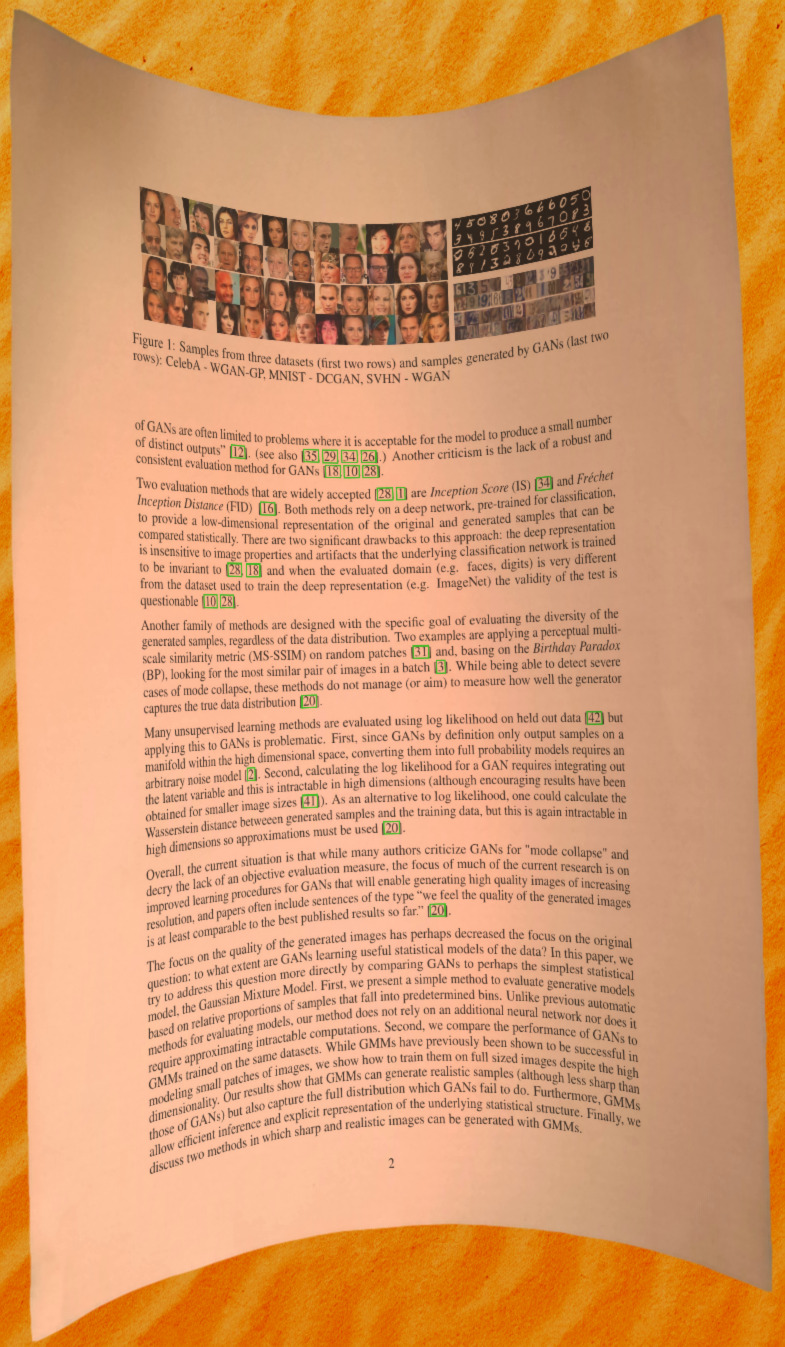}
\\\vspace{\verSpace}
    \includegraphics[height=\teaserImgHeight, width=\teaserImgWidth]{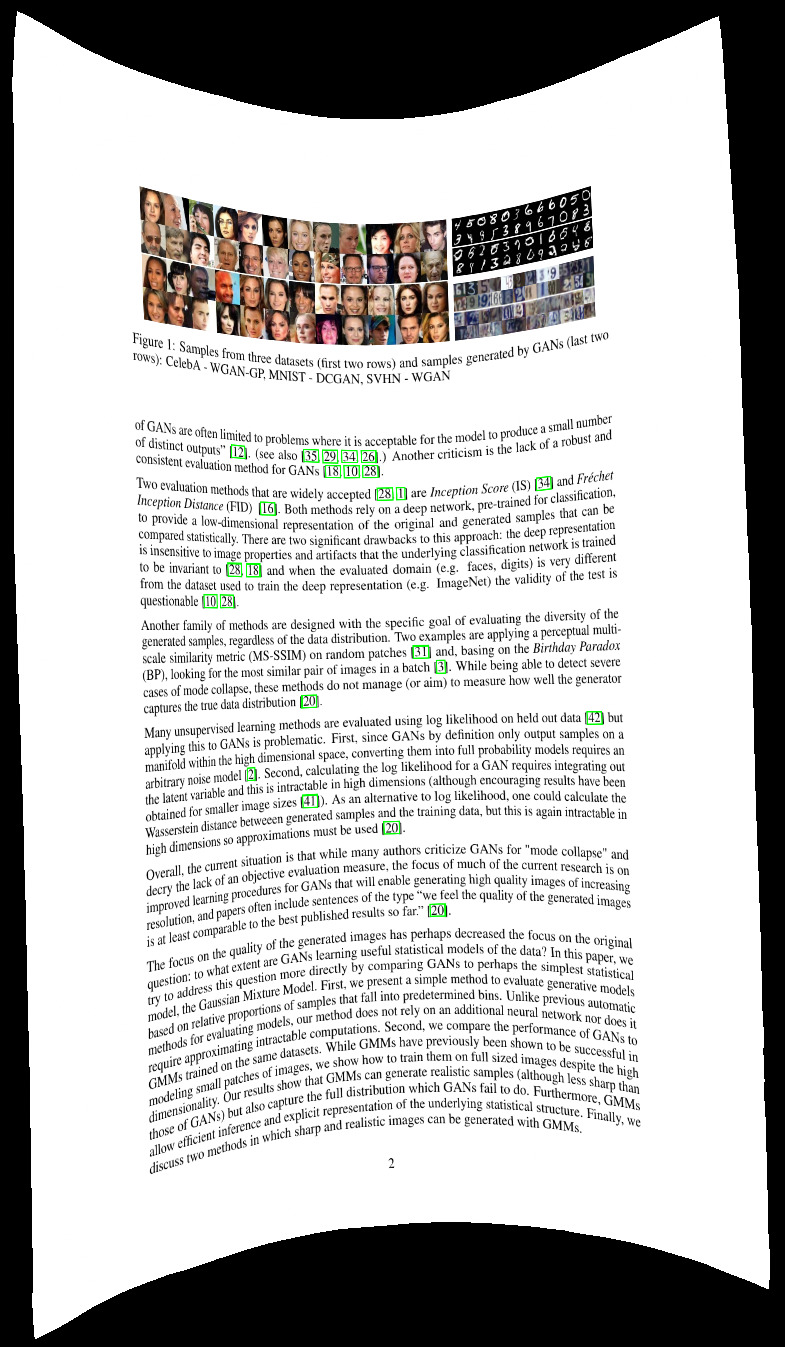}
\\\vspace{\verSpace}
    \includegraphics[height=\teaserImgHeight, width=\teaserImgWidth]{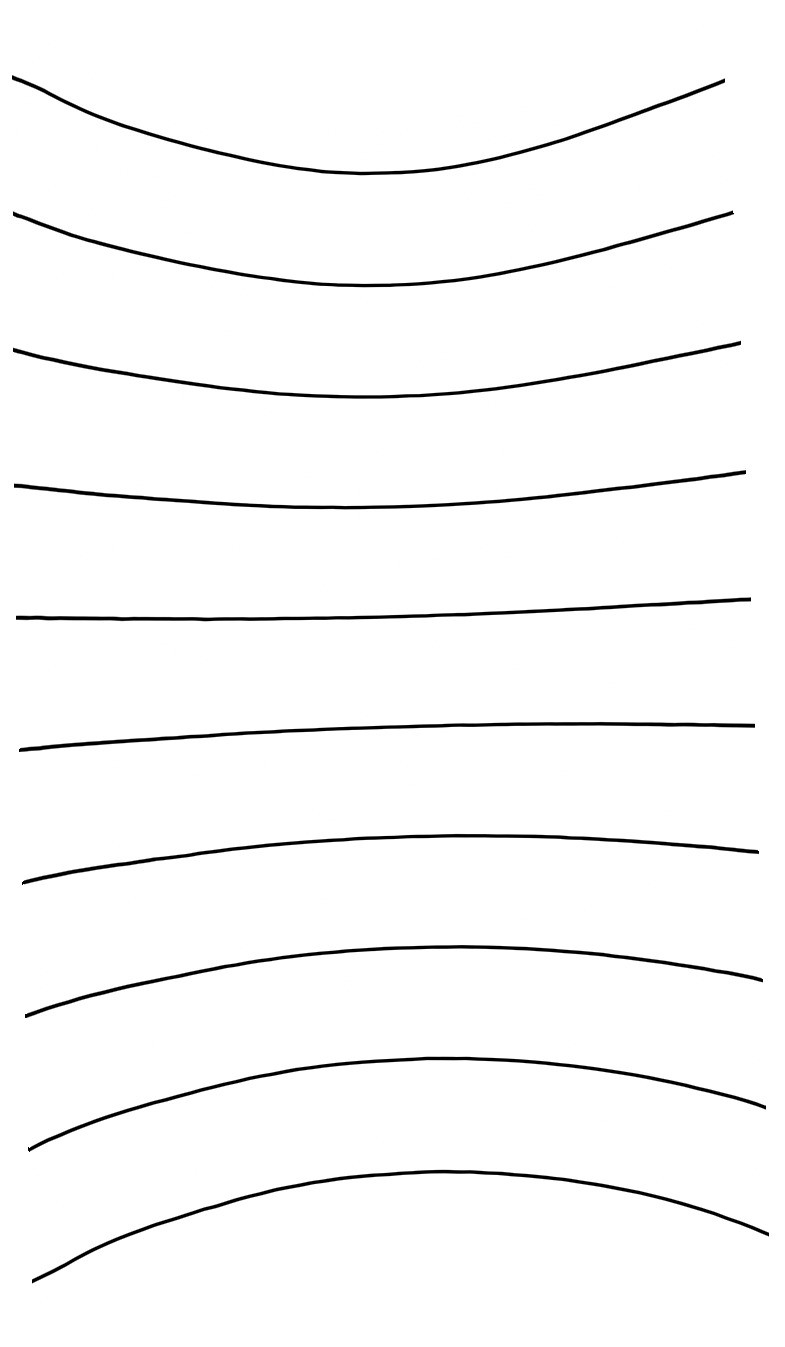}
\\\vspace{\verSpace}
    \includegraphics[height=\teaserImgHeight, width=\teaserImgWidth]{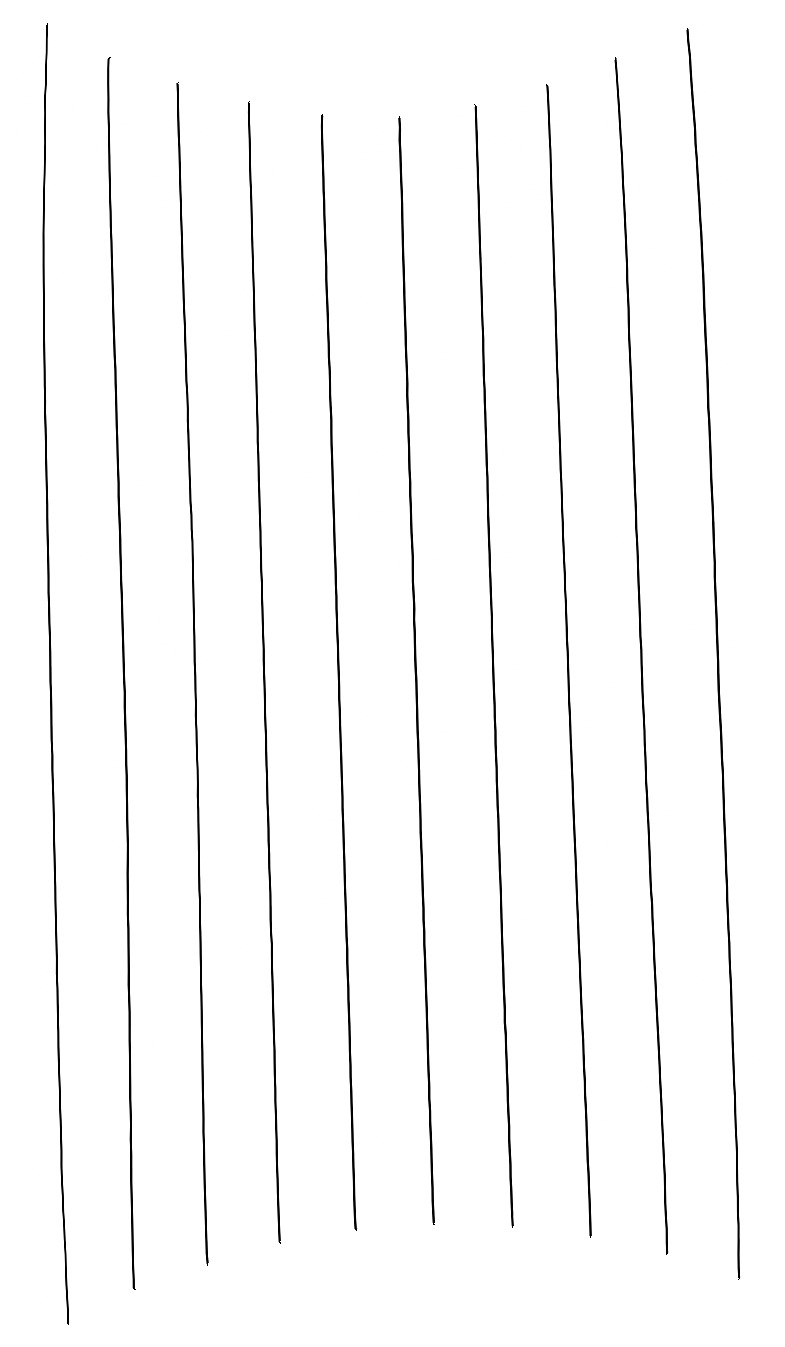}
    \caption{Input}
\end{subfigure}
\horSpace
\begin{subfigure}[t]{\teaserSubFigWidth}
    \centering
    \includegraphics[height=\teaserImgHeight, width=\teaserImgWidth]{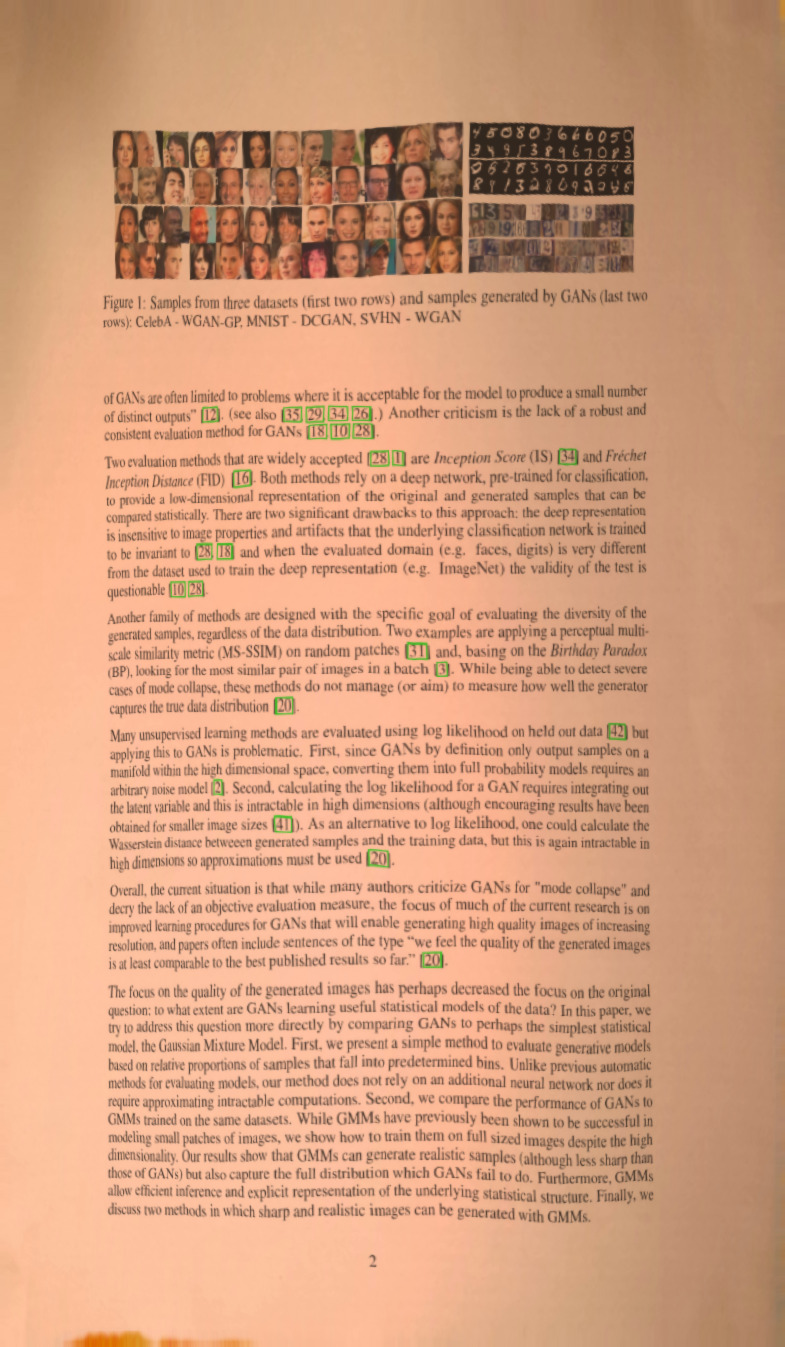}
\\\vspace{\verSpace}
    \includegraphics[height=\teaserImgHeight, width=\teaserImgWidth]{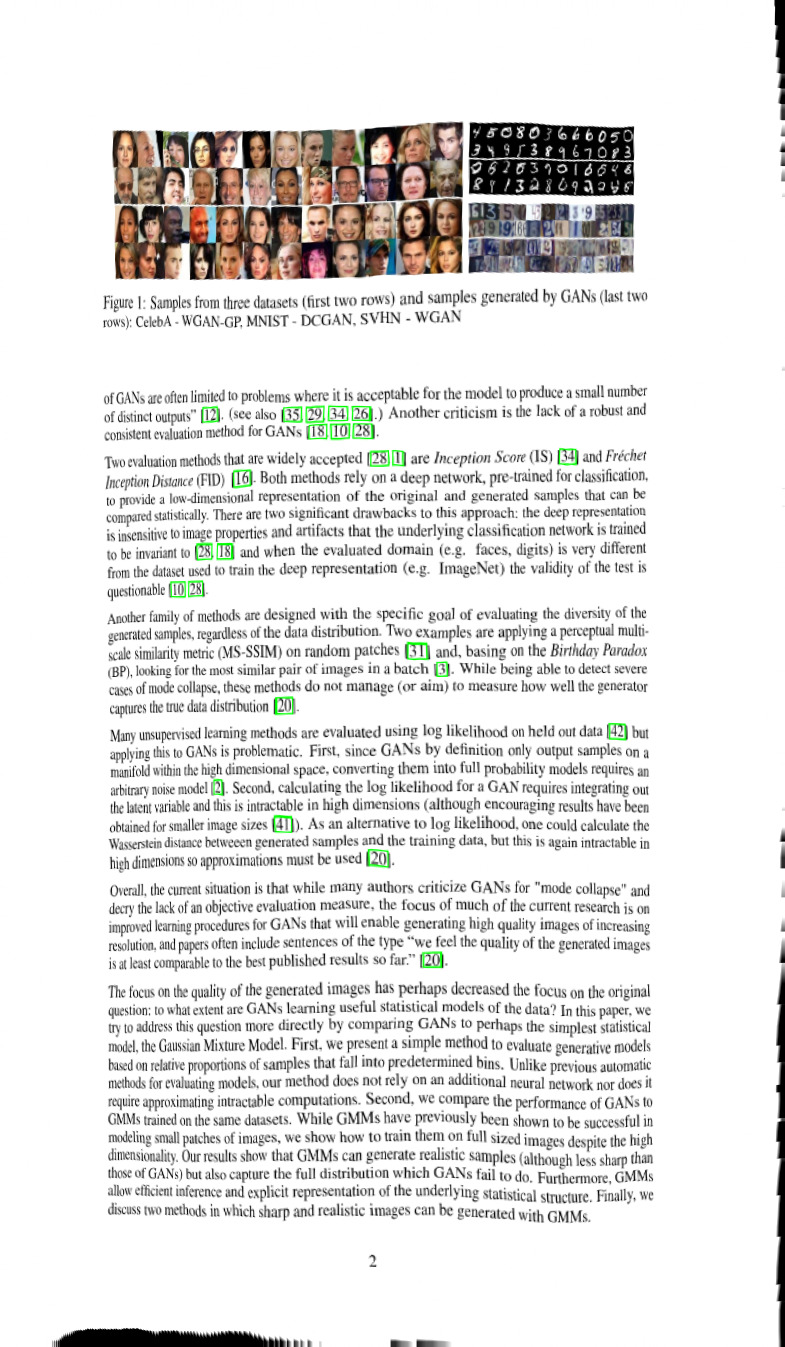}
\\\vspace{\verSpace}
    \includegraphics[height=\teaserImgHeight, width=\teaserImgWidth]{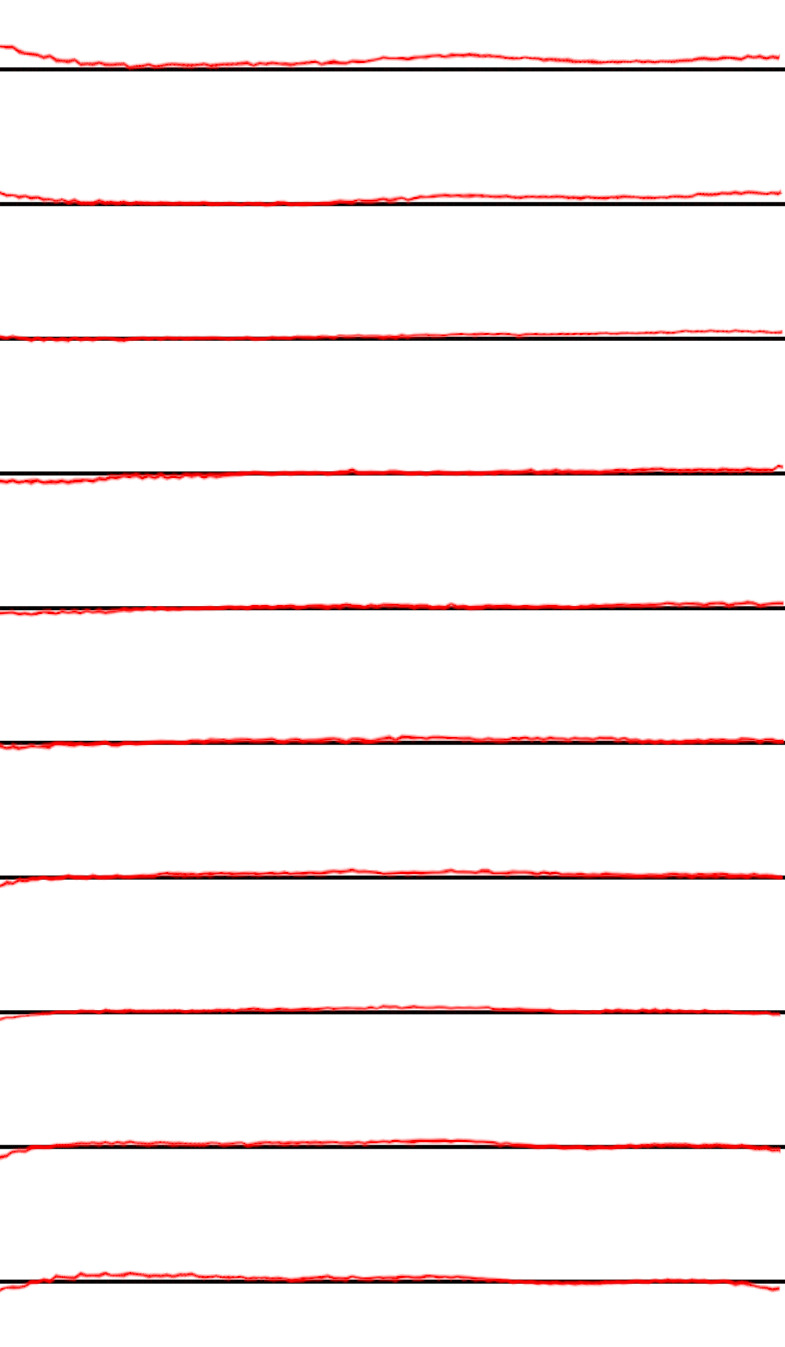}
\\\vspace{\verSpace}
    \includegraphics[height=\teaserImgHeight, width=\teaserImgWidth]{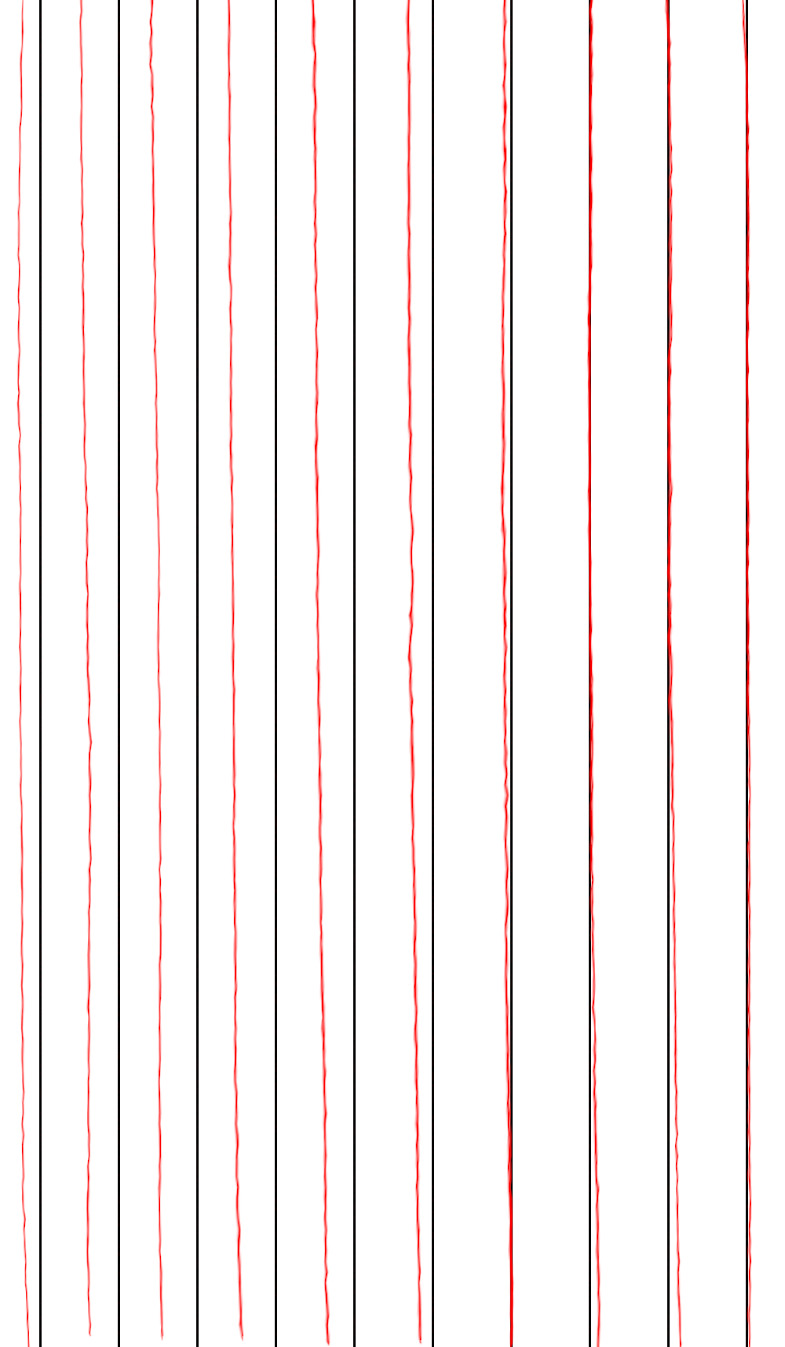}
    \caption{DewarpNet}
\end{subfigure}
\horSpace
\begin{subfigure}[t]{\teaserSubFigWidth}
    \centering
    \includegraphics[height=\teaserImgHeight, width=\teaserImgWidth]{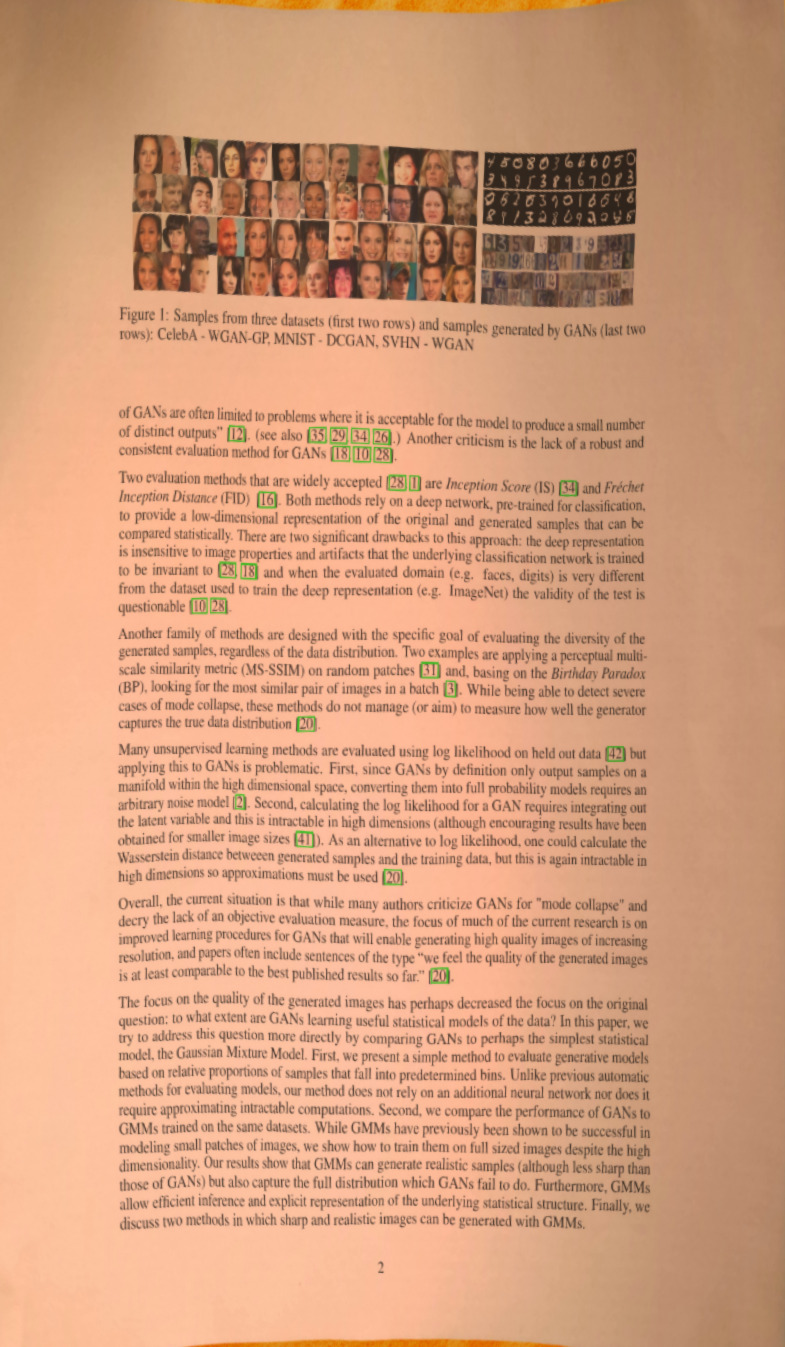}
\\\vspace{\verSpace}
    \includegraphics[height=\teaserImgHeight, width=\teaserImgWidth]{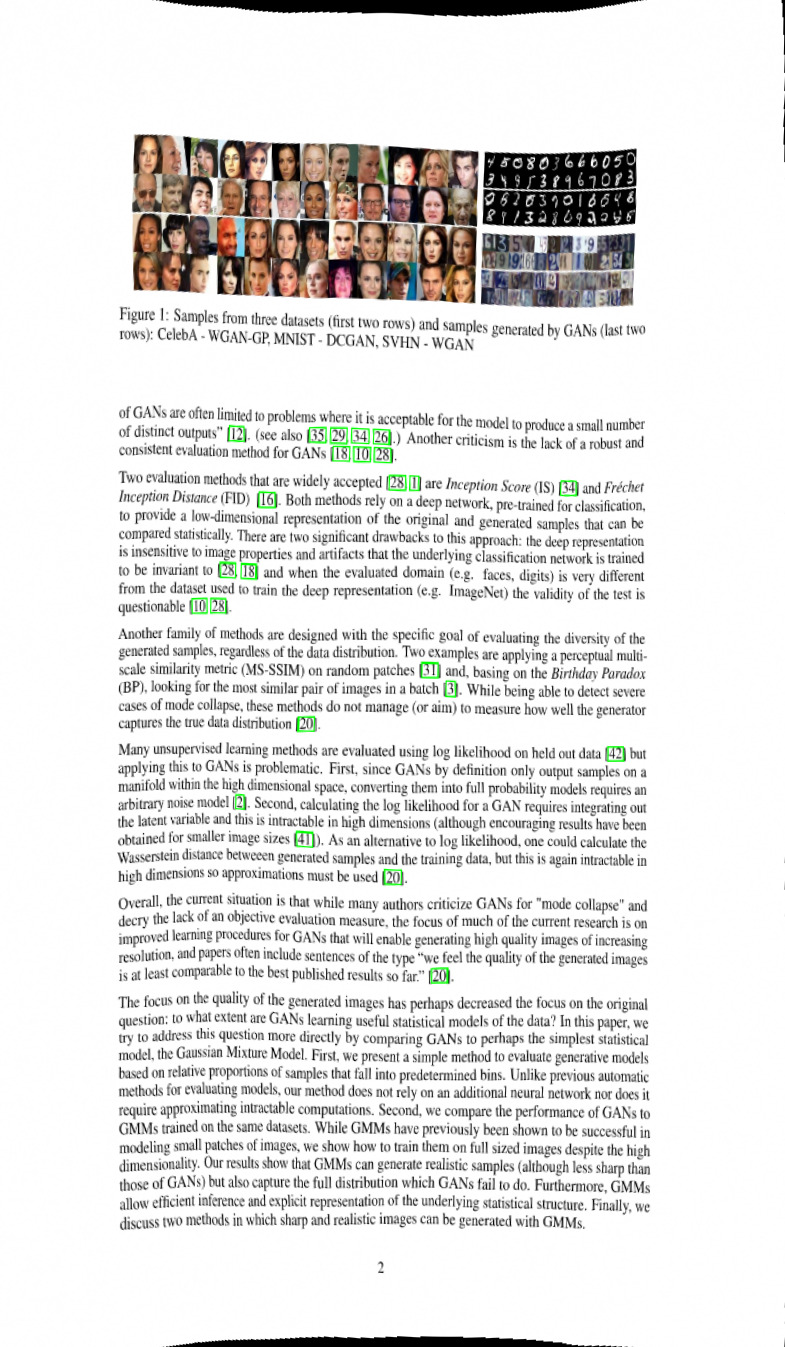}
\\\vspace{\verSpace}
    \includegraphics[height=\teaserImgHeight, width=\teaserImgWidth]{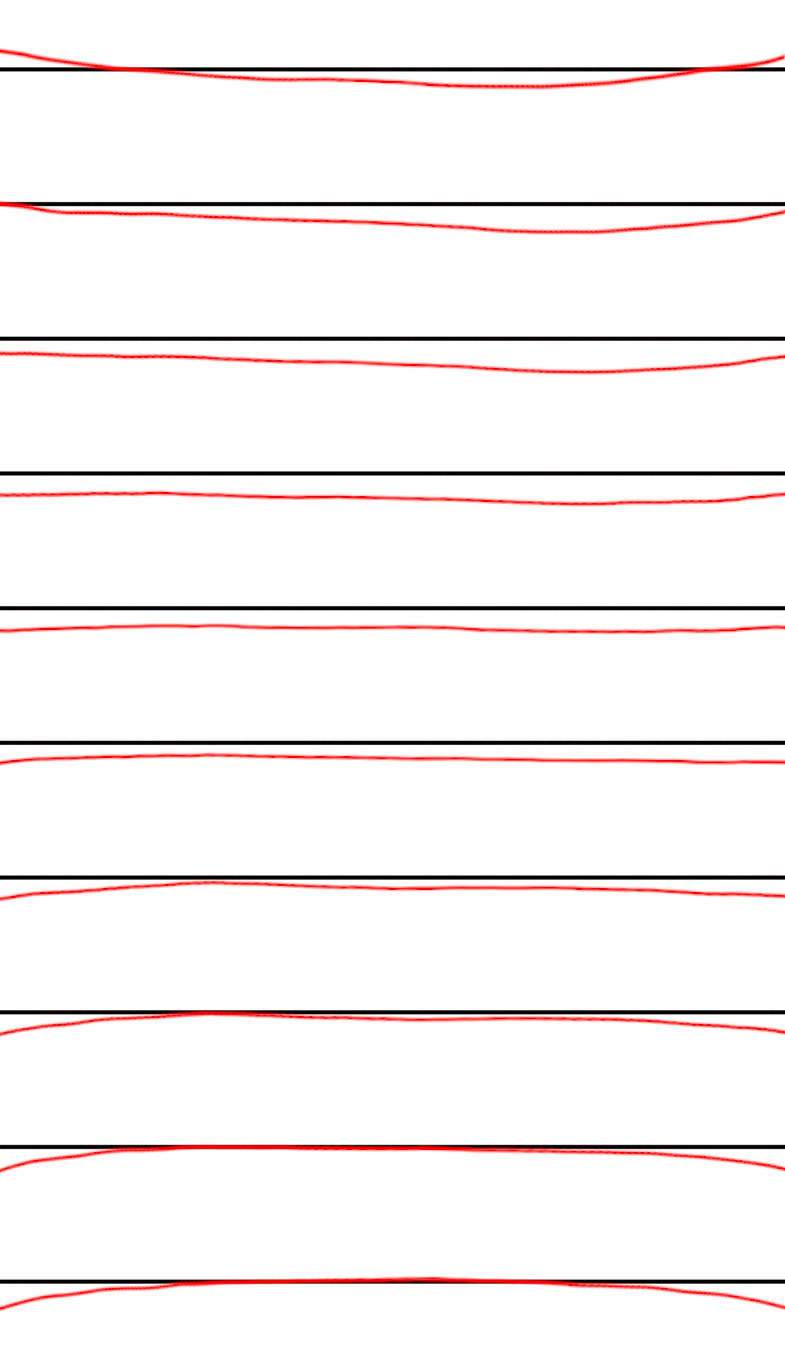}
\\\vspace{\verSpace}
    \includegraphics[height=\teaserImgHeight, width=\teaserImgWidth]{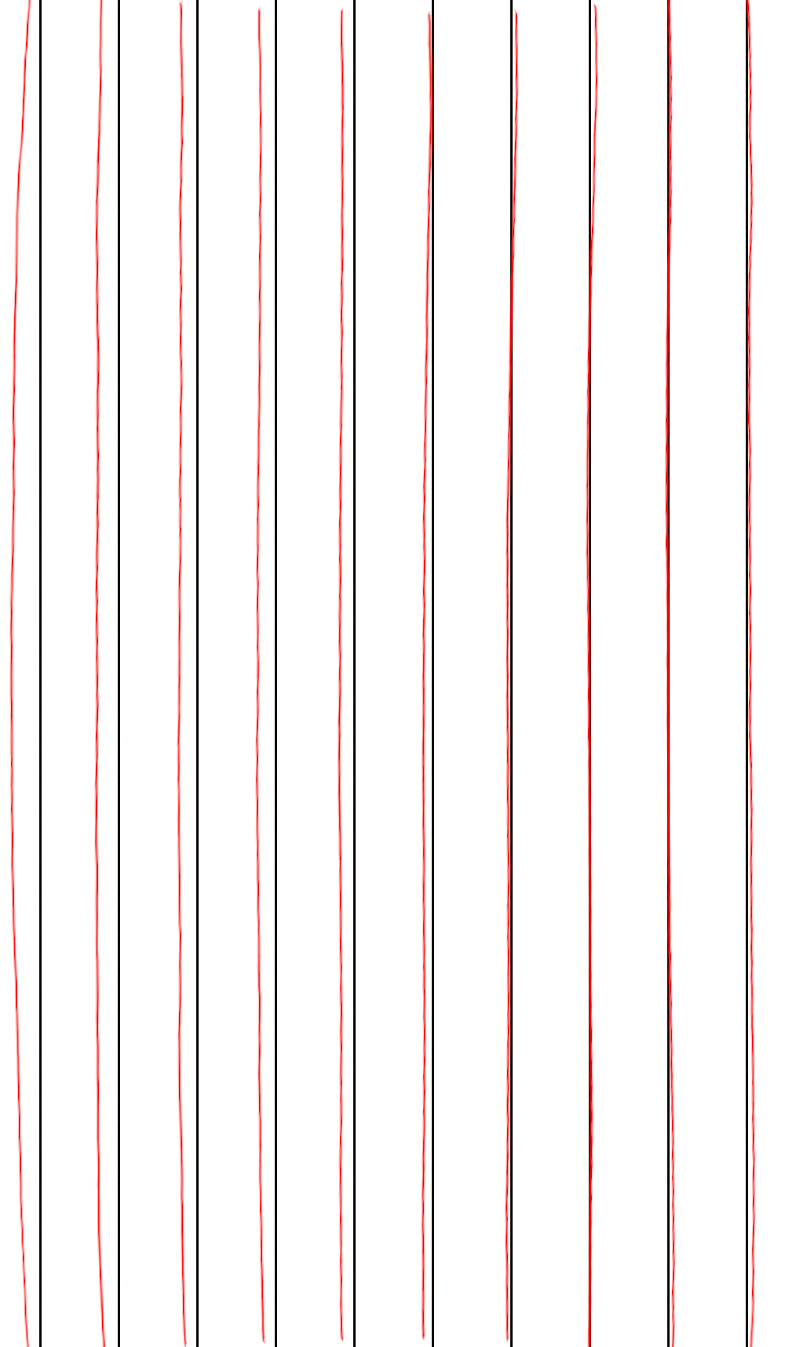}
    \caption{DDCP}
\end{subfigure}
\horSpace
\begin{subfigure}[t]{\teaserSubFigWidth}
    \centering
    \includegraphics[height=\teaserImgHeight, width=\teaserImgWidth]{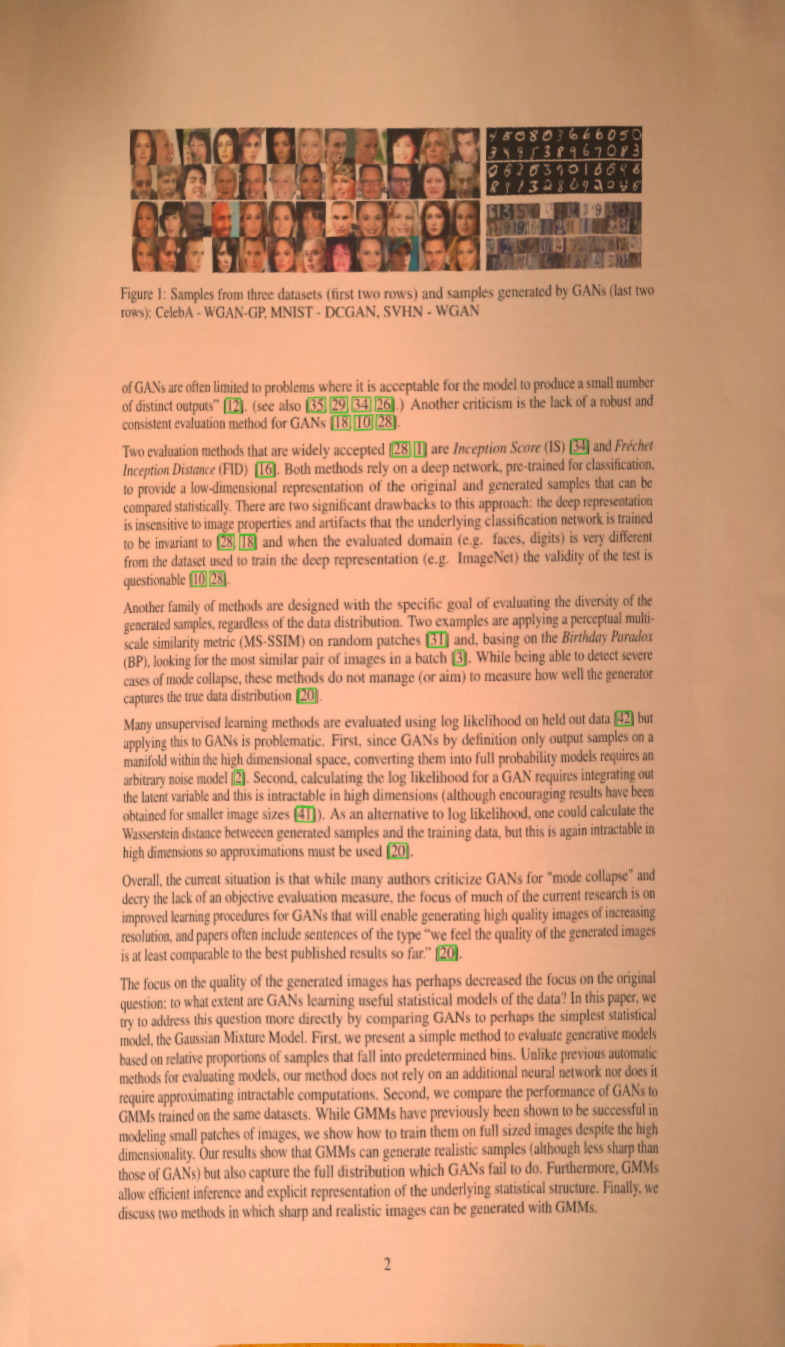}
\\\vspace{\verSpace}
    \includegraphics[height=\teaserImgHeight, width=\teaserImgWidth]{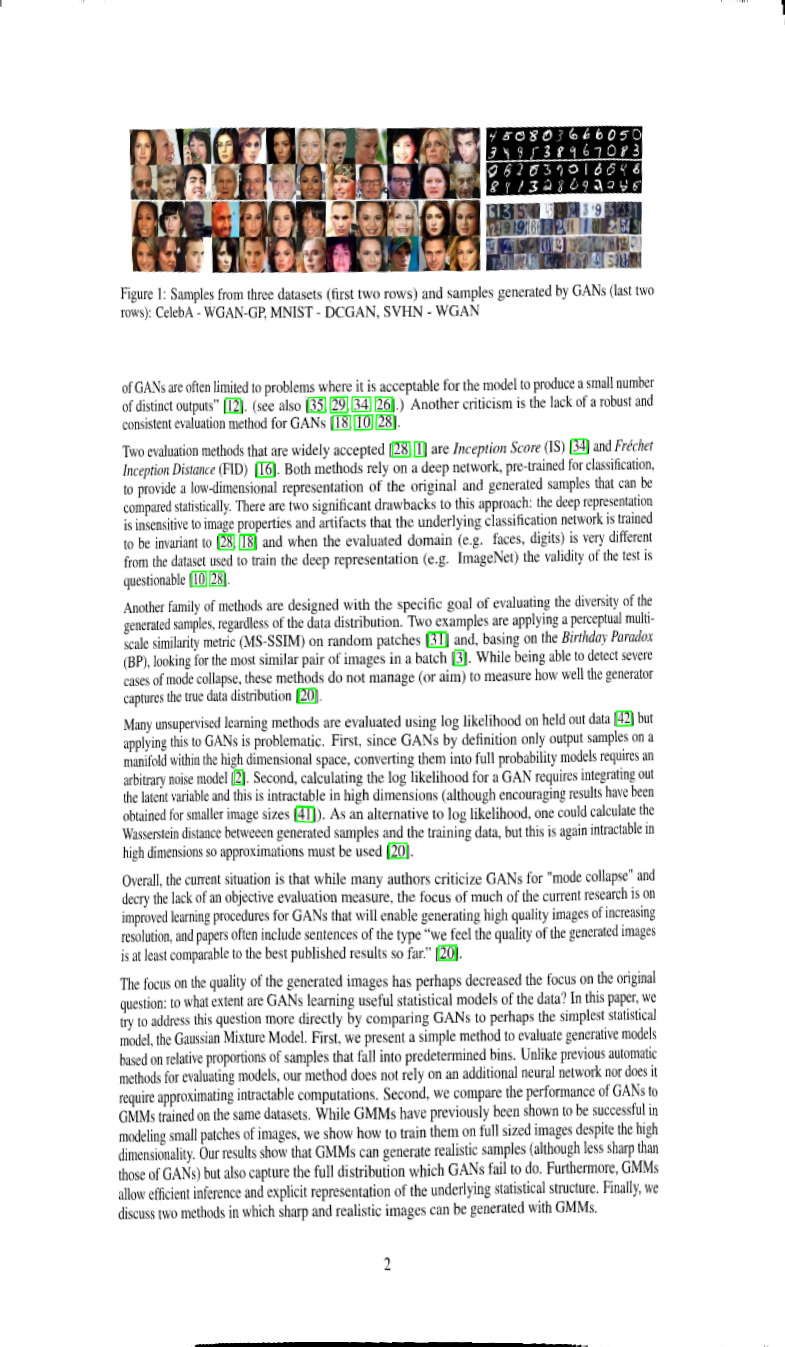}
\\\vspace{\verSpace}
    \includegraphics[height=\teaserImgHeight, width=\teaserImgWidth]{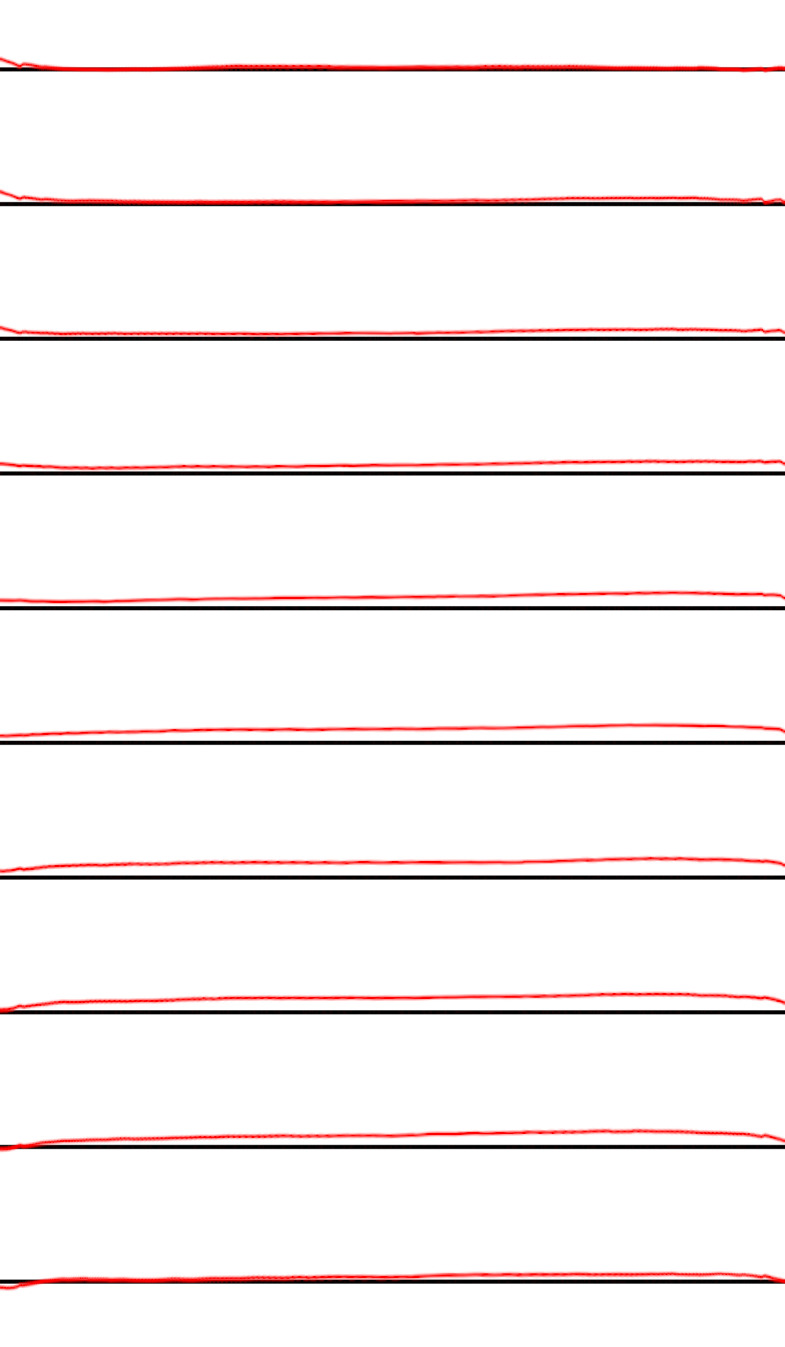}
\\\vspace{\verSpace}
    \includegraphics[height=\teaserImgHeight, width=\teaserImgWidth]{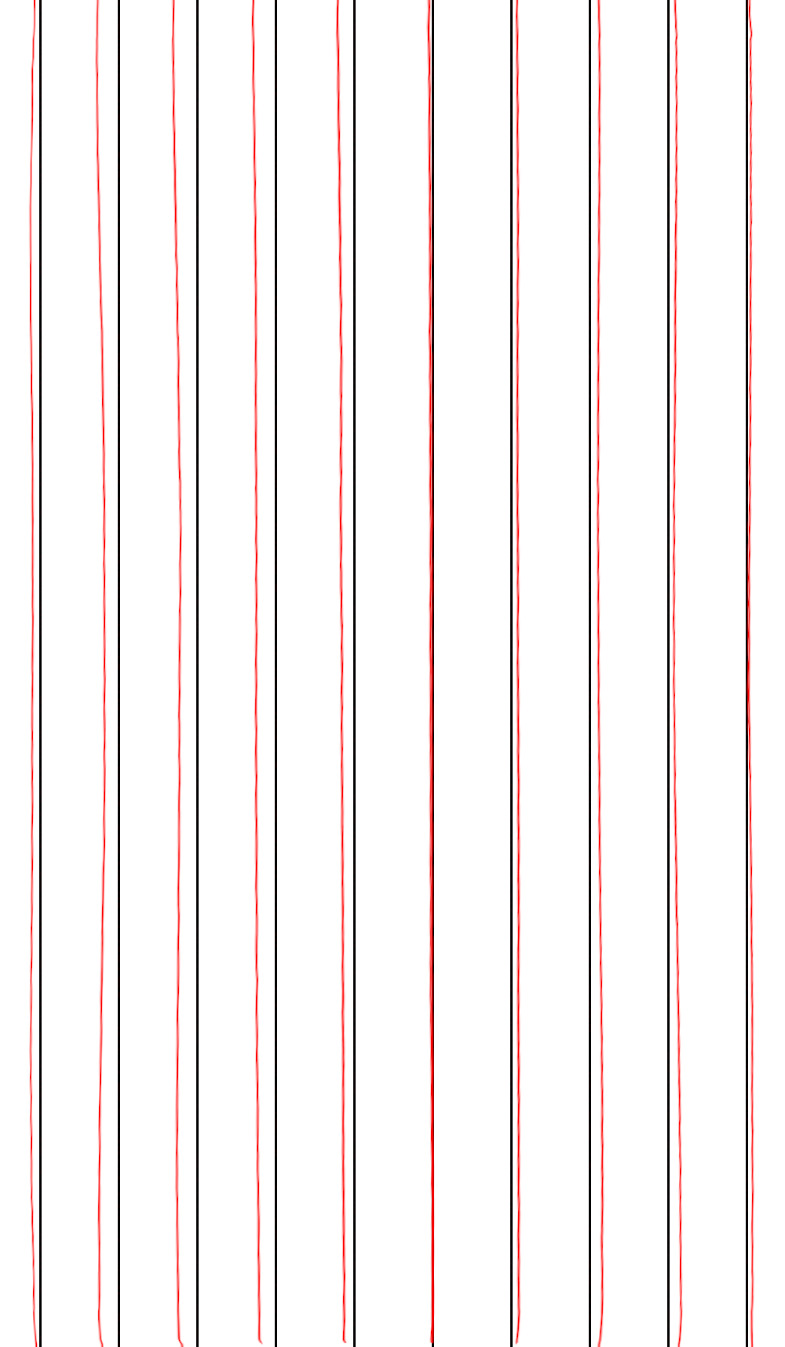}
    \caption{DocTr}
\end{subfigure}
\horSpace
\begin{subfigure}[t]{\teaserSubFigWidth}
    \centering
    \includegraphics[height=\teaserImgHeight, width=\teaserImgWidth]{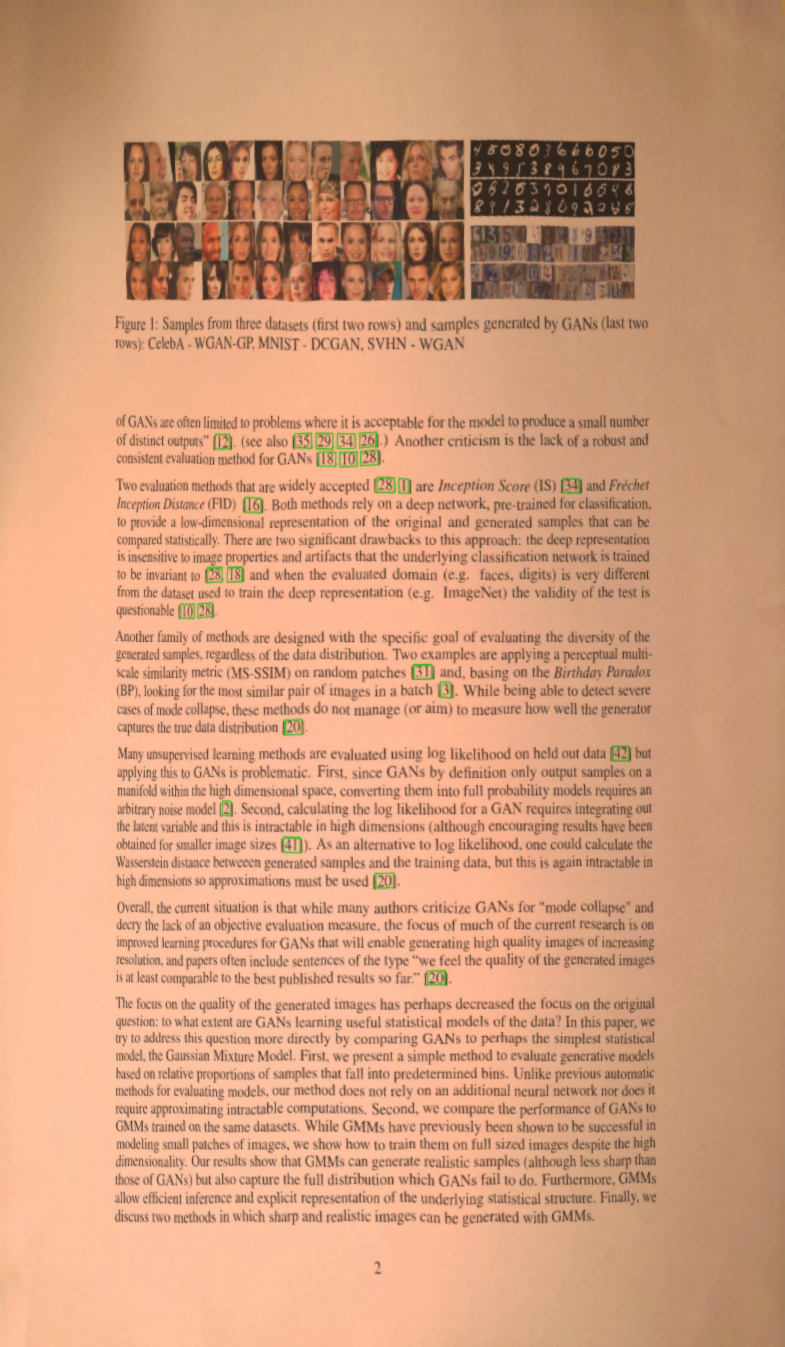}
\\\vspace{\verSpace}
    \includegraphics[height=\teaserImgHeight, width=\teaserImgWidth]{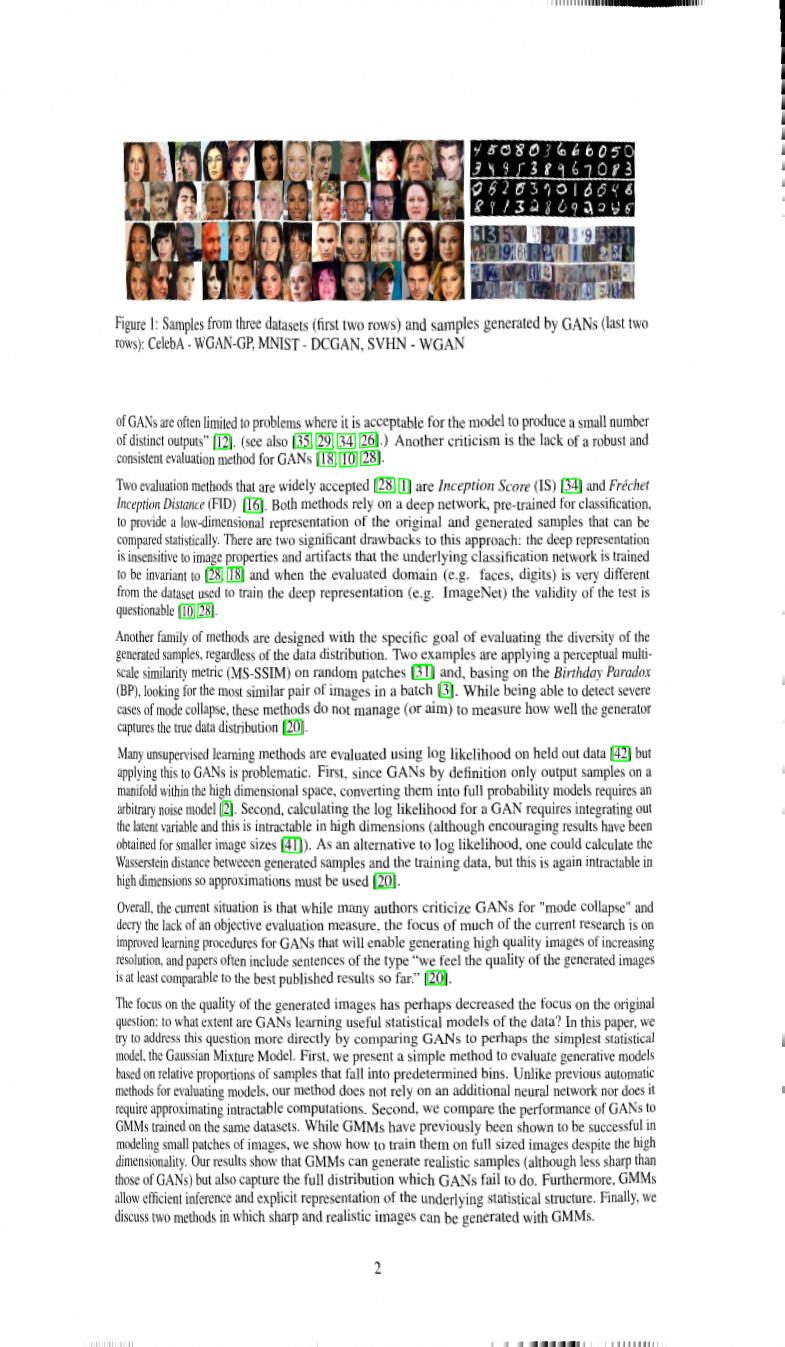}
\\\vspace{\verSpace}
    \includegraphics[height=\teaserImgHeight, width=\teaserImgWidth]{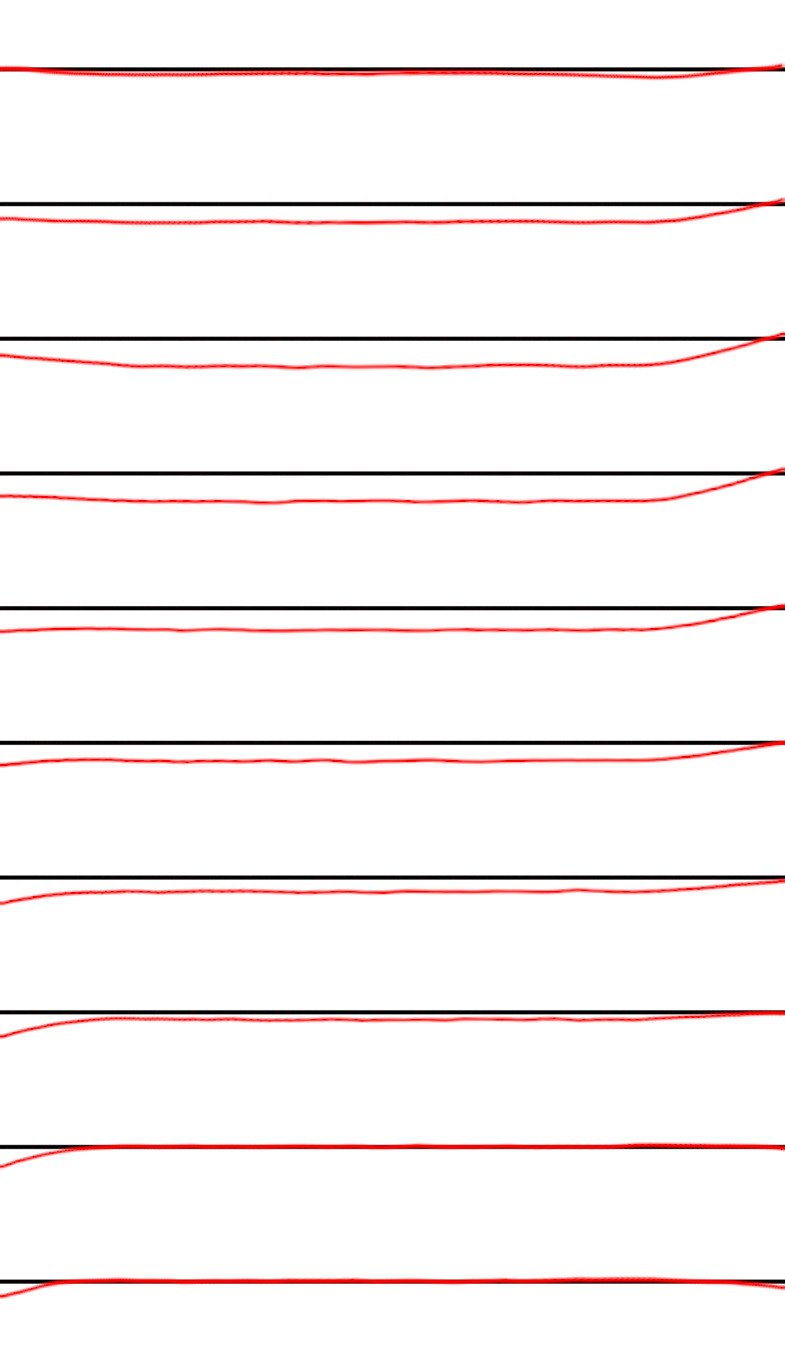}
\\\vspace{\verSpace}
    \includegraphics[height=\teaserImgHeight, width=\teaserImgWidth]{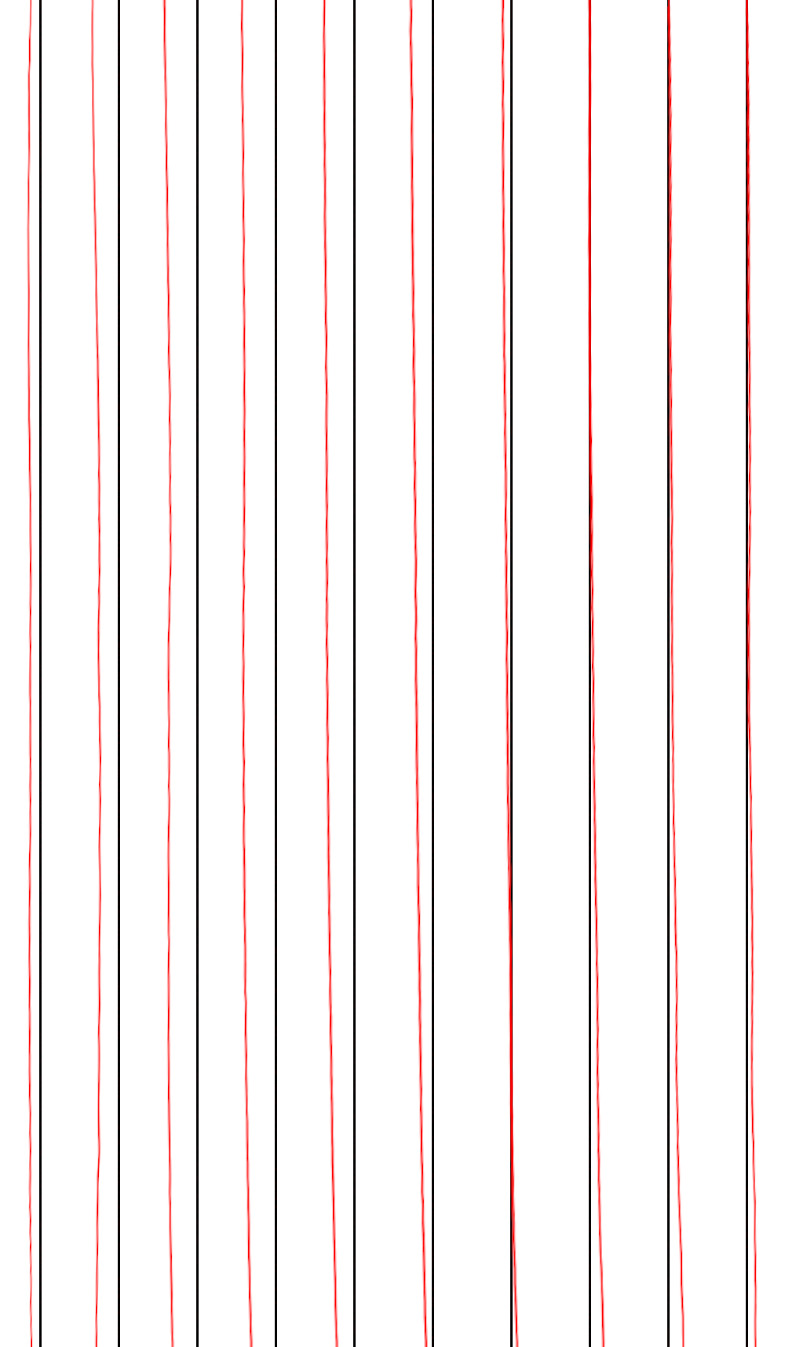}
    \caption{RDGR}
\end{subfigure}
\horSpace
\begin{subfigure}[t]{\teaserSubFigWidth}
    \centering
    \includegraphics[height=\teaserImgHeight, width=\teaserImgWidth]{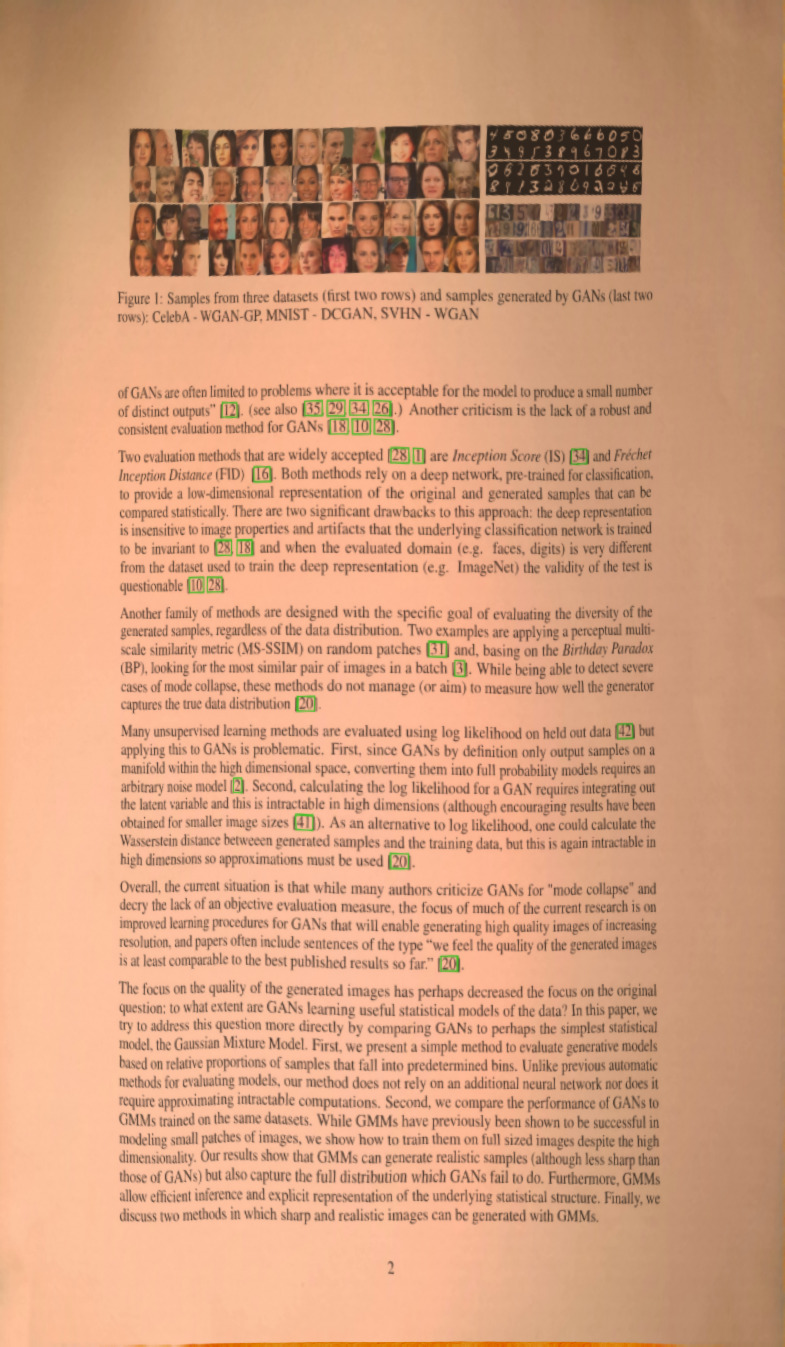}
\\\vspace{\verSpace}
    \includegraphics[height=\teaserImgHeight, width=\teaserImgWidth]{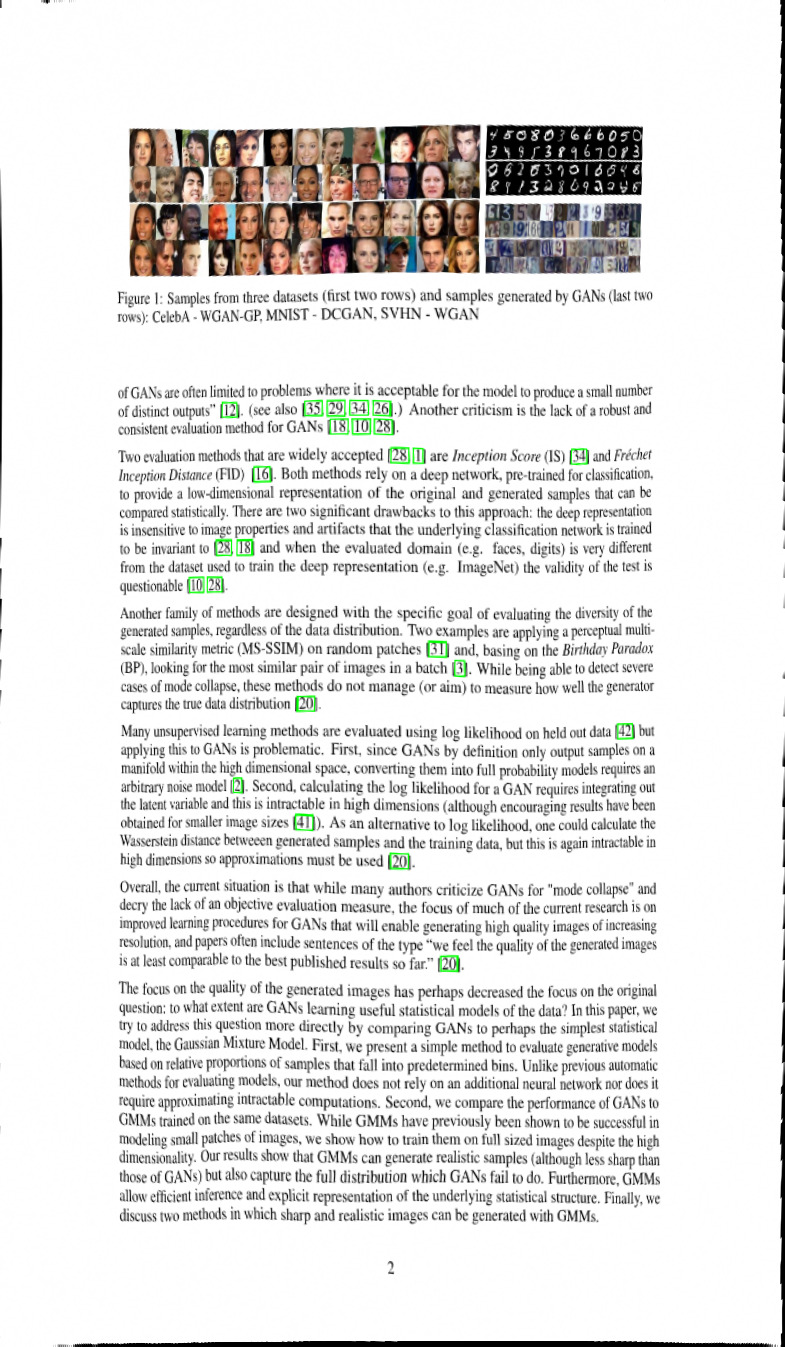}
\\\vspace{\verSpace}
    \includegraphics[height=\teaserImgHeight, width=\teaserImgWidth]{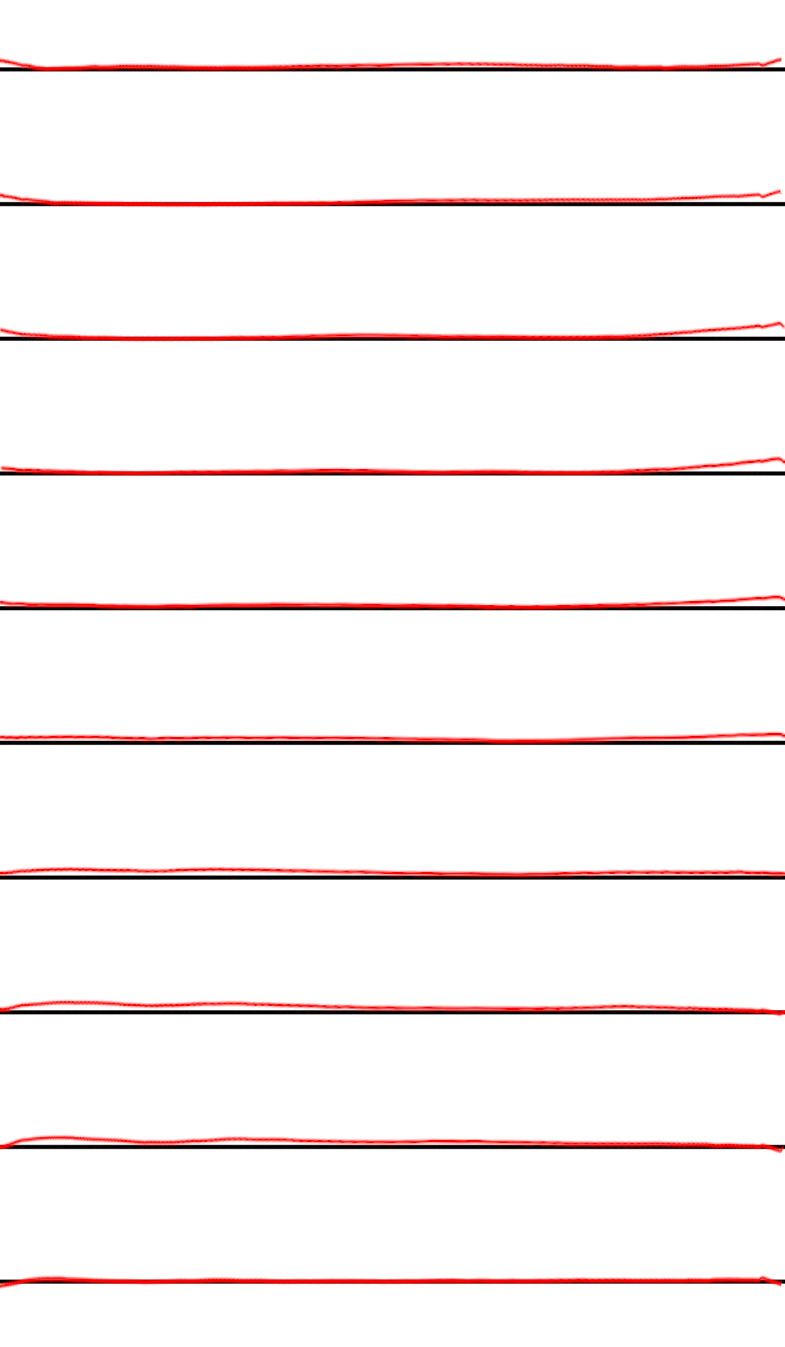}
\\\vspace{\verSpace}
    \includegraphics[height=\teaserImgHeight, width=\teaserImgWidth]{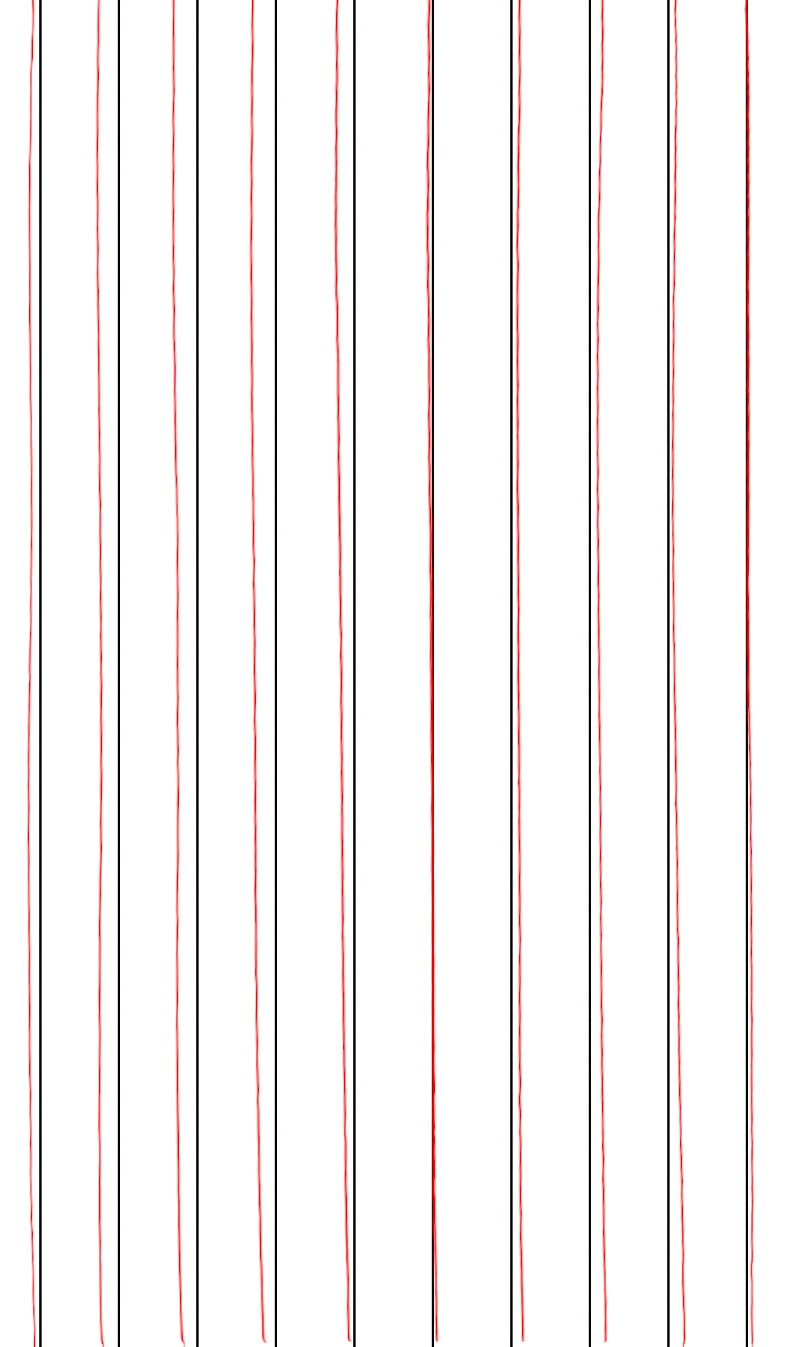}
    \caption{DocGeoNet}
\end{subfigure}
\horSpace
\begin{subfigure}[t]{\teaserSubFigWidth}
    \centering
    \includegraphics[height=\teaserImgHeight, width=\teaserImgWidth]{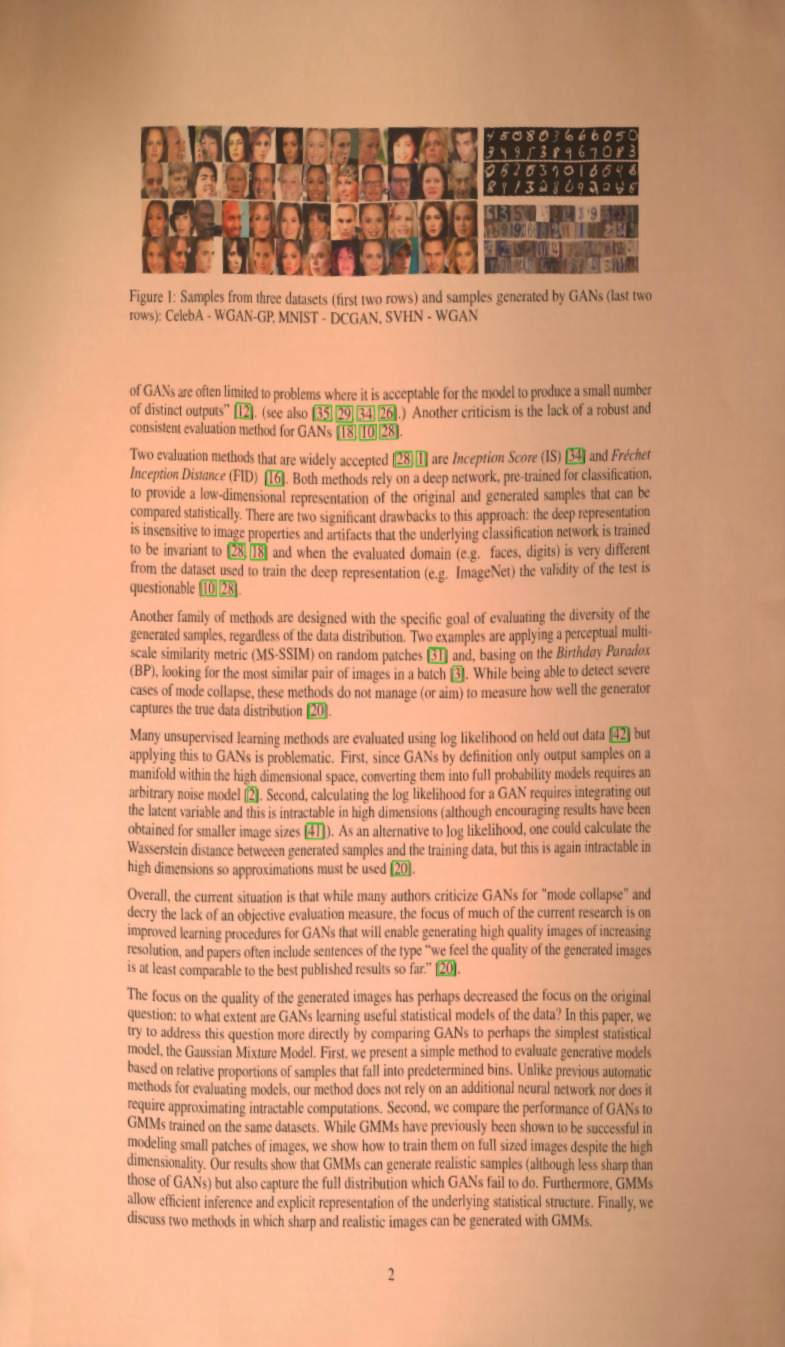}
\\\vspace{\verSpace}
    \includegraphics[height=\teaserImgHeight, width=\teaserImgWidth]{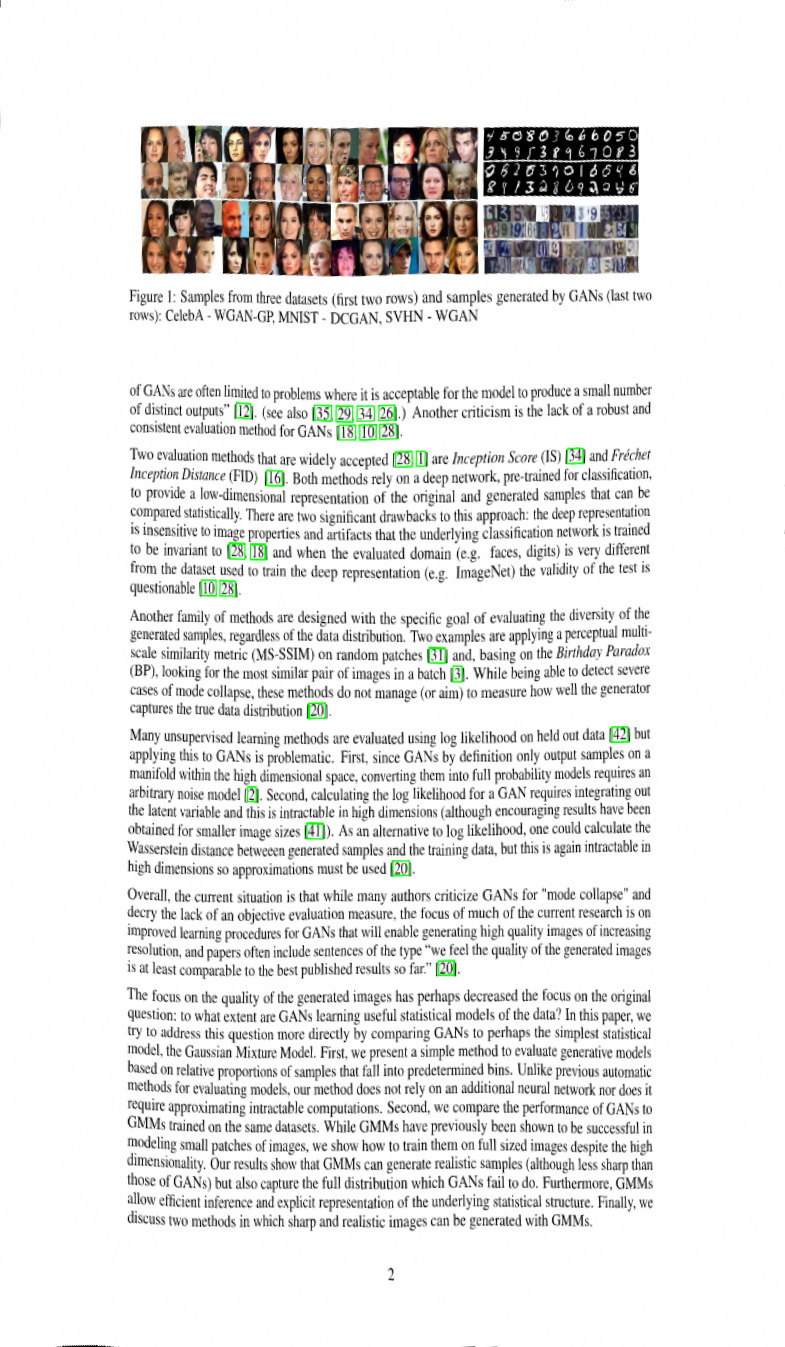}
\\\vspace{\verSpace}
    \includegraphics[height=\teaserImgHeight, width=\teaserImgWidth]{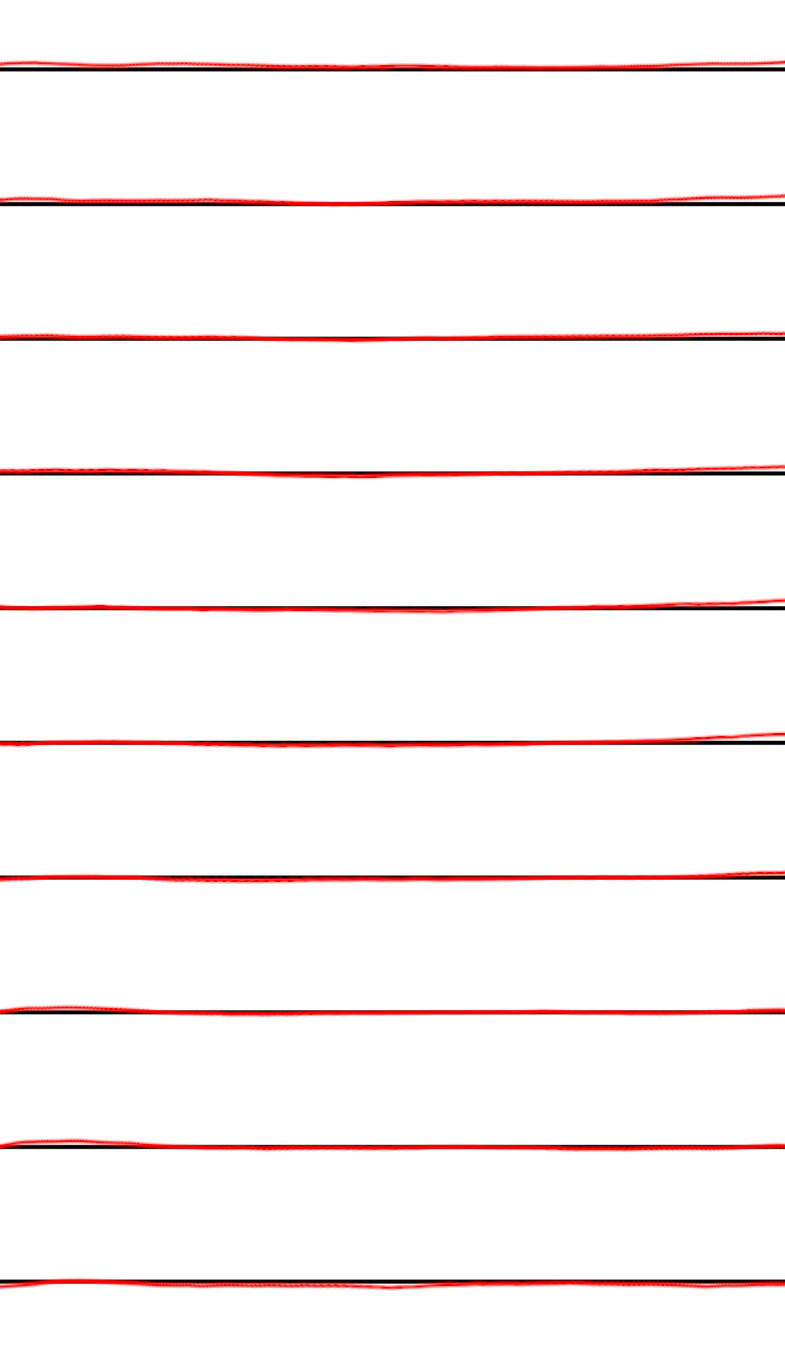}
\\\vspace{\verSpace}
    \includegraphics[height=\teaserImgHeight, width=\teaserImgWidth]{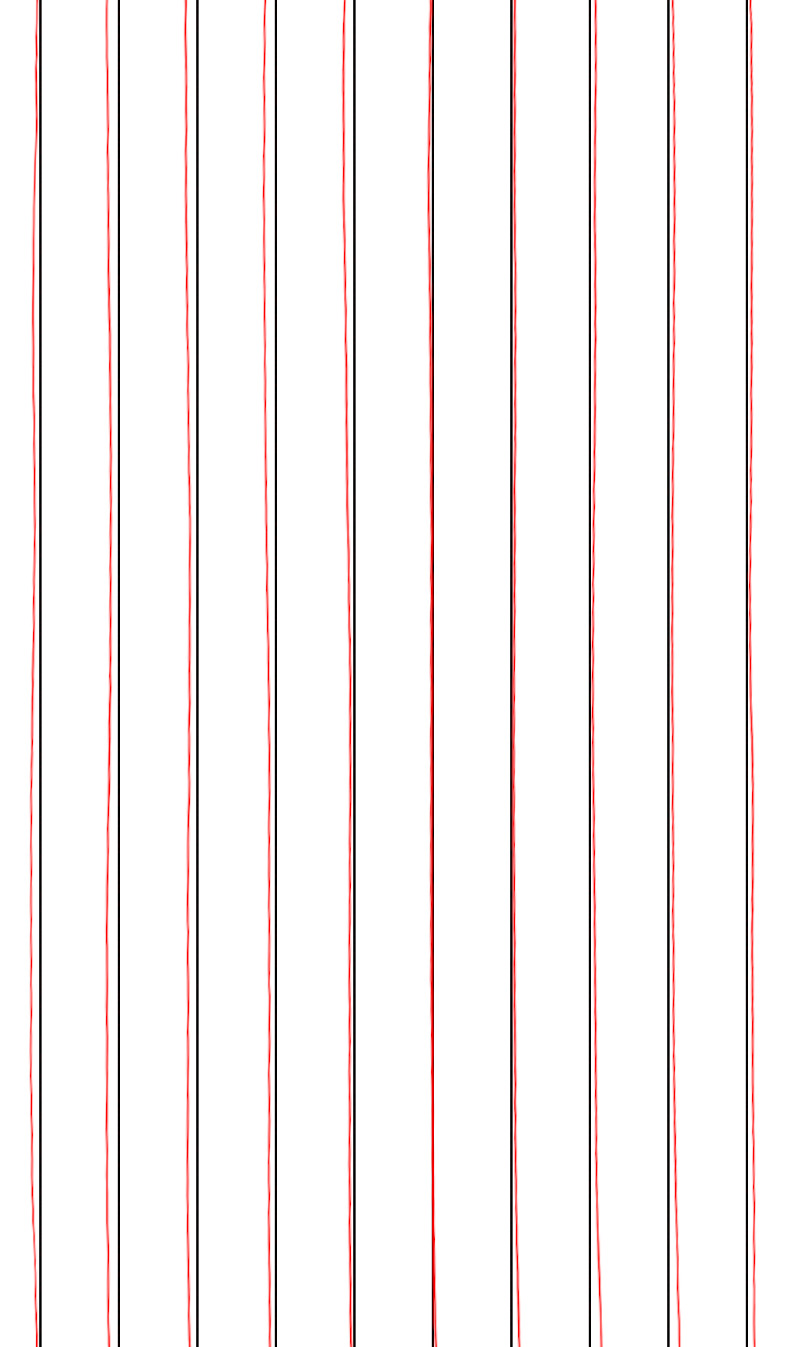}
    \caption{Ours}
\end{subfigure}
\end{center}
\centering
\caption{Results on our UVDoc benchmark. From top to bottom: shaded image, unshaded document texture, horizontal lines, vertical lines. The black lines represent the ground-truth and the red lines are the unwarped ones. From left to right: input, DewarpNet \cite{DewarpNet}, DDCP \cite{DDControlPoints}, DocTr \cite{DocTr}, RDGR \cite{RDGR}, DocGeoNet \cite{DocGeoNet}, ours.}
\label{fig:newmetricline}
\end{figure*}

\end{document}